\pdfoutput=1

\documentclass[10pt,twocolumn,letterpaper]{article}

\usepackage[pagenumbers]{cvpr} %

\providecommand{\mgt}[1]{{\color{magenta}#1}}

\usepackage{multirow}
\usepackage{makecell}   %
\usepackage{colortbl}   %
\usepackage{tabularx}   %
\usepackage{xspace}
\usepackage{pifont}     %

\providecommand{\best}[1]{\textbf{#1}}            %
\providecommand{\second}[1]{\underline{#1}}       %
\providecommand{\na}{--}                          %
\providecommand{\cmark}{\ding{51}}
\providecommand{\xmark}{\ding{55}}
\providecommand{\gr}[1]{\textcolor{gray}{#1}}     %

\providecommand{\mainref}[1]{\cref{#1}}
\providecommand{\Mainref}[1]{\Cref{#1}}

\definecolor{tabgain}{HTML}{1B7F3A}
\definecolor{tabloss}{HTML}{B03A2E}
\providecommand{\dpos}[1]{{\scriptsize\textcolor{tabgain}{(#1)}}}
\providecommand{\dneg}[1]{{\scriptsize\textcolor{tabloss}{(#1)}}}
\providecommand{\dnul}[1]{{\scriptsize\textcolor{gray}{(#1)}}}

\definecolor{oursrowbg}{HTML}{EAF3FB}
\providecommand{\oursrow}{\rowcolor{oursrowbg}}  %

\providecommand{\FigRoot}{figures}

\usepackage{twemojis} %
\usepackage{tikz}
\usetikzlibrary{arrows.meta,positioning,fit,calc,backgrounds,shapes.geometric,
  shapes.misc,decorations.pathreplacing,patterns,matrix,quotes}
\usepackage{pgfplots}
\pgfplotsset{compat=1.18}
\usepgfplotslibrary{groupplots,fillbetween}
\usepackage{xspace}
\ifdefined\DeclareCaptionJustification
  \DeclareCaptionJustification{avragged}{\rightskip=0pt plus 2.5em\relax}
\fi

\definecolor{avBlue}{HTML}{2A78D6}
\definecolor{avOrange}{HTML}{EB6834}
\definecolor{avAqua}{HTML}{1BAF7A}
\definecolor{avYellow}{HTML}{EDA100}
\definecolor{avMagenta}{HTML}{E87BA4}
\definecolor{avGreen}{HTML}{008300}
\definecolor{avViolet}{HTML}{4A3AA7}
\definecolor{avRed}{HTML}{E34948}
\definecolor{avInk}{HTML}{0B0B0B}
\definecolor{avInk2}{HTML}{52514E}    %
\definecolor{avMuted}{HTML}{898781}   %
\definecolor{avGrid}{HTML}{E1E0D9}    %
\definecolor{avAxis}{HTML}{C3C2B7}    %
\colorlet{cBaseline}{avBlue}          %
\colorlet{cOurs}{avAqua}              %
\colorlet{cThird}{avOrange}
\colorlet{cFourth}{avViolet}
\colorlet{cVideo}{avBlue}
\colorlet{cAudio}{avAqua}
\colorlet{cFused}{avViolet}
\colorlet{cTeacher}{avMuted}
\colorlet{cLoss}{avRed}
\colorlet{cPredTok}{avBlue!12}        %
\colorlet{cPredChunk}{avRed!10}       %

\tikzset{
  av/box/.style={draw=avInk2, rounded corners=2pt, line width=0.5pt, align=center,
    font=\footnotesize, inner sep=3pt, fill=white},
  av/module/.style={av/box, minimum height=7mm, minimum width=18mm, fill=black!4},
  av/encoder/.style={av/box, minimum height=9mm, minimum width=24mm, fill=avBlue!8, font=\footnotesize\bfseries},
  av/teacher/.style={av/box, minimum height=9mm, minimum width=24mm, fill=black!5, draw=avMuted},
  av/token/.style={draw=#1!70!black, fill=#1!25, minimum size=2.2mm, inner sep=0pt, line width=0.3pt},
  av/token/.default=avBlue,
  av/arrow/.style={-{Stealth[length=2.2mm,width=1.6mm]}, line width=0.6pt, color=avInk2},
  av/dashedarrow/.style={av/arrow, dashed},
  av/lossarrow/.style={{Stealth[length=2mm,width=1.5mm]}-{Stealth[length=2mm,width=1.5mm]}, line width=0.6pt, color=cLoss},
  av/label/.style={font=\scriptsize, text=avInk2, align=center},
  av/title/.style={font=\footnotesize\bfseries, text=avInk, align=center},
  av/group/.style={draw=avAxis, rounded corners=3pt, line width=0.5pt, inner sep=4pt},
}

\pgfplotsset{
  av/axis/.style={
    width=\linewidth, height=0.62\linewidth,
    axis lines*=left, axis line style={avAxis, line width=0.5pt},
    tick style={avAxis, line width=0.4pt}, tick align=outside, major tick length=2pt,
    ymajorgrids, grid style={avGrid, line width=0.35pt},
    tick label style={font=\scriptsize, text=avInk2},
    label style={font=\footnotesize, text=avInk},
    title style={font=\footnotesize\bfseries, text=avInk, yshift=-1ex},
    legend style={font=\scriptsize, draw=none, fill=none, text=avInk},
    legend cell align=left,
    every axis plot/.append style={line width=1.1pt, line join=round, line cap=round},
    scaled ticks=false,
    /pgf/number format/assume math mode=true, %
  },
  av/marker/.style={mark=*, mark size=1.9pt, mark options={solid, draw=white, line width=0.5pt}},
  av/baseline/.style={color=cBaseline, av/marker, mark=*},
  av/ours/.style={color=cOurs, av/marker, mark=diamond*, mark size=2.4pt},
  av/third/.style={color=cThird, av/marker, mark=square*, mark size=1.7pt},
  av/fourth/.style={color=cFourth, av/marker, mark=triangle*, mark size=2.2pt},
  av/seed/.style={line width=0.5pt, opacity=0.45, mark=*, mark size=1pt, mark options={solid}},
  av/refline/.style={avMuted, line width=0.6pt, dashed},
}

\providecommand{\pnum}[1]{#1}

\newcommand{\nWMvOne}{70.8}%

\usepackage{mathtools}   %
\usepackage{bm}          %
\usepackage{booktabs}
\usepackage{arydshln} %

\makeatletter
\renewcommand{\paragraph}{%
	\@startsection{paragraph}{4}{0pt}%
	{-1.2em}%
	{-0.5em}%
	{\normalfont\normalsize\bfseries}%
}
\makeatother

\newcommand{\leanparagraph}[1]{\vspace{0.1cm}\noindent\textbf{#1}~}

\usepackage{xspace}
\newcommand{\avjepa}{\mbox{JEP-AV}\xspace}

\newcommand{\mjepa}{\mbox{M-JEPA}\xspace}
\newcommand{\vjepa}{\mbox{V-JEPA}\xspace}
\newcommand{\vjepatwo}{\mbox{V-JEPA}~2\xspace}
\newcommand{\vjepatwoone}{\mbox{V-JEPA}~2.1\xspace}
\newcommand{\cavmae}{\mbox{CAV-MAE}\xspace}
\newcommand{\cavmaesync}{\mbox{CAV-MAE} Sync\xspace}
\newcommand{\spidr}{SpidR\xspace}
\newcommand{\dinosr}{DinoSR\xspace}
\newcommand{\protoloss}{ProtoLoss\xspace}

\newcommand{\Masks}{\mathcal{M}}                    %
\newcommand{\Items}{\mathcal{I}}                    %

\newcommand{\Clip}{\mathbf{x}}                     %
\newcommand{\MaskedClip}{\Clip_{\bar \Masks}}                     %
\newcommand{\mto}[2]{#1\to #2}            %

\newcommand{\params}{\theta}              %
\newcommand{\paramsT}{\bar\theta}         %
\newcommand{\paramsP}{\phi}               %
\newcommand{\paramsG}{\psi}               %
\newcommand{\Enc}{E_{\params}}            %
\newcommand{\EncT}{E_{\paramsT}}          %
\newcommand{\Pred}{P_{\paramsP}}          %
\newcommand{\ChunkPredSimple}{g_{\paramsG}}  

\newcommand{\avwm}{\mbox{AVJ-WM}\xspace}
\newcommand{\avnavbench}{\mbox{AVNav-Bench}\xspace}  %

\newcommand{\utarget}{\mathbf{u}}
\newcommand{\upred}{\mathbf{\hat{u}}}

\newcommand{\ChunkT}[1]{\mathbf{s}_{#1}}

\newcommand{\AudioToken}{\mathbf{e}^a_t}
\newcommand{\AudioTokenT}[1]{\mathbf{e}^a_{#1}}

\newcommand{\PredAudioTokenT}[1]{\mathbf{\hat{e}}^a_{#1}}

\newcommand{\VideoToken}{\mathbf{e}^v_t}
\newcommand{\VideoTokenT}[1]{\mathbf{e}^v_{#1}}

\newcommand{\PredVideoTokenT}[1]{\mathbf{\hat{e}}^v_{#1}}

\newcommand{\ActionT}[1]{\mathbf{a}_{#1}}

\newcommand{\intphysx}{\mbox{IntPhys-X}\xspace}
\newcommand{\nIPXlongPer}{\pnum{2,000}}   %
\newcommand{\nIPXsynthPer}{\pnum{5,000}}  %
\newcommand{\nIPXrealPer}{\pnum{1,000}}   %

\newcommand{\zs}[1]{\mathbf{z}^{s}_{#1}}           %
\newcommand{\zt}[1]{\mathbf{z}^{t}_{#1}}           %
\newcommand{\qs}[1]{\mathbf{q}^{s}_{#1}}           %
\newcommand{\qt}[1]{\mathbf{q}^{t}_{#1}}           %
\newcommand{\qbar}{\bar{\mathbf{q}}}                %
\newcommand{\Codebook}{\mathbf{C}}                 %

\newcommand{\taus}{\tau_s}                %
\newcommand{\taut}{\tau_t}                %
\newcommand{\lamH}{\lambda_{\mathrm{H}}}  %
\newcommand{\lamME}{\lambda_{\mathrm{ME}}}%

\newcommand{\Loss}{\mathcal{L}}

\newcommand{\LKL}{\Loss_{\mathrm{KL}}}
\newcommand{\LH}{\Loss_{\mathrm{H}}}
\newcommand{\LME}{\Loss_{\mathrm{ME}}}

\DeclareMathOperator*{\argmax}{arg\,max}

\usepackage{placeins}%

\definecolor{cvprblue}{rgb}{0.21,0.49,0.74}
\usepackage[pagebackref,breaklinks,colorlinks,allcolors=cvprblue]{hyperref}

\DeclareRobustCommand{\suppref}[1]{\cref{#1}}
\DeclareRobustCommand{\mainref}[1]{\cref{#1}}
\DeclareRobustCommand{\Mainref}[1]{\Cref{#1}}

\AtBeginDocument{%
  \crefname{appendix}{App.}{Apps.}\Crefname{appendix}{Appendix}{Appendices}
  \crefname{subappendix}{App.}{Apps.}\Crefname{subappendix}{Appendix}{Appendices}
  \crefname{subsubappendix}{App.}{Apps.}\Crefname{subsubappendix}{Appendix}{Appendices}}

\newcommand{\titlenote}{%
  \par\nopagebreak\vspace{0.6em}\noindent{\normalfont\footnotesize\rule{0.4\columnwidth}{0.4pt}\par\noindent $^{*}$Equal contribution. $^{\ddagger}$Work done while at Meta. Correspondence to: {\tt \{angelvillar,philliprust,or.tal1\}@meta.com}\par}\vspace{-12pt}}

\title{Predicting Sight and Sound for Audio-Visual Perception, \\ World Modeling and Intuitive Physics}
\vspace{-0.1cm}
\author{
	Angel Villar-Corrales$^{1,*}$ \quad Phillip Rust$^{1,*}$ \quad Or Tal$^{1,2,*}$ \quad Revant Teotia$^{3,*,\ddagger}$ \quad Dongyan Lin$^{1}$
\\
	Jiaang Li$^{4,\ddagger}$ \quad Mahi Luthra$^{1}$ \quad Yosuke Higuchi$^{1}$ \quad Jiayi Shen$^{1}$ \quad Tom Fizycki$^{1,5}$
\\
	Charles-\'Eric Saint-James$^{1}$ \quad Surya Parimi$^{1}$ \quad Bal\'azs K\'egl$^{6,\ddagger}$ \quad
	Emmanuel Dupoux$^{1,7}$ \quad Juan Pino$^{1}$
\\[0.5em]
	{\small $^{1}$Meta Superintelligence Labs \quad $^{2}$Hebrew University of Jerusalem \quad $^{3}$New York University \quad
	$^{4}$University of Copenhagen}
\\[-0.1em]
	{\small $^{5}$Universit\'e Paris Dauphine-PSL \quad $^{6}$Doctolib \quad
	$^{7}$\'Ecole des hautes \'etudes en sciences sociales}
\\[0.4em]
	{\small \url{https://github.com/facebookresearch/jep-av}}
	\vspace{-0.2cm}
}

\begin{document}
\maketitle

\begin{abstract}
Sight and sound provide complementary supervision on a shared time axis.
Yet, audio-visual (AV) self-supervised methods commonly encode the two streams separately and align them only per clip,
and few benchmarks require integrating them; we address both gaps.
We introduce \emph{\avjepa}, a joint-embedding predictive architecture whose single transformer encodes video,  audio, or both on a shared time axis.
It predicts a teacher's assignments to learned prototypes (\protoloss) within and across modalities, using token-level and $500$\,ms chunk-level targets.
We explore AV world modeling in this representation space with \avwm, an autoregressive world model building on \avjepa that predicts future AV features and plans via lightweight action adapters, and compare it with diffusion world models trained on the same features.
To evaluate cross-modal integration, we introduce two benchmarks: \emph{\intphysx}, whose intuitive-physics tasks can
only be solved by combining sight and sound, from matching the two streams to reasoning over sound--action associations
in context, and \emph{\avnavbench}, in which the world model itself plans toward image or sound goals.
\avjepa sets the state of the art among frozen self-supervised encoders on audio and AV classification and on
sound-prompted segmentation.
On \avnavbench, \avwm reaches the highest offline success on image and sound goals 9\,s ahead, scoring 39.6\% and 19.1\% \vs at most 26.9\% and
5.0\% for diffusion world models.
On \intphysx no-context scenarios, \avjepa averages 65.5\%, outperforming all representation models and on par with Gemini-4 Argon (65.3\%).
Even the best models trail humans (84--99\%) by over $20$ points on average, leaving cross-modal physical understanding an open challenge.
\titlenote
\end{abstract}

\vspace{-0.cm}
\section{Introduction}
\label{sec:intro}

\def\teaservariant{v4}%
\begin{figure}[t]
  \centering
  \providecommand{\tavB}[1]{\protect\tikz[baseline=(tavb.base)]\protect\node[circle, fill=cFused,
    inner sep=0pt, minimum size=2.9mm, font=\scriptsize\bfseries, text=white] (tavb) {#1};}%
  \captionsetup[sub]{justification=avragged}%
  \begin{subfigure}[t]{0.47\linewidth}
    \vspace{0pt}%
    \centering
    \begin{tikzpicture}[x=1mm, y=1mm, line cap=round, line join=round,
  tav/lbl/.style={font=\scriptsize, text=avInk2, inner sep=0pt},
  tav/key/.style={font=\scriptsize\bfseries, text=avInk, inner sep=0pt},
  tav/badge/.style={circle, fill=cFused, draw=white, line width=0.6pt, inner sep=0pt,
    minimum size=3.1mm, font=\scriptsize\bfseries, text=white},
  tav/arr/.style={-{Stealth[length=1.15mm,width=1mm]}, line width=0.45pt, color=avInk2},
  tav/link/.style={{Stealth[length=1mm,width=0.95mm]}-{Stealth[length=1mm,width=0.95mm]},
    line width=0.5pt, color=cLoss},
  tav/uni/.style={-{Stealth[length=0.95mm,width=0.9mm]}, line width=0.5pt, color=cLoss},
  tav/card/.style={draw=#1!75!black, fill=#1!8, line width=0.4pt, rounded corners=0.5mm,
    inner sep=0pt, minimum width=5.4mm, minimum height=5.2mm},
  tav/tok/.style={draw=cVideo!72!black, fill=cVideo!30, line width=0.35pt,
    rounded corners=0.25mm, inner sep=0pt, minimum size=1.65mm},
  tav/mask/.style={draw=avInk, fill=avInk, line width=0.35pt,
    rounded corners=0.25mm, inner sep=0pt, minimum size=1.65mm},
  tav/band/.style={fill=avGrid!45, draw=none},
]
\def\tavFrame#1#2#3#4#5{%
  \begin{scope}
    \clip[rounded corners=0.45mm] ({#1},{#2}) rectangle ++({#3},{#4});
    \fill[cVideo!9] ({#1},{#2}) rectangle ++({#3},{#4});
    \fill[cVideo!30] ({(#1)+(#5)*(#3)},{(#2)+0.70*(#4)}) circle ({0.14*(#4)});
    \fill[cVideo!50] ({#1},{#2}) -- ++({0.30*(#3)},{0.56*(#4)}) -- ++({0.18*(#3)},{-0.26*(#4)})
      -- ++({0.18*(#3)},{0.30*(#4)}) -- ({(#1)+(#3)},{(#2)+0.12*(#4)}) -- ({(#1)+(#3)},{#2}) -- cycle;
  \end{scope}
  \draw[cVideo!75!black, line width=0.4pt, rounded corners=0.45mm] ({#1},{#2}) rectangle ++({#3},{#4});}
\def\tavStack#1#2#3#4{%
  \foreach \tavI/\tavS in {3/0.22, 2/0.42, 1/0.62, 0/0.82}
    {\tavFrame{(#1)+0.6*\tavI}{(#2)+0.5*\tavI}{#3}{#4}{\tavS}}}
\def\tavWaveSeg#1#2#3#4#5#6#7{%
  \foreach \tavI in {0,...,#5} {
    \pgfmathsetmacro{\tavU}{\tavI/#5}
    \ifcase#6
      \pgfmathsetmacro{\tavEnv}{0.10 + 0.95*(\tavU>0.08)*exp(-(\tavU-0.08)/0.13)
        + 0.70*(\tavU>0.56)*exp(-(\tavU-0.56)/0.11)}
      \pgfmathsetmacro{\tavTex}{0.45 + 0.55*abs(sin(\tavI*67+#7)*cos(\tavI*29+2*#7))}
    \or
      \pgfmathsetmacro{\tavEnv}{0.40 + 0.45*exp(-((mod(\tavU*4+0.85,1)-0.15)/0.20)^2)}
      \pgfmathsetmacro{\tavTex}{0.80 + 0.20*cos(\tavI*120)}
    \or
      \pgfmathsetmacro{\tavEnv}{0.06 + 0.85*exp(-((\tavU-0.17)/0.09)^2)
        + 0.60*exp(-((\tavU-0.46)/0.07)^2) + 0.95*exp(-((\tavU-0.77)/0.10)^2)}
      \pgfmathsetmacro{\tavTex}{0.45 + 0.55*abs(sin(\tavI*67+#7)*cos(\tavI*29+2*#7))}
    \fi
    \pgfmathsetmacro{\tavAmp}{#4*max(0.08, min(1, \tavEnv*\tavTex))}
    \draw[cAudio!85!black, line width=0.42pt] ({#1+\tavU*#3},{#2-\tavAmp}) -- ({#1+\tavU*#3},{#2+\tavAmp});
  }}
\def\tavHist#1#2#3#4#5{%
  \edef\tavList{#4}%
  \foreach \tavH [count=\tavK] in \tavList {
    \pgfmathsetmacro{\tavX}{#1-2.06+(\tavK-1)*0.88}
    \ifnum\tavK=#5\relax\colorlet{tavFill}{#3}\else\colorlet{tavFill}{#3!35}\fi
    \fill[tavFill] (\tavX,{#2-1.95}) rectangle ++(0.6,\tavH);
  }
  \draw[#3!75!black, line width=0.3pt] ({#1-2.25},{#2-1.95}) -- ({#1+2.25},{#2-1.95});}
\def\tavWaveTokV#1#2#3#4{%
  \fill[cAudio!14] ({#1-1.4},{#2-2.5}) rectangle ({#1+1.4},{#2+2.5});
  \foreach \tavC in {1,...,24}
    {\draw[cAudio!55, line width=0.12pt] ({#1-1.4},{#2-2.5+\tavC*0.2}) -- ++(2.8,0);}
  \ifnum#3>0\relax
    \fill[avInk] ({#1-1.4},{#2-2.5+(#3-1)*0.2}) rectangle ({#1+1.4},{#2-2.5+#4*0.2});
  \fi
  \draw[cAudio!80!black, line width=0.4pt] ({#1-1.4},{#2-2.5}) rectangle ({#1+1.4},{#2+2.5});}

\def\tavXv{6.77}       %
\def\tavXcv{14.045}     %
\def\tavXca{23.445}     %
\def\tavXa{29.345}      %
\def\tavBx{3.495}      %
\def\tavBX{31.245}      %
\def\tavYb{18.9}       %

\tavStack{3.6}{3.4}{5.6}{4.0}
\tavWaveSeg{18.5}{7.6}{4.1}{2.0}{11}{0}{10}
\tavWaveSeg{22.95}{7.6}{4.1}{2.0}{11}{1}{0}
\tavWaveSeg{27.4}{7.6}{4.1}{2.0}{11}{2}{40}
\node[tav/lbl, rotate=90, anchor=south] at (3.15,5.4) {video};
\node[tav/badge] at (14.35,3.95) {1};
\node[tav/lbl, anchor=base west] at (15.95,3.2)
  {sound\,\textperiodcentered\,music\,\textperiodcentered\,speech};
\draw[tav/arr] (7.3,9.2) -- (7.3,11.5);
\draw[tav/arr] (25.0,9.85) -- (25.0,11.5);

\node[draw=cFused, fill=cFused!14, line width=0.5pt, rounded corners=1.2mm, minimum width=27.75mm,
  minimum height=4.6mm, inner sep=0pt, font=\scriptsize\bfseries, text=cFused!80!black]
  (tavEnc) at ({(\tavBx+\tavBX)/2},14.0) {one joint encoder};
\node[tav/badge] at (tavEnc.west) {2};
\draw[tav/arr] (\tavXv,16.55) -- (\tavXv,{\tavYb-0.15});
\draw[tav/arr] (\tavXa,16.55) -- (\tavXa,{\tavYb-0.15});
\node[tav/mask] at (13.7,17.65) {};
\node[tav/lbl, anchor=west] at (14.95,17.65) {masked};

\begin{scope}[on background layer]
  \foreach \tavK in {0,2} {\fill[tav/band] (\tavBx,{\tavYb+5.8*\tavK}) rectangle (\tavBX,{\tavYb+5.8*(\tavK+1)});}
\end{scope}
\draw[-{Stealth[length=0.9mm,width=0.8mm]}, line width=0.4pt, color=avMuted] (2.3,{\tavYb+0.4}) -- (2.3,{\tavYb+17.4-0.4});
\node[tav/lbl, text=avMuted, rotate=90, anchor=south] at (1.95,{\tavYb+8.7}) {time};
\foreach \tavK/\tavMv/\tavSv/\tavMa/\tavMb/\tavSa/\tavHv/\tavPv/\tavHa/\tavPa in {
    0/3/0/12/21/4/{0.5,3.2,0.65,0.4,0.55}/2/{0.55,0.45,3.2,0.6,0.45}/3,
    1/3/1/0/0/0/{0.45,0.6,0.55,3.2,0.5}/4/{0.5,0.65,0.45,0.55,3.1}/5,
    2/3/1/4/13/22/{0.55,3.0,0.75,0.45,0.4}/2/{0.45,0.55,3.0,0.65,0.55}/3} {
  \pgfmathsetmacro{\tavYc}{\tavYb+2.9+5.8*\tavK}    %
  \pgfmathsetmacro{\tavYt}{\tavYc-0.3}              %
  \foreach \tavCol in {1,2,3} {
    \pgfmathsetmacro{\tavXt}{\tavXv+(\tavCol-2)*1.95}
    \ifnum\tavCol=\tavMv\relax
      \node[tav/mask] at (\tavXt,{\tavYt+0.975}) {};
      \node[tav/mask] at (\tavXt,{\tavYt-0.975}) {};
    \else
      \node[tav/tok] at (\tavXt,{\tavYt+0.975}) {};
      \node[tav/tok] at (\tavXt,{\tavYt-0.975}) {};
    \fi
  }
  \tavWaveTokV{\tavXa}{\tavYc}{\tavMa}{\tavMb}
  \fill[cVideo!14] ({\tavXv+2.975},{\tavYt-2.05}) -- ({\tavXv+2.975},{\tavYt+2.05})
    -- ({\tavXcv-2.7},{\tavYc+1.0}) -- ({\tavXcv-2.7},{\tavYc-1.0}) -- cycle;
  \fill[cAudio!20] ({\tavXa-1.55},{\tavYc-2.5}) -- ({\tavXa-1.55},{\tavYc+2.5})
    -- ({\tavXca+2.7},{\tavYc+1.0}) -- ({\tavXca+2.7},{\tavYc-1.0}) -- cycle;
  \node[tav/card=cVideo] (tavV\tavK) at (\tavXcv,\tavYc) {};
  \node[tav/card=cAudio] (tavA\tavK) at (\tavXca,\tavYc) {};
  \tavHist{\tavXcv}{\tavYc+0.35}{cVideo}{\tavHv}{\tavPv}
  \tavHist{\tavXca}{\tavYc+0.35}{cAudio}{\tavHa}{\tavPa}
  \draw[tav/link, shorten <=0.2mm, shorten >=0.2mm] (tavV\tavK.east) -- (tavA\tavK.west);
  \ifnum\tavSv>0\relax
    \pgfmathsetmacro{\tavXs}{\tavXv+(\tavSv-2)*1.95}
    \pgfmathsetmacro{\tavXe}{\tavXv+(\tavMv-2)*1.95}
    \pgfmathsetmacro{\tavOut}{\tavSv<\tavMv ? 70 : 110}
    \pgfmathsetmacro{\tavIn}{\tavSv<\tavMv ? 110 : 70}
    \draw[tav/uni] (\tavXs,{\tavYt+1.87}) to[out=\tavOut, in=\tavIn] (\tavXe,{\tavYt+1.87});
  \fi
  \ifnum\tavSa>0\relax
    \pgfmathsetmacro{\tavYs}{\tavYc-2.5+(\tavSa-0.5)*0.2}
    \pgfmathsetmacro{\tavYe}{\tavYc-2.5+(\tavMa+\tavMb-1)*0.1}
    \pgfmathsetmacro{\tavOut}{\tavSa<\tavMa ? 28 : -28}
    \pgfmathsetmacro{\tavIn}{\tavSa<\tavMa ? -28 : 28}
    \draw[tav/uni] ({\tavXa+1.47},\tavYs) to[out=\tavOut, in=\tavIn] ({\tavXa+1.47},\tavYe);
  \fi
}

\node[tav/badge] at ({(\tavXcv+\tavXca)/2},35.5) {3};

\node[tav/key, anchor=base] at (\tavXv,37.4) {video tokens};
\node[tav/key, anchor=base] at (\tavXa,37.4) {waveform tokens};
\end{tikzpicture}%
    \caption{\avjepa pretraining.}
    \label{fig:teaser_avjepa_model}
  \end{subfigure}\hfill
  \begin{subfigure}[t]{0.51\linewidth}
    \vspace{0pt}%
    \centering
    \begin{tikzpicture}[x=1mm, y=1mm, line cap=round, line join=round,
  tavw/lbl/.style={font=\scriptsize, text=avInk2, inner sep=0pt},
  tavw/arr/.style={-{Stealth[length=1.15mm,width=1mm]}, line width=0.45pt, color=avInk2},
  tavw/act/.style={-{Stealth[length=1.15mm,width=1mm]}, line width=0.45pt, color=avOrange!85!black},
  tavw/state/.style={draw=cFused!75!black, fill=cFused!22, line width=0.45pt, rounded corners=0.6mm,
    minimum width=4.0mm, minimum height=3.6mm, inner sep=0pt, font=\scriptsize, text=avInk},
  tavw/pred/.style={tavw/state, fill=cFused!9, dash pattern=on 1.1pt off 0.8pt},
  tavw/loop/.style={-{Stealth[length=1.1mm,width=0.95mm]}, line width=0.55pt, color=cFused!70!black,
    dash pattern=on 1.6pt off 1pt},
  tavw/action/.style={draw=avOrange, fill=avOrange!18, line width=0.45pt, rounded corners=0.6mm,
    minimum width=6.0mm, minimum height=3.6mm, inner sep=0pt, font=\scriptsize, text=avInk},
  tavw/wm/.style={draw=avInk2, fill=black!4, line width=0.5pt, rounded corners=0.8mm,
    minimum width=16mm, minimum height=4.4mm, inner sep=0pt, font=\scriptsize\bfseries, text=avInk},
  tavw/out/.style={draw=avAxis, fill=white, line width=0.5pt, rounded corners=0.9mm,
    minimum height=4.0mm, inner sep=0pt, align=center, font=\scriptsize\bfseries, text=avInk},
]
\path (0,2.6) (42.5,38.88);
\def\tavFrame#1#2#3#4#5{%
  \begin{scope}
    \clip[rounded corners=0.45mm] ({#1},{#2}) rectangle ++({#3},{#4});
    \fill[cVideo!9] ({#1},{#2}) rectangle ++({#3},{#4});
    \fill[cVideo!30] ({(#1)+(#5)*(#3)},{(#2)+0.70*(#4)}) circle ({0.14*(#4)});
    \fill[cVideo!50] ({#1},{#2}) -- ++({0.30*(#3)},{0.56*(#4)}) -- ++({0.18*(#3)},{-0.26*(#4)})
      -- ++({0.18*(#3)},{0.30*(#4)}) -- ({(#1)+(#3)},{(#2)+0.12*(#4)}) -- ({(#1)+(#3)},{#2}) -- cycle;
  \end{scope}
  \draw[cVideo!75!black, line width=0.4pt, rounded corners=0.45mm] ({#1},{#2}) rectangle ++({#3},{#4});}
\def\tavWaveSeg#1#2#3#4#5#6#7{%
  \foreach \tavI in {0,...,#5} {
    \pgfmathsetmacro{\tavU}{\tavI/#5}
    \ifcase#6
      \pgfmathsetmacro{\tavEnv}{0.10 + 0.95*(\tavU>0.08)*exp(-(\tavU-0.08)/0.13)
        + 0.70*(\tavU>0.56)*exp(-(\tavU-0.56)/0.11)}
      \pgfmathsetmacro{\tavTex}{0.45 + 0.55*abs(sin(\tavI*67+#7)*cos(\tavI*29+2*#7))}
    \or
      \pgfmathsetmacro{\tavEnv}{0.40 + 0.45*exp(-((mod(\tavU*4+0.85,1)-0.15)/0.20)^2)}
      \pgfmathsetmacro{\tavTex}{0.80 + 0.20*cos(\tavI*120)}
    \or
      \pgfmathsetmacro{\tavEnv}{0.06 + 0.85*exp(-((\tavU-0.17)/0.09)^2)
        + 0.60*exp(-((\tavU-0.46)/0.07)^2) + 0.95*exp(-((\tavU-0.77)/0.10)^2)}
      \pgfmathsetmacro{\tavTex}{0.45 + 0.55*abs(sin(\tavI*67+#7)*cos(\tavI*29+2*#7))}
    \fi
    \pgfmathsetmacro{\tavAmp}{#4*max(0.08, min(1, \tavEnv*\tavTex))}
    \draw[cAudio!85!black, line width=0.42pt] ({#1+\tavU*#3},{#2-\tavAmp}) -- ({#1+\tavU*#3},{#2+\tavAmp});
  }}

\foreach \tavI/\tavS in {3/0.22, 2/0.42, 1/0.62, 0/0.82}
  {\tavFrame{9.5+0.6*\tavI}{3.1+0.5*\tavI}{5.6}{4.0}{\tavS}}
\tavWaveSeg{19.0}{5.85}{4.1}{1.8}{11}{0}{10}
\tavWaveSeg{23.45}{5.85}{4.1}{1.8}{11}{2}{40}
\node[tavw/lbl, anchor=east] at (9.2,5.1) {video};
\node[tavw/lbl, anchor=west] at (28.2,5.85) {audio};
\draw[tavw/arr] (13.2,8.85) -- (13.2,10.25);
\draw[tavw/arr] (23.3,8.0) -- (23.3,10.25);

\node[draw=cFused, fill=cFused!14, line width=0.5pt, rounded corners=1.2mm, minimum width=34mm,
  minimum height=4.0mm, inner sep=0pt, font=\scriptsize]
  at (18,12.35) {\textcolor{cFused!80!black}{\textbf{\avjepa}}\,\textcolor{avInk2}{(frozen)}};
\draw[tavw/arr] (3.85,14.5) -- (3.85,15.9);

\foreach \tavK/\tavX/\tavIn/\tavStyle/\tavA/\tavOut in {
    1/7.0/{s_t}/state/{a_t}/{\hat{s}_{t+1}},
    2/18.0/{}/pred/{a_{t+1}}/{\hat{s}_{t+2}},
    3/29.0/{}/pred/{a_{t+2}}/{\hat{s}_{t+3}}} {
  \node[tavw/\tavStyle] at ({\tavX-3.15},17.8) {$\tavIn$};
  \node[tavw/action] at ({\tavX+2.15},17.8) {$\tavA$};
  \draw[tavw/arr] (\tavX,19.75) -- (\tavX,21.05);
  \draw[tavw/arr] (\tavX,25.8) -- (\tavX,27.1);
  \node[tavw/pred, minimum width=6.4mm] at (\tavX,29.0) {$\tavOut$};
}
\node[draw=avInk2, fill=black!4, line width=0.5pt, rounded corners=1.2mm, minimum width=34mm,
  minimum height=4.5mm, inner sep=0pt, font=\scriptsize]
  at (18,23.4) {\textbf{\textcolor{avInk}{WM}}\,\textcolor{avInk2}{(world model)}};
\begin{scope}[on background layer]
  \draw[tavw/loop] (10.25,29.0) to[out=-40, in=90, looseness=1.0] (14.85,19.65);
  \draw[tavw/loop] (21.25,29.0) to[out=-40, in=90, looseness=1.0] (25.85,19.65);
\end{scope}
\node[tavw/lbl, text=cFused!70!black, anchor=west] at (33.3,29.0) {rollout};

\draw[decorate, decoration={brace, amplitude=0.9mm}, avMuted, line width=0.45pt] (3.3,31.5) -- (32.7,31.5);
\node[tavw/out, minimum width=13.5mm] (tavNav) at (9.75,36.75) {navigation};
\node[tavw/out, minimum width=19.5mm] (tavPhys) at (31.25,36.75) {intuitive physics};
\draw[tavw/arr] (16.6,32.6) -- (12.2,34.6);
\draw[tavw/arr] (19.4,32.6) -- (26.8,34.6);
\end{tikzpicture}%
    \caption{\avwm rollouts.}
    \label{fig:teaser_avjepa_wm}
  \end{subfigure}
  \par\smallskip
  \begin{subfigure}[t]{0.49\linewidth}
    \vspace{0pt}%
    \centering
    \begin{tikzpicture}[x=1mm, y=1mm, line cap=round, line join=round]
\input{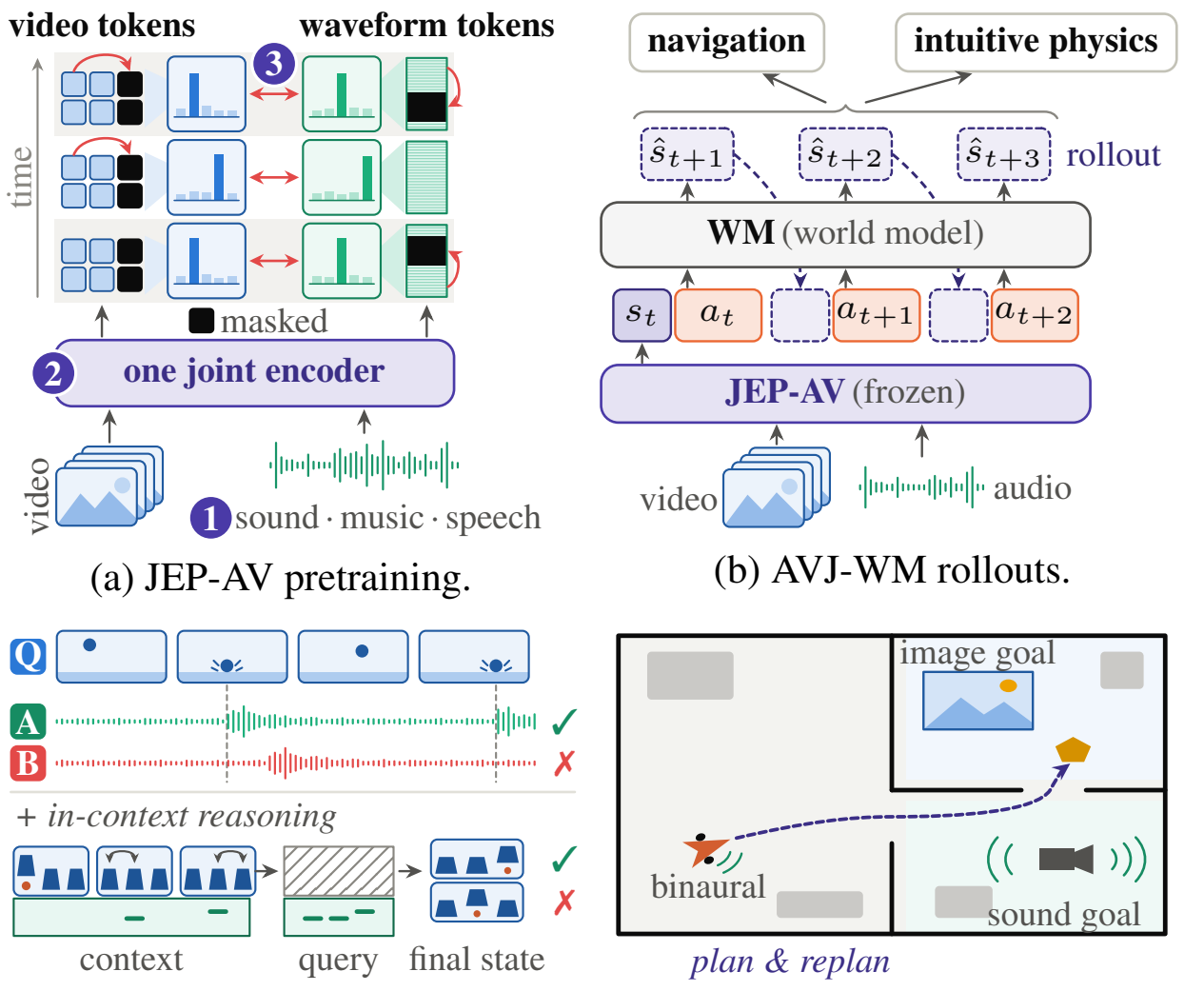}
\def\isxTH{2.6}\def\isxTS{0.48}\def\isxTW{0.55}
\path (0,0) rectangle (40,24);
\node[isx/ab=cVideo, minimum size=2.5mm] at (1.6,21.8) {Q};
\foreach \isxK/\isxBxx/\isxH/\isxC in {0/0.30/0.85/0, 1/0.45/0/1, 2/0.58/0.6/0, 3/0.70/0/1}
  {\isxFrame{3.6+\isxK*8.4}{20.0}{7.6}{3.6}{\isxBxx}{\isxH}{\isxC}}
\def\isxTa{15.42}  %
\def\isxTb{34.12}  %
\draw[isx/guide] (\isxTa,19.8) -- (\isxTa,13.1);
\draw[isx/guide] (\isxTb,19.8) -- (\isxTb,13.1);
\node[isx/ab=cAudio!80!black, minimum size=2.5mm] at (1.6,17.3) {A};
\isxWave{3.6}{17.3}{33.2}{1.5}{cAudio}{92}{\isxTa}{\isxTb}{0}
\node[isx/lbl] at (38.9,17.3) {\isxCorrect};
\node[isx/ab=cLoss, minimum size=2.5mm] at (1.6,14.4) {B};
\isxWave{3.6}{14.4}{33.2}{1.5}{cLoss}{92}{\isxTa}{\isxTb}{8}
\node[isx/lbl] at (38.9,14.4) {\isxWrong};
\draw[avGrid, line width=0.5pt] (0.4,12.4) -- (39.6,12.4);
\node[isx/lbl, anchor=base west, inner sep=0pt, font=\scriptsize\itshape] at (0.6,10.0) {+ in-context reasoning};%
\foreach \isxK/\isxU/\isxS/\isxT in {0/1/0/0,1/0/1/2,2/0/2/3} {\isxCups{0.6+\isxK*5.8}{5.2}{5.2}{3.4}{\isxU}{\isxS}{\isxT}}
\isxTrack{0.6}{17.0}{2.4}
\foreach \isxK/\isxL in {1/2,2/3} {\isxTone{3.2+\isxK*5.8}{2.4}{\isxL}}
\draw[isx/arr] (17.4,6.9) -- (19.0,6.9);
\fill[white] (19.4,5.2) rectangle (27.0,8.6); \fill[isx/hatch] (19.4,5.2) rectangle (27.0,8.6);
\draw[avMuted, line width=0.4pt] (19.4,5.2) rectangle (27.0,8.6);
\isxTrack{19.4}{27.0}{2.4}
\foreach \isxX/\isxL in {21.4/2,23.2/2,25.0/3} {\isxTone{\isxX}{2.4}{\isxL}}
\draw[isx/arr] (27.4,6.9) -- (29.0,6.9);
\isxCups{29.6}{6.4}{6.4}{2.7}{3}{0}{0}
\node[isx/lbl] at (38.9,7.75) {\isxCorrect};
\isxCups{29.6}{3.4}{6.4}{2.7}{2}{0}{0}
\node[isx/lbl] at (38.9,4.75) {\isxWrong};
\node[isx/lbl, anchor=base] at (8.8,0.0) {context};
\node[isx/lbl, anchor=base] at (23.2,0.0) {query};
\node[isx/lbl, anchor=base] at (32.8,0.0) {final state};
\end{tikzpicture}%
    \caption{\intphysx: AV intuitive physics.}
    \label{fig:teaser_intphysx}
  \end{subfigure}\hfill
  \begin{subfigure}[t]{0.49\linewidth}
    \vspace{0pt}%
    \centering
    \begin{tikzpicture}[x=1mm, y=0.74mm, line cap=round, line join=round,  %
  nav/wall/.style={draw=avInk, line width=0.9pt},
  nav/lbl/.style={font=\scriptsize, text=avInk2, inner sep=0.5pt},
  nav/furn/.style={fill=avMuted!45, draw=none, rounded corners=0.3mm},
]
\path (0,0) rectangle (40,33);
\fill[avGrid!40] (1,4) rectangle (39,32);
\fill[cVideo!6] (21,18.5) rectangle (39,32);
\fill[cAudio!7] (21,4) rectangle (39,16.5);
\fill[nav/furn] (3,26) rectangle (9,30.5);
\fill[nav/furn] (12,5.5) rectangle (18,8);
\fill[nav/furn] (34.5,27) rectangle (37.5,30.5);
\fill[nav/furn] (23,5.5) rectangle (27,8.5);
\draw[nav/wall] (1,4) rectangle (39,32);
\draw[nav/wall] (20,4) -- (20,12.5) (20,17.5) -- (20,32);
\draw[nav/wall] (20,17.5) -- (29,17.5) (33.5,17.5) -- (39,17.5);
\begin{scope}[shift={(7,12)}, rotate=20]
  \fill[avOrange!90!black] (1.9,0) -- (-1.3,1.3) -- (-0.5,0) -- (-1.3,-1.3) -- cycle;
  \fill[avInk] (0,1.15) circle (0.35); \fill[avInk] (0,-1.15) circle (0.35);
\end{scope}
\fill[avInk2] (30.3,10.0) rectangle ++(2.4,2.0);
\fill[avInk2] (32.7,10.3) -- (34.0,9.6) -- (34.0,12.4) -- (32.7,11.7) -- cycle;
\foreach \r in {1.6,2.6,3.6} {\draw[cAudio!80!black, line width=0.5pt] ($(34.0,11.0)+(-35:\r)$) arc[start angle=-35, end angle=35, radius=\r];}
\foreach \r in {2.4,3.6} {\draw[cAudio!80!black, line width=0.5pt] ($(30.3,11.0)+(145:\r)$) arc[start angle=145, end angle=215, radius=\r];}
\node[nav/lbl, anchor=north] at (32.0,7.4) {sound goal};
\foreach \r in {1.8,2.6} {\draw[cAudio!80!black, line width=0.45pt] ($(7,12)+(-60:\r)$) arc[start angle=-60, end angle=-10, radius=\r];}
\node[nav/lbl, anchor=north west] at (3.0,10.0) {binaural};
\fill[avYellow!90!black] (32.6,21.2) +(90:1.3) -- +(162:1.3) -- +(234:1.3) -- +(306:1.3) -- +(18:1.3) -- cycle;
\begin{scope}[shift={(22.2,23.2)}]
  \fill[white] (0,0) rectangle (7.6,5.4);
  \fill[cVideo!20] (0,0) rectangle (7.6,5.4);
  \fill[cVideo!55] (0,0) -- (2.6,3.0) -- (4.2,1.4) -- (5.6,2.6) -- (7.6,0.6) -- (7.6,0) -- cycle;
  \fill[avYellow] (5.9,4.1) circle (0.6);
  \draw[cVideo!75!black, line width=0.45pt] (0,0) rectangle (7.6,5.4);
\end{scope}
\node[nav/lbl, anchor=base] at (26.0,29.4) {image goal};
\draw[-{Stealth[length=1.2mm,width=1.1mm]}, cFused!80!black, line width=0.7pt, dash pattern=on 1.4pt off 0.9pt]
  (9.2,13.0) .. controls (14,14.5) and (17,15) .. (20,15) .. controls (25,15) and (30,15.5) .. (31,17.5)
  .. controls (31.6,19.0) and (32.0,19.6) .. (32.4,20.0);
\node[nav/lbl, anchor=north, text=cFused!80!black, font=\scriptsize\itshape] at (13.0,1.0+2.0) {plan \& replan};
\end{tikzpicture}%
    \caption{\avnavbench: AV navigation.}
    \label{fig:teaser_avnav}
  \end{subfigure}
  \vspace{-0.2cm}
  \caption{
    (a)~Video and waveform\,\tavB{1} share one encoder\,\tavB{2}, trained with intra- and cross-modal \protoloss{}\,\tavB{3}.
    (b)~WM over \avjepa features with rollouts for planning.
    (c)~\intphysx: cross-modal understanding of intuitive physics.
    (d)~\avnavbench: planning and navigation to image / sound goals.
   }
  \label{fig:teaser_avjepa}
  \vspace{-0.5cm}
\end{figure}

Human perception is natively multimodal: infants attend to events specified jointly by sight and
sound~\cite{spelke1979bimodal,bahrick2000intersensory}, a seen face can change the speech adults
hear~\cite{mcgurk1976hearing}, and sound tells us where to look or where to go.
Unfolding on a shared time axis, the two streams could supervise one another~\cite{smith2005embodied}.
Exploiting this requires models that learn from both streams jointly and predict how they unfold, and benchmarks in which neither stream alone
suffices.

Self-supervised learning (SSL) exploits this structure only in part.
Image and video SSL~\cite{he2022masked,oquab2023dinov2,tong2022videomae,assran2025vjepa2} ignores the audio track, and
AV SSL~\cite{gong2023contrastive,huang2023mavil,kim2024equiav,araujo2025cavmaesync} typically makes three design choices
that weaken cross-modal learning.
First, separate encoders let the two streams interact only after encoding.
Second, spectrogram front-ends fix the time-frequency resolution of audio in advance~\cite{zeghidour2021leaf}.
Last, clip-level objectives pool each stream into one embedding, never aligning individual audio and video events.
Shared encoders~\cite{owens2018audio,akbari2021vatt,lin2024avsiam,teotia2026mjepa}, raw
waveforms~\cite{owens2018audio,akbari2021vatt,haliassos2023raven} and segment-level
targets~\cite{araujo2025cavmaesync,iashin2024synchformer} have each been explored, but rarely together; existing AV
JEPAs still encode spectrograms and regress clip- or view-level targets~\cite{teotia2026mjepa,robson2026leavjepa}.

We introduce \textbf{\avjepa}, a JEPA~\cite{lecun2022path,assran2023selfsupervised} that addresses all three choices at
once: a single vision transformer encodes video and audio waveforms on a shared time
axis~\cite{su2024roformer,xu2025qwen25omni}, and its targets are dense in time.
A student predicts an EMA teacher's targets with two changes to the standard JEPA recipe: it matches soft assignments to learned prototypes (\protoloss) rather than directly regressing features, and it predicts time-aligned  500\,ms cross-modal chunks rather than pooled clip embeddings, so it associates AV events.

Building on these representations, predicting how a scene unfolds and acting on it require a world model.
Most world models~\cite{bar2025nwm,zhou2024dinowm} are vision-only and miss what sound reveals: occluded events, distance and direction~\cite{chen2020soundspaces}.
We therefore explore world modeling in \avjepa's representation space, asking whether its frozen AV features can serve as the state of a world model that plans. We train \textbf{\avwm}, a causal world model with modules designed for these features, that predicts future AV chunks and plans through lightweight action adapters; we compare it with diffusion world models trained on the same frozen encoder.

Intuitive-physics benchmarks are video-only~\cite{riochet2022intphys,bordes2025intphys2}, and AV benchmarks
rely on generation~\cite{li2026flatsounds} or question answering~\cite{yu2022pacs}, leaving cross-modal physical
understanding untested.
We introduce \textbf{\intphysx}, in which no item is solvable from a single modality; a model matches a query in one modality to the option from the same event in the other, testing
the information the two streams share; its in-context part extends this to associating sounds with actions in context and reasoning across modalities.

Likewise, navigation is the standard test of whether a video world model adapts to actions and plans~\cite{bar2025nwm, murlabadia2026vjepa21}, yet these benchmarks are silent and current AV navigation benchmarks~\cite{chen2020soundspaces,chen2022soundspaces2} target policies, not world models.
We therefore introduce \textbf{AVNav-Bench}, a simulated navigation benchmark with image or binaural sound goals, in which the world model itself chooses the actions.

\avjepa sets the SoTA among frozen self-supervised encoders on audio and AV classification and segmentation and,
pretrained on video only, outperforms matched \vjepatwo and \vjepatwoone models on most video benchmarks.
On \avnavbench, \avjepa's frozen features serve as a world-model state: the same diffusion world model plans on them
within 3 points of V-JEPA~2.1's.
\avwm's offline lead over diffusion world models grows with goal distance, from close at 2\,s to 13 points on goals over 9\,s ahead,
and it improves almost $4\times$ on sound goals.
On \intphysx, \avjepa leads all representation models, on par with Gemini-4 Argon, yet all models remain far below humans: the best model in each scenario trails them by over 20 points on average.

\noindent In summary, our contributions (\cref{fig:teaser_avjepa}) are fourfold:
\begin{itemize}[leftmargin=1.2em,itemsep=1pt,topsep=2pt]
    \item \textbf{\avjepa:} a single encoder for audio and video, trained with intra- and cross-modal \emph{\protoloss}, setting the SoTA among frozen self-supervised encoders on audio and AV classification and segmentation (\cref{sec: avjepa,sec:av_main}).
	\item \textbf{\avwm:} a focused exploration of WM over frozen \avjepa features, predicting future AV chunks and
	planning via action-conditioned rollouts, compared with diffusion WMs on the same features (\cref{sec:avwm,sec:avnavresults}).  %
	\item \textbf{\intphysx:} to our knowledge, the first intuitive-physics benchmark that requires cross-modal
	inference by design, including in-context scenarios
	(\cref{sec:bench_intphysx,sec:intphysx}).  %
	\item \textbf{AVNav-Bench:} a new AV navigation benchmark where the world model itself plans to image or sound goals, offline and in closed loop~(\cref{sec:avwmbench,sec:avwmplanningsec}).
\end{itemize}

\vspace{-0.2cm}
\section{Related Work}
\label{sec:related}
\vspace{-0.2cm}

\leanparagraph{Audio-Visual Self-Supervised Learning.}
AV SSL learns audio and visual representations from unlabeled video, leveraging the correspondence between sight and sound.
Early methods exploited pretext tasks, such as correspondence~\cite{arandjelovic2017look,arandjelovic2018objects}, 
cross-modal distillation~\cite{aytar2016soundnet}, sound separation~\cite{zhao2018soundpixels} and synchronization~\cite{owens2018audio,korbar2018cooperative,chung2016outoftime}.
Later methods scaled to larger data with contrastive~\cite{morgado2021avid,alayrac2020mmv,kim2024equiav},
clustering~\cite{alwassel2020xdc,asano2020selavi} or masked modeling~\cite{gong2023contrastive,huang2023mavil,georgescu2023avmae,sarkar2022xkd} objectives, or 
leveraging language priors~\cite{girdhar2023imagebind, vyas2025pushing}.
These methods often align separate audio and visual encoders at clip level, capturing \emph{what} happens across modalities but not \emph{when}.
Finer alignment comes from contrasting short segments~\cite{araujo2025cavmaesync,iashin2024synchformer},
and shared encoders~\cite{akbari2021vatt,lin2024avsiam,nagrani2021mbt,jaegle2021perceiver}, but the two are rarely combined.
In contrast, \avjepa jointly processes video and audio with a single encoder, temporally dense targets and a discrete target objective.

\leanparagraph{Joint-embedding predictive architectures.}
JEPAs~\cite{lecun2022path} predict the latent representation of masked content from visible context, typically by regressing EMA-teacher features, 
for images~\cite{assran2023selfsupervised}, video~\cite{bardes2024revisiting,assran2025vjepa2,murlabadia2026vjepa21} and audio~\cite{fei2024ajepa}.
Other JEPAs replace regression with contrastive~\cite{chen2026vl}, classification~\cite{quelennec2025matpac} or discrete target prediction~\cite{baek2025discretejepa,ioannides2026sjepa}.
Closest to ours, \mjepa~\cite{teotia2026mjepa} and LeAVJEPA~\cite{robson2026leavjepa} encode video and spectrograms with a single encoder and regress clip- or view-level cross-modal targets.
\avjepa instead encodes raw waveforms, places audio and video tokens on a shared time axis, and applies dense cross-modal supervision per 500\,ms chunk with \protoloss, predicting the teacher's prototype assignments rather than regressing its features.

\leanparagraph{Intuitive physics and violation of expectation.}
Video benchmarks of intuitive physics~\cite{riochet2022intphys,bordes2025intphys2,jassim2024grasp,weihs2022inflevel}
adopt the violation-of-expectation (VoE) paradigm from infant research~\cite{baillargeon1985object,spelke1992origins},
measuring plausibility by a model's surprise.
Video JEPAs pass such tests, whereas multimodal LLMs often stay near chance~\cite{garrido2025intuitive}.
Sound also reveals an object's shape and material~\cite{zhang2017shape,gao2023objectfolderbench,clarke2023realimpact}, 
but audio-visual benchmarks rely on question answering~\cite{yu2022pacs,patraucean2023perception,gong2024avodyssey}
or check whether generated sound obeys physics~\cite{xie2025phyavbench,li2026flatsounds}.
\intphysx is, to our knowledge, the first intuitive-physics benchmark across sight and sound in which only combining both streams identifies the physically consistent pairing, 
and it extends the test to cross-modal in-context reasoning.%

\leanparagraph{World models for navigation.}
World models support planning by predicting future scene representations under candidate actions~\cite{ha2018worldmodels,lecun2022path}.
NWM~\cite{bar2025nwm} plans over image latents, whereas DINO-WM~\cite{zhou2024dinowm}, DINO-world~\cite{baldassarre2025dinoworld} and \vjepatwo~\cite{assran2025vjepa2} plan in the feature space of a frozen encoder, \vjepatwo after post-training an action-conditioned predictor on under 62 hours of robot video.
Navigation policies~\cite{shah2021recon,shah2023gnm,sridhar2024nomad} instead map observations and goals directly to actions.
AV agents learn to find sounding objects in simulation~\cite{chen2020soundspaces,chen2022soundspaces2,shi2025enmus}, and AV-CDiT~\cite{wang2025avwm} predicts video and binaural audio to aid such an agent.
\avwm enables AV frozen-feature planning. We pretrain it without actions on video with sound, add small action adapters, and plan to image or sound goals on \avnavbench without a policy.

\vspace{-0.1cm}
\section{\avjepa}  %
\vspace{-0.1cm}
\label{sec:method}
\label{sec: avjepa}

\begin{figure}[t]
  \centering
  \input{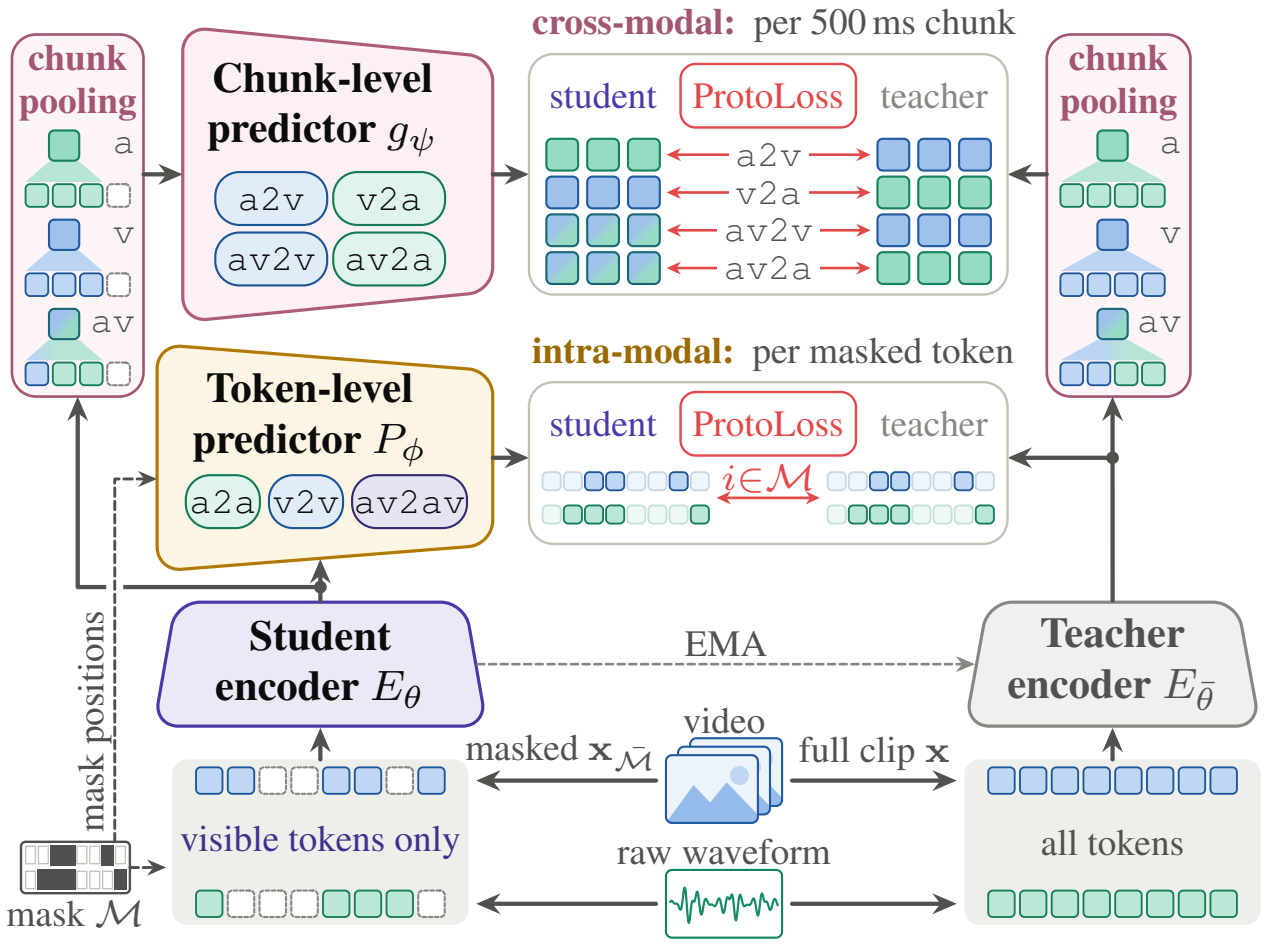}%
  \vspace{-0.2cm}
  \caption{
	  Overview of \avjepa.
	  The student predicts EMA teacher targets from masked inputs: per token within modalities and per 500\,ms chunk across modalities, both with \protoloss.
  }
  \label{fig:main}\label{fig:main_v2_arch}
  \vspace{-0.3cm}
\end{figure}

\avjepa (\cref{fig:main_v2_arch}) follows the JEPA framework~\cite{lecun2022path}: a student $\Enc$ encodes masked video, audio or both on a shared time axis (\cref{sec:method_encoder}), and predictors $\Pred$ and $\ChunkPredSimple$ infer the representations an EMA teacher $\EncT$ computes from the full clip.
Targets are masked tokens within each modality and time-aligned 500\,ms chunks across modalities (\cref{sec:prediction}).
Instead of regressing the teacher's features, the student matches its assignments to learned prototypes (\cref{sec:method_protoloss}).%

\subsection{Audio-Visual Encoding on a Shared Time Axis}  %
\label{sec:method_encoder}

\avjepa encodes audio and video using a single ViT~\cite{dosovitskiy2021vit} rather than separate encoders per
modality, so the modalities interact in every layer on a shared time axis.

\leanparagraph{Tokenization.}
Video is split into tubelets~\cite{tong2022videomae,bardes2024revisiting}, one temporal slice of spatio-temporal tokens per 500\,ms.
Audio is tokenized from waveform by a strided convolutional module~\cite{baevski2020wav2vec,poli2025spidr}
emitting one token per 20\,ms.
Learned additive embeddings indicate each token's modality.

\leanparagraph{Shared time axis.}
Multimodal rotary position embeddings (M-RoPE)~\cite{su2024roformer,xu2025qwen25omni} place all tokens on one time
axis in 20\,ms steps: audio token $j$ sits at $(j,0,0)$ and video token $(h,w)$ of slice $c$ at $(25c,h,w)$.
Co-occurring sounds and frames thus get nearby time coordinates, which rotary attention relates directly through
relative positions, while video tokens keep their spatial layout.

\leanparagraph{Masking.}
Each modality is masked independently, according to its structure: video with spatial blocks spanning all temporal slices, audio with contiguous 200\,ms spans.
Content hidden in one modality may remain visible in the other, so cross-modal prediction can then recover masked content.

\subsection{Prediction Within and Across Modalities}  %
\label{sec:prediction}

For input modes $m\in\{a,v,av\}$ (audio, video or both), prediction term $\Loss_{m_1\to m_2}$ maps the student's encoding of the masked input in mode $m_1$
to the teacher's encoding of the full input in mode $m_2$.
The key choice is target granularity: we predict at the finest level where source and target correspond, \ie
tokens within a mode and chunks across modes.%

\leanparagraph{Intra-modal token prediction.}
Within a mode ($m_1{=}m_2$), student and teacher tokens have a one-to-one correspondence. Following~\cite{assran2023selfsupervised,bardes2024revisiting}, a transformer predictor receives the student's tokens with learnable mask tokens and predicts the teacher's representation at every masked position.
The $\Loss_{\mto{a}{a}}$ and $\Loss_{\mto{v}{v}}$ terms learn the structure of each modality on its own, so the encoder can operate effectively when only one is available, as for sounds whose source is off-screen.
In $\Loss_{\mto{av}{av}}$, the student uses visible context across both modalities to predict the masked tokens.

\leanparagraph{Dense cross-modal chunk prediction.}
Across modalities, tokens lack a one-to-one correspondence: each video token spans 25 audio tokens in time, and a
sound's source may cover several video patches.
Clip-level targets, as in \mjepa, avoid this imbalance by pooling each modality over the whole clip into a single vector,
discarding when events happen.
To preserve temporal alignment, we pool tokens (mean or attention pooling) within each 500\,ms chunk $c$, \ie one temporal slice of video tokens and the co-occurring audio tokens,  the smallest window aligning audio and video.
This yields one target of mode $m_2$ per chunk on the shared time axis, which an MLP $\ChunkPredSimple$ predicts from the pooled visible student tokens of source mode $m_1$ in the same chunk.
Four cross-modal terms use these targets: $\Loss_{\mto{a}{v}}$ and $\Loss_{\mto{v}{a}}$ predict each modality from the other, while
$\Loss_{\mto{av}{a}}$ and $\Loss_{\mto{av}{v}}$ make the joint representation retain both modalities.

\subsection{Prototype Prediction Objective}  %
\label{sec:method_protoloss}
\begingroup
\setlength{\abovedisplayskip}{3pt}\setlength{\belowdisplayskip}{3pt}%
\setlength{\abovedisplayshortskip}{1pt}\setlength{\belowdisplayshortskip}{3pt}%

Every prediction term compares student predictions with teacher targets, per masked token within a modality or per
chunk across modalities.
Rather than directly regressing teacher features~\cite{assran2025vjepa2,teotia2026mjepa}, we propose \protoloss: %
a discrete prototype prediction objective that expresses both predictions and targets as soft assignments to $K$ learned prototypes, so the student predicts a distribution over prototypes rather than regressing an embedding.
Discrete targets encourage the student to capture semantically meaningful structure shared within and across
modalities, while staying invariant to fine-grained variations that need not be predicted; with chunk-level targets, feature
regression degrades representations whereas \protoloss trains stably (\cref{tab:ablations_single_col_v1}).

Formally, let $\Codebook \in \mathbb{R}^{K \times D}$ be a codebook of $K$ unit-norm learned prototypes: one per modality for the intra-modal terms, and one shared by all cross-modal terms.
For item $i$ (masked token or chunk) with normalized student prediction $\zs{i}$ and teacher target $\zt{i}$, the codebook assignments are
\begin{equation}
  \qs{i} = \sigma(\operatorname{BN}(\Codebook \zs{i})/\tau_s),\;\;
  \qt{i} = \sigma(\operatorname{BN}(\Codebook\operatorname{sg}(\zt{i}))/\tau_t),
  \label{eq:assign}
\end{equation}
where $\sigma$ is the softmax, $\operatorname{sg}$ the stop-gradient, $\operatorname{BN}$ a batch normalization, and
$\tau_t<\tau_s$ makes the teacher assignment sharper.
$\operatorname{BN}$ standardizes each prototype's logit over the batch, so no prototype is systematically favored.
The student matches the teacher by minimizing the Kullback--Leibler (KL) divergence between their assignments,
averaged over the set $\Items$ of items in the term:%
\begin{equation}
  \mathcal{L}_{\mathrm{KL}} = \frac{1}{|\Items|}\sum_{i\in \Items}\mathrm{KL}\big(\operatorname{sg}(\qt{i})\,\|\, \qs{i} \big).
  \label{eq:kl}
\end{equation}

Matching alone is trivially satisfied by assigning all items to one prototype, so two entropy terms on the teacher assignments shape the prototype distribution.
The \emph{conditional entropy} $\mathcal{L}_{\mathrm{H}}=\frac{1}{|\Items|}\sum_{i\in\Items}H(\qt{i})$ makes each teacher
assignment peaked, and \emph{mean-entropy maximization} (MEMax~\cite{assran2022msn}),
$\mathcal{L}_{\mathrm{ME}} = -H(\qbar)$ with $\qbar = \frac{1}{|\Items|}\sum_{i\in \Items} \qt{i}$, encourages uniform
prototype usage; together they maximize the mutual information between items and prototypes~\cite{hu2017imsat}.
Similar prototype objectives have been proposed for SSL~\cite{darcet2025capi,poli2025spidr}, usually balanced by
Sinkhorn--Knopp or online clustering, which \protoloss replaces with entropy terms.
JEM~\cite{seitzer2026scalable} and GSLM2~\cite{higuchi2026gslm} concurrently propose the same objective for images and speech.
Each term minimizes
\begin{equation}
  \mathcal{L}_{\mathrm{proto}} = \mathcal{L}_{\mathrm{KL}} + \lambda_{\mathrm{H}}\,\mathcal{L}_{\mathrm{H}}
    + \lambda_{\mathrm{ME}}\,\mathcal{L}_{\mathrm{ME}} \, .
  \label{eq:protoloss}
\end{equation}
The training loss sums $\mathcal{L}_{\mathrm{proto}}$ over all seven (3 intra-modal + 4 cross-modal) prediction terms with equal weights.
\endgroup


\vspace{-0.1cm}
\section{Audio-Visual World Modeling}  %
\label{sec:avwm}
\begin{figure}[t]
	\centering
	\begin{tikzpicture}[x=1mm, y=1mm, line cap=round, line join=round,
  avwm/tok/.style={draw=avInk2, line width=0.45pt, rounded corners=1.1pt,
    minimum width=9mm, minimum height=4mm, inner sep=0pt, font=\footnotesize, text=avInk},
  avwm/mod/.style={draw=avInk2, line width=0.5pt, rounded corners=1.8pt, fill=black!4,
    minimum width=56mm, minimum height=4.8mm, inner sep=1pt, align=center,
    font=\footnotesize, text=avInk},
  avwm/arr/.style={-{Stealth[length=1.3mm,width=1mm]}, line width=0.45pt, draw=avInk2},
  avwm/note/.style={font=\scriptsize, text=avInk2, align=center, inner sep=0pt},
  avwm/title/.style={font=\footnotesize\bfseries, text=avInk, anchor=west, inner sep=0pt},
  avwm/frame/.pic={
    \fill[pic actions] (0,0) rectangle (6,6);
    \draw[avBlue!32, line width=0.25pt] (0,0) grid[step=1.5] (6,6);
    \draw[avBlue!55!black, line width=0.4pt] (0,0) rectangle (6,6);
  },
  avwm/snow/.pic={
    \foreach \aa in {0,60,120}{
      \draw[avMuted, line width=0.45pt] (\aa:1.15) -- (\aa+180:1.15);
    }
  },
  avwm/flame/.pic={
    \fill[avOrange] (0,-1.05) .. controls (0.85,-0.25) and (0.45,0.45) .. (0,1.15)
      .. controls (-0.45,0.45) and (-0.85,-0.25) .. (0,-1.05);
    \fill[avYellow] (0,-0.95) .. controls (0.4,-0.45) and (0.2,0.1) .. (0,0.55)
      .. controls (-0.2,0.1) and (-0.4,-0.45) .. (0,-0.95);
  },
  avwm/wave/.pic={
    \fill[white] (0,0) rectangle (6,2.8);
    \draw[cAudio!75!black, line width=0.4pt]
      plot[domain=0:1, samples=50]
      ({0.3+5.4*\x}, {1.4 + 0.85*sin(360*7*\x)*exp(-((\x-0.45)/0.35)^2)});
    \draw[cAudio!55!black, line width=0.4pt] (0,0) rectangle (6,2.8);
  },
]
\tikzset{avwm/roll/.style={-{Stealth[length=1.2mm,width=0.95mm]}, dashed, line width=0.5pt,
  draw=cFused!70!black}}
\foreach \xa/\ya/\xb/\yb in {10.9/46.4/19.1/26.2, 25.9/46.4/35.0/25.3, 35.0/47.3/44.1/26.2}{
  \draw[avwm/roll] (\xa,\ya) to[out=-20, in=150, looseness=1.1] (\xb,\yb);
}
\node[avwm/mod, fill=black!6, draw=avMuted, minimum width=57mm, minimum height=4.4mm] (enc) at (29.0,18.0) {\textbf{\avjepa Encoder}};
\node[avwm/mod, minimum width=57mm, minimum height=4.4mm] (cmp) at (29.0,30.4) {\textbf{AV-Combiner}};
\node[avwm/mod, minimum width=57mm, minimum height=4.4mm] (wm) at (29.0,42.2) {\textbf{Causal World Model}};
\foreach \c/\i/\j/\s/\n in {7.5/1/2/12/1, 22.5/2/3/22/2, 47.5/T/T{\!+\!}1/46/3}{
  \pic[fill=avBlue!15] at (\c-3.6,7.8) {avwm/frame};
  \pic[fill=avBlue!7] at (\c-2.8,7.0) {avwm/frame};
  \pic at (\c-3,3.2) {avwm/wave};
  \node[avwm/note, anchor=north] at (\c,2.7) {$\i$};
  \draw[avwm/arr] (\c,13.8) -- (\c,15.8);
  \node[avwm/tok, minimum width=6.2mm, fill=cAudio!20] (ea\n) at (\c-3.4,24.2) {\scriptsize $\AudioTokenT{\i}$};
  \node[avwm/tok, minimum width=6.2mm, fill=avBlue!16] (ev\n) at (\c+3.4,24.2) {\scriptsize $\VideoTokenT{\i}$};
  \node[avwm/tok, minimum width=7.6mm, minimum height=3.4mm, fill=cFused!\s] (s\n) at (\c,36.3) {$\ChunkT{\i}$};
  \node[avwm/tok, minimum width=6.2mm, fill=cAudio!20] (pa\n) at (\c-3.4,48.4) {\scriptsize $\PredAudioTokenT{\j}$};
  \node[avwm/tok, minimum width=6.2mm, fill=avBlue!16] (pv\n) at (\c+3.4,48.4) {\scriptsize $\PredVideoTokenT{\j}$};
  \draw[avwm/arr] (\c,20.2) -- (\c,22.2);
  \draw[avwm/arr] (\c,26.2) -- (\c,28.2);
  \draw[avwm/arr] (\c,32.6) -- (s\n.south);
  \draw[avwm/arr] (s\n.north) -- (\c,40.0);
  \draw[avwm/arr] (\c,44.4) -- (\c,46.4);
}
\foreach \yy in {24.2,36.3,48.4} \node[font=\scriptsize, text=avInk2] at (35.0,\yy) {$\cdots$};
\node[inner sep=0pt] at (55.2,18.0) {\twemoji[height=2.6mm]{snowflake}};
\node[inner sep=0pt] at (55.2,30.4) {\twemoji[height=2.6mm]{fire}};
\node[inner sep=0pt] at (55.2,42.2) {\twemoji[height=2.6mm]{fire}};

\draw[avMuted, line width=0.6pt, dash pattern=on 1.05746mm off 1.05746mm] (57.315,44.215) -- (61.2686,50.9314)
  arc[start angle=135, end angle=90, radius=1.6] -- (80.8,51.4) arc[start angle=90, end angle=0, radius=1.6]
  -- (82.4,15.2) arc[start angle=0, end angle=-90, radius=1.6] -- (62.4,13.6)
  arc[start angle=270, end angle=225, radius=1.6] -- (57.315,40.185);
\begin{scope}[shift={(60.5,0)}]
\draw[avAxis, fill=black!3, line width=0.4pt, rounded corners=1.4pt] (4.0,20.4) rectangle (19.0,35.2);
\node[font=\tiny, text=avInk2, inner sep=0pt] at (2.15,27.8) {$\times L$};
\node[draw=avMuted, fill=black!6, rounded corners=1.4pt, line width=0.5pt, minimum width=13.4mm,
  minimum height=5.4mm, inner sep=0pt, font=\scriptsize, text=black, align=center] (zblk) at (11.5,24.1)
  {Transformer\\[-1pt]Block};
\node[draw=avOrange, fill=avOrange!14, dashed, rounded corners=1.4pt, line width=0.6pt, minimum width=9.6mm,
  minimum height=5.4mm, inner sep=0pt, font=\scriptsize, text=black, align=center] (zada) at (11.5,31.3)
  {\textbf{Action}\\[-1pt]\textbf{Adapter}};
\node[inner sep=0pt] at (5.2,31.3) {\twemoji[height=2.2mm]{fire}};
\node[draw=cAudio!75!black, fill=cAudio!40, rounded corners=1.4pt, line width=0.5pt, minimum width=8mm,
  minimum height=5.4mm, inner sep=0pt, font=\scriptsize, text=black, align=center] (zha) at (6.9,40.9)
  {Audio\\[-1pt]Head};
\node[draw=avBlue!70!black, fill=avBlue!32, rounded corners=1.4pt, line width=0.5pt, minimum width=8mm,
  minimum height=5.4mm, inner sep=0pt, font=\scriptsize, text=black, align=center] (zhv) at (16.1,40.9)
  {Video\\[-1pt]Head};
\node[avwm/tok, minimum width=7.6mm, minimum height=3.4mm, fill=cFused!30] (zs) at (11.5,16.6) {$\ChunkT{t}$};
\node[avwm/tok, minimum width=5.0mm, minimum height=3.4mm, fill=avOrange!18, draw=avOrange, dashed] (za) at (18.9,16.6) {$\ActionT{t}$};
\node[avwm/tok, minimum width=6.2mm, fill=cAudio!20] at (6.9,48.4) {\scriptsize $\PredAudioTokenT{t+1}$};
\node[avwm/tok, minimum width=6.2mm, fill=avBlue!16] at (16.1,48.4) {\scriptsize $\PredVideoTokenT{t+1}$};
\draw[avwm/arr] (zs.north) -- (zblk.south);
\draw[avwm/arr] (zblk.north) -- (zada.south);
\draw[avInk2, line width=0.45pt] (zada.north) -- (11.5,36.4);
\draw[avwm/arr] (11.5,36.4) -| (zha.south);
\draw[avwm/arr] (11.5,36.4) -| (zhv.south);
\fill[avInk2] (11.5,36.4) circle (0.3);
\draw[avwm/arr] (zha.north) -- (6.9,46.4);
\draw[avwm/arr] (zhv.north) -- (16.1,46.4);
\draw[avwm/arr, draw=avOrange, dashed] (20.4,18.3) -- (20.4,31.3) -- (zada.east);

\draw[avAxis, line width=0.4pt, rounded corners=1.2pt] (0.3,12.05) rectangle (18.6,1.2);
\node[inner sep=0pt] at (2.0,10.4) {\twemoji[height=2.6mm]{snowflake}};
\node[font=\scriptsize, text=avInk2, anchor=west, inner sep=0pt] at (3.8,10.4) {frozen};
\node[inner sep=0pt] at (2.0,6.65) {\twemoji[height=2.6mm]{fire}};
\node[font=\scriptsize, text=avInk2, anchor=west, inner sep=0pt] at (3.8,6.65) {trainable};
\draw[avOrange, fill=avOrange!14, dashed, line width=0.5pt, rounded corners=0.6pt] (0.9,2.0) rectangle (3.1,3.8);
\node[font=\scriptsize, text=avInk2, anchor=west, inner sep=0pt] at (3.8,2.9) {planning only};
\end{scope}
\end{tikzpicture}%
	\vspace{-0.2cm}
	\caption{
		\textbf{Audio-visual world model.}
		The AV-Combiner converts encoded audio $\AudioTokenT{t}$ and video $\VideoTokenT{t}$ tokens (\avjepa outputs) into a
		chunk state $\ChunkT{t}$.
		A causal world model predicts the next chunk's tokens $\PredAudioTokenT{t+1}, \PredVideoTokenT{t+1}$, fed back for
		autoregressive rollouts.
		For planning, an adapter injects the action $\ActionT{t}$ after each transformer block.
	}
	\label{fig:avworldmodel}
    \vspace{-0.8cm}
\end{figure}
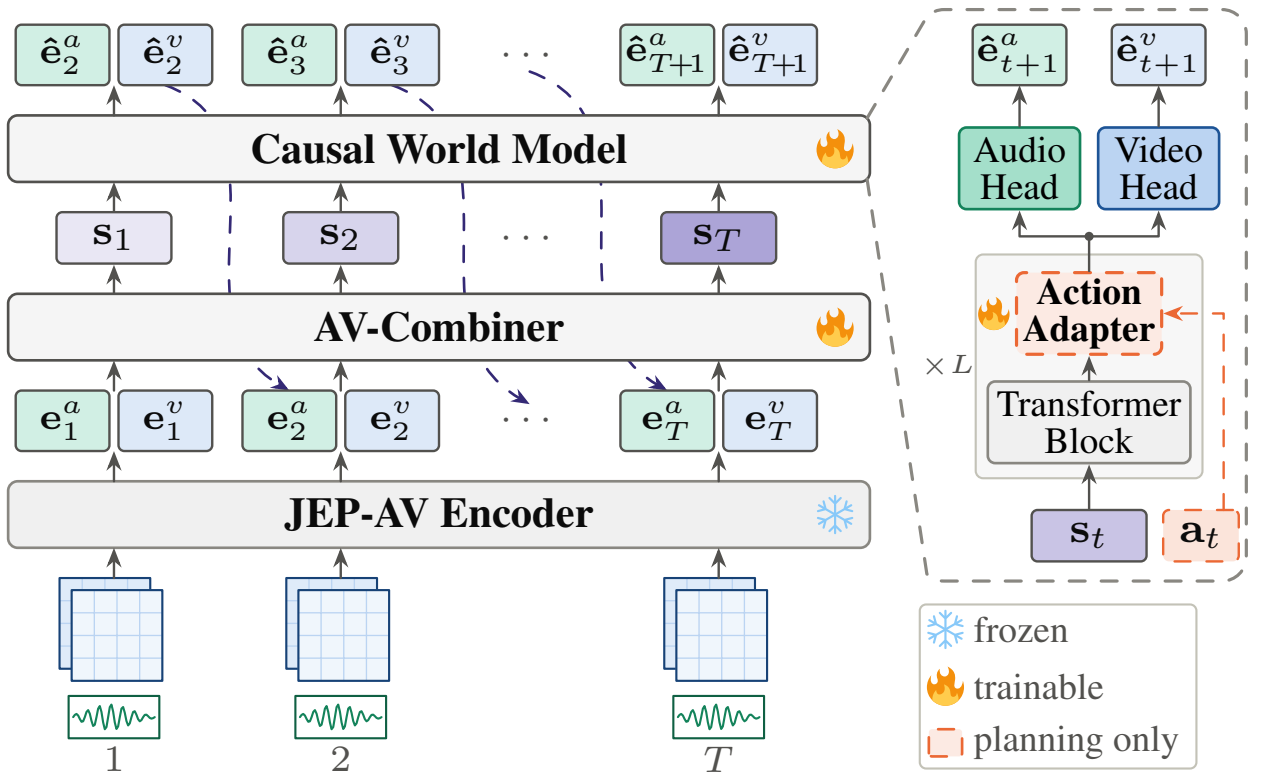
We ask whether \avjepa's frozen chunk features can serve as the state of a world model for planning, and design
\avwm (\cref{fig:avworldmodel}) accordingly.
Four properties constrain the design: \mgt{(i)} the features are high-dimensional (221 tokens per 0.5\,s
chunk) and span two streams, either of which may be absent; \mgt{(ii)} goals and events may depend on both streams, as when a
sounding object is out of view; \mgt{(iii)} planning evaluates many candidate trajectories over several seconds, which
requires efficient prediction steps and limited error accumulation; and \mgt{(iv)} action-labeled AV data
are scarce, whereas action-free AV video is abundant.

\leanparagraph{Windowed chunk encoding.}
Chunks of 0.5\,s, on which \avjepa aligns both streams (\cref{sec:method}), serve as the WM's steps.
\avjepa encodes 4\,s clips bidirectionally, but the WM needs chunk-level causality, so
whole-clip features cannot be reused, while encoding each chunk in isolation loses context.
As a trade-off, the frozen encoder processes each chunk together with the preceding 1.5\,s of video and 0.5\,s of
audio, which worked well in preliminary experiments. %
A learned AV-Combiner~\cite{piergiovanni2024mirasol} then compresses each chunk's video ($\VideoToken$) and audio
($\AudioToken$) tokens into a 64-token chunk state $\ChunkT{t}$, keeping the predictor's sequences short \mgt{(i--iii)}.

\leanparagraph{Causal next-chunk prediction.}
A causal transformer, bidirectional within a chunk and causal across chunks, predicts the next chunk's AV tokens from the history of chunk states:
$\left( \PredVideoTokenT{2:T+1} \, , \, \PredAudioTokenT{2:T+1} \right) = \text{\avwm} \left( \ChunkT{1:T} \right) $.
Since \avjepa's features are already abstract, \avwm regresses them directly rather than pixels: the combiner and
predictor are jointly trained with teacher forcing to minimize
{\setlength{\abovedisplayskip}{2pt}\setlength{\belowdisplayskip}{2pt}%
\begin{equation}
  \mathcal{L}_{\mathrm{pred}} = \sum_{t=1}^{T} \Big[\ell\big(\PredVideoTokenT{t+1}, \VideoTokenT{t+1}\big) + \ell\big(\PredAudioTokenT{t+1}, \AudioTokenT{t+1}\big)\Big],
  \label{eq:wm_pred}
\end{equation}}%
where $\ell \left( \utarget, \upred \right) =\frac{1}{2} \big( \|\upred - \utarget \|_1 + 1 - \cos \left( \upred, \utarget \right) \big)$ combines an $\ell_1$ and a cosine distance between layer-normalized tokens.
At test time, the model feeds its own predictions back as inputs to roll out several steps; each step takes one forward
pass and reuses the computation of earlier steps \mgt{(iii)}.
Regression thus trades the ability to sample several plausible futures for one-pass prediction steps.%

\leanparagraph{Auxiliary objectives.}
Six auxiliary terms target \mgt{(i)}--\mgt{(iii)}.
To limit error accumulation \mgt{(iii)}, $\Loss_{\mathrm{roll}}$ applies $\ell$ to the model's own predictions fed back for
1--8\,s~\cite{assran2025vjepa2}. As in multi-token prediction~\cite{gloeckle2024multitoken,teoh2026mtp}, horizon heads used only in training also
predict, from each chunk, the model's own predictions 1--8\,s ahead in a single step ($\Loss_{\mathrm{hor}}$), which
could make distant predictions cheaper \mgt{(iii)}.
To keep the compressed chunk states informative \mgt{(i)}, variance and decorrelation terms ($\Loss_{\mathrm{var}}$,
$\Loss_{\mathrm{cov}}$)~\cite{bardes2022vicreg} keep every dimension active and non-redundant across the batch, and a
reconstruction term ($\Loss_{\mathrm{rec}}$) rebuilds the encoder tokens from a pooled projection of them.
A consistency term ($\Loss_{\mathrm{cons}}$) pulls video-only and audio-only states towards audio-visual ones, so the
model can infer a missing stream \mgt{(i, ii)}.
The total loss is $\Loss_{\mathrm{WM}} = \Loss_{\mathrm{pred}} + \sum_{k}\lambda_k\Loss_k$ over these six terms
(\suppref{sec:suppl_avwm_arch}).%

\leanparagraph{Action adapters.}
Following DINO-world~\cite{baldassarre2025dinoworld}, we pretrain \avwm without actions on video with sound and then
adapt it to actions \mgt{(iv)}: a residual adapter after each causal block (\cref{fig:avworldmodel}, right) maps the hidden
state and the action $\ActionT{t}$ through a small MLP with a zero-initialized output, so training starts from the
action-free model.
The adapters and predictor are fine-tuned on action-labeled clips, with the encoder and the AV-Combiner frozen
(\suppref{sec:suppl_avwm_arch}).

\leanparagraph{Planning.}
We roll out candidate action sequences and score each by $\ell$ between its final prediction and the goal; for
sound goals, the model encodes and predicts each ear separately and compares both with the goal \mgt{(ii)}.
Offline, we rank a fixed pool, as in NWM~\cite{bar2025nwm}; in closed loop, we replan via model
predictive control~\cite{zhou2024dinowm} (\suppref{sec:suppl_avnav_planner}).%

\leanparagraph{Relation to diffusion world models.}
NWM~\cite{bar2025nwm} plans with a conditional diffusion transformer (CDiT) over image latents, V-JEPA~2.1 trains it on
frozen encoder features~\cite{murlabadia2026vjepa21}, and AV-CDiT~\cite{wang2025avwm} extends it to binaural audio with
a three-stage recipe.
Diffusion can sample distinct futures, but each imagined step takes several denoising passes (8 on JEPA features), at
odds with \mgt{(iii)}, and is challenging in high-dimensional feature spaces~\cite{murlabadia2026vjepa21} \mgt{(i)}.
We therefore use these models as baselines: we reimplement AV-CDiT and train CDiT and AV-CDiT variants on frozen
\avjepa features (\suppref{sec:suppl_avnav_baselines}); \cref{sec:avnavresults} compares them with \avwm. %

\providecommand{\vjepatworep}{\mbox{V-JEPA}~2\textsuperscript{rep.}\xspace}
\providecommand{\vjepatwoonerep}{\mbox{V-JEPA}~2.1\textsuperscript{rep.}\xspace}
\providecommand{\tabdashrule}[1]{\noalign{\vskip 1.5pt}\multispan{#1}{\color{black!40}\xleaders\hbox to 3.2pt{\hss\vrule
  height 0.25pt depth 0.25pt width 1.8pt\hss}\hskip 0pt plus 1fill\relax}\cr\noalign{\vskip 1.5pt}}
\ifdefined\ttbA\else\newsavebox{\ttbA}\newsavebox{\ttbB}\newsavebox{\ttbC}\fi
\providecommand{\ttsize}[2]{\multirow{-#1}{*}{\rotatebox[origin=c]{90}{\scriptsize\textit{#2}}}}
\providecommand{\tabdashruleskip}[1]{\noalign{\vskip 1.5pt}\omit&\multispan{#1}{\color{black!40}\xleaders\hbox to 3.2pt{\hss\vrule
  height 0.25pt depth 0.25pt width 1.8pt\hss}\hskip 0pt plus 1fill\relax}\cr\noalign{\vskip 1.5pt}}
\begin{table*}[t]
  \centering
  \caption{
    \textbf{Downstream encoder evaluation.}
    \textbf{(a)}~Audio event classification with audio, video and AV inputs.
    \textbf{(b)}~Visual classification from video-only inputs.
    \textbf{(c)}~Sound-prompted segmentation (ADE20K-Sounds) and AV retrieval (AudioSet).
    Classification with attentive probes. Dashed rules separate larger-data video references.
    \avjepa leads in audio and AV classification and segmentation, improves most matched video baselines, and is competitive in retrieval.
	Best two results are bold and underlined, respectively.
  }
  \label{tab:frozen_comparison}
  \fontsize{7.5}{9}\selectfont  %
  \setlength{\tabcolsep}{2pt}
  \vspace{-0.25cm}
  \sbox{\ttbA}{%
  \begin{tabular}{@{}c@{\hspace{3pt}}l c c c c c@{}}
    \toprule
    & & \multicolumn{3}{c}{AS-20K} & FSD & ESC \\
    \cmidrule(lr){3-5} \cmidrule(lr){6-6} \cmidrule(l){7-7}
    & Method & A & V & A-V & mAP & Acc. \\
    \midrule
    & \cavmae~\cite{gong2023contrastive} & 19.4 & 18.1 & 34.6 & 46.1 & 77.5
    \\
    & EquiAV~\cite{kim2024equiav} & 34.3 & 18.6 & 38.6 & 57.9 & 93.2
    \\
    & \cavmaesync~\cite{araujo2025cavmaesync} & 29.3 & 16.2 & 30.5 & 55.5 & 89.2
    \\
    & LeAVJEPA~\cite{robson2026leavjepa} & 27.8 & 14.8 & 31.0 & \na & 88.0
    \\
    & \mjepa~\cite{teotia2026mjepa} & 38.8 & 21.8 & {41.3} & 62.3 & 95.8
    \\
    & \quad + \protoloss & 38.0 & 22.1 & 41.8 & {62.1} & 95.4
    \\
    \oursrow \cellcolor{white} & \avjepa & \second{40.1} & \second{22.9} & \second{43.2} & \second{66.4} & \best{96.7}
    \\
    \oursrow \cellcolor{white}\ttsize{8}{ViT-B} & \quad + Attn.~Pool & \best{40.5} & \best{23.8} & \best{43.3} & \best{67.4} & \second{96.4}
    \\
    \midrule
    & LeAVJEPA~\cite{robson2026leavjepa} & 31.6 & 18.3 & 36.0 & \na & 91.3
    \\
    & \mjepa~\cite{teotia2026mjepa} & 40.7 & 25.9 & 43.8 & {64.3} & 96.5
    \\
    & \quad + \protoloss & 40.5 & 26.2 & 44.7 & 64.7 & 96.3
    \\
    \oursrow \cellcolor{white} & \avjepa & \second{41.7} & \best{27.1} & \best{45.6} & \second{68.0} & \second{97.3}
    \\
    \oursrow \cellcolor{white}\ttsize{5}{ViT-L} & \quad + Attn.~Pool & \best{42.1} & \second{26.5} & \second{45.3} & \best{68.7} & \best{97.4}
    \\
    \bottomrule
    \vspace{-0.1cm}
  \end{tabular}}
  \sbox{\ttbB}{%
  \begin{tabular}{@{}c@{\hspace{3pt}}l l c c c c c@{}}
    \toprule
    & & & \multicolumn{3}{c}{Appearance} & \multicolumn{2}{c}{Motion} \\
    \cmidrule(lr){4-6} \cmidrule(l){7-8}
    & Method & Data & IN1K & K400 & COIN & SSv2 & D48 \\
    \midrule
    & \vjepatworep & VM2M & \second{75.2} & \second{78.1} & \second{78.6} & \second{65.9} & 85.6
    \\
    & \vjepatwoonerep & VM2M & 74.5 & 76.4 & 78.5 & 64.8 & \second{85.9}
    \\
    \oursrow \cellcolor{white} & \avjepa & VM2M & \best{76.4} & \best{79.5} & \best{79.2} & \best{68.2} & \best{87.8}
    \\
    \tabdashruleskip{7}
    \ttsize{4.15}{ViT-B} & SALT~\cite{li2025salt} & V-3.6M & 74.8 & 70.9 & 80.5 & 66.1 & 78.7
    \\
    \midrule
    & \vjepa~\cite{bardes2024revisiting} & VM2M & 74.8 & 80.8 & \na & 69.5 & \na
    \\
    & InternVideo~\cite{wang2022internvideo} & UH1.35M & \na & 78.6 & \na & 67.4 & 69.6
    \\
    & \vjepatworep & VM2M & 77.4 & 82.1 & 81.9 & 71.6 & \best{89.1}
    \\
    & \vjepatwoonerep & VM2M & \second{78.5} & \second{82.2} & \second{85.2} & \second{71.8} & 87.5
    \\
    \oursrow \cellcolor{white} & \avjepa & VM2M & \best{78.7} & \best{83.9} & \best{86.6} & \best{73.5} & \second{88.1}
    \\
    \tabdashruleskip{7}
    & SALT~\cite{li2025salt} & V-3.6M & 79.0 & 85.4 & 85.3 & 74.9 & 82.5
    \\
    & \vjepatwo~\cite{assran2025vjepa2} & VM22M & 83.5 & 85.1 & 86.8 & 73.7 & 89.0
    \\
    \ttsize{8.15}{ViT-L} & \vjepatwoone~\cite{murlabadia2026vjepa21} & VM163M & 81.8 & 84.3 & \na & 74.2 & \na
    \\
    \bottomrule
  \end{tabular}}%
  \sbox{\ttbC}{%
  \begin{tabular}{@{}l c c c@{}}
    \toprule
    & Segm. & \multicolumn{2}{c}{Retr. R@1} \\
    \cmidrule(lr){2-2} \cmidrule(l){3-4}
    Method & mAP & A$\to$V & V$\to$A \\
    \midrule
    Center-Prior  & 23.2 & \na & \na 
    \\
    Random Retrieval & \na  & 0.06 & 0.06
    \\
    DenseAV~\cite{hamilton2024separating} & 37.1 & 27.6 & 26.3
    \\
    TACO~\cite{malard2026taco} & 35.8 & \na & \na
    \\
    ImageBind~\cite{girdhar2023imagebind} & 18.3 & 20.8 & 22.1
    \\
    \cavmae\tiny{Scale++}~\cite{gong2023contrastive} & 25.9 & 17.9 & 19.9
    \\
    EquiAV~\cite{kim2024equiav} & 25.9 & 29.5 & 30.3
    \\
    \cavmaesync~\cite{araujo2025cavmaesync} & 22.6 & 27.9 & \best{35.2}
    \\
    \mjepa (B) & 35.0 & 25.8 & 27.0
    \\
    \mjepa (L)~\cite{teotia2026mjepa} & 31.8 & \best{33.7} & \second{33.8}
    \\
    LeAVJEPA (L)~\cite{robson2026leavjepa} & \na & 19.1 & 18.7
    \\
    \quad + InfoNCE head & \na & 24.4 & 25.6
    \\
    \oursrow \avjepa (B) & \best{44.1} & 24.2 & 22.6
    \\
    \oursrow \avjepa (L) & \second{42.8} & \second{31.5} & 30.7
    \\
    \bottomrule
  \end{tabular}}%
  \begin{subtable}[t]{\wd\ttbA}
    \centering
    \caption{Audio \& AV event classification.}
    \label{tab:frozen_comparison_audio}
    \usebox{\ttbA}
  \end{subtable}%
  \hfill
  \begin{subtable}[t]{\wd\ttbB}
    \centering
    \caption{Image classification \& video action recognition.}
    \label{tab:frozen_comparison_video}
    \usebox{\ttbB}
  \end{subtable}%
  \hfill
  \begin{subtable}[t]{\wd\ttbC}
    \centering
    \caption{Cross-modal segment. \& retrieval.}
    \label{tab:frozen_comparison_segm}
    \usebox{\ttbC}
  \end{subtable}
  \vspace{-0.25cm}
\end{table*}

\section{Evaluation Benchmarks}
\label{sec:benchmarks}

We introduce two complementary benchmarks: \intphysx tests
cross-modal intuitive physics, while \avnavbench evaluates WM planning toward image or sound goals.

\subsection{IntPhys-X: Cross-Modal Intuitive Physics}
\label{sec:bench_intphysx}
\label{sec:intphysx_benchmark}

\intphysx tests intuitive physics that requires cross-modal inference by design: no task can be solved from a
single modality.
Each item pairs a query in one modality with two options in the other, only one of which stems from the same
event.
The benchmark has two parts, which test complementary abilities: capturing the information that the two
streams share, and reasoning from one stream to a complementary picture of the world state in the other.
\suppref{sec:suppl_intphysx_construction} details how the dataset was built and \suppref{sec:suppl_intphysx_scenarios} each scenario.

\begin{figure}[t]
	\centering
	\begin{tikzpicture}[x=1mm, y=0.9mm, line cap=round, line join=round]
\input{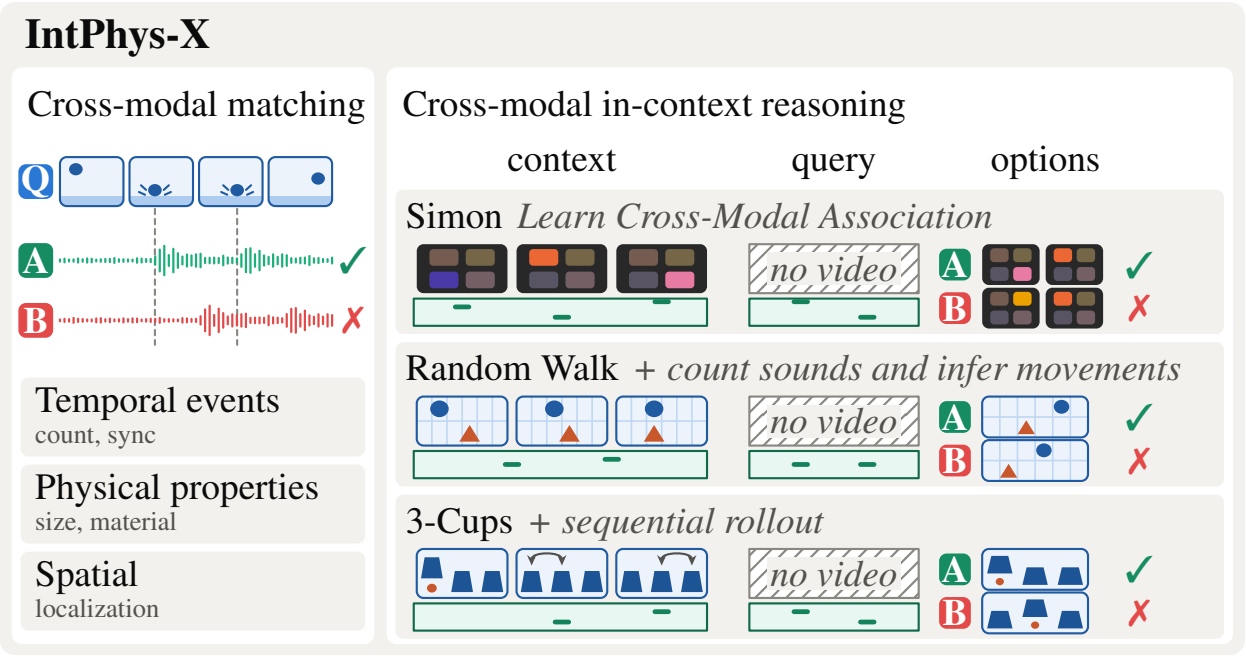}
\def\isxTH{2.0}\def\isxTS{0.38}\def\isxTW{0.45}
\fill[isx/card] (0,0) rectangle (82.4,48.0);
\node[isx/title] at (1.6,44.6) {IntPhys-X};
\fill[white, rounded corners=2pt] (0.8,0.8) rectangle (24.8,43.2);
\fill[white, rounded corners=2pt] (25.6,0.8) rectangle (81.6,43.2);
\node[isx/lbl, anchor=base west, inner sep=0pt, font=\scriptsize, text=avInk] at (1.8,39.6) {Cross-modal matching};
\node[isx/ab=cVideo, minimum size=2.3mm] at (2.4,34.8) {Q};
\foreach \isxK/\isxBxx/\isxH/\isxC in {0/0.25/0.9/0, 1/0.40/0/1, 2/0.60/0/1, 3/0.78/0.5/0}
  {\isxFrame{4.0+\isxK*4.6}{33.0}{4.2}{3.6}{\isxBxx}{\isxH}{\isxC}}
\def\isxTa{10.28}  %
\def\isxTb{15.72}  %
\draw[isx/guide] (\isxTa,32.8) -- (\isxTa,22.8);
\draw[isx/guide] (\isxTb,32.8) -- (\isxTb,22.8);
\node[isx/ab=cAudio!80!black, minimum size=2.3mm] at (2.4,29.0) {A};
\isxWave{4.0}{29.0}{18.0}{1.4}{cAudio}{60}{\isxTa}{\isxTb}{0}
\node[isx/lbl] at (23.4,29.0) {\isxCorrect};
\node[isx/ab=cLoss, minimum size=2.3mm] at (2.4,24.6) {B};
\isxWave{4.0}{24.6}{18.0}{1.4}{cLoss}{60}{\isxTa}{\isxTb}{10}
\node[isx/lbl] at (23.4,24.6) {\isxWrong};
\foreach \isxY/\isxG/\isxS in {14.6/Temporal events/{count, sync}, 8.2/Physical properties/{size, material}, 1.8/Spatial/localization} {
  \fill[avGrid!45, rounded corners=1.5pt] (1.4,\isxY) rectangle (24.2,\isxY+5.8);
  \node[isx/lbl, anchor=base west, inner sep=0pt, font=\scriptsize, text=avInk] at (2.3,\isxY+3.3) {\isxG};
  \node[isx/lbl, anchor=base west, inner sep=0pt, font=\tiny] at (2.3,\isxY+1.0) {\isxS};}
\begin{scope}[shift={(25.6,0)}]
\node[isx/lbl, anchor=base west, inner sep=0pt, font=\scriptsize, text=avInk] (isxS2) at (1.0,39.6) {Cross-modal in-context reasoning};%
\node[isx/lbl, anchor=base, font=\scriptsize, text=avInk] at (11.6,35.6) {context};
\node[isx/lbl, anchor=base, font=\scriptsize, text=avInk] at (29.6,35.6) {query};
\node[isx/lbl, anchor=base, font=\scriptsize, text=avInk] at (43.6,35.6) {options};
\def\isxFW{6.0}\def\isxFS{6.6}\def\isxFH{3.6}
\def\isxRowC#1#2#3{%
  \fill[avGrid!45, rounded corners=1.5pt] (0.6,#1) rectangle (55.4,{#1+10.6});
  \node[isx/lbl, anchor=base west, inner sep=0pt, font=\scriptsize, text=avInk] (isxRn) at (1.2,{#1+7.8}) {#2};
  \node[isx/lbl, anchor=base west, inner sep=0pt, font=\scriptsize\itshape] at ($(isxRn.base east)+(0.8,0)$) {#3};
  \isxTrack{1.8}{21.2}{#1+0.6}
  \isxNoVideo{24.0}{35.2}{#1+3.0}{\isxFH}
  \isxTrack{24.0}{35.2}{#1+0.6}
  \node[isx/ab=cAudio!80!black, minimum size=2.1mm] at (37.6,{#1+5.1}) {A};
  \node[isx/lbl] at (49.8,{#1+5.1}) {\isxCorrect};
  \node[isx/ab=cLoss, minimum size=2.1mm] at (37.6,{#1+1.9}) {B};
  \node[isx/lbl] at (49.8,{#1+1.9}) {\isxWrong};}
\def\isxY{23.6}
\isxRowC{\isxY}{Simon}{Learn Cross-Modal Association}
\foreach \isxK/\isxL in {0/3,1/1,2/4} {\isxSimon{2.0+\isxK*\isxFS}{\isxY+3.0}{\isxFW}{\isxFH}{\isxL}}
\foreach \isxK/\isxL in {0/3,1/1,2/4} {\isxTone{5.0+\isxK*\isxFS}{\isxY+0.6}{\isxL}}
\foreach \isxX/\isxL in {27.4/4,31.8/1} {\isxTone{\isxX}{\isxY+0.6}{\isxL}}
\foreach \isxK/\isxL in {0/4,1/1} {\isxSimon{39.4+\isxK*4.2}{\isxY+3.6}{3.8}{3.0}{\isxL}}
\foreach \isxK/\isxL in {0/2,1/1} {\isxSimon{39.4+\isxK*4.2}{\isxY+0.4}{3.8}{3.0}{\isxL}}
\def\isxY{12.4}
\isxRowC{\isxY}{Random Walk}{+ count sounds and infer movements}
\foreach \isxK/\isxA/\isxB in {0/1/3,1/2/3,2/2/2} {\isxBoard{2.0+\isxK*\isxFS}{\isxY+3.0}{\isxFW}{\isxFH}{\isxA}{\isxB}}
\foreach \isxK/\isxL in {1/2,2/3} {\isxTone{1.7+\isxK*\isxFS}{\isxY+0.6}{\isxL}}
\foreach \isxX/\isxL in {27.4/2,31.8/2} {\isxTone{\isxX}{\isxY+0.6}{\isxL}}
\isxBoard{39.4}{\isxY+3.6}{7.0}{3.0}{4}{2}
\isxBoard{39.4}{\isxY+0.4}{7.0}{3.0}{3}{1}
\def\isxY{1.2}
\isxRowC{\isxY}{3-Cups}{+ sequential rollout}
\foreach \isxK/\isxU/\isxS/\isxT in {0/1/0/0,1/0/1/2,2/0/2/3} {\isxCups{2.0+\isxK*\isxFS}{\isxY+3.0}{\isxFW}{\isxFH}{\isxU}{\isxS}{\isxT}}
\foreach \isxK/\isxL in {1/1,2/3} {\isxTone{5.0+\isxK*\isxFS}{\isxY+0.6}{\isxL}}
\foreach \isxX/\isxL in {27.4/3,31.8/1} {\isxTone{\isxX}{\isxY+0.6}{\isxL}}
\isxCups{39.4}{\isxY+3.6}{7.0}{3.0}{1}{0}{0}
\isxCups{39.4}{\isxY+0.4}{7.0}{3.0}{2}{0}{0}
\end{scope}
\end{tikzpicture}%
    \vspace{-0.25cm}
	\caption{Illustration of all \intphysx scenarios.\vspace{-0.5cm}}%
	\label{fig:intphysx_structure}
	\vspace{0cm}
\end{figure}

\leanparagraph{Cross-modal matching without context (\cref{fig:intphysx_structure}, left).}  %
In the first part, each item is built from a minimal pair of clips differing in
one attribute, \eg two vs.\ three drum strikes, so only the information shared by the two streams identifies the
matching pair.
Finding matching pairs thus makes this part well suited to evaluating the cross-modal capabilities of representation
models, whose query and option embeddings can be compared directly.
Its five scenarios each isolate one regularity, in three groups: temporal events (\emph{count}, \emph{synchronization}),
physical properties (\emph{size}, \emph{material}) and spatial cues (\emph{localization}).
Each scenario has $5$k synthetic and $1$k real examples, except localization (only synthetic).

\leanparagraph{Cross-modal in-context reasoning (\cref{fig:intphysx_structure}, right).}  %
The second part tests adaptation and cross-modal reasoning across three increasingly difficult scenarios, where AV context links actions to unfamiliar sounds; the model must learn this mapping and infer the world state in one modality from the other.
In \emph{Simon}, the model matches sequences of colored lights and sounds, but can reject an option from a single mismatching event.
The remaining scenarios require the final state, removing this shortcut.
In \emph{Random Walk},  an audio query describes actions moving characters on a grid, and the model selects their final positions; counting actions suffices regardless of order.
In \emph{3-Cups}, an audio query describes movements of cups hiding a ball; as order matters,  the model must imagine a full rollout to pick the final state.

\leanparagraph{Evaluation protocols.}
No model is trained on \intphysx, which provides a small dev set for calibrating inference methods: representation
models compare embeddings by similarity, multimodal LLMs answer a multiple-choice prompt, and \avwm
scores options by violation of expectation, \ie how much it reduces prediction error (\suppref{sec:suppl_intphysx_protocol}).

\subsection{AVNav-Bench: Audio-Visual Navigation}
\label{sec:avwmbench}

\avnavbench is a simulated indoor navigation benchmark with binaural sound from SoundSpaces~2.0~\cite{chen2022soundspaces2},
built from walks through 150 HSSD houses~\cite{khanna2024hssd} split into training, validation and test houses.
A goal is either an image of the goal pose or the 1--3\,s sound of a named object.
In the \emph{offline} mode \cite{bar2025nwm}, a model plans once from 1\,s of context toward image
goals 2, 4 and 6\,s ahead, and toward \emph{end goals}, \ie an image or a sound over 9\,s ahead at the end of the walk,
where the plan must also stop.
In the \emph{closed-loop} mode, the model acts in the simulator and replans every second, on the 2\,s image goals and
on \emph{medium} and \emph{hard} goals in other rooms, 5--10\,m and 10--25\,m away along the shortest path.
We define success geometrically and identically for every model, so latent and pixel WMs as well as policies
can be evaluated.
To our knowledge, its closed loop is the first AV navigation setting in which the WM itself decides, with no
policy in the loop.
\suppref{sec:suppl_avnav} details the construction, settings, success criteria and relation to prior benchmarks.
\section{Experiments}
\label{sec:av_results}

\subsection{Setup}
\label{sec:setup}

\leanparagraph{\avjepa.}
We pretrain \avjepa at ViT-B and ViT-L (86M and 304M params) encoder sizes on AudioSet-2M~\cite{gemmeke2017audio} for AV experiments and
on VideoMix2M~\cite{bardes2024revisiting} for video-only ones.
All encoders are evaluated frozen zero-shot or with attentive probes.
Besides published results, we compare with \vjepatwo~\cite{assran2025vjepa2} and \vjepatwoone~\cite{murlabadia2026vjepa21} reproductions trained with the same data as \avjepa, isolating the effect of our model design and objective.
Further details in~\suppref{sec:suppl_setup}.

\leanparagraph{\avwm.}
We train the action-free WM, a ViT-H predictor (630M params) with the AV-Combiner (39M) on top, on the frozen
\avjepa-L encoder with
10\,s clips from AudioSet~\cite{gemmeke2017audio} and Kinetics~\cite{kay2017kinetics}, and 64\,s windows of
instructional ~\cite{miech2019howto100m} and egocentric
videos~\cite{grauman2022ego4d,grauman2024egoexo4d,wang2023holoassist,chavan2025indego,ma2024nymeria,detone2026nymeriaplus}
($\sim$100k hours in total).
We also train AVJ-WM-B, a ViT-B predictor (85M; and 14M combiner), on \avjepa-B with a shorter schedule.
For navigation, we fine-tune both with action adapters and compare with policies trained by imitation or
reinforcement learning (RL) ~\cite{shah2023gnm,sridhar2024nomad,chen2020soundspaces,chen2021savi} and with the diffusion WMs CDiT~\cite{bar2025nwm} and AV-CDiT~\cite{wang2025avwm}, which we also train on \avjepa features; all are trained
in the 92 AVNav-Bench training houses, on their 9,702 walks or, for RL, in the simulator. Further implementation, dataset, baselines and evaluation details are in \suppref{sec:suppl_setup} and \suppref{sec:supplworldmodel}.

\leanparagraph{\intphysx.}
The no-context scenarios target representation models.
We compare \avjepa with two open-source AV encoders: PE-AV~\cite{vyas2025pushing}, a contrastive encoder that is
larger and trained on more data (1.1B parameters, 114M clips), and \cavmaesync~\cite{araujo2025cavmaesync}, similar
in scale to \avjepa-B.
To isolate the effect of our modeling choices, we add \mjepa~\cite{teotia2026mjepa}, a JEPA reproduced on the same
data that encodes spectrograms and predicts clip-level cross-modal targets.
We also report MLLMs (Gemini, Qwen3-Omni~\cite{xu2025qwen3omni}, AV-Flamingo~\cite{ghosh2026avflamingo})
and \avwm for reference.
Models with mono audio input receive the right channel.

\subsection{Audio-Visual Representations}
\label{sec:av_main}

\leanparagraph{Audio and AV classification.}
\Cref{tab:frozen_comparison_audio} shows that \avjepa outperforms all baselines across benchmarks and scales.
On AS-20K, it improves over \mjepa, the strongest baseline, by 1.0--1.9 mAP with audio, video and AV inputs, and the gains transfer to FSD50K and ESC-50.
Attentive pooling during pretraining (+ Attn.~Pool), instead of mean pooling, further improves audio classification.
Notably, adding \protoloss to \mjepa improves its video and AV processing.
However, broader gains only emerge when combined with a shared time axis and chunk-level targets (\cref{tab:ablations_single_col_v1}).

\leanparagraph{Visual recognition.}
With video-only pretraining (\cref{tab:frozen_comparison_video}), 
\avjepa generally outperforms matched \vjepatwo and \vjepatwoone models  at both scales, with gains of up to 2.3 points on motion-centric SSv2.
The benefits of \protoloss thus extend to video-only representation learning, without audio as additional supervision.
\avjepa even approaches V-JEPA variants trained on $> 10 \times$ more data.

\leanparagraph{Cross-modal segmentation and retrieval.}
\Cref{tab:frozen_comparison_segm} evaluates AV correspondence via sound-prompted segmentation on ADE20K-Sounds~\cite{hamilton2024separating} and cross-modal retrieval on AudioSet.
\avjepa achieves the best segmentation mAP, outperforming specialized baselines~\cite{hamilton2024separating}, suggesting that dense cross-modal prediction grounds sounds in the visual scene.
In retrieval, \avjepa-L ranks second for A$\to$V and third for V$\to$A.

\begin{table}[t]
  \centering
  \providecommand{\abcInput}[1]{{%
    \renewenvironment{table}[1][]{}{}%
    \renewcommand{\caption}[2][]{}%
    \renewcommand{\label}[1]{}%
    \renewcommand{\vspace}[1]{}%
    \input{#1}}}
  \ifdefined\abcFitBox\else\newsavebox\abcFitBox\fi
  \providecommand{\abcGrab}[1]{{%
    \let\abcOrigTabular\tabular \let\abcOrigEndTabular\endtabular
    \renewenvironment{tabular}{\global\setbox\abcFitBox\hbox\bgroup\abcOrigTabular}{\abcOrigEndTabular\egroup}%
    \setbox0\hbox{#1}}}
  \providecommand{\abcPlace}{%
    \ifdim\wd\abcFitBox>\linewidth\resizebox{\linewidth}{!}{\usebox\abcFitBox}\else\usebox\abcFitBox\fi}
  \caption{
    \textbf{\avjepa ablation study} 
    with ViT-B and shorter schedule.
    Deltas relative to \mjepa.
    Shaded rows are \avjepa variants.
  }
  \vspace{-0.2cm}
  \label{tab:ablations_single_col_v1}
  \abcGrab{{%
    \renewenvironment{table}[1][]{}{}%
    \renewcommand{\caption}[2][]{}%
    \renewcommand{\label}[1]{}%
    \renewcommand{\vspace}[1]{}%
    
\begin{table}[t]
	\centering
	\providecommand{\abstep}[2]{\hspace*{#1em}+ #2}
	\providecommand{\abNameExtra}{0pt}
	\providecommand{\abGroupExtra}{0pt}
	\caption{
		\textbf{Evaluating \avjepa design choices.}
		Each row varies one design choice relative to the row above.
		Deltas relative to \mjepa~\cite{teotia2026mjepa}.
		We highlight the full \avjepa variants.
	}
	\label{tab:ablation_cumulative_compact}
	\footnotesize
	\vspace{-0.2cm}
	\setlength{\tabcolsep}{6.pt}
	\let\originalDneg\dneg
	\newcommand{\dnegsmall}[1]{\scalebox{0.95}{\originalDneg{$#1$}}}
	\let\originalDpos\dpos
	\newcommand{\dpossmall}[1]{\scalebox{0.95}{\originalDpos{$#1$}}}
	\let\originalDnul\dnul
	\newcommand{\dnulsmall}[1]{\scalebox{0.95}{\originalDnul{$#1$}}}
	\renewcommand{\oursrow}{\rowcolor{oursrowbg}[2.5pt][4pt]}
	\begin{tabular}{
			@{}l<{\hspace{\abNameExtra}}@{\hspace{6pt}}
			r@{\hspace{2.5pt}}l<{\hspace{\abGroupExtra}}@{\hspace{4pt}}
			r@{\hspace{2.5pt}}l<{\hspace{\abGroupExtra}}@{\hspace{4pt}}
			r@{\hspace{2.5pt}}l@{}
		}
		\toprule
		& \multicolumn{6}{c}{AS-20K mAP (\%)} \\
		\cmidrule(l){2-7}
		Variant
		& \multicolumn{2}{c<{\hspace{\abGroupExtra}}}{A}
		& \multicolumn{2}{c<{\hspace{\abGroupExtra}}}{V}
		& \multicolumn{2}{c}{A-V} \\
		\midrule
		
		\mjepa (spectrogram)
		& {38.8} & \dnulsmall{+0.0}
		& {21.0} & \dnulsmall{+0.0}
		& {41.1} & \dnulsmall{+0.0} \\

		\abstep{0.5}{\protoloss}
        & {38.0} & \dnegsmall{-0.8}
		& {21.3} & \dpossmall{+0.3}
		& {41.8} & \dpossmall{+0.7} \\
		
		\midrule
		
		\mjepa (raw waveform)
		& 35.7 & \dnegsmall{-3.1}
		& 19.6 & \dnegsmall{-1.4}
		& {37.7} & \dnegsmall{-3.4} \\
		
		\abstep{0.5}{M-RoPE}
		& 37.7 & \dnegsmall{-1.1}
		& 19.4 & \dnegsmall{-1.6}
		& {40.6} & \dnegsmall{-0.5} \\
		
		\abstep{1.}{Chunk targets ($\ell_1$)}
		& 31.4 & \dnegsmall{-7.4}
		& 17.2 & \dnegsmall{-3.8}
		& 34.2 & \dnegsmall{-6.9} \\
		
		\abstep{1.}{\protoloss}
		& 38.1 & \dnegsmall{-0.7}
		& 19.6 & \dnegsmall{-1.4}
		& {41.0} & \dnegsmall{-0.1} \\
		
		\abstep{1.5}{Chunk targets}
		& 39.4 & \dpossmall{+0.6}
		& \second{22.2} & \dpossmall{+1.2}
		& 42.2 & \dpossmall{+1.1} \\
		
		\oursrow
		\abstep{2.0}{7 terms (\textbf{\avjepa})}
		& \second{39.4} & \dpossmall{+0.6}
		& {22.1} & \dpossmall{+1.1}
		& {\best{42.6}} & \dpossmall{+1.5} \\
		\oursrow
		\abstep{2.5}{attentive pooling}
		& \best{40.1} & \dpossmall{+1.3}
		& \best{22.9} & \dpossmall{+1.9}
		& \best{42.6} & \dpossmall{+1.5} \\
	
		\bottomrule
	\end{tabular}
	
\end{table}
}}%
  \ifdim\wd\abcFitBox<\linewidth\edef\abGroupExtra{\the\dimexpr(\linewidth-\wd\abcFitBox)/3\relax}\let\abNameExtra\abGroupExtra\fi
  \abcGrab{{%
    \renewenvironment{table}[1][]{}{}%
    \renewcommand{\caption}[2][]{}%
    \renewcommand{\label}[1]{}%
    \renewcommand{\vspace}[1]{}%
    }}%
  \abcPlace
\vspace{-0.2cm}
\end{table}

\leanparagraph{Ablation Study.}
\Cref{tab:ablations_single_col_v1} ablates the core design choices of \avjepa, starting from the M-JEPA recipe: spectrogram inputs, $\ell_1$ feature regression, and nine prediction terms with clip-level cross-modal targets.
Using waveform inputs costs up to 3.4 mAP; a shared time axis with M-RoPE recovers 2.0--2.9 mAP for audio and AV, and \protoloss closes part of the remaining gap.
Chunk-level targets improve performance with \protoloss, but lead to degraded representations with feature regression.
Conversely, \protoloss yields larger gains with chunk-level than clip-level targets, indicating a strong interaction between target granularity and the prediction objective.
Seven prediction terms match nine (\suppref{sec:suppl_av_ablations}), and attentive pooling further improves performance.
In summary, no single change suffices: only by combining \protoloss and chunk-level targets on a shared time axis does the model achieve SoTA.

\leanparagraph{Further results and analysis.}
In the Appendix, we ablate the prediction terms, \protoloss and pretraining modalities (\suppref{sec:suppl_av_results}), scale the \avjepa encoders (\suppref{sec:suppl_scaling}), show qualitative examples (\suppref{sec:suppl_av_qual}) and results on speech (\suppref{sec:suppl_speech}).

\subsection{\texorpdfstring{IntPhys-X Results}{IntPhys-X Results}}
\label{sec:intphysx}
\DeclareRobustCommand{\ipxSim}{\makebox[0.95em][l]{\tikz[baseline=0pt]\draw[line width=0.45pt] (0,0) rectangle (0.6em,0.6em);}}%
\DeclareRobustCommand{\ipxPrompt}{\makebox[0.95em][l]{\tikz[baseline=0pt]\draw[line width=0.45pt] (0.3em,0.3em) circle (0.32em);}}%
\DeclareRobustCommand{\ipxVoE}{\makebox[0.95em][l]{\tikz[baseline=0pt]\draw[line width=0.45pt] (0,0) -- (0.66em,0) -- (0.33em,0.62em) -- cycle;}}%
\ifdefined\ipxSideBySide\captionsetup{type=table}\else\begin{table}[t]\fi
  \centering
  \caption{\textbf{IntPhys-X accuracy.} Chance at 50\% $\pm$ 1.39 / 3.10 / 2.19, 95\% CI for Synthetic / Real / In-context.
    Humans score 94.8\% / 86.5\% on average for no-context (synth + real) / in-context scenarios; most models are far lower.
    Inference: \ipxSim similarity, \ipxPrompt text prompt, \ipxVoE violation of expectation.}

  \label{tab:intphysx_results}
  \vspace{-0.25cm}
  \ifdefined\ipxSideBySide\vfill\fi  %
  \setlength{\tabcolsep}{2.5pt}
  \resizebox{\linewidth}{!}{%
  \begin{tabular}{@{}c@{\hspace{3pt}}l ccccc cccc ccc@{}}
    \toprule
    & & \multicolumn{5}{c}{Synthetic}
    & \multicolumn{4}{c}{Real}
    & \multicolumn{3}{c}{In-context} \\
    \cmidrule(lr){3-7}\cmidrule(lr){8-11}\cmidrule(l){12-14}
    & Model & Cnt & Sync & Size & Mat & Loc
          & Cnt & Sync & Size & Mat
          & Smn & R.W. & Cups \\
    \midrule
     & 2\,fps & 88.1 & 72.1 & 87.2 & 95.4 & 92.7 & 88.8 & 83.8 & 89.0 & 91.7 & \na & \na & \na \\
     & 5\,fps & 98.5 & 92.9 & 86.4 & 97.1 & 92.9 & 97.7 & 92.7 & 89.3 & 92.4 & \na & \na & \na \\
    \multirow{-3}{*}{\rotatebox[origin=c]{90}{\footnotesize\textit{Human}}} & 30\,fps & 98.7 & 95.0 & 89.9 & 96.6 & 94.5 & 99.3 & 94.5 & 91.3 & 93.5 & 90.3 & 84.7 & 84.4 \\
    \midrule
     & \ipxSim{}PE-AV-L & 55.9 & 55.7 & 50.6 & \best{49.9} & 49.1 & 64.9 & 61.8 & \second{57.7} & \best{67.7} & \na & \na & \na \\
     & \ipxSim\cavmaesync & 48.2 & 47.8 & 50.3 & \second{48.9} & 49.7 & 41.8 & 45.3 & 48.0 & 53.8 & \na & \na & \na \\
     & \ipxSim\mjepa-B & \second{73.7} & 77.6 & \second{55.2} & 47.4 & \best{52.6} & \second{76.5} & \best{80.8} & 46.9 & 56.0 & \na & \na & \na \\
     & \ipxSim\mjepa-L & 63.7 & 76.8 & 50.2 & 47.8 & 51.4 & 67.2 & 69.3 & 51.4 & 52.2 & \na & \na & \na \\
    \oursrow \cellcolor{white} & \ipxSim\avjepa-B & 71.3 & \second{78.2} & \best{60.2} & 48.1 & \second{51.7} & 76.3 & 76.2 & 53.3 & \second{66.8} & \na & \na & \na \\
    \oursrow \cellcolor{white}\multirow{-6}{*}{\rotatebox[origin=c]{90}{\footnotesize\textit{Representation}}} & \ipxSim\avjepa-L & \best{74.2} & \best{83.7} & 50.9 & 45.5 & 50.0 & \best{76.9} & \second{78.4} & \best{63.4} & 66.6 & \na & \na & \na \\
    \midrule
     & \ipxPrompt{}Gemini-4 Argon & \best{55.8} & \best{55.2} & \best{65.8} & \best{58.5} & \second{49.9} & \second{78.2} & \second{76.2} & \best{69.5} & \best{79.0} & \best{66.1} & \best{60.9} & \second{50.8} \\
     & \ipxPrompt{}Gemini 3.1 Pro & \second{52.4} & \second{51.6} & \second{56.1} & 51.4 & \second{49.9} & 72.3 & 65.4 & \second{64.3} & \second{78.7} & 52.3 & \second{54.0} & \second{50.8} \\
     & \ipxPrompt{}Gemini 3.8 Flash & 51.7 & 50.3 & 51.3 & 50.8 & 49.8 & 59.7 & 58.6 & 52.4 & 68.1 & 44.7 & 49.5 & 49.9 \\
     & \ipxPrompt{}Qwen3-Omni & 50.4 & 49.5 & 50.0 & 49.1 & 49.6 & 55.4 & 50.6 & 49.8 & 61.1 & 34.0 & 51.7 & \best{51.2} \\
     & \ipxPrompt{}AV-Flamingo & 50.0 & 49.9 & 50.1 & 49.9 & \best{50.2} & 51.2 & 50.4 & 50.6 & 54.3 & 44.1 & 49.6 & 49.8 \\
    \oursrow \cellcolor{white}\multirow{-6}{*}{\rotatebox[origin=c]{90}{\footnotesize\textit{MLLMs \& WM}}} & \ipxVoE\avwm & 48.0 & 43.5 & 50.1 & \second{51.6} & 49.6 & \best{80.7} & \best{88.2} & 50.2 & 57.5 & \second{54.5} & 50.0 & \best{51.2} \\
    \bottomrule
  \end{tabular}}
\vspace{-0.2cm}
\ifdefined\ipxSideBySide\else\end{table}\fi

\label{sec:intphysx_results}

{\color{black}%
\leanparagraph{Overview.}
IntPhys-X remains far from solved (\cref{tab:intphysx_results}).
On the no-context scenarios, \avjepa-L averages 65.5\%, ahead of all representation models and on par with Gemini-4
Argon (65.3\%).
In a human study with at least 100 unique annotators and 1,250 annotations per scenario (\suppref{sec:protocol_subjective}), humans reach 86--99\% on every no-context scenario at 5 and 30\,fps, and the best model in each column trails them by 6--42 points.
Even for humans, the frame rate matters: at 2\,fps, synchronization accuracy drops to 72.1\%, since judging it requires
fine timing.
This drop highlights a challenge for current models, some of which see video at only a few frames per second
(1.6--5\,fps; \suppref{sec:suppl_intphysx_protocol}).

\leanparagraph{Temporal events.}  %
Joint encoding suits temporal alignment: \avjepa and \mjepa encode both streams with one encoder and predict one from
the other, so audio and video events interact before pooling, and they lead all representation models on count and
synchronization (up to 83.7\% on synthetic and 80.8\% on real items).
Contrastive models, which align pooled embeddings, fall behind: PE-AV trails by 12--28 points despite its scale, and
\cavmaesync stays near or below chance.
\avwm, scored by violation of expectation, is the best model on real count and synchronization (80.7\% and 88.2\%)
but stays near or below chance on the out-of-distribution synthetic clips (\suppref{sec:suppl_intphysx_protocol}).
Multimodal LLMs, which see few frames per second, stay near chance on synthetic temporal events (at most 55.8\%).

\leanparagraph{Physical properties.}  %
Inferring these properties from sound remains open: on synthetic size and material, best models trail humans by 24--38 points.
Scores rise on real recordings, most for MLLMs (up to 79.0\% \vs 67.7\% for representation models), but MLLMs drop
on synthetic minimal pairs, which differ only in the critical attribute (\eg Argon: 79.0\% to 58.5\% on material).
We hypothesize that they recall how recognized objects typically sound rather than infer the property from
cross-modal evidence, which real recordings alone would hide and \intphysx expose.
Localization stays at chance (at most 52.6\% \vs 94.5\% for humans).

\leanparagraph{In-context reasoning.}  %
Models associate sounds with actions in context when the answer can be read off individual sounds, but remain near chance when action order matters.
Only Gemini-4 Argon clearly exceeds chance on Simon (66.1\%) and Random Walk (60.9\%), where associating or counting individual sounds suffices.
It finds these shortcuts inconsistently: on Simon, it reports checking only the step in which the two options differ in about half of the items, and on Random Walk it reaches 75.7\% only when the prompt hints that counting the sounds
suffices (\suppref{sec:suppl_intphysx_protocol}).
On 3-Cups, which requires a full cross-modal rollout of the scene, no model beats chance, whereas humans solve all three scenarios (84.4--90.3\% at 30\,fps), leaving cross-modal world modeling over time an open challenge
(\suppref{sec:suppl_avwm_context}).
}

\begin{table}[ht!]
  \color{black} %
  \centering
  \caption{\textbf{AVNav-Bench: success \% on the test houses.}
    We group policies before WMs, which all use our planner. --:~n/a or infeasible. \textsuperscript{\dag}RL with simulator access; all other models use the training walks. \emph{Data}: pretraining on NWM's navigation data (NWM\textsuperscript{+} adds Ego4D) or our AV data. WMs on SD-VAE latents start from NWM's released checkpoints; we train all others.
    \emph{Time}: plan for a 2\,s closed-loop episode on one GPU.
    On med./hard, video-only models count sound goals as failures. Details in \suppref{sec:suppl_avnav}.}%
  \label{tab:avnavbench}
  \scriptsize
  \setlength{\tabcolsep}{1.0pt}
  \vspace{-0.2cm}
  \resizebox{\linewidth}{!}{%
  \begin{tabular}{@{}c@{\hspace{2pt}}l@{\hspace{4pt}}l@{\hspace{4pt}}l *{8}{c} r@{}}
    \toprule
    & & & & \multicolumn{5}{c}{Offline} & \multicolumn{3}{c}{Closed loop} & \\
    \cmidrule(lr){5-9} \cmidrule(lr){10-12}
    & & & & \multicolumn{3}{c}{Image goal} & \multicolumn{2}{c}{End goal} & & & & Time \\
    \cmidrule(lr){5-7} \cmidrule(lr){8-9}
    & Model & Latents & Data & 2\,s & 4\,s & 6\,s & Image & Sound & 2\,s & Med. & Hard & (s) \\
    \midrule
    & \multicolumn{3}{@{}l}{Goal-blind walk} & 49.0 & 32.5 & 9.2 & 0.0 & 0.0 & 51.7 & 0.0 & 0.0 & \na \\
    & \multicolumn{3}{@{}l}{Random actions} & 26.1 & 18.8 & 9.4 & 0.7 & 2.2 & 27.5 & 0.1 & 0.0 & \na \\
    & \multicolumn{3}{@{}l}{Random cost} & 10.7 & 5.7 & 3.5 & 2.2 & 3.9 & 1.5 & 0.0 & 0.0 & \na \\%
    \cmidrule(l){2-13}
    & \multicolumn{3}{@{}l}{Perfect cost} & 98.0 & 96.2 & 85.8 & 84.8 & 85.7 & 100 & 46.5 & 11.0 & \na \\%
    \multirow{-5}{*}{\rotatebox[origin=c]{90}{\scriptsize\textit{Reference}}} & \multicolumn{3}{@{}l}{Perfect pose} & 100 & 100 & 100 & 100 & 100 & 100 & 100 & 100 & \na \\
    \midrule
    & GNM~\cite{shah2023gnm} & \na & NWM & 78.4 & 29.6 & 11.4 & 4.0 & \na & 94.3 & \na & \na & \na \\
    & NoMaD~\cite{sridhar2024nomad} & \na & NWM & 56.2 & 53.5 & 24.6 & 1.3 & \na & 68.5 & \na & \na & \na \\
    \cmidrule(l){2-13}
    & CDiT-B~\cite{bar2025nwm} & SD-VAE & NWM & 70.9 & 65.0 & 54.2 & 24.6 & \na & 89.7 & \na & \na & 570 \\%
    & CDiT-XL & SD-VAE & NWM\textsuperscript{+} & 78.5 & 71.4 & 61.1 & 26.9 & \na & 93.1 & \na & \na & 2,203 \\%
    & CDiT~\cite{murlabadia2026vjepa21} & \vjepatwoone-B & \na & \best{83.8} & 69.7 & 49.4 & 12.2 & \na & \best{96.1} & 1.1 & 0.3 & 22.7 \\%
    \multirow{-6}{*}{\rotatebox[origin=c]{90}{\scriptsize\textit{Video only}}} & CDiT & \avjepa-B & \na & 80.9 & 69.3 & 49.7 & 15.1 & \na & \second{95.4} & 0.9 & 0.0 & 22.6 \\%
    \midrule
    & AV-Nav\textsuperscript{\dag}~\cite{chen2020soundspaces} & \na & \na & \na & \na & \na & \na & \na & 57.2 & 13.0 & 7.0 & \na \\%
    & SAVi\textsuperscript{\dag}~\cite{chen2021savi} & \na & \na & \na & \na & \na & \na & \na & 73.1 & 25.2 & 21.3 & \na \\%
    \cmidrule(l){2-13}
    & AV-CDiT~\cite{wang2025avwm} & SD-VAE, DAC & NWM & 66.4 & 60.7 & 46.1 & 17.6 & 2.4 & 90.0 & \na & \na & 325 \\
    & AV-CDiT & \avjepa-B & \na & 81.2 & 71.9 & 55.0 & 12.7 & 4.5 & 95.3 & 1.2 & 0.2 & 26.7 \\%
    & AV-CDiT & \avjepa-B & Ours & 81.5 & 70.9 & 53.2 & 13.2 & 5.0 & 94.9 & 0.8 & 0.0 & 35.7 \\%
    \oursrow \cellcolor{white} & AVJ-WM-B & \avjepa-B & Ours & \second{82.7} & \best{77.6} & \second{66.2} & \second{32.4} & \second{18.0} & 93.7 & 1.4 & 0.1 & \best{6.7} \\%
    \oursrow \cellcolor{white}\multirow{-7}{*}{\rotatebox[origin=c]{90}{\scriptsize\textit{Audio-visual}}} & AVJ-WM & \avjepa-L & Ours & 82.4 & \second{76.6} & \best{68.3} & \best{39.6} & \best{19.1} & 92.0 & 1.7 & 0.0 & \second{14.2} \\%
    \bottomrule
  \end{tabular}}
  
 \vspace{-0.2cm}
\end{table}

\subsection{\texorpdfstring{AVNav-Bench Results}{AVNav-Bench Results}}
\label{sec:avwmplanningsec}
\label{sec:avnavresults}%

On AVNav-Bench (\cref{sec:avwmbench}), all world models use our planner (\cref{sec:avwm}), replanning every second
in the closed loop, and every choice is made on validation houses (\suppref{sec:suppl_avnav_construction}).

\leanparagraph{Latent space.}
\avjepa's latents are as good a space for planning as \vjepatwoone's: the same CDiT plans about equally well on both,
and the AV-CDiT, which also uses sound, surpasses the \vjepatwoone{} CDiT at longer horizons.
This holds although \avjepa-B saw only AudioSet's 2M clips, whereas \vjepatwoone-B is distilled from a 1.8B-parameter
model trained on 163M videos and images.

\leanparagraph{Offline planning.}
Our world models' lead grows with goal distance (\cref{tab:avnavbench}).
At 2\,s, the best diffusion world models match AVJ-WM. At 6\,s, AVJ-WM leads them by 7 points, and on end
goals, 37 steps away, it reaches 39.6\% against at most 26.9\%.
Even AVJ-WM-B beats NWM's largest model, CDiT-XL with extra Ego4D training, with a quarter of its parameters.
Pretraining is key to this lead: without it, AVJ-WM-B falls to 27.8\% at 2\,s, and with a tenth of the training walks it
loses less than the AV-CDiT, which does not gain from pretraining on our data (\suppref{tab:avnavdata}).%

\leanparagraph{Closed loop.}
Nearby goals are largely solved: on the 2\,s tier, our world models succeed in over 90\% of the episodes, close to
the strongest baselines.
AVJ-WM-B plans an episode 3--5$\times$ faster than the CDiTs on the same features and over $300\times$
faster than CDiT-XL, because it predicts each step in one forward pass rather than by iterative denoising.
Goals in other rooms remain open for every method; even the best, a policy trained by RL in the simulator, solves about
a quarter of them.
With perfect costs, our planner reaches 46.5\% on medium goals, but only 1.7\% with AVJ-WM's costs. This suggests that the
world model's costs limit planning across rooms more than the planner.

\leanparagraph{Sound goals.}
Our WMs succeed more than three times as often as the best
CDiT, which barely beats a planner with random costs.
The sound cost of AVJ-WM's predictions rises with the distance to the source in 75\% of the walks, against 65\% for the
AV-CDiT's predictions (\suppref{fig:avwmnavpred}), which may partially explain this.%

\leanparagraph{Standard benchmarks.}
AVJ-WM also does well beyond our benchmarks;
under NWM's evaluation \cite{bar2025nwm}, it outperforms NWM's released CDiTs on SCAND~\cite{karnan2022scand} and
TartanDrive~\cite{triest2022tartandrive} and has the lowest relative trajectory error on RECON~\cite{shah2021recon}
(\suppref{tab:avwmplanning}).
On IntPhys~v1.1, our rebuild of IntPhys~\cite{riochet2022intphys} without its rendering shortcut
(\suppref{sec:intphysv11}), it reaches \nWMvOne\%, on par with the best models we evaluate, including \vjepatwo up
to \mbox{ViT-g}, from its prediction error alone and without benchmark-specific training (\suppref{tab:avwmphysicsbreakdown}).

\leanparagraph{Further analysis.}
\suppref{sec:suppl_avnav_analysis} analyses these results further, ablates the fine-tuning recipe and shows executed
paths, candidate plans, predictions and counterfactual rollouts, and \suppref{sec:supplavwmablations} ablates the
pretraining recipe and the model size.

\section{Conclusion}
\label{sec:conclusion}
Sight and sound unfold on a shared time axis, so each can supervise the other.
We introduced \avjepa, a single encoder for video and audio waveforms trained with intra- and cross-modal \protoloss, which sets the SoTA among frozen self-supervised encoders on audio and AV classification and segmentation.
Its features also serve as the state of an AV navigation world model: \avwm's offline lead over diffusion world models grows with goal distance and is largest on sound goals.
We further introduced \intphysx for cross-modal intuitive physics and \avnavbench for world-model planning toward image or sound goals; on \intphysx, \avjepa leads all representation models, yet every model remains far below humans.
Overall, learning from sight and sound jointly yields strong AV representations that also enable planning, while cross-modal physical understanding remains an open challenge.

\section*{Acknowledgment}
We would like to thank Mido Assran and Nicolas Ballas for helpful discussions and support on this project.

\section*{LLM Usage Statement}
We used the AI assistants Claude Opus 5.5 (Anthropic) and GPT Sol 6.1 (OpenAI) as a writing aid, for drafting and revising prose and the \LaTeX{} code of figures and tables, and as part of our experiment tracking and reporting tooling.
All research claims, experimental designs, measurements and numbers reported here were produced by the authors and verified against the evaluation harness outputs.

{
    \small
    \bibliographystyle{ieeenat_fullname}
    \bibliography{main}
}

\clearpage

\setcounter{table}{0}\setcounter{figure}{0}\setcounter{equation}{0}
\renewcommand{\thetable}{S\arabic{table}}\renewcommand{\theHtable}{S\arabic{table}}
\renewcommand{\thefigure}{S\arabic{figure}}\renewcommand{\theHfigure}{S\arabic{figure}}
\renewcommand{\theequation}{S\arabic{equation}}\renewcommand{\theHequation}{S\arabic{equation}}

\section*{Overview}

This appendix is organized as follows.
\begin{itemize}
	\item \textbf{\Cref{sec:suppl_limitations}: Limitations and Future Work.} Discussion of limitations of our current approaches, and potential future research directions.
\end{itemize}
\begin{itemize}
  \item \textbf{\Cref{sec:suppl_related}: Extended related work.} Extends \mainref{sec:related} of the main paper, covering AV SSL, JEPAs, prototype-based targets, intuitive physics and world models.
\end{itemize}
\noindent\textbf{\avjepa.}
\begin{itemize}
  \item \textbf{\Cref{sec:suppl_setup}: \avjepa Implementation details.} Pretraining, datasets, architecture, baselines and  evaluations.
  \item \textbf{\Cref{sec:suppl_av_results}: Additional results.} Full frozen-encoder comparison tables, analysis and ablation studies.
  \item \textbf{\Cref{sec:suppl_av_qual}: Qualitative audio-visual results.} Sound-prompted localization and t-SNE embeddings.
  \item \textbf{\Cref{sec:suppl_scaling}: Scaling experiments.} Experiments scaling \avjepa to larger encoders and training datasets.
  \item \textbf{\Cref{sec:suppl_speech}: Speech experiments.} \avjepa trained on speech, compared with speech SSL models.
\end{itemize}

\vspace{0.15cm}
\noindent\textbf{\avwm.}
\begin{itemize}
  \item \textbf{\Cref{sec:supplworldmodel}: World-model details and results.} \avwm on IntPhys~v1.1, InfLevel and RECON,
    protocols, ablations, architecture, training recipe and qualitative results.
\end{itemize}

\vspace{0.15cm}
\noindent\textbf{\intphysx.}
\begin{itemize}
  \item \textbf{\Cref{sec:suppl_intphysx}: \intphysx benchmark.} Construction, scenarios with examples,
    preprocessing and inference for each model family, and the human evaluation with its quality assurance and results
    by query direction.
\end{itemize}

\vspace{0.15cm}
\noindent\textbf{AVNav-Bench.}
\begin{itemize}
  \item \textbf{\Cref{sec:suppl_avnav}: AVNav-Bench.} Construction, goals and settings, scoring, baselines, full results
    and relation to prior benchmarks.
\end{itemize}

\appendix
\section{Limitations and Future Work}
\label{sec:suppl_limitations}

This work explores audio-visual representation learning on a shared time axis, latent world modeling 
and cross-modal evaluation; \cref{sec:suppl_future} outlines the directions it opens.

\subsection{{Limitations}}
\label{sec:suppl_limits}

\paragraph{{\avjepa.}}
Web videos rarely carry reliable spatial audio, so \avjepa and the models built on it process mono audio and cannot
learn interaural cues.

\paragraph{World model.}
\avwm is an initial exploration of audio-visual world modeling, and its design sets its current limits. It is a
deterministic predictor, so its point predictions average over uncertain outcomes and provide no likelihoods. It
predicts in the latent space of a frozen \avjepa encoder that encodes 4\,s clips bidirectionally
(\mainref{sec:avwm}), so the two are not trained jointly. Planning searches over multi-step
rollouts, which is compute-intensive (\cref{tab:avnavplantime}). Our scaling study stops at \avjepa-L encoders and
ViT-H-sized predictors (\cref{sec:supplavwmablations}). Finally, most of its training data are action-free; actions
enter only through navigation fine-tuning, our only action domain.

\paragraph{\intphysx.}
The in-context scenarios bind arbitrary sounds to actions within the context clip, so by design the whole
sequence, not only the binding, lies outside any training distribution. Solving them requires inferring the binding
from the context and applying it to the continuation, which favors models that can reason over the context or reflect
on their own answers. Surprise-based evaluation is therefore inherently challenging: a model's surprise reflects the
novelty of the scene as much as the consistency between the query and an option.

\paragraph{\avnavbench evaluation.}
\avnavbench renders houses and their binaural acoustics in simulation, which enables controlled, reproducible
comparisons; transfer to real-world navigation remains untested. Its episodes test image and sound goals in isolation,
each from one static source. Distant goals in other rooms, which lie beyond the current view and the planning
horizon, remain challenging (\mainref{sec:avnavresults}).

\subsection{Future Work}
\label{sec:suppl_future}

\paragraph{\avjepa.}
Our speech experiments (\cref{sec:suppl_speech}) are preliminary. Making language a core component of
the model, and learning representations that serve both general audio and speech, are promising directions. Encoding
spatial audio from stereo, binaural or multichannel recordings would let the model use audio for spatial understanding,
including the Localization scenario of \intphysx and \avnavbench.

\paragraph{World model.}
Current limitations motivate several future explorations. On the model side, a stochastic or generative latent
predictor could represent uncertain outcomes and provide likelihoods for surprise-based evaluation, and training the
encoder and the world model jointly, likely with a chunk-causal encoder, may help both; further scaling could show
whether better prediction carries over to planning. On the action side, \avwm is trained largely on action-free video:
extracting actions for these data, \eg latent actions or camera poses, and studying how training with them generalizes
is a promising direction, including for tasks beyond navigation such as manipulation, where sounds often signal
contact. For planning, the horizon heads already match multi-step rollout prediction in one forward pass
(\cref{sec:supplavwmablations}); we hypothesize that, conditioned on actions, they could reduce the compute of planning
and help plan across longer distances.

\paragraph{\intphysx.}
Our benchmark remains unsolved: every model stays well below humans, and we do not yet understand why even very strong
models struggle on scenarios that humans solve easily. Analyzing which cues models rely on and where they fail is an
open direction that the benchmark's controlled minimal pairs make measurable.

\paragraph{\avnavbench.}
Moving or competing sources and combined image and sound goals would be valuable additions. Complementing local
world-model planning with an exploration policy that seeks informative observations and sub-goals is a promising
direction for distant goals.

\section{Extended Related Work}
\label{sec:suppl_related}

This section extends the related work of the main paper (\mainref{sec:related}) with audio-visual SSL (\cref{sec:suppl_rw_avssl}), 
JEPAs (\cref{sec:suppl_rw_jepa}), discrete and prototype-based targets (\cref{sec:suppl_rw_proto}), 
intuitive physics (\cref{sec:suppl_rw_physics}) and world models (\cref{sec:suppl_rw_wm}).

\subsection{Audio-visual self-supervised learning}
\label{sec:suppl_rw_avssl}

Audio-visual SSL uses the correspondence between sight and sound as supervision.
Early pretext tasks distilled visual predictions into an audio network~\cite{aytar2016soundnet} or relied on audio-visual correspondence~\cite{arandjelovic2017look,arandjelovic2018objects,harwath2018jointly}, sound separation~\cite{zhao2018soundpixels}, sound prediction from silent video~\cite{owens2016visually} and synchronization~\cite{chung2016outoftime,owens2018audio,korbar2018cooperative}.
Later methods use contrastive~\cite{morgado2021avid,alayrac2020mmv,kim2024equiav,sarkar2023crisscross}, clustering~\cite{alwassel2020xdc,asano2020selavi} and masked-modeling objectives~\cite{gong2023contrastive,huang2023mavil,georgescu2023avmae,sarkar2022xkd,su2024vab}, the latter building on masked autoencoding~\cite{he2022masked} and often adding a contrastive loss.
ImageBind~\cite{girdhar2023imagebind} and PE-AV~\cite{vyas2025pushing} further combine contrastive learning with language priors or supervision.
Most of these methods use separate encoders and relate the modalities through pooled clip-level embeddings, capturing \emph{what} happens in the other modality but not \emph{when}.

Finer alignment has mostly been pursued for specific tasks: \cavmaesync~\cite{araujo2025cavmaesync} and Synchformer~\cite{iashin2024synchformer} contrast short audio and visual segments from separate encoders, TIM~\cite{chalk2024tim} queries videos with time intervals, and Mirasol3B~\cite{piergiovanni2024mirasol} processes
time-aligned audio-video chunks, a design that \avwm follows (\mainref{sec:avwm}).
Existing joint encoders, which use shared backbones~\cite{akbari2021vatt,lee2021avbert,lin2024avsiam,teotia2026mjepa}, bottleneck attention~\cite{nagrani2021mbt} or latent cross-attention~\cite{jaegle2021perceiver}, nearly all encode audio as spectrograms and do not align the modalities on a common time grid.
In contrast, \avjepa feeds raw waveforms and video to a single encoder on a common 20\,ms time grid and predicts the other modality's teacher representation for every 500\,ms chunk, capturing both \emph{what} happens and \emph{when}.

\subsection{Joint-embedding predictive architectures}
\label{sec:suppl_rw_jepa}

JEPAs~\cite{lecun2022path} predict the representation of masked content from the visible context instead of reconstructing the input.
I-JEPA~\cite{assran2023selfsupervised} predicts the EMA-teacher features of masked image blocks; \vjepa~\cite{bardes2024revisiting} extends it to video with spatio-temporal masks, \vjepatwo~\cite{assran2025vjepa2} scales the model and adds action-conditioned planning, and \vjepatwoone~\cite{murlabadia2026vjepa21} adds dense multi-level targets.
data2vec~\cite{baevski2022data2vec,baevski2023efficient} applies a similar recipe to speech, vision and language, and SALT~\cite{li2025salt} replaces the EMA teacher with a frozen pixel-reconstruction model.
JEPAs have also been applied to other modalities, including audio~\cite{fei2024ajepa}, speech~\cite{ioannides2026sjepa}, vision--language~\cite{chen2026vl} and medical imaging~\cite{radhachandran2026us, yang2026clin}.

Closest to ours, \mjepa~\cite{teotia2026mjepa} encodes video and spectrograms with a single shared encoder and regresses EMA-teacher features with an $\ell_1$ loss: multi-level features of masked tokens within a modality, but one mean-pooled clip-level embedding across modalities.
Concurrently, Robson~\etal~\cite{robson2026leavjepa} train an early-fusion ViT on mel-spectrograms without a teacher, aligning a clip's global embedding with local, modality-specific views.
\avjepa keeps \mjepa's single encoder but replaces clip-level regression with prototype prediction per token and per 500\,ms chunk.

Existing JEPA objectives differ in the loss and in how they avoid collapse.
Most regress continuous teacher features with an $\ell_1$ or $\ell_2$ loss~\cite{assran2023selfsupervised,bardes2024revisiting,baevski2022data2vec} and avoid collapse 
with an asymmetric predictor and an EMA teacher~\cite{grill2020byol,chen2021simsiam}.
Others constrain the embedding statistics: C-JEPA~\cite{mo2024cjepa} and MC-JEPA~\cite{bardes2023mcjepa} add VICReg~\cite{bardes2022vicreg} variance and covariance terms.
LeJEPA~\cite{balestriero2025lejepa} enforces an isotropic Gaussian with SIGReg, removing the need for a teacher, predictor or stop-gradient, an approach extended to video~\cite{kuhn2026levjepa}, audio-visual data~\cite{robson2026leavjepa} and world modeling~\cite{maes2026leworldmodel}.
A third group replaces regression altogether: VL-JEPA~\cite{chen2026vl} predicts text embeddings with a contrastive loss,
MATPAC~\cite{quelennec2025matpac,quelennec2025matpacpp} adds a teacher--student classification task,
Discrete-JEPA~\cite{baek2025discretejepa} learns discrete semantic tokens,
and S-JEPA~\cite{ioannides2026sjepa}, the closest to our \protoloss, matches the soft posteriors of a Gaussian mixture at masked speech positions with a KL divergence.
\avjepa instead matches the teacher's soft assignments to learned prototypes, a distribution that can represent several plausible targets where regression commits to one, and two entropy terms complement the EMA teacher in preventing collapse.

\subsection{Discrete and prototype-based targets}
\label{sec:suppl_rw_proto}

Predicting cluster or prototype assignments is a long-standing alternative to feature regression in image SSL, from
$k$-means pseudo-labels~\cite{caron2018deepcluster} and Sinkhorn-balanced assignments~\cite{asano2020sela,caron2020swav}
to matching a teacher's softmax over learned prototypes~\cite{caron2021emerging,zhou2022ibot,assran2022msn,oquab2023dinov2,simeoni2025dinov3,darcet2025capi}
and predicting the codes of a pretrained tokenizer~\cite{bao2022beit}.
Discrete targets are also common in
speech~\cite{baevski2020wav2vec,hsu2021hubert,chen2022wavlm,qian2022contentvec,shi2024multiresolution,liu2023dinosr,poli2025spidr},
general audio~\cite{chen2023beats,li2024mert,zhu2025muq} and audio-visual learning~\cite{alwassel2020xdc,asano2020selavi,shi2022avhubert}.
Concurrently to our work, JEM~\cite{seitzer2026scalable} and GSLM2~\cite{higuchi2026gslm} derive the same objective as \protoloss for image and speech SSL, balancing prototype targets with two entropy terms inspired by IMSAT~\cite{hu2017imsat}.

\subsection{Intuitive physics and violation of expectation}
\label{sec:suppl_rw_physics}

Infants look longer at events that break physical laws~\cite{baillargeon1985object,spelke1992origins} and relate sounds
to the events that produce them~\cite{spelke1979bimodal,bahrick2000intersensory}, and adults notice audio-visual
asynchrony~\cite{dixon1980detection}; violation of expectation (VoE) turns such surprise into a test of physical knowledge.
Video benchmarks bring VoE to rendered and real
scenes~\cite{riochet2022intphys,bordes2025intphys2,jassim2024grasp,weihs2022inflevel} (\cref{tab:intphys_benchmarks}),
alongside object-centric VoE models~\cite{smith2019adept,piloto2022plato} and outcome-prediction or question-answering
benchmarks~\cite{bear2021physion,yi2020clevrer,motamed2025physicsiq}.
Under VoE, intuitive physics emerges in video JEPAs, whereas pixel-generative models and multimodal LLMs stay close to
chance~\cite{garrido2025intuitive}; however, all of these benchmarks---summarized in \cref{tab:intphys_benchmarks}--- are visual only.

Sound carries physical evidence too: impact sounds reveal shape and material~\cite{zhang2017shape,owens2016visually},
multisensory object datasets support learning such
properties~\cite{gao2021objectfolder,gao2022objectfolder2,gao2023objectfolderbench,clarke2023realimpact}, and sound helps
to count repetitive actions~\cite{zhang2021repetitive}.
Audio-visual benchmarks probe physical or acoustic knowledge through question
answering~\cite{yu2022pacs,patraucean2023perception,li2022musicavqa,gong2024avodyssey}, or test whether generated sound
obeys physics~\cite{xie2025phyavbench,li2026flatsounds,li2026acoustitrace}.
\intphysx instead poses VoE minimal pairs that require both modalities without a language interface, so that self-supervised encoders, world models and multimodal
LLMs~\cite{team2023gemini,xu2025qwen25omni} are scored on the same pairs.

\begin{table}[t]
  \centering
  \providecommand{\ipxbRR}{\raggedright\arraybackslash}%
  \caption{\textbf{Intuitive-physics benchmarks.} Violation-of-expectation (VoE)    benchmarks are video-only, and audio-visual physics benchmarks ask questions  instead of testing expectations.
  	IntPhys-X tests cross-modal physical expectation. Real: contains real (non-rendered) recordings.
  	Not listed: benchmarks that test whether \emph{generated} sound obeys
    physics~\cite{xie2025phyavbench,li2026flatsounds,li2026acoustitrace}.}
  \label{tab:intphys_benchmarks}
  \footnotesize
  \setlength{\tabcolsep}{2.6pt}
  \begin{tabular}{@{}l c c c >{\ipxbRR}p{1.42in}@{}}
    \toprule
    Benchmark & Audio & Real & VoE & Properties tested \\
    \midrule
    IntPhys-v1~\cite{riochet2022intphys} & \xmark & \xmark & \cmark
      & Permanence, shape constancy, spatio-temporal continuity \\
    IntPhys-v2~\cite{bordes2025intphys2} & \xmark & \xmark & \cmark
      & Permanence, immutability, spatio-temporal continuity \\
    GRASP~\cite{jassim2024grasp} & \xmark & \xmark & \cmark
      & Eight principles, e.g., continuity, solidity, gravity \\
    InfLevel~\cite{weihs2022inflevel} & \xmark & \cmark & \cmark
      & Continuity, solidity, gravity \\
    PACS~\cite{yu2022pacs} & \cmark & \cmark & \xmark
      & Physical commonsense attributes of objects (QA) \\
    Perception Test~\cite{patraucean2023perception} & \cmark & \cmark & \xmark
      & Physics as one of four skill areas (QA, tracking) \\
    \midrule
    \oursrow IntPhys-X (ours) & \cmark & \cmark & \cmark
      & Counting, synchronization, size, localization, material \\
    \bottomrule
  \end{tabular}
	\vspace{-0.2cm}
\end{table}

\subsection{World models for planning and navigation}
\label{sec:suppl_rw_wm}

World models predict how an environment evolves, passively or conditioned on actions~\cite{ha2018worldmodels,lecun2022path}.
Agents plan or learn behavior inside learned latent dynamics~\cite{hafner2019planet,hafner2020dreamer,hafner2025dreamerv3,schrittwieser2020muzero,hansen2024tdmpc2},
robots plan by predicting the video of candidate actions~\cite{finn2017foresight}, and generative models simulate
interactive environments in pixels~\cite{yang2024unisim,bruce2024genie,alonso2024diamond}.
Without action labels, video can still train such models by inferring actions~\cite{baker2022vpt}, learning latent actions~\cite{bruce2024genie},
or pretraining action-free dynamics that are later conditioned on actions~\cite{seo2022apv,baldassarre2025dinoworld}, as
in \avwm.

For navigation, Pathdreamer~\cite{koh2021pathdreamer} generates future indoor views to guide vision-and-language
navigation, and NWM~\cite{bar2025nwm} plans by predicting the future frames of candidate trajectories with a conditional
diffusion transformer (CDiT), a model that \vjepatwoone~\cite{murlabadia2026vjepa21} trains on its own features.
AV-CDiT~\cite{wang2025avwm} adds binaural audio and uses its model as a lookahead for a pretrained audio-visual
navigation policy. DINO-WM~\cite{zhou2024dinowm}, DINO-world~\cite{baldassarre2025dinoworld} and
\vjepatwo-AC~\cite{assran2025vjepa2} plan in the feature space of a frozen encoder, and PLDM~\cite{sobal2026learning}
plans with latent dynamics learned from reward-free offline trajectories. \avwm brings frozen-feature planning to audio
and video (\mainref{sec:avwm}); we compare it with NWM and AV-CDiT reimplemented on the same frozen latents.

Navigation policies instead map observations and an optional goal image directly to
actions~\cite{shah2023gnm,shah2023vint,sridhar2024nomad}; NWM is trained on their robot
datasets~\cite{shah2021recon,karnan2022scand,triest2022tartandrive,hirose2023sacson}, whose test splits we also plan on
(\cref{sec:suppl_avwm_nwm_eval}). Audio-visual navigation agents, mostly trained with reinforcement learning in
simulators that render spatial audio~\cite{chen2022soundspaces2,gao2023sonicverse}, find sounding targets in 3D
scenes~\cite{gan2020look,chen2020soundspaces,chen2021avwan}, including targets that stop
sounding~\cite{chen2021savi,zeng2026savnce}, move~\cite{younes2023catch} or compete with other
sounds~\cite{yu2022saavn,shi2025enmus}; some agents ask for language help~\cite{paul2022avlen,liu2024caven} or pursue
long-horizon missions~\cite{chen2026lhavln}. In \avnavbench, all models instead learn from the same fixed walks, and
the world model itself plans toward image or sound goals without a policy (\cref{sec:suppl_avnav_prior}).

\section{\avjepa Implementation Details}
\label{sec:suppl_setup}

We describe the training data, architecture, pretraining recipes and frozen-encoder evaluations used by \avjepa.
Models are pretrained on AudioSet-2M~\cite{gemmeke2017audio} (audio-visual), VideoMix2M~\cite{bardes2024revisiting} (video only), or both.

\subsection{Pretraining datasets}
\label{sec:suppl_data}

\paragraph{AudioSet-2M.}
We use the 1.81M available audio-video clips from the unbalanced training split of AudioSet~\cite{gemmeke2017audio}.
For pretraining, we select 16 frames sampled at 4\,fps from a random position in the clip, along with the co-occurring 4\,s audio waveform.
Frames share a random-resized crop to 224\,px and a random horizontal flip.
Audio is resampled to 16\,kHz mono, without normalization or augmentation.

\paragraph{VideoMix2M.}
Following \vjepa~\cite{bardes2024revisiting}, we create VideoMix2M by combining Kinetics-710~\cite{kay2017kinetics},
SSv2~\cite{goyal2017something} and HowTo100M~\cite{miech2019howto100m}, sampled at train time with probabilities 0.335, 0.100 and 0.565.
We use the same 16-frame sampling and spatial augmentations as for AudioSet-2M.

\subsection{Architecture}
\label{sec:suppl_arch}

\paragraph{Encoders and tokenization.}
All encoders are pre-LN ViTs~\cite{dosovitskiy2021vit} of different encoder size, ranging from \emph{base} to \emph{giant} scale (\cref{tab:model_scales}).
Videos are tokenized into $2{\times}16{\times}16$ video tubelets~\cite{bardes2024revisiting}.
Audio is tokenized with a convolutional front-end, following wav2vec~2.0 and SpidR~\cite{baevski2020wav2vec,poli2025spidr},
producing one token every 20\,ms (200 tokens per 4\,s).
Learned additive modality embeddings distinguish audio and video.
Audio-visual models use M-RoPE~\cite{su2024roformer, xu2025qwen25omni} to augment each token with temporal and spatial information, allocating half of each head's rotary dimensions to time and a quarter to each spatial axis; video-only models use 3D RoPE~\cite{wang2024qwen2vl}.

\begin{table}[t]
  \centering
  \caption{
  	\textbf{\avjepa hyperparameters, and encoder and predictor configurations.}
    Encoder parameter counts include tokenizers but exclude predictors, codebooks and optional attentive poolers.
  }
  \vspace{-0.2cm}
  \label{tab:model_scales}
  \fontsize{7.5}{9}\selectfont
  \setlength{\tabcolsep}{2pt}
  \begin{tabular}{
  		@{}p{0.34\columnwidth}
  		@{\hspace{6pt}}
  		*{4}{>{\centering\arraybackslash}p{\dimexpr(0.66\columnwidth-6pt-6\tabcolsep)/4\relax}}
  		@{}
  	}
    \toprule
    Configuration & ViT-B & ViT-L & ViT-H & ViT-g \\
    \midrule
    \multicolumn{5}{@{}l@{}}{\cellcolor{black!8}\textbf{Encoder (A, V and AV models)}} \\
    Blocks & 12 & 24 & 32 & 40 \\
    Width $D$ & 768 & 1024 & 1280 & 1408 \\
    \# Attn. heads & 12 & 16 & 16 & 22 \\
    FFN width & 3072 & 4096 & 5120 & 6144 \\
    Parameters & 90.8M & 308.6M & 636.5M & 1.02B \\
    \midrule
    \multicolumn{5}{@{}l@{}}{\cellcolor{black!8}\textbf{Token transformer ($\Pred$; A, V and AV models)}} \\
    Depth (AS) & {12} & {12} & {12} & {24} \\
    Width / \# Attn. heads & \multicolumn{4}{c@{}}{384 / 12} \\
    FFN width & 1536 & 1536 & 1536 & 1675 \\
    Input projection & \multicolumn{4}{c@{}}{affine LN  $+ \; \textrm{Linear}(D\!\rightarrow\!384)$} \\
    Output projection & \multicolumn{4}{c@{}}{affine LN  $+ \; \textrm{Linear}(384\!\rightarrow\!D)$} \\
    Sharing & \multicolumn{4}{c@{}}{one predictor for the three AV token-level terms.} \\
    \midrule
    \multicolumn{5}{@{}l@{}}{\cellcolor{black!8}\textbf{Chunk MLP ($\ChunkPredSimple$; AV models only)}} \\
 	Architecture & \multicolumn{4}{c@{}}{affine LN $+ \, \textrm{MLP}(D\!\rightarrow\!2D\!\rightarrow\!2D\!\rightarrow\!D)$} \\
    Hidden width $2D$ & 1536 & 2048 & 2560 & 2816 \\
    Parameters & 4.73M & 8.40M & 13.12M & 15.88M \\
    Activation & \multicolumn{4}{c@{}}{GELU on hidden layers} \\    
    Sharing & \multicolumn{4}{c@{}}{one MLP for all cross-modal terms and chunks.} \\
    \midrule
    \multicolumn{5}{@{}l@{}}{\cellcolor{black!8}\textbf{Chunk pooling and targets (AV models only)}} \\
    Chunking & \multicolumn{4}{c@{}}{500\,ms $\rightarrow$ eight chunks per 4\,s clip.} \\    
 	Pooling & \multicolumn{4}{c@{}}{mean or attention pooling}\\    
	Attn.~pooler design & \multicolumn{4}{c@{}}{one cross-attention block with learned query.}\\  
    Attn.~pooler width & 768 & 1024 & 1280 & 1408 \\
    Attn.~pooler heads & 12 & 16 & 16 & 16 \\
	Attn.~pooler sharing & \multicolumn{4}{c@{}}{separate A/V/AV poolers; EMA-copies in teacher.}\\  
    \midrule
    \multicolumn{5}{@{}l@{}}{\cellcolor{black!8}\textbf{Codebooks and \protoloss}} \\
    \# Prototypes (A/V/Shared) & \multicolumn{4}{c@{}}{4096 \, / \, 4096 \, / \, 1024} \\
    Student temp. $\taus$ & \multicolumn{4}{c@{}}{1} \\
    Teacher temp. $\taut$ & \multicolumn{4}{c@{}}{$1/6$ (AV); $1/10$ (video-only)} \\
    Prediction weight & \multicolumn{4}{c@{}}{1} \\
    Entropy weight $\lamH$ & \multicolumn{4}{c@{}}{0.5} \\
    MEMax weight $\lamME$ & \multicolumn{4}{c@{}}{10} \\
    \bottomrule
  \end{tabular}
  \vspace{-0.2cm}
\end{table}

\paragraph{Predictors and targets.}
\Cref{tab:model_scales} consolidates the token predictor ($\Pred$), chunk predictor ($\ChunkPredSimple$), pooling and codebook configurations.
All models use 12-layer token-level predictors, except the ViT-g models, which profit from the extra capacity provided by a more expressive, 24-layer predictor.
These token-level predictors process visible context and masked-position queries for intra-modal predictions, with shared weights for audio, video and joint inputs.
Cross-modal prediction instead pools features into 500\,ms time-aligned chunks and applies one target-conditioned MLP independently to each chunk, without cross-chunk attention.
In some experiments (\emph{+ AttnPool}), we replace mean pooling with learned per-chunk attention; teacher poolers are EMA copies of the student poolers.

\paragraph{Codebooks and prototypes.}
Each codebook is a learnable $K{\times}D$ prototype matrix, implemented as a bias-free $1{\times}1$ convolution, with $K=4096$ for the audio and video codebooks, and $K=1024$ for the shared cross-modal chunk targets.
Before projection, we apply affine-free LayerNorm to both teacher targets and student predictions, and $\ell_2$-normalize each prototype to unit norm.
The same prototype matrix projects both student and teacher tokens to $K$ logits, followed by non-affine batch normalization and temperature-scaled softmax.

\subsection{Pretraining recipes}
\label{sec:suppl_recipe}

\paragraph{Objective and masking.}
Audio-visual pretraining sums three token-level terms (\texttt{a2a}, \texttt{v2v}, \texttt{av2av}) and four chunk-level terms (\texttt{a2v}, \texttt{v2a}, \texttt{av2a}, \texttt{av2v}).
The student and the teacher process each clip in three input modes: audio-only, video-only and audio-visual.
Video masking follows \vjepa~\cite{bardes2024revisiting}: eight spatial blocks
covering 15\% of a frame each, extended through time (about 63\% masked tokens).
Audio uses overlapping 200\,ms spans from 16 random starts (about 57\% masked tokens).
Video-only training averages losses over this short-range mask and a long-range mask of two 70\%-area blocks (about 82\% masked tokens).

In our \protoloss, we define the student temperature as $\taus = 1$, the teacher temperature as $\taut = 1/6$ for audio-visual and $\taut = 1/10$ for video-only training.
Sharpening the teacher's prototype assignments improves downstream performance (\cref{tab:ablations_combined_protoloss}).

Conditional-entropy and MEMax weights are set to $\lamH = 0.5$ and $\lamME = 10$, respectively.
The entropy regularizers $\LH$ and $\LME$ act on detached teacher features and update only the codebooks, while the student encoder is updated solely via the prototype prediction objective $\LKL$. Ablations in \cref{tab:ablations_combined_protoloss} show that both regularizers are important for strong performance.

\paragraph{Combined pretraining.}
We combine paired audio-visual data with video-only data to learn multi-modal representations and cross-modal grounding alongside broader visual and motion representations.
AudioSet supplies the paired signal, while VideoMix2M expands visual coverage without requiring aligned audio; both streams train the same shared encoder.
For this mixed-dataset pretraining, separate GPU groups process audio-visual and video-only examples, sharing all parameters and one optimizer step.
The overall objective is $\mathcal{L}_{\mathrm{AS}}+w_{\mathrm{V}}\mathcal{L}_{\mathrm{VM}}$, with each stream's loss averaged separately, and a weight of $w_{\mathrm{V}} = 10$.
Codebook batch statistics and MEMax are computed within each stream.

\paragraph{Optimization.}
Recipes are mostly shared across encoder scales.
AudioSet-2M models use AdamW~\cite{loshchilov2019decoupled},
$\beta=(0.9,0.995)$, weight decay 0.04 and gradient clipping at 3.
The learning rate warms up over 35k steps to $5.25{\times}10^{-4}$ and then stays constant; training lasts 270k steps with batch 448 (121M clips) across 16 GPUs.
Teacher EMA momentum ramps from 0.999, reaching 0.99995 at the last step.
VideoMix2M models use $\beta=(0.9,0.999)$, the same weight decay, a 12k-step warm-up to $3{\times}10^{-4}$, and 360k steps with batch 512 (184M clips) across 16 GPUs.
Combined pretraining uses the same AdamW settings and learning-rate schedule as AudioSet-only training, with batches of 448 audio-visual and 1344 video clips per update for 240k steps.
Teacher EMA momentum is fixed at 0.99925.
In all cases, we use bf16 computation with fp32 weights.
For efficiency purposes, we run most ablation and analysis experiments for 180k optimization steps.

\subsection{Baselines}
\label{sec:suppl_baselines}

\paragraph{M-JEPA.}
We reproduce \mjepa~\cite{teotia2026mjepa} following the official recipe\footnote{\url{https://github.com/facebookresearch/MJEPA}},
and using the same AudioSet-2M clips, video preprocessing, masking and evaluation protocol as \avjepa.
Its audio input is a 10\,s clip, represented as a 128-bin log-mel spectrogram with $16{\times}16$ patches.
The video input is a 4\,s window from the same recording, comprising 16 frames sampled at approximately 4\,fps and cropped to $224{\times}224$ pixels.
Video frames are embedded as $2{\times}16{\times}16$ tubelets, and both modalities use sinusoidal positional embeddings.
M-JEPA regresses teacher features with an $\ell_1$ loss over nine prediction terms, using multi-level token targets and clip-mean cross-modal targets with separate MLP heads.
Its cosine learning-rate schedule follows the official implementation, with peak $6.25{\times}10^{-4}$.

\paragraph{V-JEPA 2 and V-JEPA 2.1.}
For a fair data- and recipe-matched comparison, we reproduce \vjepatwo~\cite{assran2025vjepa2} and \vjepatwoone~\cite{murlabadia2026vjepa21}  following the official implementation\footnote{\url{https://github.com/facebookresearch/vjepa2}}, using the same VideoMix2M data, encoder configurations and optimization recipe as our video-only \avjepa models.
Unlike our prototype-based prediction, \vjepatwo regresses normalized teacher features at masked positions with an $\ell_1$ loss, whereas \vjepatwoone additionally uses multi-level teacher targets and a context loss on visible tokens.

\subsection{Evaluation protocols}
\label{sec:suppl_eval}

Unless stated otherwise, evaluations use the frozen EMA teacher's normalized last-layer tokens.
Classification uses the attentive probing protocol of \vjepatwo~\cite{assran2025vjepa2}: three self-attention blocks and one cross-attention block with a learned query, followed by a linear classifier.
Only the probe is trained, using AdamW, cosine learning rate decay, and a sweep of learning rates and weight decay values.
For multi-label tasks, we train with the binary cross-entropy loss and report mAP, whereas for single-label tasks we use cross-entropy and measure the top-1 accuracy.
Results and ablations are averaged over two pretraining seeds.
For each seed, we select one encoder checkpoint for the video suite, based on SSv2 accuracy, and one for the audio and audio-visual suites, based on AudioSet-20K audio (A) mAP; each checkpoint is then kept fixed
across all tasks of the corresponding evaluation suite.
We evaluate our reproduced baselines with the same protocol, using the same probe hyperparameter sweep and checkpoint selection as for \avjepa.

\paragraph{AudioSet-20K classification and audio transfer.}
Following the standard protocol, we train our evaluation probes on AudioSet's balanced split (20.7k clips)
and evaluate on 19.0k clips over 527 sound classes~\cite{gemmeke2017audio}, with audio (A), video (V), or a joint
input (A-V).
For audio, we use the full 10\,s waveform, whereas for video we concatenate tokens from three 16-frame segments at 4\,fps and 224\,px.
A-V uses aligned 4\,s windows at 224\,px: one random window for training and three windows with averaged predictions for evaluation.
Audio and video probes train for 50 epochs, while the A-V results use M-JEPA's 20-epoch probe recipe with label mixup and probe EMA.
The audio probe uses input mixup, whereas the video probe uses spatial augmentations.
To evaluate how well the audio representations transfer beyond AudioSet, we apply the same frozen attentive probing to FSD50K~\cite{fonseca2022fsd50k} (200 sound classes, mAP) and ESC-50~\cite{piczak2015esc} (50 sound classes, mean accuracy).

\paragraph{Image and action recognition.}
We use frozen attentive probes on Kinetics-400~\cite{kay2017kinetics}, SSv2~\cite{goyal2017something}, COIN~\cite{tang2019coin} and Diving-48~\cite{li2018resound}.
Following \vjepatwo's protocol~\cite{assran2025vjepa2}, inputs are 256\,px,
with 16 frames for Kinetics-400 and SSv2, and 32 for COIN and Diving-48.
Evaluation uses 8, 2, 8 and 4 temporal segments, respectively, each with three spatial crops.
Tokens from each segment are concatenated and crop predictions averaged.
Diving-48 concatenates features from the last, third-, fifth- and seventh-to-last encoder blocks.
We additionally evaluate on ImageNet-1K~\cite{deng2009imagenet} with an attentive probe at 256\,px, presenting each image as a two-frame clip.
All five tasks report top-1 classification accuracy.

\paragraph{Sound-prompted segmentation.}
We follow DenseAV's ADE20K-Sounds protocol~\cite{hamilton2024separating}, which provides 106 image-sound pairs over 20 classes, and tests cross-modal grounding and localization.
The sound is the query and the image's sounding-object mask is the target.
Without training a segmentation head, we form a spatial map by cosine similarity between a pooled audio representation and image-patch features.
Before computing cosine similarities, we standardize video and audio embeddings by removing the four leading principal components from each modality.
We fit both PCA bases on 85 unlabeled development clips that are not part of ADE20K-Sounds, and fix the number of removed components on these clips, without tuning on ADE20K-Sounds.
For baselines, we re-evaluate them using their standard readouts when checkpoints are available, whereas we use published results otherwise.
We score the maps with class-averaged pixel AP (mAP) and foreground IoU (mIoU).

\paragraph{Cross-modal retrieval.}
We follow CAV-MAE's AudioSet-retrieval evaluation subset~\cite{gong2023contrastive}, composed of 1718 audio-video pairs.
Each clip supplies 10\,s of audio and 16 frames spanning the same interval.
Audio and video are encoded separately.
\avjepa ranks mean-centered, pooled encoder representations by cosine similarity, without a trained predictor or retrieval head.
An audio query ranks the full video gallery and vice versa, with the paired clip as the correct match.
We report Recall@1, 5 and 10 in both directions ($\textrm{A}\rightarrow\textrm{V}$ and  $\textrm{V}\rightarrow\textrm{A}$).
The \mjepa baseline instead maps the query through its trained cross-modal
predictor before comparing with the other modality's pooled features~\cite{teotia2026mjepa}.

\section{Additional Audio-Visual Results}
\label{sec:suppl_av_results}

\subsection{Additional Results}

\providecommand{\vjepatworep}{\mbox{V-JEPA}~2\textsuperscript{rep.}\xspace}
\providecommand{\vjepatwoonerep}{\mbox{V-JEPA}~2.1\textsuperscript{rep.}\xspace}
\providecommand{\tabdashrule}[1]{\noalign{\vskip 1.5pt}\multispan{#1}{\color{black!40}\xleaders\hbox to 3.2pt{\hss\vrule
  height 0.25pt depth 0.25pt width 1.8pt\hss}\hskip 0pt plus 1fill\relax}\cr\noalign{\vskip 1.5pt}}
\providecommand{\suppvideosizeheader}[2]{%
  \rowcolor{black!8}%
  \multicolumn{#1}{@{}l@{}}{\rule[-4pt]{0pt}{14pt}\hspace{3pt}\textbf{#2}}\\}
\begin{table}[t]
  \centering
  \caption{
    \textbf{Image classification and video action recognition} with video-only pretraining: top-1 accuracy (\%) of frozen encoders with attentive probes
    on \mbox{ImageNet-1K}~\cite{deng2009imagenet}, Kinetics-400~\cite{kay2017kinetics},
    COIN~\cite{tang2019coin}, SSv2~\cite{goyal2017something} and Diving-48~\cite{li2018resound}.
    \textsuperscript{rep.}: V-JEPA~2/2.1 reproductions, trained on the same data as \avjepa.
    \avjepa and \textsuperscript{rep.} results average two seeds.
    Published results follow their original protocols or are reported by~\cite{assran2025vjepa2}.
    \textbf{Bold}/\underline{underline}: best/second per encoder size.
    Rows below the dashed rule use more pretraining data and are not ranked.
	\avjepa outperforms matched baselines on K400, COIN and SSv2 at every encoder size, and narrows the gap to V-JEPA variants trained on $>10\times$ more data.
  }
  \vspace{-0.2cm}
  \label{tab:cls_by_scale_final}
  \footnotesize
  \setlength{\tabcolsep}{2.5pt}
  \begin{tabular}{@{}l l c c c c c@{}}
    \toprule
    & & \multicolumn{3}{c}{Appearance} & \multicolumn{2}{c}{Motion} \\
    \cmidrule(lr){3-5} \cmidrule(l){6-7}
    Method & Data & IN1K & K400 & COIN & SSv2 & D48 \\
    \midrule
    \noalign{\vskip -\belowrulesep}%
    \suppvideosizeheader{7}{ViT-B\quad(86M)}
    \vjepatworep & VM2M & \second{75.2} & \second{78.1} & \second{78.6} & \second{65.9} & 85.6 \\
    \vjepatwoonerep & VM2M & 74.5 & 76.4 & 78.5 & 64.8 & \second{85.9} \\
    \oursrow \avjepa & VM2M & \best{76.4} & \best{79.5} & \best{79.2} & \best{68.2} & \best{87.8} \\
    \tabdashrule{7}
    SALT~\cite{li2025salt} & V-3.6M & 74.8 & 70.9 & 80.5 & 66.1 & 78.7 \\
    \suppvideosizeheader{7}{ViT-L\quad(304M)}
    VideoMAE~\cite{tong2022videomae} & K400/SSv2 & 71.1 & 77.8 & \na & 65.5 & \na \\
    OmniMAE~\cite{girdhar2023omnimae} & IN+SSv2 & 75.1 & 65.6 & \na & 60.6 & \na \\
    Hiera-L~\cite{ryali2023hiera} & K400 & 68.9 & 75.5 & \na & 64.2 & \na \\
    MVD~\cite{wang2023masked} & IN+K400 & 73.3 & 79.4 & \na & 66.5 & \na \\
    \vjepa~\cite{bardes2024revisiting} & VM2M & 74.8 & 80.8 & \na & 69.5 & \na \\
    InternVideo~\cite{wang2022internvideo} & UH1.35M & \na & 78.6 & \na & 67.4 & 69.6 \\
    \vjepatworep & VM2M & 77.4 & 82.1 & 81.9 & 71.6 & \best{89.1} \\
    \vjepatwoonerep & VM2M & \second{78.5} & \second{82.2} & \second{85.2} & \second{71.8} & 87.5 \\
    \oursrow \avjepa & VM2M & \best{78.7} & \best{83.9} & \best{86.6} & \best{73.5} & \second{88.1} \\
    \tabdashrule{7}
    SALT~\cite{li2025salt} & V-3.6M & 79.0 & 85.4 & 85.3 & 74.9 & 82.5 \\
    \vjepatwo~\cite{assran2025vjepa2} & VM22M & 83.5 & 85.1 & 86.8 & 73.7 & 89.0 \\
    \vjepatwoone~\cite{murlabadia2026vjepa21} & VM163M & 81.8 & 84.3 & \na & 74.2 & \na \\
    \suppvideosizeheader{7}{ViT-H\quad(632M)}
    VideoMAE~\cite{tong2022videomae} & K400 & 72.3 & 79.8 & \na & 66.2 & \na \\
    OmniMAE~\cite{girdhar2023omnimae} & IN+SSv2 & 76.3 & 71.4 & \na & 65.4 & \na \\
    Hiera-H~\cite{ryali2023hiera} & K400 & 71.4 & 77.0 & \na & 64.7 & \na \\
    \vjepa~\cite{bardes2024revisiting} & VM2M & 75.9 & 82.0 & \na & 71.4 & \na \\
    \vjepatworep & VM2M & 78.0 & 82.5 & 83.4 & 71.9 & 87.7 \\
    \vjepatwoonerep & VM2M & \best{79.7} & \second{83.8} & \second{86.4} & \second{72.9} & \best{88.6} \\
    \oursrow \avjepa & VM2M & \second{79.0} & \best{84.4} & \best{87.0} & \best{74.2} & \second{88.4} \\
    \tabdashrule{7}
    SALT~\cite{li2025salt} & V-3.6M & 79.6 & 86.0 & 87.0 & 75.4 & 86.4 \\
    \vjepatwo~\cite{assran2025vjepa2} & VM22M & 83.8 & 85.3 & 87.9 & 74.0 & 89.8 \\
    \suppvideosizeheader{7}{ViT-g\quad(1B)}
    VideoMAE~V2~\cite{wang2023videomae} & UH1.35M & 71.4 & 71.2 & \na & 61.2 & \na \\
    \mjepa~\cite{teotia2026mjepa} & AS+VM2M & \na & \underline{85.0} & \na & \underline{73.9} & \na \\
    \vjepatworep & VM2M & 77.7 & {84.1} & 83.9 & 72.1 & \best{89.9} \\
    \vjepatwoonerep & VM2M & \second{79.4} & 83.1 & 85.5 & {73.2} & 88.3 \\
    \oursrow \avjepa & VM2M & \best{79.6} & \best{85.2} & \best{88.0} & \best{75.2} & \second{88.5}  \\
    \tabdashrule{7}
    SALT~\cite{li2025salt} & V-3.6M & 79.7 & 86.8 & 87.0 & 76.2 & 85.5 \\
    \vjepatwo~\cite{assran2025vjepa2} & VM22M & 84.6 & 86.6 & 90.7 & 75.3 & 90.1 \\
    \vjepatwoone~\cite{murlabadia2026vjepa21} & VM163M & 84.8 & 87.0 & \na & 76.9 & 89.0 \\
    InternVideo2-1B~\cite{wang2024internvideo2} & IV-25.5M & 85.8 & 89.4 & 93.8 & 69.7 & 86.4 \\
    \bottomrule
  \end{tabular}
  \vspace{-0.3cm}
\end{table}

This section compares frozen \avjepa encoders with a broad set of baselines on visual recognition, sound-prompted segmentation and cross-modal retrieval.

\paragraph{Visual recognition.}
\Cref{tab:cls_by_scale_final} evaluates image classification (IN1K) and video action recognition (K400, COIN, SSv2
and Diving-48) with frozen encoders and attentive probes.
To compare pretraining recipes on equal data, we pretrain \avjepa on video only, using VideoMix2M, the same pretraining dataset as V-JEPA, and our \vjepatwo and \vjepatwoone reproductions.
\avjepa outperforms all data-matched V-JEPA variants on K400, COIN and SSv2 at every encoder size, with the largest gains on motion-centric SSv2 (1.3--2.3 points).
\avjepa also narrows the gap to models pretrained on far more data: \avjepa huge (-H) and giant (-g) match the SSv2 scores of \vjepatwo, which uses 11$\times$ more data (74.2 vs.\ 74.0, and 75.2 vs.\ 75.3).

\paragraph{Sound-prompted segmentation.}
\Cref{tab:ade20k_soundprompted} evaluates spatial AV alignment on ADE20K-Sounds: given a sound, the model
must highlight the sounding object in an image.
For \avjepa, the map is the cosine similarity between the pooled audio token and each image-patch token, after normalizing and preprocessing the encoded features.
Without localization training, both \avjepa encoders outperform all baselines, with \avjepa-B achieving the best mAP and mIoU.
\mjepa, which predicts clip-level rather than chunk-level cross-modal targets, reaches 35.0 mAP at ViT-B, 9.1 points below \avjepa.
We hypothesize that larger models may favor contextual and semantic information over local correspondence, benefiting classification without necessarily improving dense tasks, like sound-prompted localization.

\paragraph{Cross-modal retrieval.}
\Cref{tab:audioset_retrieval} reports zero-shot audio-to-video and video-to-audio retrieval on AudioSet (Recall@1, 5 and 10).
 \avjepa-L ranks first at R@5 and R@10 for audio-to-video retrieval among the small and medium encoder models, ahead of contrastive baselines such as \cavmaesync.
Video-to-audio retrieval is weaker, with \cavmaesync and \mjepa-L outperforming our proposed method.
\avjepa-L outperforms LeAVJEPA, our closest baseline, by over 10 R@1 points in both directions, and still leads by over 5 R@1 points when LeAVJEPA adds a contrastive InfoNCE head.
Scaling \avjepa monotonically improves performance across all retrieval metrics in both $\textrm{A} \rightarrow \textrm{V}$ and $\textrm{V} \rightarrow \textrm{A}$ settings, though still trailing \cavmaesync and \mjepa-L in video-to-audio retrieval.

\begin{table}[t]
  \centering
    \caption{
    \textbf{Sound-prompted segmentation} on ADE20K-Sounds~\cite{hamilton2024separating} (mAP and mIoU, \%).
    We use published results where checkpoints are not available, and evaluate all other models with their official readouts; \avjepa results average two seeds.
    Encoder: visual + audio backbones if different.
    \textcolor{gray}{Gray}: center prior (a fixed Gaussian at the image center, no audio).
    \textbf{Bold}/\underline{underline}: best/second best.
    Without any localization training, both \avjepa encoders outperform all baselines, including dedicated dense audio-visual models.
  }
  \vspace{-0.2cm}
  \label{tab:ade20k_soundprompted}
  \footnotesize
  \setlength{\tabcolsep}{4pt}
  \begin{tabular}{@{}l l c c@{}}
    \toprule
    Method & Encoder & mAP & mIoU \\
    \midrule
    \textcolor{gray}{Center prior} & \textcolor{gray}{\na} & \textcolor{gray}{23.2} & \textcolor{gray}{20.2} \\
    DAVENet~\cite{harwath2018jointly} & VGG16 + CNN & 16.8 & 17.0 \\
    DenseAV~\cite{hamilton2024separating} & ViT-S + HuBERT-L & 37.1 & 26.5 \\
    Ryu et~al.~\cite{ryu2025seeing} & ViT-S + HuBERT-L & 36.7 & 27.0 \\
    TACO~\cite{malard2026taco} & ConvNeXt-L & 35.8 & 27.7 \\
    ImageBind~\cite{girdhar2023imagebind} & ViT-H + ViT-B & 18.3 & 18.1 \\
    \cavmae{}\textsuperscript{Scale++}~\cite{gong2023contrastive} & ViT-B & 25.9 & 22.7 \\
    EquiAV~\cite{kim2024equiav} & ViT-B & 25.9 & 22.9 \\
    \cavmaesync~\cite{araujo2025cavmaesync} & ViT-B & 22.6 & 22.7 \\
    \mjepa~\cite{teotia2026mjepa} & ViT-B & 35.0 & 27.2 \\
    \mjepa~\cite{teotia2026mjepa} & ViT-L & 31.8 & 25.0 \\
    \oursrow \avjepa & ViT-B & \best{44.1} & \best{32.2} \\
    \oursrow \avjepa & ViT-L & \second{42.8} & \second{30.9} \\
    \bottomrule
  \end{tabular}
\end{table}

\providecommand{\rk}[1]{{\scriptsize #1}}
\begin{table}[t]
  \centering
  \caption{
    \textbf{Cross-modal retrieval} on AudioSet~\cite{gong2023contrastive} (Recall@$K$, \%).
    \avjepa averages two seeds.
    Encoder: visual/audio backbones.
    \textcolor{gray}{Gray}: random chance.
    \textbf{Bold}/\underline{underline}: best/second best.
    \avjepa achieves the best audio-to-video R@5 and R@10, and improves with scale in both directions.
  }
  \vspace{-0.2cm}
  \label{tab:audioset_retrieval}
  \footnotesize
  \setlength{\tabcolsep}{2.0pt}
  \begin{tabular}{@{}l l c c c c c c@{}}
    \toprule
    & & \multicolumn{3}{c}{Audio $\to$ Video} & \multicolumn{3}{c}{Video $\to$ Audio} \\
    \cmidrule(lr){3-5} \cmidrule(l){6-8}
    Method & Encoder & \rk{R@1} & \rk{R@5} & \rk{R@10} & \rk{R@1} & \rk{R@5} & \rk{R@10} \\
    \midrule
    \textcolor{gray}{Random} & \textcolor{gray}{\na} & \textcolor{gray}{0.06} & \textcolor{gray}{0.29} & \textcolor{gray}{0.58}
      & \textcolor{gray}{0.06} & \textcolor{gray}{0.29} & \textcolor{gray}{0.58} \\
    \cavmae{}\textsuperscript{Scale++}~\cite{gong2023contrastive} & ViT-B & 17.9 & 39.3 & 50.1 & 19.9 & 40.9 & 50.6 \\
    ImageBind~\cite{girdhar2023imagebind} & ViT-H/B & 20.8 & 42.6 & 51.6 & 22.1 & 43.2 & 52.6 \\
    EquiAV~\cite{kim2024equiav} & ViT-B & 29.5 & 51.7 & 61.5 & 30.3 & 51.9 & 61.9 \\
    \cavmaesync~\cite{araujo2025cavmaesync} & ViT-B & 27.9 & 52.4 & 62.2 & \best{35.2} & \second{58.3} & \second{67.6} \\
    DenseAV~\cite{hamilton2024separating} & ViT-S/HuB. & 27.6 & 49.9 & 58.8 & 26.3 & 49.5 & 58.4 \\
    \mjepa~\cite{teotia2026mjepa} & ViT-B & 25.8 & 50.7 & 60.4 & 27.0 & 51.3 & 60.9 \\
    \mjepa~\cite{teotia2026mjepa} & ViT-L & \best{33.7} & 56.8 & 66.1 & \second{33.8} & \best{59.0} & \best{68.3} \\
    LeAVJEPA~\cite{robson2026leavjepa} & ViT-L & 19.1 & 40.5 & 50.4 & 18.7 & 42.1 & 51.9 \\
    \textcolor{black}{\quad + InfoNCE head} & \textcolor{black}{ViT-L} & \textcolor{black}{24.4}
      & \textcolor{black}{46.9} & \textcolor{black}{57.7} & \textcolor{black}{25.6} & \textcolor{black}{48.0} & \textcolor{black}{57.1} \\
    \oursrow \avjepa & ViT-B & 24.2 & 48.7 & 60.8 & 22.6 & 49.2 & 59.3 \\
    \oursrow \avjepa & ViT-L & 31.5 & 57.2 & \second{67.0} & 30.7 & 55.7 & 64.8 \\
    \oursrow \avjepa & ViT-H & 31.9 & \second{57.8} & \second{67.0} & 31.5 & 56.2 & 65.7 \\
    \oursrow \avjepa & ViT-g & \second{32.4} & \best{58.1} & \best{67.5} & 31.8 & 57.0 & 67.0 \\
    \bottomrule
  \end{tabular}
\end{table}

\begin{table*}[t]
  \centering
  \providecommand{\abcInput}[1]{{%
    \renewenvironment{table}[1][]{}{}%
    \renewcommand{\caption}[2][]{}%
    \renewcommand{\label}[1]{}%
    \renewcommand{\vspace}[1]{}%
    \input{#1}}}
  \ifdefined\abcFitBox\else\newsavebox\abcFitBox\fi
  \providecommand{\abcGrab}[1]{{%
    \let\abcOrigTabular\tabular \let\abcOrigEndTabular\endtabular
    \renewenvironment{tabular}{\global\setbox\abcFitBox\hbox\bgroup\abcOrigTabular}{\abcOrigEndTabular\egroup}%
    \setbox0\hbox{#1}}}
  \ifdefined\abscA\else\newsavebox\abscA\newsavebox\abscB\newsavebox\abscC\fi
  \ifdefined\abscD\else\newsavebox\abscD\fi
  \abcGrab{{%
    \renewenvironment{table}[1][]{}{}%
    \renewcommand{\caption}[2][]{}%
    \renewcommand{\label}[1]{}%
    \renewcommand{\vspace}[1]{}%
    
\begin{table}[t]
 	\centering
  	\providecommand{\alosT}[1]{\rotatebox{60}{\scriptsize #1}}
  	\providecommand{\alosC}{{\scriptsize\cmark}}
  	\let\originalDneg\dneg
  	\newcommand{\dnegsmall}[1]{\scalebox{0.95}{\originalDneg{$#1$}}}
  	\let\originalDpos\dpos
  	\newcommand{\dpossmall}[1]{\scalebox{0.95}{\originalDpos{$#1$}}}
  	\let\originalDnul\dnul
  	\newcommand{\dnulsmall}[1]{\scalebox{0.95}{\originalDnul{$#1$}}}
  	\renewcommand{\oursrow}{\rowcolor{oursrowbg}[2.5pt][6pt]}
  	\caption{
  		\textbf{Which prediction terms matter?} 
  		\avjepa variants trained with different sets of prediction terms.
		Cross-modal terms are essential, and anchoring on the fused pass (\emph{av2a} and \emph{av2v}) leads to best performance.
 	}
	\label{tab:ablation_losses}
  	\footnotesize
  	\setlength{\tabcolsep}{6pt}
	 	\begin{tabular}{
	 			*{6}{c@{\hspace{0.pt}}}c
	 			@{\hspace{4.pt}}c
	 			@{\hspace{7pt}}                 %
	 			r@{\hspace{4.pt}}l               %
	 			@{\hspace{7pt}}
	 			r@{\hspace{4.pt}}l               %
	 			@{\hspace{7pt}}
	 			r@{\hspace{4.pt}}l@{}            %
	 		}
	    \toprule
	    & \multicolumn{6}{c}{\scriptsize Cross-modal term}& & \multicolumn{6}{c}{AS-20K mAP (\%) $\uparrow$} 
	    \\
	    \cmidrule(lr){2-7} \cmidrule(l){9-14}
	    \alosT{Tok.} & \alosT{\texttt{a2v}} & \alosT{\texttt{v2a}} & \alosT{\texttt{a2av}} & \alosT{\texttt{v2av}} & \alosT{\texttt{av2a}}
        & \alosT{\texttt{av2v}}  & \#	& \multicolumn{2}{c}{A} & \multicolumn{2}{c}{V} & \multicolumn{2}{c}{A-V} 
	    \\
	    \midrule
	    \alosC & & & & & & & 3
	      & 29.4 & \dnulsmall{+0.0} & 17.0 & \dnulsmall{+0.0} & {31.3} & \dnulsmall{+0.0}
	     \\
	    \alosC & \alosC & \alosC & & & & & 5
	      & 37.8 & \dpossmall{+8.4} & 21.3 & \dpossmall{+4.3} & {38.5} & \dpossmall{+7.2}
	      \\
	    \alosC & \alosC & \alosC & \alosC & \alosC & & & 7
	      & 37.1 & \dpossmall{+7.7} & 20.8 & \dpossmall{+3.8} & {36.7} & \dpossmall{+5.4}
	      \\
	    \oursrow \alosC & \alosC & \alosC & & & \alosC & \alosC & 7
	      & \best{39.4} & \best{\dpossmall{+10.0}} & \second{22.1} & \dpossmall{+5.1} & {\best{42.6}} & \best{\dpossmall{+11.3}}
	      \\
	    \alosC & \alosC & \alosC & \alosC & \alosC & \alosC & \alosC & 9
   		& \best{39.4} & \best{\dpossmall{+10.0}} & \best{22.2} & \best{\dpossmall{+5.2}} & {\second{42.2}} & \dpossmall{+10.9}
	     \\
	    \bottomrule
	  \end{tabular}
 \end{table}
}}%
  \sbox\abscA{\usebox\abcFitBox}%
  \abcGrab{{%
    \renewenvironment{table}[1][]{}{}%
    \renewcommand{\caption}[2][]{}%
    \renewcommand{\label}[1]{}%
    \renewcommand{\vspace}[1]{}%
    
\begin{table}[t]
  \centering
  \let\originalDneg\dneg
  \newcommand{\dnegsmall}[1]{\scalebox{0.95}{\originalDneg{$#1$}}}
  \let\originalDpos\dpos
  \newcommand{\dpossmall}[1]{\scalebox{0.95}{\originalDpos{$#1$}}}
  \let\originalDnul\dnul
  \newcommand{\dnulsmall}[1]{\scalebox{0.95}{\originalDnul{$#1$}}}
  \renewcommand{\oursrow}{\rowcolor{oursrowbg}[2.5pt][7pt]}
  \providecommand{\amodC}{{\scriptsize\cmark}}
  \providecommand{\amodX}{{\scriptsize\textcolor{gray}{\xmark}}}
  \caption{
    Audio-visual pre-training improves both modalities.
    Parentheses show gains over unimodal pre-training.
  }
  \label{tab:ablation_modality}
  \footnotesize
  \setlength{\tabcolsep}{12pt}
  \begin{tabular}{
  		@{}
  		l@{\hspace{5pt}}
  		c@{\hspace{4pt}}
  		c@{\hspace{5pt}}
		c@{\hspace{9pt}}
  		c@{\hspace{7pt}}
  		@{}
  	}
    \toprule
    & \multicolumn{2}{c}{Pre-training} & \multicolumn{2}{c}{AS-20K mAP (\%) $\uparrow$} \\
    \cmidrule(lr){2-3} \cmidrule(l){4-5}
    \avjepa Variant & Audio & Video & \multicolumn{1}{c}{A} & \multicolumn{1}{c}{V} \\
    \midrule
    Audio-only (\texttt{a2a}) & \amodC & \amodX & 30.9 & \na \\
    Video-only (\texttt{v2v}) & \amodX & \amodC & \na & 16.1 \\
    \oursrow Audio-Visual & \amodC & \amodC & \best{39.4} \dpossmall{+8.5} & \best{22.1} \dpossmall{+6.0} \\
    \bottomrule
  \end{tabular}
\end{table}
}}%
  \sbox\abscB{\usebox\abcFitBox}%
  \abcGrab{{%
    \renewenvironment{table}[1][]{}{}%
    \renewcommand{\caption}[2][]{}%
    \renewcommand{\label}[1]{}%
    \renewcommand{\vspace}[1]{}%
    
\begin{table}[t]
  \centering
  \providecommand{\amodC}{{\scriptsize\cmark}}
  \providecommand{\amodX}{{\scriptsize\textcolor{red}{\xmark}}}
  \caption{
    \textbf{\protoloss hyperparameters.}
    \avjepa using different \protoloss variants, including no-EMA teacher, entropy $\lambda_{\mathrm{H}}$ and MEMax weights   $\lambda_{\mathrm{ME}}$, and teacher temperature $\tau_t$ (\cref{eq:assign,eq:protoloss}).
    Shaded row is the \avjepa.
    We report AudioSet performance, codebook usage and mean largest logit.
  }
  \label{tab:ablation_loss_weights}
  \footnotesize
  \setlength{\tabcolsep}{4pt}
  \begin{tabular}{@{}ccccccccc@{}}
    \toprule
    \multicolumn{1}{c}{} &
    \multicolumn{3}{c}{\protoloss} &
    \multicolumn{2}{c}{Codebook} &
    \multicolumn{3}{c}{AS-20K mAP (\%) $\uparrow$} 
    \\
    \cmidrule(r){2-4} \cmidrule(lr){5-6} \cmidrule(l){7-9}
    EMA &
    $\lambda_{\mathrm{H}}$ & $\lambda_{\mathrm{ME}}$ & $\tau_t$ & Usage (\%) & Max.\ prob. & A & V & A-V   \\
    \midrule
    \oursrow
    \amodC & 0.5 & 10 & $1/6$ & 99.5 & 0.827 & \best{39.4} & \best{22.1} & \best{42.6} \\
    \amodX & {0.5} & {10} & {$1/6$}  & 99.5 & 0.815 & 36.0 & 19.9 & 38.8  \\
    \amodC & \textcolor{red}{0}  & {10} & {$1/6$} & 94.7 & 0.023 & 34.5 & 17.8 & 38.4  \\
    \amodC & {0.5} & \textcolor{red}{0}     & {$1/6$} & 2.3 & 0.969 & 33.2 & 17.4 & 36.4  \\
    \amodC & {0.5} & {10} & \textcolor{red}{1}    & 88.7 & 0.252 & \second{37.2} & \second{21.3} & \second{40.9}  \\
    \bottomrule
  \end{tabular}
  \vspace{-0.1cm}
\end{table}
}}%
  \sbox\abscC{\usebox\abcFitBox}%
  \abcGrab{{%
    \renewenvironment{table}[1][]{}{}%
    \renewcommand{\caption}[2][]{}%
    \renewcommand{\label}[1]{}%
    \renewcommand{\vspace}[1]{}%
    \begin{table}[t]
  \centering
  \caption{
  	\textbf{Spectrogram chunking and prediction loss.}
    Two-seed means at final checkpoints: 260k for clip-level \mjepa baselines, 270k for chunked variants and \avjepa.
    A-V uses the offline AV++ evaluation; FSD50K reports mAP and ESC-50 reports accuracy (\%).
    Target denotes cross-modal target granularity.
    }
  \label{tab:ablation_chunking}
  \footnotesize
  \setlength{\tabcolsep}{3.5pt}
  \begin{tabular}{lcc<{\hspace{5pt}}rrrrr}
    \toprule
    & & & \multicolumn{3}{c}{AS-20K} & FSD & ESC \\
    \cmidrule(lr){4-6}\cmidrule(lr){7-7}\cmidrule(l){8-8}
    Model & Loss & Target & A & V & A-V & mAP & Acc. \\
    \midrule
    \mjepa & $\ell_1$ & Clip & 38.8 & 21.8 & 41.3 & 62.3 & 95.8 \\
    \mjepa & \protoloss & Clip & 38.0 & 22.1 & 41.8 & 62.1 & 95.4 \\
    \mjepa & $\ell_1$ & Chunk & 2.4 & 1.1 & 1.2 & 5.4 & 11.8 \\
    \mjepa & \protoloss & Chunk & \underline{39.7} & \textbf{23.1}& \underline{43.1} & \underline{64.5} & \underline{96.3} \\
    \oursrow \avjepa & \protoloss & Chunk & \textbf{40.1} & \underline{22.9} & \textbf{43.2} & \textbf{66.4} & \textbf{96.7} \\
    \bottomrule
  \end{tabular}
\end{table}
}}%
  \sbox\abscD{\usebox\abcFitBox}%
  \edef\abscPanelWidth{\the\dimexpr(\textwidth-\columnsep)/2\relax}%
  \dimen0=\wd\abscA
  \ifdim\wd\abscB>\dimen0 \dimen0=\wd\abscB\fi
  \ifdim\wd\abscC>\dimen0 \dimen0=\wd\abscC\fi
  \ifdim\wd\abscD>\dimen0 \dimen0=\wd\abscD\fi
  \dimen2=\abscPanelWidth
  \ifdim\dimen0>\dimen2
    \edef\abscWA{\the\dimexpr\wd\abscA*\numexpr\dimen2\relax/\numexpr\dimen0\relax\relax}%
    \edef\abscWB{\the\dimexpr\wd\abscB*\numexpr\dimen2\relax/\numexpr\dimen0\relax\relax}%
    \edef\abscWC{\the\dimexpr\wd\abscC*\numexpr\dimen2\relax/\numexpr\dimen0\relax\relax}%
    \edef\abscWD{\the\dimexpr\wd\abscD*\numexpr\dimen2\relax/\numexpr\dimen0\relax\relax}%
  \else
    \edef\abscWA{\the\wd\abscA}\edef\abscWB{\the\wd\abscB}\edef\abscWC{\the\wd\abscC}\edef\abscWD{\the\wd\abscD}%
  \fi
  \caption{
    \textbf{\avjepa ablations} with ViT-B encoders: AS-20K mAP (\%) with audio (A), video (V) or audio-visual (A-V) input; all results average two seeds.
    \textbf{(a)}~Cross-modal prediction terms; Tok.: the three token-level terms; \#: number of terms; deltas are relative to the token-level-only baseline (first row).
    \textbf{(b)}~Unimodal (audio- or video-only) versus audio-visual pretraining; deltas are relative to the unimodal models.
    \textbf{(c)}~\protoloss without EMA teacher, or with a different conditional-entropy weight $\lamH$, MEMax weight
    $\lamME$ or teacher temperature $\taut$ (changed value in \textcolor{red}{red}); Usage: share of prototypes in use; Max.\ prob.: mean largest teacher assignment probability.
    \textbf{(d)}~Prediction objective ($\ell_1$ or \protoloss) and cross-modal target granularity (clip or chunk) for \mjepa.
    (a)--(c) use a shorter schedule and are evaluated after 180k iterations; (d) reports full schedule (270k iterations).
    Shaded: default \avjepa recipe.
    \textbf{Bold}/\underline{underline}: best/second best.
  }
  \vspace{-0.2cm}
  \label{tab:ablations_combined}
  \begin{subtable}[t]{\abscPanelWidth}
    \vspace{0pt}
    \centering
    \caption{Ablation of prediction loss terms.}
    \label{tab:ablations_combined_terms}
    \resizebox{\abscWA}{!}{\usebox\abscA}
  \end{subtable}%
  \hfill
  \begin{subtable}[t]{\abscPanelWidth}
    \vspace{0pt}
    \centering
    \caption{Effect of pretraining modalities.}
    \label{tab:ablations_combined_modality}
    \resizebox{\abscWB}{!}{\usebox\abscB}
  \end{subtable}
  \par\vspace{8pt}
  \begin{subtable}[t]{\abscPanelWidth}
    \vspace{0pt}
    \centering
    \caption{Effect of \protoloss hyperparameters and EMA teacher.}
    \label{tab:ablations_combined_protoloss}
    \resizebox{\abscWC}{!}{\usebox\abscC}
  \end{subtable}%
  \hfill
  \begin{subtable}[t]{\abscPanelWidth}
    \vspace{0pt}
    \centering
    \caption{Impact of prediction loss and target granularity on M-JEPA.}
    \label{tab:ablations_combined_chunking}
    \resizebox{\abscWD}{!}{\usebox\abscD}
  \end{subtable}
\end{table*}

\subsection{Additional Ablations}
\label{sec:suppl_av_ablations}
In \cref{tab:ablations_combined}, we report additional analyses and ablation studies evaluating the effect of different prediction loss terms, pretraining modalities, \protoloss hyperparameters, and prediction objective and target granularity on M-JEPA.
All variants use ViT-B encoders and report AS-20K classification mAP with audio (A), video (V) or audio-visual (A-V)
input.
We highlight as shaded rows the default \avjepa recipe.
For efficiency some ablations are run with a shorter schedule of 180k training iterations, in contrast to the default 270k.
All results are the average of two runs using different random seeds.

\paragraph{Prediction terms (\cref{tab:ablations_combined_terms}).}
\avjepa combines three token-level terms within each input mode (\texttt{a2a}, \texttt{v2v}, \texttt{av2av}) with
chunk-level cross-modal terms, where \texttt{x2y} predicts the targets of mode \texttt{y} from the masked input of mode
\texttt{x}.
Cross-modal prediction is essential: adding \texttt{a2v} and \texttt{v2a} to the token-level terms improves performance in all three
input modes by a large margin.
The direction of the additional terms matters: predicting each modality from the joint input (\texttt{av2a}, \texttt{av2v}) gives the strongest results, whereas predicting the joint input from either modality alone (\texttt{a2av}, \texttt{v2av}) reduces performance.
We hypothesize that joint-to-unimodal prediction encourages the fused representation to preserve both streams, while unimodal-to-joint prediction demands information unavailable from the source and may compete with other objectives.
Using all nine terms brings no further gain: it matches the seven-term recipe on audio and video, while slightly trailing with AV inputs (42.2 vs.\ 42.6 mAP), despite two additional terms.
Effective cross-modal supervision thus depends on prediction direction, not simply on the number of objectives.

\paragraph{Pretraining modalities (\cref{tab:ablations_combined_modality}).}
\Cref{tab:ablations_combined_modality} compares \avjepa variants pretrained with a single modality (audio or video only), using solely the \texttt{a2a}
or \texttt{v2v} prediction objectives, respectively.
Joint AV pretraining improves both modalities: audio classification rises from 30.9 to 39.4 mAP (+8.5) and
video classification from 16.1 to 22.1 mAP (+6.0).
Learning from both modalities thus benefits each of them, even when only one is available at test time.

\paragraph{\protoloss hyperparameters (\cref{tab:ablations_combined_protoloss}).}
\protoloss trains the student to match the teacher's soft assignments to learned prototypes, with two entropy terms on
the teacher assignments: a conditional-entropy term (weighted by $\lamH$) that makes each assignment peaked, and
mean-entropy maximization (MEMax, weighted by $\lamME$) that encourages uniform prototype usage.
In \cref{tab:ablations_combined_protoloss} we analyze the effect of these regularizations and other hyperparameters in \protoloss.
Namely, we measure the downstream performance of the model variants, as well as the average codebook  usage and the mean largest teacher assignment probability.
Both entropy regularization terms are necessary and prevent different failure modes.
Without MEMax ($\lamME{=}0$), the teacher collapses onto 2.3\% of the prototypes, whereas without the conditional-entropy term ($\lamH{=}0$), its assignments become nearly flat (mean largest probability 0.023); each failure costs 4.2--6.2 mAP.
Importantly, these two terms alone are enough to prevent collapse: without an EMA teacher, \protoloss still uses 99.5\% of the prototypes with sharp assignments (0.815), and the model learns strong representations that outperform both entropy ablations.
This contrasts with regression-based JEPAs, which rely on the EMA teacher to avoid collapse~\cite{grill2020byol,chen2021simsiam}.
The EMA teacher and sharp targets remain beneficial: removing the EMA teacher costs 2.2--3.8 mAP, and raising the teacher temperature from $\taut{=}1/6$ to 1 costs 0.8--2.2 mAP.

\paragraph{Prediction objective and target resolution (\cref{tab:ablations_combined_chunking}).}
To validate the generality of our method, we examine how the prediction objective ($\ell_1$ or \protoloss) interacts with cross-modal target granularity (clip- or chunk-level) on M-JEPA, spectrogram-input models.
Replacing clip-level targets with time-aligned 500\,ms chunks yields stronger representations when trained with \protoloss, improving audio, video and AV classification, as well as transfer to FSD50K and ESC-50.
In contrast, the same chunked recipe with $\ell_1$ regression suffers a severe performance drop across all tasks, consistent with representation collapse.
We hypothesize that \protoloss avoids this failure through its entropy regularizers, which discourage both concentration on a few prototypes and uninformative, diffuse assignments (see \cref{tab:ablations_combined_protoloss}).
Combining M-JEPA, \protoloss and chunk-level targets already approaches \avjepa's performance: the spectrogram variant slightly leads on video (23.1 \vs 22.9 mAP), while waveform-based \avjepa improves audio (40.1 \vs 39.7) and marginally improves AV classification (43.2 \vs 43.1).
Its advantages are more pronounced on audio transfer, reaching 66.4 \vs 64.5 mAP on FSD50K and 96.7 \vs 96.3\% accuracy on ESC-50.
These results support combining dense temporal supervision with a non-collapsing prediction objective, and suggest that finer-grained waveform tokens on the shared audio-video time axis may further improve audio representation learning.

\section{Qualitative Audio-Visual Results}
\label{sec:suppl_av_qual}

\begin{figure}[t!]
  \centering
  \begin{minipage}{0.85\linewidth}%
  \centering
  \input{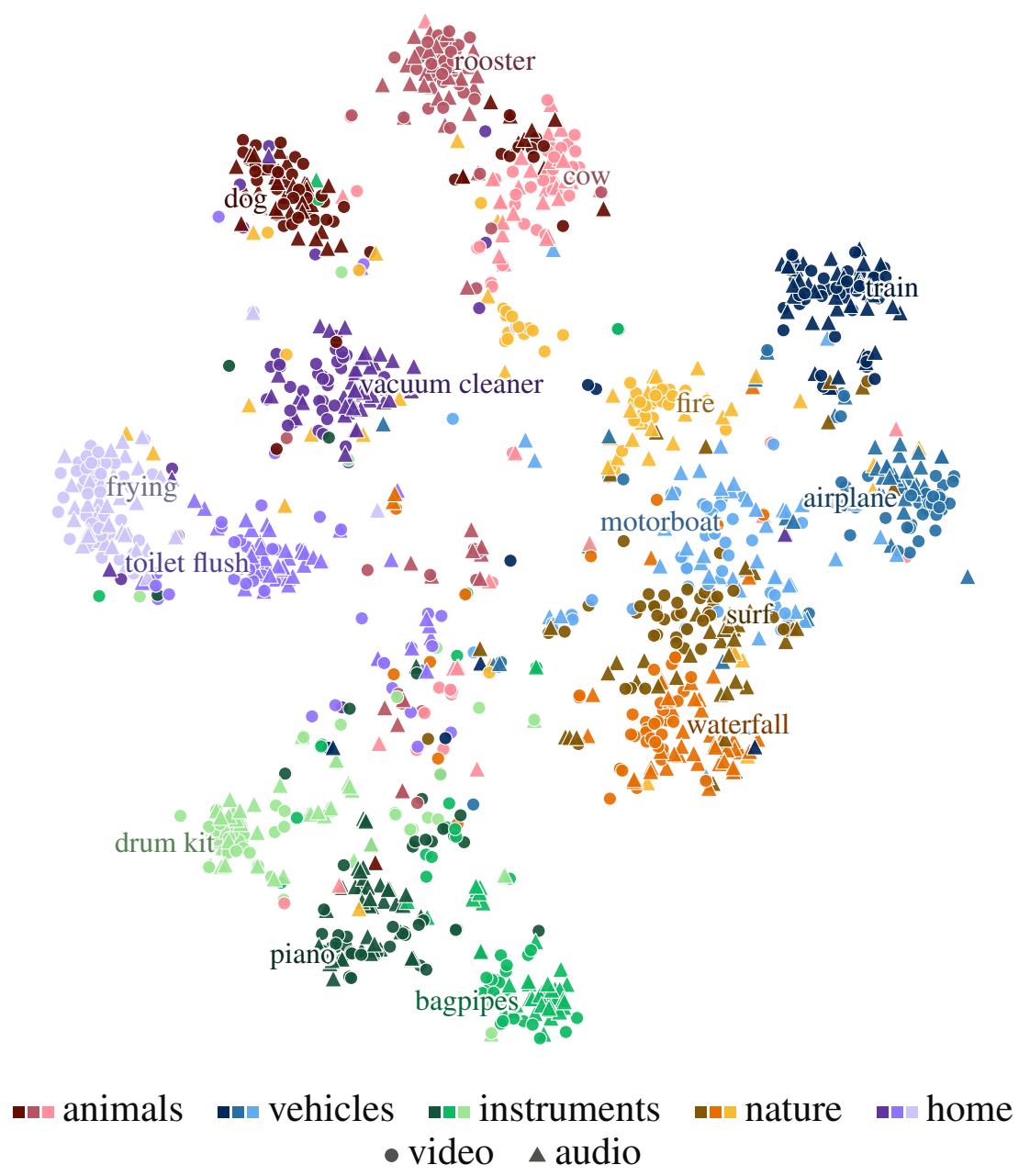}
  \end{minipage}
  \vspace{-0.2cm}
  \caption{
  	\textbf{Joint t-SNE} of \avjepa audio ($\triangle$) and video ($\circ$) embeddings for 675 AudioSet clips from 15 classes (45 per class).
    Audio and video of the same class cluster together, and related classes (shared color family) form neighboring regions.
  }
  \label{fig:av_tsne_semantic}
  \vspace{-0.4cm}
\end{figure}

\paragraph{Joint audio-visual embedding.}
\cref{fig:av_tsne_semantic} visualizes \avjepa embeddings with a joint t-SNE~\cite{vandermaaten2008visualizing} of AudioSet evaluation clips~\cite{gemmeke2017audio}, with audio and video encoded separately.
Videos and audios, depicted with circles and triangles respectively, from the same category occupy shared regions of the embedding space, while semantically similar classes (\eg water-related categories such as waterfall, surf or motorboat) neighbor each other.

\begin{figure*}[t!]
  \centering
  \begin{subfigure}{\linewidth}
    \centering
    \includegraphics[width=\linewidth]{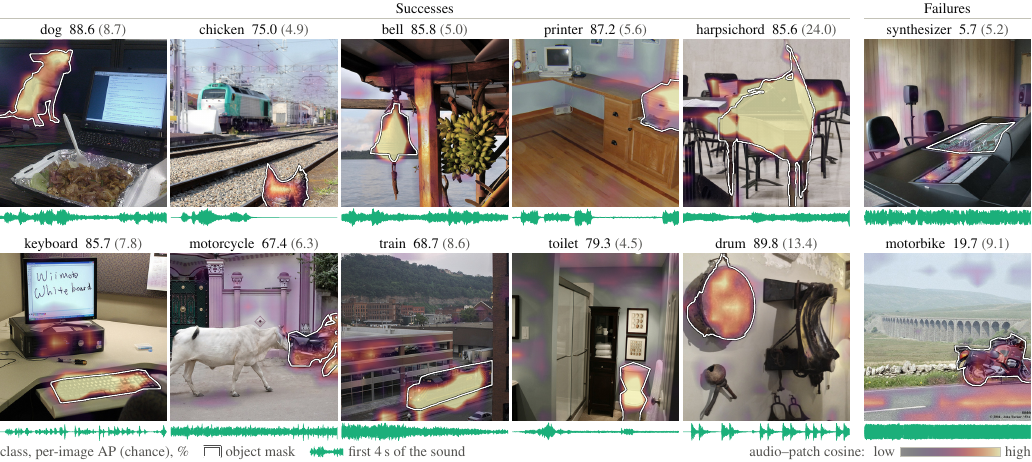}
    \caption{Sound-prompted segmentation results.}
    \label{fig:av_seg_examples_v5}
  \end{subfigure}
  \vspace{0.1cm}
  \begin{subfigure}{\linewidth}
    \centering
    \includegraphics[width=\linewidth]{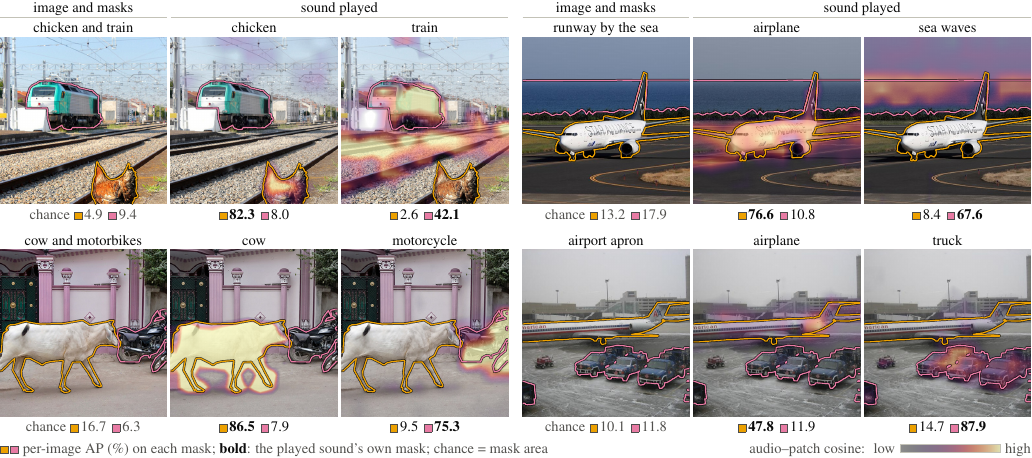}
    \caption{One image, two sounds.}
    \label{fig:av_seg_prompting_v4}
  \end{subfigure}
  \vspace{-0.5cm}
  \caption{
  	\textbf{Sound-prompted localization.}
  	Heatmaps show the cosine similarity between the pooled audio embedding and every image-patch token, upsampled for visualization; we report the per-image AP (\%) of each heatmap against the ground-truth object mask, together with chance-level AP.
  	(a)~\avjepa localizes the sounding object, even when it is small and not the most salient part of the scene; the rightmost column shows failure cases.
  	(b)~Prompting the same image with the sound of either object moves the \avjepa map onto that object.
  	The map scores higher AP on the played sound's mask, showing that localization is driven by sound rather than visual priors.
   }
  \label{fig:av_seg}
\end{figure*}

\paragraph{Sound-prompted localization.}
Since audio and video share one embedding space, \avjepa localizes sounds without any training: we score each image patch by its cosine similarity to the pooled audio embedding, and upsample the outcome for visualization.
\Cref{fig:av_seg_examples_v5} shows the resulting segmentations, including failure cases (quantitative results in \cref{tab:ade20k_soundprompted}).
\avjepa localizes small sounding objects, even when they are not the most salient or largest object in the scene.
In \cref{fig:av_seg_prompting_v4}, prompting the same image with the sounds of different objects moves the heatmap from one object to the other (42--88 AP on the played sound's mask \vs 3--15 AP on the other), confirming that localization is driven by the sound rather than by visual saliency.

\providecommand{\pnum}[1]{#1}%
\providecommand{\nTSNWclips}{\pnum{675}}%
\providecommand{\nTSNWpoints}{\pnum{1,350}}%
\providecommand{\nTSNWclasses}{\pnum{15}}%
\providecommand{\nTSNWperClass}{\pnum{45}}%
\providecommand{\nTSNWknnFull}{\pnum{76.3\%}}%
\providecommand{\nTSNWknnBal}{\pnum{75.7\%}}%
\providecommand{\nTSNWknnPlain}{\pnum{69.1\%}}%
\providecommand{\nTSNWchance}{\pnum{6.5\%}}%
\providecommand{\nTSNWfamFull}{\pnum{86.7\%}}%
\providecommand{\nTSNWfamBal}{\pnum{86.2\%}}%
\providecommand{\nTSNWfamPlain}{\pnum{80.8\%}}%
\providecommand{\nTSNWmixFull}{\pnum{0.08}}%
\providecommand{\nTSNWmixBal}{\pnum{0.39}}%
\providecommand{\nTSNWmixPlain}{\pnum{0.12}}%
\providecommand{\nTSNWperplexity}{\pnum{30}}%
\providecommand{\nTSNWfitbank}{\pnum{997}}%

\providecommand{\pnum}[1]{#1}%
\providecommand{\nSegTwoPairs}{\pnum{106}}%
\providecommand{\nSegTwoClasses}{\pnum{20}}%
\providecommand{\nSegTwoMAPfour}{\pnum{47.5}}%
\providecommand{\nSegTwoMAPfourCI}{\pnum{[37.2,\,57.7]}}%
\providecommand{\nSegTwoMAPten}{\pnum{48.6}}%
\providecommand{\nSegTwoMAPtenCI}{\pnum{[39.0,\,57.9]}}%
\providecommand{\nSegTwoSwapFour}{\pnum{18.5}}%
\providecommand{\nSegTwoCentre}{\pnum{23.2}}%
\providecommand{\nSegTwoChance}{\pnum{16.4}}%
\providecommand{\nSegTwoObjFour}{\pnum{35.1}}%
\providecommand{\nSegFiveImgMissSounds}{\pnum{5}}%
\providecommand{\nSegFiveImgMissBound}{\pnum{7}}%
\providecommand{\nSegFiveClipMissOthers}{\pnum{3}}%
\providecommand{\nSegFiveClipMissLo}{\pnum{61.2}}%
\providecommand{\nSegFiveClipMissHi}{\pnum{86.2}}%
\providecommand{\nSegFiveSwapMean}{\pnum{20.0}}%
\providecommand{\nSegFiveSeedOneMAP}{\pnum{40.1}}%
\providecommand{\nSegFiveSeedOneObj}{\pnum{34.5}}%
\providecommand{\nSegFiveExtHighInstr}{\pnum{2}}%
\providecommand{\nSegFiveExtHighInstrList}{\pnum{the bell clip on the harpsichord, 74.0; the dog clip on the drum, 63.5}}%
\providecommand{\nSegFiveExtHigh}{\pnum{4}}%
\providecommand{\nSegFiveExtPairs}{\pnum{72}}%
\providecommand{\nSegFiveExtHighAP}{\pnum{50}}%
\providecommand{\nSegFiveMaskLo}{\pnum{4.5}}%
\providecommand{\nSegFiveMaskHi}{\pnum{8.7}}%
\providecommand{\nSegFiveOwnMean}{\pnum{79.7}}%
\providecommand{\nSegFiveObjMean}{\pnum{14.2}}%
\providecommand{\nSegFiveChanceMean}{\pnum{6.4}}%
\providecommand{\nSegFiveClipMissObj}{\pnum{80.6}}%
\providecommand{\nSegFiveInstrMaskLo}{\pnum{13.4}}%
\providecommand{\nSegFiveInstrMaskHi}{\pnum{24.0}}%
\providecommand{\nSegFiveInstrObjHarpsichord}{\pnum{44.3}}%
\providecommand{\nSegFiveInstrObjDrum}{\pnum{44.5}}%
\providecommand{\nSegFiveBoxPad}{\pnum{15}}%
\providecommand{\nSegFivePoolN}{\pnum{1400}}%
\providecommand{\nSegFivePoolClasses}{\pnum{44}}%
\providecommand{\nSegFivePoolPass}{\pnum{178}}%
\providecommand{\nSegFivePoolAP}{\pnum{50}}%
\providecommand{\nSegFiveInstrPoolN}{\pnum{1673}}%
\providecommand{\nSegFiveInstrPoolGC}{\pnum{977}}%
\providecommand{\nSegFiveInstrPoolHV}{\pnum{696}}%
\providecommand{\nSegFiveInstrPoolPhotos}{\pnum{1299}}%
\providecommand{\nSegFiveInstrPoolShown}{\pnum{harpsichord 23 of 42 with annotated masks; drum 7 of 100 with GrabCut}}%
\providecommand{\nSegFiveInstrPoolClasses}{\pnum{16}}%
\providecommand{\nSegFiveInstrPoolPass}{\pnum{698}}%
\providecommand{\pnum}[1]{#1}%
\providecommand{\nPromptFourESCsounds}{\pnum{chicken, train, airplane, sea waves, cow}}%
\providecommand{\nPromptFourASsounds}{\pnum{motorcycle, truck}}%
\providecommand{\nPromptFourESCclips}{\pnum{40}}%
\providecommand{\nPromptFourASclips}{\pnum{20}}%
\providecommand{\nPromptFourAgree}{\pnum{297}}%
\providecommand{\nPromptFourClips}{\pnum{320}}%
\providecommand{\nPromptFourAgreeLow}{\pnum{16 of 20}}%
\providecommand{\nPromptFourAgreeLowSound}{\pnum{motorcycle}}%
\providecommand{\nPromptFourAgreeLowScene}{\pnum{cow and motorbikes}}%
\providecommand{\nPromptFourCocoMaskLo}{\pnum{4.9}}%
\providecommand{\nPromptFourCocoMaskHi}{\pnum{16.7}}%
\providecommand{\nPromptFourPooledClips}{\pnum{80}}%

\section{Model Scaling}
\label{sec:suppl_scaling}
We study how \avjepa scales along two axes: encoder size and pretraining data.
\Cref{sec:suppl_single_dataset_scaling} scales the encoder capacity with the pretraining data held fixed, and compares it with V-JEPA~2, V-JEPA~2.1 and \mjepa baselines on video, audio and audio-visual benchmarks.
\Cref{sec:suppl_multidb_scaling} then pretrains a single encoder jointly on AudioSet and VideoMix2M (VM2M), and compares it with same-size experts trained on only one of the two datasets.

\providecommand{\scalingnum}[1]{\csname scvalue@#1\endcsname}
\expandafter\def\csname scvalue@capacity-A-L\endcsname{40.18}
\expandafter\def\csname scvalue@capacity-V-L\endcsname{28.75}
\expandafter\def\csname scvalue@capacity-AV-L\endcsname{43.24}
\expandafter\def\csname scvalue@capacity-AVpp-L\endcsname{45.21}
\expandafter\def\csname scvalue@capacity-P12-AVpp-L\endcsname{45.23}
\expandafter\def\csname scvalue@capacity-A-H\endcsname{41.45}
\expandafter\def\csname scvalue@capacity-V-H\endcsname{29.45}
\expandafter\def\csname scvalue@capacity-AV-H\endcsname{44.07}
\expandafter\def\csname scvalue@capacity-AVpp-H\endcsname{46.13}
\expandafter\def\csname scvalue@capacity-P12-AVpp-H\endcsname{46.44}
\expandafter\def\csname scvalue@capacity-A-g\endcsname{41.94}
\expandafter\def\csname scvalue@capacity-V-g\endcsname{30.46}
\expandafter\def\csname scvalue@capacity-AV-g\endcsname{44.35}
\expandafter\def\csname scvalue@capacity-AVpp-g\endcsname{46.90}
\expandafter\def\csname scvalue@capacity-P12-AVpp-g\endcsname{46.26}
\expandafter\def\csname scvalue@depth-delta-L\endcsname{+1.31}
\expandafter\def\csname scvalue@depth-delta-H\endcsname{+1.07}
\expandafter\def\csname scvalue@depth-delta-g\endcsname{+0.19}
\expandafter\def\csname scvalue@specialist-A-L\endcsname{-0.48}
\expandafter\def\csname scvalue@specialist-V-L\endcsname{+3.05}
\expandafter\def\csname scvalue@specialist-A-H\endcsname{-0.29}
\expandafter\def\csname scvalue@specialist-V-H\endcsname{+2.21}
\expandafter\def\csname scvalue@weight-SS-5\endcsname{69.05}
\expandafter\def\csname scvalue@weight-A-5\endcsname{38.56}
\expandafter\def\csname scvalue@weight-SS-10\endcsname{72.46}
\expandafter\def\csname scvalue@weight-A-10\endcsname{38.68}
\expandafter\def\csname scvalue@weight-SS-20\endcsname{72.55}
\expandafter\def\csname scvalue@weight-A-20\endcsname{37.66}
\expandafter\def\csname scvalue@best-AVpp-g-P24\endcsname{46.91}

\begin{figure*}[t]
  \centering
  \captionsetup{skip=2pt}
  \captionsetup[subfigure]{skip=1pt}
  \begin{subfigure}{\textwidth}
    \centering
    \includegraphics[width=\linewidth]{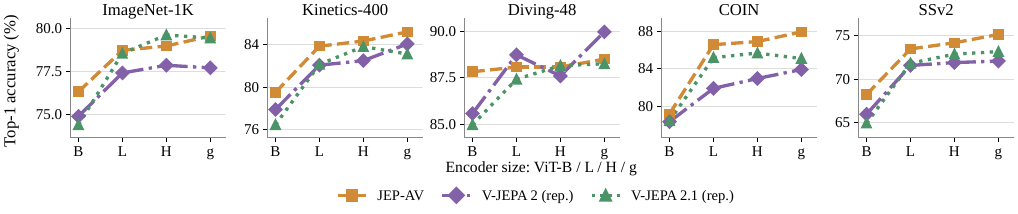}
    \caption{Video-only pretraining on VM2M.}
    \label{fig:unimodal_encoder_scaling_video}
  \end{subfigure}
  \par\medskip
  \begin{subfigure}{\textwidth}
    \centering
    \includegraphics[width=\linewidth]{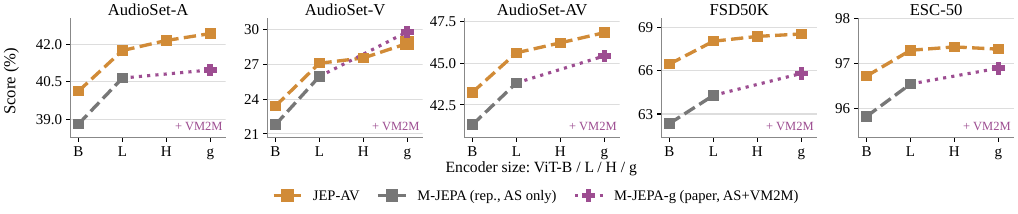}
    \caption{Audio-visual pretraining on AudioSet only.}
    \label{fig:unimodal_encoder_scaling_audio}
  \end{subfigure}
  \par
  \caption{
  	\textbf{Encoder scaling with single-dataset pretraining.}
	Comparison of \avjepa with our V-JEPA~2/2.1 and \mjepa reproductions from ViT-B to ViT-g, pretrained on (a)~VM2M or (b)~AudioSet; the purple dotted line is the published \mjepa-g~\cite{teotia2026mjepa}, which also uses VM2M.
	We report mAP for AudioSet and FSD50K, and top-1 accuracy (\%) otherwise. Results are the average of two seeds.
	\avjepa improves steadily with scale and outperforms the reproductions on most benchmarks and encoder sizes.
   }
  \label{fig:unimodal_encoder_scaling}
\end{figure*}

\subsection{Single-dataset encoder scaling}
\label{sec:suppl_single_dataset_scaling}
We first scale the encoder from ViT-B (86M parameters) to ViT-g (1B parameters) while keeping the pretraining data fixed: VM2M video only for the vision benchmarks (\cref{fig:unimodal_encoder_scaling_video}), and AudioSet audio-visual pretraining for the AudioSet and audio-transfer benchmarks (\cref{fig:unimodal_encoder_scaling_audio}).
We compare against our V-JEPA~2 and V-JEPA~2.1 reproductions trained on VM2M, our \mjepa reproductions trained on AudioSet at ViT-B and ViT-L, and the published \mjepa-g~\cite{teotia2026mjepa}, which is trained on AudioSet and VM2M.
 
\paragraph{Video representations.}
\avjepa improves SSv2 from 68.2\% at ViT-B to 75.2\% at ViT-g, and outperforms both V-JEPA baselines at every size on SSv2, K400 and COIN.
K400, COIN and ImageNet-1K also improve steadily with scale, with most of the gain between ViT-B and ViT-L.
Results on the other two benchmarks are mixed.
First, V-JEPA~2.1 achieves the best performance on ImageNet-1K at ViT-H (79.7 vs.\ 79.0\%) and is comparable with \avjepa at ViT-g (79.4 vs.\ 79.6\%).
We hypothesize that \vjepatwoone's objective, designed to learn dense spatial features~\cite{murlabadia2026vjepa21}, benefits static image recognition, whereas \avjepa's temporal prediction particularly benefits SSv2.
Second, on Diving-48, \avjepa leads clearly at ViT-B (+1.9) but improves little with scale (87.8 to 88.5\%), while a V-JEPA reproduction leads at every larger size.

\paragraph{Audio and audio-visual representations.}
On AudioSet, audio mAP rises from 40.1 at ViT-B to 42.4 at ViT-g, video mAP from 23.4 to 28.8, and joint audio-visual mAP from 43.3 to 46.8.
FSD50K improves monotonically with scale, demonstrating that scaling properties carry over beyond AudioSet, whereas ESC-50 improves only slightly, as it is close to saturation.
\avjepa outperforms our \mjepa reproduction on all five benchmarks at both ViT-B and ViT-L.
At ViT-L, it already matches or exceeds the published \mjepa-g on AudioSet audio and AV classification, FSD50K and ESC-50, even though \mjepa-g uses a larger encoder and additional VM2M video data.
\mjepa-g leads only on AudioSet video classification (29.8 vs.\ 28.8 mAP at ViT-g), likely due to this additional video data used during pretraining.

\begin{figure*}[t]
  \centering
  \captionsetup{skip=2pt}
  \includegraphics[width=\textwidth]{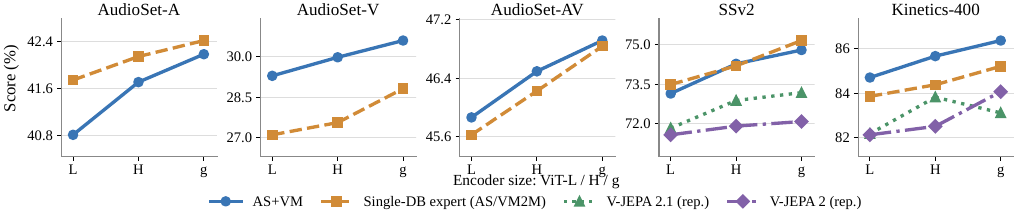}
  \par
  \caption{
  	\textbf{Effect of pretraining with a combined dataset.}
    Comparison of a single \avjepa encoder pretrained on AudioSet and VM2M (\texttt{AS+VM}) with same-size single-dataset experts (AudioSet-only for AudioSet, VM2M-only for SSv2 and K400) from ViT-L to ViT-g; our V-JEPA~2/2.1 reproductions are shown on SSv2 and K400 for reference.
    We report mAP for AudioSet and top-1 accuracy (\%) otherwise; \texttt{AS+VM} results use a single seed, and the other models average two seeds.
    The combined encoder improves AudioSet video and AV classification and K400 over the experts, matches them on SSv2, and has a small audio cost that shrinks with scale.
	}
  \label{fig:multidb_scaling}
\end{figure*}

\subsection{Combined-dataset pretraining}
\label{sec:suppl_multidb_scaling}
Next, we pretrain a single \avjepa encoder on AudioSet and VM2M combined, and compare with same-size experts pretrained on only one of the datasets: AudioSet-2M for the AudioSet evaluations, and VM2M for SSv2 and K400.
The results, depicted in \cref{fig:multidb_scaling}, show that adding VM2M to the pretraining data largely improves the tasks that involve video.
On AudioSet, video-only mAP improves by about 2 points at every encoder size, and joint audio-visual mAP also improves at all scales (+0.2/+0.3/+0.1).
On K400, the combined model outperforms the VM2M-only expert by $\sim$1 point, while SSv2 stays on par (73.1/74.3/74.8\% vs.\ 73.5/74.2/75.2\%).
Both models outperform our V-JEPA~2/2.1 reproductions on SSv2 and K400 at every size.
In exchange, AudioSet audio mAP drops slightly, but the gap to the AudioSet-only expert narrows with scale, from 0.9 points at ViT-L to 0.2 at ViT-g.
Thus, a single encoder trained on both datasets comes close to the single-dataset experts on audio and SSv2, while improving video and audio-visual classification on AudioSet and K400.

\section{Speech Experiments}
\label{sec:suppl_speech}
\providecommand{\speechnum}[1]{\csname speechvalue@#1\endcsname}
\expandafter\def\csname speechvalue@abx-baseline\endcsname{6.16}
\expandafter\def\csname speechvalue@abx-ours\endcsname{4.80}
\expandafter\def\csname speechvalue@map-baseline\endcsname{45.23}
\expandafter\def\csname speechvalue@map-ours\endcsname{58.76}
\expandafter\def\csname speechvalue@ABX_reduction_pp\endcsname{1.36}
\expandafter\def\csname speechvalue@ABX_relative_reduction_percent\endcsname{22.02}
\expandafter\def\csname speechvalue@MAP_gain_pp\endcsname{13.54}
\expandafter\def\csname speechvalue@SpidR_ABX_gap_baseline\endcsname{2.24}
\expandafter\def\csname speechvalue@SpidR_ABX_gap_AVJEPA\endcsname{0.89}
\expandafter\def\csname speechvalue@SpidR_MAP_gap_AVJEPA\endcsname{2.12}

To measure the generality of our model beyond video and audio representations, we investigate whether a \avjepa variant can learn useful linguistic representations from speech.

\paragraph{Evaluation setup.}
We pretrain an audio-only \avjepa on the LibriSpeech~\cite{panayotov2015librispeech} dataset, featuring 960\,h of read English speech, using only the \texttt{a2a} objective, and compare it with an \mjepa baseline and several speech SSL methods.
We evaluate frozen EMA-teacher features at 50\,Hz with utterance-level instance normalization, without training a downstream probe.
Phonetic ABX~\cite{schatz2016abx} measures triphone discriminability within and across speakers on LibriSpeech dev-clean and dev-other (lower error is better), whereas word MAP@R~\cite{carlin2011rapid} measures spoken-word retrieval (higher is better).
We select one encoder layer by its mean error across the four ABX conditions and report MAP@R from the same layer and checkpoint, avoiding metric-specific layer or checkpoint selection.

\begin{table}[!t]
\caption{
	\textbf{Speech unit evaluation} of audio-visual and speech SSL encoders pretrained on LibriSpeech 960\,h.
	We report ABX error (\%, $\downarrow$) and word MAP@R (\%, $\uparrow$) on dev-clean (C) and dev-other (O); W/A: within/across speakers; Layer: encoder layer with the lowest mean ABX error, also used for MAP@R.
	\mjepa and \avjepa are ViT-B models trained with the same optimization schedule and average 2 seeds; HuBERT, \dinosr and \spidr results are taken from~\cite{poli2025spidr}.
	\avjepa outperforms \mjepa and \dinosr on every metric, surpasses HuBERT on MAP@R, and approaches the strongest specialist, \spidr.
}
\vspace{-0.2cm}
\label{tab:speech_initial_pilot}
\centering\footnotesize
\setlength{\tabcolsep}{3pt}
\begin{tabular}{
		@{}lc*{6}{>{\raggedleft\arraybackslash}p{24.456pt}}@{}
	}
\toprule
 & & \multicolumn{4}{c}{ABX $\downarrow$} & \multicolumn{2}{c}{MAP $\uparrow$} \\
\cmidrule(lr){3-6}\cmidrule(l){7-8}
Model & Layer & C-W & C-A & O-W & O-A & C & O \\
\midrule
\multicolumn{8}{@{}l@{}}{\cellcolor{black!8}\textbf{Audio-Visual SSL models pretrained on speech}} \\
\mjepa & 10 & 4.50 & 5.25 & 6.27 & 8.62 & 54.73 & 35.72 \\
\oursrow \avjepa & 11 & 3.62 & 4.22 & 4.79 & 6.59 & 66.38 & 51.15 \\
\midrule
\multicolumn{8}{@{}l@{}}{\cellcolor{black!8}\textbf{Speech SSL baselines}} \\
HuBERT & 11 & 3.38 & 4.01 & 4.26 & 6.49 & 46.07 & 33.37 \\
\dinosr & 5 & 4.05 & 4.72 & 5.11 & 7.29 & 63.02 & 45.86 \\
\spidr & 6 & 3.32 & 3.66 & 3.74 & 4.95 & 66.50 & 55.26 \\
\bottomrule
\end{tabular}
\end{table}

\paragraph{Evaluation results.}
As shown in \cref{tab:speech_initial_pilot}, the audio-only \avjepa recipe improves all four ABX metrics and both MAP word-retrieval splits over the \mjepa baseline.
Mean ABX error decreases from \speechnum{abx-baseline}\% to \speechnum{abx-ours}\%, a 22\% relative reduction, while mean word MAP@R increases from \speechnum{map-baseline}\% to \speechnum{map-ours}\%.
These improvements support the effectiveness of combining our prototype-prediction objective with raw-waveform processing for learning phonetic and word-level structure.
\avjepa also narrows the gap to specialized speech models, outperforming popular baselines like HuBERT~\cite{hsu2021hubert} on the MAP scores and DinoSR~\cite{liu2023dinosr} on ABX.
These results suggest that our SSL recipe learns broadly useful representations, including phonetic and word-level features approaching the quality of those learned by speech-specialist models.

\FloatBarrier

\FloatBarrier
{\color{black}\section{World-model details and results}
\label{sec:supplworldmodel}
\renewcommand{\dbltopfraction}{0.9}\setcounter{dbltopnumber}{3}\renewcommand{\topfraction}{0.9}\setcounter{topnumber}{3}\renewcommand{\textfraction}{0.07}\renewcommand{\dblfloatpagefraction}{0.6}\renewcommand{\floatpagefraction}{0.6}%

\makeatletter\def\ftype@table{1}\makeatother

\subsection{Architecture and objectives}
\label{sec:suppl_avwm_arch}

We build \avwm on the frozen \avjepa encoder, whose pretraining (\mainref{sec:method}) is Stage~1, in two further stages (\mainref{fig:avworldmodel}). Stage~2 pretrains an
action-free predictor of the next chunk on unlabeled video, and Stage~3 adds action adapters for fine-tuning on walks
with recorded actions (\cref{sec:avwmrecipe,sec:supplavwmnwmplanning,sec:suppl_avnav_ours}).

\makeatletter\def\ftype@table{2}\makeatother%
\begin{table}[t]
  \color{black}
  \captionsetup{font={color=black}, labelfont={color=black}}
  \centering
  \caption{\textbf{AVJ-WM configurations.} AVJ-WM is WM-H on the \avjepa-L encoder, and AVJ-WM-B is WM-B on the
    \avjepa-B encoder, without QK-norm in the readout.}
  \vspace{-0.2cm}
  \label{tab:avwmconfigs}
  \fontsize{7.5}{9}\selectfont
  \setlength{\tabcolsep}{2pt}
  \begin{tabular}{
      @{}p{0.34\columnwidth}
      @{\hspace{6pt}}
      *{3}{>{\centering\arraybackslash}p{\dimexpr(0.66\columnwidth-6pt-4\tabcolsep)/3\relax}}
      @{}
    }
    \toprule
    Configuration & WM-B & WM-L & WM-H \\
    \midrule
    \multicolumn{4}{@{}l@{}}{\cellcolor{black!8}\textbf{Frozen encoder and chunks}} \\
    Encoder & \multicolumn{3}{c@{}}{\avjepa-L or -B (\cref{tab:model_scales}), last block} \\
    Chunk & \multicolumn{3}{c@{}}{0.5\,s, 2 frames at 4\,fps, 16\,kHz mono audio} \\
    Frame size & \multicolumn{3}{c@{}}{$224 \times 224$} \\
    Encoder windows & \multicolumn{3}{c@{}}{video 2\,s, audio 1\,s, ending at the chunk} \\
    Tokens per chunk & \multicolumn{3}{c@{}}{196 video ($14 \times 14$), 25 audio} \\
    \midrule
    \multicolumn{4}{@{}l@{}}{\cellcolor{black!8}\textbf{AV-Combiner}} \\
    Blocks & \multicolumn{3}{c@{}}{2} \\
    Latent tokens & \multicolumn{3}{c@{}}{64} \\
    Width & 768 & 1024 & 1280 \\
    Parameters & 14.2M & 25.3M & 39.4M \\
    \midrule
    \multicolumn{4}{@{}l@{}}{\cellcolor{black!8}\textbf{Causal predictor}} \\
    Blocks & 12 & 24 & 32 \\
    Width & 768 & 1024 & 1280 \\
    \# Attn.\ heads & 12 & 16 & 16 \\
    MLP ratio & \multicolumn{3}{c@{}}{4} \\
    Position encoding & \multicolumn{3}{c@{}}{RoPE~\cite{su2024roformer}} \\
    Context & \multicolumn{3}{c@{}}{64\,s (128 chunks), 10\,s on AudioSet and Kinetics} \\
    Dropout, drop path & \multicolumn{3}{c@{}}{0, 0} \\
    Parameters & 85.1M & 302.3M & 629.7M \\
    \midrule
    \multicolumn{4}{@{}l@{}}{\cellcolor{black!8}\textbf{Readout}} \\
    Design & \multicolumn{3}{c@{}}{1 cross-attention block, 221 learned queries} \\
    QK-norm~\cite{henry2020qknorm} & \multicolumn{3}{c@{}}{yes} \\
    Output & \multicolumn{3}{c@{}}{one linear projection per stream} \\
    \bottomrule
  \end{tabular}
  \vspace{-0.2cm}
\end{table}
\begin{table}[t]
  \color{black}
  \captionsetup{font={color=black}, labelfont={color=black}}
  \centering
  \caption{\textbf{AVJ-WM Stage~2 hyperparameters.} We use the short schedule for our size study and AVJ-WM-B.}
  \vspace{-0.2cm}
  \label{tab:avwmhyper}
  \fontsize{7.5}{9}\selectfont
  \setlength{\tabcolsep}{2pt}
  \begin{tabular}{
      @{}p{0.34\columnwidth}
      @{\hspace{6pt}}
      *{2}{>{\centering\arraybackslash}p{\dimexpr(0.66\columnwidth-6pt-2\tabcolsep)/2\relax}}
      @{}
    }
    \toprule
    Configuration & AVJ-WM & Short schedule \\
    \midrule
    \multicolumn{3}{@{}l@{}}{\cellcolor{black!8}\textbf{Objective (weights, \cref{eq:avwm_stage2})}} \\
    Prediction & \multicolumn{2}{c@{}}{1} \\
    Rollout & \multicolumn{2}{c@{}}{0.25, 2--16 chunks fed back} \\
    Rollout starts & \multicolumn{2}{c@{}}{4 per 64\,s window} \\
    Horizon heads & \multicolumn{2}{c@{}}{0.125 per head} \\
    Horizons & \multicolumn{2}{c@{}}{2, 4, 8 and 16 chunks} \\
    Chunk states & \multicolumn{2}{c@{}}{variance 1, decorrelation 1} \\
    Pooled input tokens & \multicolumn{2}{c@{}}{variance 1, decorrelation 3} \\
    Reconstruction & \multicolumn{2}{c@{}}{0.1} \\
    Consistency & \multicolumn{2}{c@{}}{0.25, ramped over 2,000 updates} \\
    Consistency target & \multicolumn{2}{c@{}}{EMA of the combiner, decay 0.998} \\
    Audio input dropout & \multicolumn{2}{c@{}}{50\% of clips} \\
    \midrule
    \multicolumn{3}{@{}l@{}}{\cellcolor{black!8}\textbf{Optimization}} \\
    Optimizer & \multicolumn{2}{c@{}}{AdamW, $\beta = (0.9, 0.999)$, $\epsilon = 10^{-8}$} \\
    Weight decay & \multicolumn{2}{c@{}}{0.04} \\
    Peak learning rate & $5 \times 10^{-4}$ & $3 \times 10^{-4}$ \\
    Warm-up from 1\% & 1,000 updates & 250 updates \\
    Exp.\ cool-down to 1\% & last 4,226 updates & last 2,574 updates \\
    Updates & 42,260 & 28,314 \\
    Gradient clipping & \multicolumn{2}{c@{}}{norm 10} \\
    Precision & \multicolumn{2}{c@{}}{bfloat16} \\
    \midrule
    \multicolumn{3}{@{}l@{}}{\cellcolor{black!8}\textbf{Batch per update}} \\
    AudioSet clips (10\,s) & 512 & 64 \\
    Kinetics clips (10\,s) & 256 & 32 \\
    Long windows (64\,s) & 416 & 52 \\
    Video per update & 9.5\,h & 1.2\,h \\
    Video in total & about 403k\,h & about 34k\,h \\
    \midrule
    \multicolumn{3}{@{}l@{}}{\cellcolor{black!8}\textbf{Augmentation}} \\
    Video & \multicolumn{2}{c@{}}{square crop (30--100\% of the frame), flip} \\
    AudioSet video & \multicolumn{2}{c@{}}{flip only} \\
    Audio & \multicolumn{2}{c@{}}{gain of $-6$ to $+6$\,dB} \\
    \bottomrule
  \end{tabular}
  \vspace{-0.2cm}
\end{table}
\begin{table}[t]
  \color{black}
  \captionsetup{font={color=black}, labelfont={color=black}}
  \centering
  \caption{\textbf{AVJ-WM Stage~2 training data.} Selected hours per source and each source's share of the video
    seen in training.}
  \vspace{-0.2cm}
  \label{tab:avwmtraining}
  \fontsize{7.5}{9}\selectfont
  \setlength{\tabcolsep}{2pt}
  \begin{tabular}{
      @{}p{0.38\columnwidth}
      @{\hspace{6pt}}
      *{3}{>{\centering\arraybackslash}p{\dimexpr(0.62\columnwidth-6pt-4\tabcolsep)/3\relax}}
      @{}
    }
    \toprule
    Source & Samples & Selected hours & Share of training video \\
    \midrule
    HowTo100M~\cite{miech2019howto100m} & 64\,s windows & 90,473 & 67.2\% \\
    Ego4D~\cite{grauman2022ego4d} & 64\,s windows & 3,200 & 8.6\% \\
    Nymeria~\cite{ma2024nymeria,detone2026nymeriaplus} & 64\,s windows & 322 & 0.9\% \\
    IndEgo~\cite{chavan2025indego} & 64\,s windows & 156 & 0.4\% \\
    HoloAssist~\cite{wang2023holoassist} & 64\,s windows & 127 & 0.3\% \\
    Ego-Exo4D v2, ego view~\cite{grauman2024egoexo4d} & 64\,s windows & 99 & 0.3\% \\%
    AudioSet~\cite{gemmeke2017audio} & 10\,s clips & 4,305 & 14.9\% \\
    Kinetics~\cite{kay2017kinetics} & 10\,s clips & 1,179 & 7.5\% \\
    \bottomrule
  \end{tabular}
  \vspace{-0.2cm}
\end{table}
\makeatletter\def\ftype@table{1}\makeatother%
\paragraph{Windowed chunk encoding.}
Each 0.5\,s chunk holds two video frames and the matching waveform. The frozen encoder encodes it in
context, from a 2\,s video window (the chunk and the 1.5\,s before it) and, in a separate pass, a 1\,s audio window, and
we keep the chunk's own 196 video and 25 audio tokens. Separate passes let each stream use its own window and
leave one stream's tokens unchanged when the other is missing; the AV-Combiner then fuses the two. In preliminary experiments, this improved the features over
encoding each chunk alone, especially for motion, at a fraction of the cost of a 4\,s window per chunk.

\paragraph{AV-Combiner.}
A learned AV-Combiner~\cite{piergiovanni2024mirasol} compresses the 221 tokens into a chunk state $s_t$ of 64
latent tokens. It projects the tokens to the predictor's width and adds learned position embeddings and an audio type
embedding; learned missing-audio tokens replace absent or hidden audio. Each latent starts as a learned embedding plus
the mean of one of 64 equal windows over the tokens, video first (56 windows cover only video, 7 only audio and one
both), and two transformer blocks let every latent attend to all tokens and latents. The consistency term below teaches the combiner to infer a missing stream from the other.

\paragraph{Causal predictor and readout.}
The predictor is a transformer over up to 128 chunk states (64\,s), with bidirectional attention within a chunk,
causal attention across chunks and rotary position embeddings~\cite{su2024roformer}. A readout cross-attends 221 learned queries, one per target token, to the 64 predicted latents, similar to prior work~\cite{jaegle2022perceiverio,baldassarre2025dinoworld}. QK-norm~\cite{henry2020qknorm}
keeps the attention scores from growing in long runs, and one linear projection per stream maps the readout's outputs to
video and audio tokens. The targets are the frozen encoder's tokens of the next chunk, with stop-gradient. In rollouts, predicted tokens are fed back through the combiner as if observed, and the rollout term below trains the combiner and the predictor on such inputs.

\paragraph{Stage~2 objective.}
The prediction term $\Loss_{\mathrm{pred}}$ (\mainref{eq:wm_pred}) is averaged over each stream's tokens and
over the chunks where that stream is observed, and summed over the two streams. The six auxiliary terms of \mainref{sec:avwm} are
\begin{equation}
  \begin{aligned}
    \Loss_{\mathrm{roll}} &= \mathrm{mean}_{2 \le n \le N}\; \ell\big(\hat{e}^{(n)}_{t+n},\, e_{t+n}\big), \\
    \Loss_{\mathrm{hor}} &= \textstyle\sum_{k}\, \ell\big(\hat{e}^{[k]}_{t+k},\, \mathrm{sg}[\hat{e}_{t+k}]\big), \\
    \Loss_{\mathrm{var}}(x) &= \tfrac{1}{D}\textstyle\sum_d \max\big(0, 1 - \mathrm{std}(x_d)\big), \\
    \Loss_{\mathrm{cov}}(x) &= \mathrm{mean}_{i \neq j}\, \mathrm{Cov}(x)_{ij}^2, \\
    \Loss_{\mathrm{rec}} &= \big\|g(\tilde{e}_t) - e_t\big\|_1, \\
    \Loss_{\mathrm{cons}} &= \tfrac12 \textstyle\sum_{m \in \{v, a\}} d\big(s^m_t,\, \mathrm{sg}[\bar{s}_t]\big),
  \end{aligned}
  \label{eq:avwm_terms}
\end{equation}
averaged over chunks and clips (sg: stop-gradient). As in other latent world
models~\cite{hansen2024tdmpc2,assran2025vjepa2,sobal2026learning,zholus2026robojepa}, $\hat{e}^{(n)}_{t+n}$ predicts chunk
$t+n$ after $n-1$ of the model's own predictions have been fed back, detached; the maximum rollout length $N$ grows from
2 chunks over the first 500 updates. As in multi-token prediction~\cite{gloeckle2024multitoken,teoh2026mtp}, one head per
horizon $k \in \{2, 4, 8, 16\}$ chunks predicts $\hat{e}^{[k]}_{t+k}$, the model's own prediction, from the predictor's
output at chunk $t$ (up to $k = 8$ on ten-second clips). The VICReg~\cite{bardes2022vicreg} variance and decorrelation
terms act on the $D$-dimensional chunk states $s$ and on the projected encoder tokens. For these terms, we pool the
projected tokens to $7 \times 7$ video and 5 audio tokens per chunk ($\tilde{e}$), while the combiner reads all of them.
From $\tilde{e}$, a linear head $g$ reconstructs 196 randomly drawn video tokens and all 25 audio tokens (on
layer-normalized tokens). The consistency term compares the states $s^v_t$ and
$s^a_t$ computed from video or audio alone, with learned missing tokens for the other stream, with the state $\bar{s}_t$
that an exponential moving average of the combiner computes from both, through the cosine distance $d$ after subtracting
the batch mean. We also hide the audio input of half the clips while still predicting their audio. The Stage~2 loss is
\begin{equation}
  \begin{aligned}
    \Loss_{\mathrm{Stage\,2}} ={}& \Loss_{\mathrm{pred}} + \lambda_{\mathrm{roll}}\Loss_{\mathrm{roll}} + \lambda_{\mathrm{hor}}\Loss_{\mathrm{hor}} \\
    &+ \lambda_{\mathrm{rec}}\Loss_{\mathrm{rec}} + \lambda_{\mathrm{cons}}\Loss_{\mathrm{cons}} \\
    &+ \textstyle\sum_{x \in \{s,\, \tilde{e}\}} \big(\lambda^{x}_{\mathrm{var}}\Loss_{\mathrm{var}}(x) + \lambda^{x}_{\mathrm{cov}}\Loss_{\mathrm{cov}}(x)\big),
  \end{aligned}
  \label{eq:avwm_stage2}
\end{equation}
with the weights $\lambda$ of \cref{tab:avwmhyper}, and everything except the encoder is trained.

\begin{figure*}[t]
  \color{black}
  \captionsetup{font={color=black}, labelfont={color=black}}
  \centering
  \includegraphics[width=0.94\textwidth]{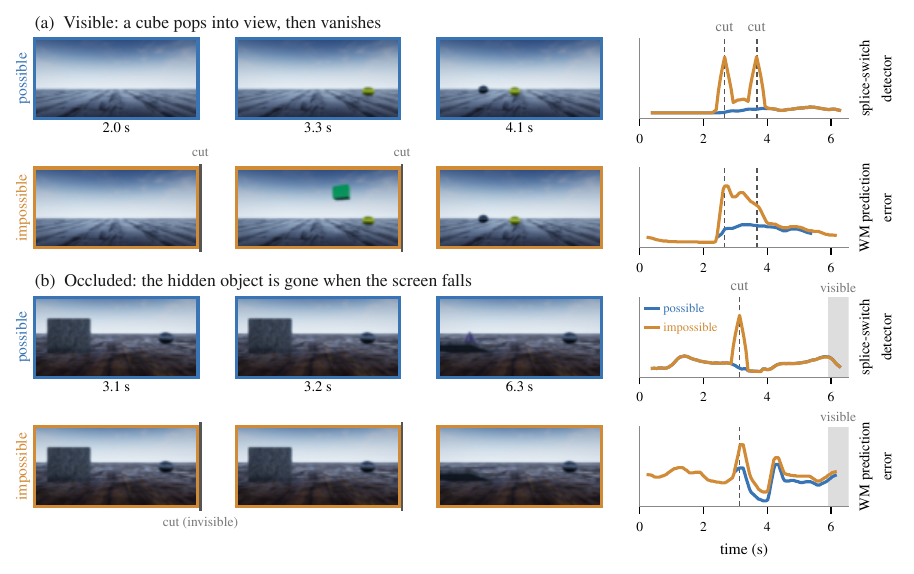}
  \caption{\textbf{The IntPhys~v1 splice artifact.} The splice-switch detector's measure (change of the 6-frame average background; right) and our world model's prediction error both jump at the cuts between two renders (ticks), not at the physical event. \textbf{(a)}~A visible cut, where a cube pops into view and vanishes. \textbf{(b)}~An occluded cut, where only the render noise switches, seconds before the screen falls.}
  \label{fig:intphyssplice}
\end{figure*}
\paragraph{Stage~3 action adapters.}
Stage~3 inserts the residual action adapters of \mainref{sec:avwm} after each causal block and trains them with the
predictor blocks and the readout, keeping the encoder and the AV-Combiner frozen. Beyond the prediction and rollout terms, its full objective depends on the benchmark; we give it with our navigation recipe (\cref{sec:suppl_avnav_ours}).

\begin{table*}[t]
  \color{black} %
  \captionsetup{font={color=black}, labelfont={color=black}}
  \centering
  \caption{\textbf{Intuitive physics on vision-only benchmarks} (paired accuracy, \%). AVJ-WM is best on IntPhys~v1.1, and every model stays near chance on IntPhys~2 and on InfLevel's solidity and gravity. IntPhys~v1.1 is our rebuild of v1 without the splice artifact (\cref{sec:intphysv11}); readouts as in \cref{sec:avwmprotocols}, world models after action-free pretraining (three seeds for AVJ-WM-B, one for AVJ-WM). Enc./Pred.: encoder/predictor size (encoders with their own pretraining predictor); Perm.: permanence; Shape: shape constancy; Cont.: continuity; Imm.: immutability; Sol.: solidity; Grav.: gravity. $^\dagger$Released checkpoint, evaluated by us. $^\ddagger$V-JEPA-H of Garrido et al., pretrained on VideoMix2M~\cite{bardes2024revisiting}, with their fixed readout and the official relative score, which compares a scene's two implausible clips with its two plausible ones. With this readout, the released \mbox{V-JEPA}~2.1-G also reaches 100, at 95.0\% paired accuracy. \textbf{Bold}/\underline{underline}: best/second best, excluding the published row and the splice-switch detector, unless all models are within the coin-flip band.}
  \label{tab:avwmphysicsbreakdown}
  \label{tab:avwminflevel}
  \footnotesize
  \setlength{\tabcolsep}{3.4pt}
  \begin{tabular}{@{}l cc c cccc ccccc ccc@{}}
    \toprule
    & & & IntPhys~v1 & \multicolumn{4}{c}{IntPhys~v1.1} & \multicolumn{5}{c}{IntPhys~2 Main} & \multicolumn{3}{c@{}}{InfLevel-Lab} \\
    \cmidrule(lr){4-4}\cmidrule(lr){5-8}\cmidrule(lr){9-13}\cmidrule(l){14-16}
    Model & Enc. & Pred. & All & All & Perm. & Shape & Cont. & All & Perm. & Imm. & Cont. & Sol. & Cont. & Sol. & Grav. \\
    \midrule
    V-JEPA-H~\cite{garrido2025intuitive}, published$^\ddagger$ & H & -- & 99.7--100 & -- & -- & -- & -- & -- & -- & -- & -- & -- & -- & -- & -- \\
    \mbox{V-JEPA}~2$^\dagger$ & L & -- & 70.3 & 58.4 & 59.5 & 51.8 & 64.0 & 52.6 & 51.7 & 48.3 & 56.7 & 53.4 & 57.0 & 47.8 & 48.6 \\
    \mbox{V-JEPA}~2$^\dagger$ & H & -- & 87.3 & \second{69.4} & \best{72.2} & 52.6 & \second{83.4} & 53.8 & 50.8 & 50.8 & 59.2 & 54.1 & 63.4 & 46.9 & 49.9 \\
    \mbox{V-JEPA}~2$^\dagger$ & g & -- & 64.3 & 64.7 & 67.3 & 53.9 & 73.0 & 51.8 & 55.0 & 49.2 & 56.7 & 47.3 & 45.4 & 49.1 & 49.7 \\
    \mbox{V-JEPA}~2.1$^\dagger$ & G & -- & 92.4 & 64.6 & 58.4 & 61.8 & 73.6 & 47.2 & 51.7 & 45.8 & 49.2 & 43.2 & \best{80.6} & 50.7 & 50.5 \\
    \midrule
    Splice-switch detector & -- & -- & 99.8 & -- & -- & -- & -- & -- & -- & -- & -- & -- & -- & -- & -- \\
    \avjepa encoder & B & -- & 95.5 & 64.8 & 63.4 & 55.1 & 76.0 & 49.6 & 50.0 & 53.3 & 45.8 & 49.3 & 78.0 & 52.7 & 51.6 \\
    \avjepa encoder & L & -- & \second{98.4} & 69.3 & 64.3 & 59.8 & \best{83.9} & 50.0 & 55.8 & 50.0 & 48.3 & 46.6 & \second{78.1} & 50.7 & 49.6 \\%
    AV-CDiT~\cite{wang2025avwm}, pretrained & B & B & -- & 68.6 & \second{69.8} & \best{73.0} & 62.9 & 51.2 & 50.0 & 50.8 & 47.5 & 55.5 & 60.9 & 52.7 & 50.3 \\
    \oursrow AVJ-WM-B (ours) & B & B & 96.4 & 66.0 & 62.2 & 60.6 & 75.2 & 49.3 & 51.4 & 46.7 & 47.5 & 51.1 & 73.2 & 51.3 & 49.9 \\%
    \oursrow AVJ-WM (ours) & L & H & \best{99.4} & \best{70.8} & 66.4 & \second{64.3} & 81.6 & 49.0 & 48.3 & 46.7 & 52.5 & 48.6 & 65.6 & 51.3 & 48.8 \\
    \midrule
    Coin flip, 95\% band & & & $\pm$1.2 & $\pm$1.2 & $\pm$2.1 & $\pm$2.1 & $\pm$2.1 & $\pm$4.3 & $\pm$9.2 & $\pm$9.2 & $\pm$9.2 & $\pm$8.2 & $\pm$3.0 & $\pm$4.7 & $\pm$2.9 \\
    \bottomrule
  \end{tabular}
\end{table*}
\subsection{Training recipe}
\label{sec:avwmrecipe}

For AVJ-WM, we train the WM-H predictor (\cref{tab:avwmconfigs}) on the frozen \avjepa-L encoder with the hyperparameters of \cref{tab:avwmhyper}.
Our size study (\cref{sec:supplavwmablations}) and AVJ-WM-B use a shorter schedule with an eighth of the batch and a lower peak learning rate (\cref{tab:avwmhyper}).
\cref{tab:avwmtraining} describes the sources of our ten-second clips and 64\,s windows of long video; we rectify the Aria Glasses fisheye recordings (Ego-Exo4D, IndEgo,
Nymeria) to a pinhole view with a $90^{\circ}$ field of view.

\begin{table*}[t]
  \color{black}
  \captionsetup{font={color=black}, labelfont={color=black}}
  \centering
  \caption{\textbf{Visual planning on NWM's datasets} (ATE and RPE in waypoint units, lower is better; \cref{sec:supplavwmnwmplanning}). Our world models plan better than NWM's released CDiTs on SCAND and TartanDrive. $^\dagger$Released NWM checkpoint, evaluated by us. \textcolor{gray}{Gray}: released GNM and NoMaD weights and results reported by other papers under their own protocols. We average three seeds for the rows we train. \textbf{Bold}/\underline{underline}: best/second best among the other rows.}
  \label{tab:avwmplanning}
  \footnotesize
  \setlength{\tabcolsep}{3.6pt}
  \begin{tabular}{@{}cl l cc cc cc cc@{}}
    \toprule
    & & & \multicolumn{2}{c}{RECON} & \multicolumn{2}{c}{SCAND} & \multicolumn{2}{c}{TartanDrive} & \multicolumn{2}{c@{}}{HuRoN} \\
    \cmidrule(lr){4-5} \cmidrule(lr){6-7} \cmidrule(lr){8-9} \cmidrule(l){10-11}
    & Model & Latents & ATE & RPE & ATE & RPE & ATE & RPE & ATE & RPE \\
    \midrule
     & Walking straight ahead & -- & 1.916 & 0.536 & 2.967 & 0.623 & 5.752 & 1.188 & 3.246 & 0.852 \\
    \multirow{-2}{*}{\rotatebox[origin=c]{90}{\scriptsize\textit{Ref.}}} & \gr{Our planner, perfect cost (privileged)} & -- & \gr{0.569} & \gr{0.283} & \gr{0.461} & \gr{0.209} & \gr{3.473} & \gr{0.786} & \gr{4.514} & \gr{0.998} \\
    \midrule
     & GNM~\cite{shah2023gnm}, retrained & -- & 1.541 & 0.701 & 1.700 & 0.540 & \second{3.548} & 1.389 & \second{2.256} & \second{0.948} \\
     & NoMaD~\cite{sridhar2024nomad}, retrained & -- & 1.315 & 0.350 & 2.564 & 0.549 & \best{2.590} & \best{0.605} & \best{1.930} & \best{0.592} \\
     & \gr{GNM, released} & -- & \gr{1.514} & \gr{0.698} & \gr{1.878} & \gr{0.571} & \gr{4.903} & \gr{1.456} & \gr{2.120} & \gr{0.917} \\
     & \gr{NoMaD, released} & -- & \gr{1.124} & \gr{0.306} & \gr{2.690} & \gr{0.586} & \gr{3.892} & \gr{0.842} & \gr{1.523} & \gr{0.507} \\
    \cmidrule(l){2-11}
     & CDiT-S~\cite{bar2025nwm}$^\dagger$ & SD-VAE & 1.206 & 0.352 & 1.057 & 0.289 & 5.827 & 1.216 & 5.449 & 1.178 \\
     & CDiT-B$^\dagger$ & SD-VAE & \second{1.137} & 0.348 & 1.043 & 0.288 & 5.739 & 1.199 & 5.593 & 1.205 \\
     & CDiT-L$^\dagger$ & SD-VAE & \best{1.126} & 0.349 & 1.110 & 0.296 & 5.750 & 1.202 & 5.720 & 1.230 \\
     & CDiT-XL$^\dagger$ & SD-VAE & 1.186 & 0.346 & 1.212 & 0.314 & 5.765 & 1.206 & 5.553 & 1.196 \\
     & \gr{CDiT-XL, as reported} & \gr{SD-VAE} & \gr{1.13} & \gr{0.35} & \gr{1.28} & \gr{0.33} & \gr{5.63} & \gr{1.18} & -- & -- \\
     & \gr{SkipWorld~\cite{yang2026skipworld}, as reported} & \gr{SD-VAE} & \gr{1.19} & \gr{0.34} & \gr{0.98} & \gr{0.28} & \gr{5.74} & \gr{1.20} & \gr{4.86} & \gr{1.04} \\
     & \gr{RAE-NWM~\cite{zhang2026raenwm}, as reported} & \gr{DINOv2} & \gr{1.36} & \gr{0.37} & \gr{1.14} & \gr{0.28} & -- & -- & \gr{2.91} & \gr{0.70} \\
     & CDiT~\cite{murlabadia2026vjepa21} & \vjepatwoone-B & 1.316 & 0.340 & 1.016 & 0.280 & 5.782 & 1.201 & 5.556 & 1.192 \\
     & \gr{CDiT~\cite{murlabadia2026vjepa21}, as reported} & \gr{\vjepatwoone-G} & \gr{1.179} & \gr{0.340} & \gr{1.038} & \gr{0.285} & \gr{5.687} & \gr{1.187} & \gr{4.054} & \gr{0.938} \\
    \multirow{-14}{*}{\rotatebox[origin=c]{90}{\scriptsize\textit{Video only}}} & CDiT & \avjepa-B & 1.339 & 0.348 & \second{1.002} & \second{0.278} & 5.743 & 1.192 & 5.473 & 1.177 \\
    \midrule
     & AV-CDiT~\cite{wang2025avwm}, pretrained & \avjepa-B & 1.340 & 0.347 & \best{0.959} & \best{0.271} & 5.660 & 1.179 & 5.305 & 1.141 \\
    \oursrow \cellcolor{white} & AVJ-WM-B (ours) & \avjepa-B & 1.227 & \second{0.335} & 1.036 & \second{0.278} & 4.766 & \second{1.012} & 5.160 & 1.101 \\
    \oursrow \cellcolor{white}\multirow{-3}{*}{\rotatebox[origin=c]{90}{\scriptsize\textit{AV}}} & AVJ-WM (ours) & \avjepa-L & 1.218 & \best{0.330} & 1.005 & \best{0.271} & 4.909 & 1.039 & 5.266 & 1.119 \\
    \bottomrule
  \end{tabular}
\end{table*}
\begin{figure*}[t]
  \color{black}
  \captionsetup{font={color=black}, labelfont={color=black}}
  \centering
  \includegraphics[width=\textwidth]{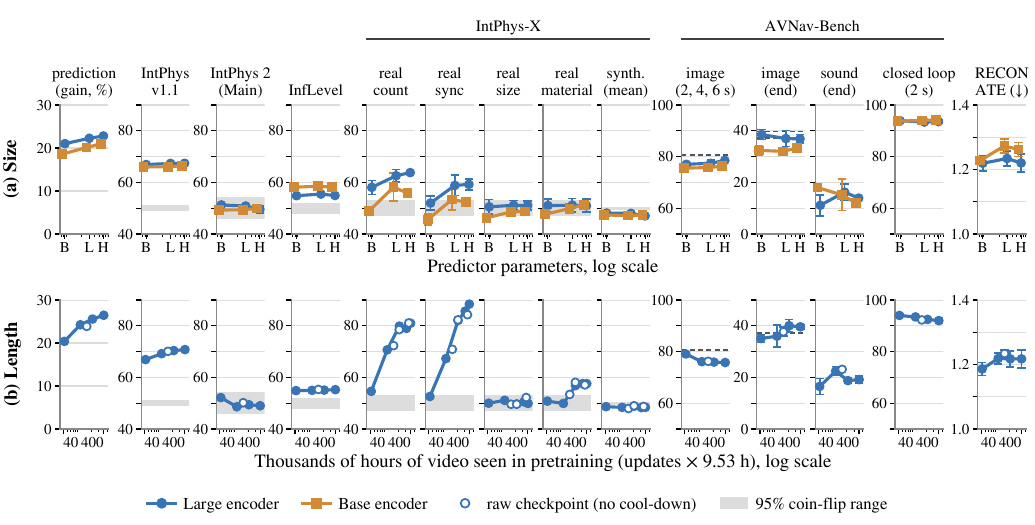}%
  \caption{\textbf{Scaling the world model.} Prediction improves with size and training, and cross-modal physics mostly with training, while scaling benefits are mixed in planning. \textbf{(a)}~Predictor size (\cref{tab:avwmconfigs}) on the \avjepa-L encoder (circles) and the \avjepa-B encoder (squares). \textbf{(b)}~Training length of AVJ-WM, in hours of video seen. Prediction gain is the error reduction relative to copying the last
    observed chunk. Error bars show 1 SD, dashed lines exact predictions and gray bands chance. All accuracy panels share one 50-point scale. Details in \cref{sec:supplavwmablations}.}
  \label{fig:avwmladder}
\end{figure*}
\begin{figure}[t]
  \color{black}
  \captionsetup{font={color=black}, labelfont={color=black}}
  \centering
  \includegraphics[width=\columnwidth]{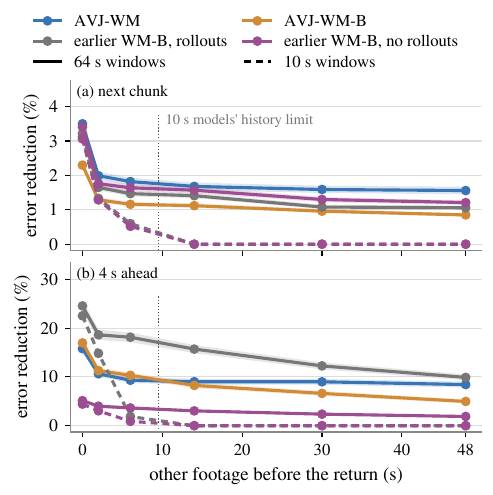}%
  \caption{\textbf{Returning to a scene.} Having seen a scene still lowers AVJ-WM's prediction error 48\,s later, whereas earlier WM-B variants trained on 10\,s windows lose this benefit once the scene leaves their history (dotted line). Error reduction on the scene's continuation relative to a control clip without the scene, against the length of the intervening footage, \textbf{(a)}~for the next chunk and \textbf{(b)}~4\,s ahead; 95\% intervals over 978 clips. The earlier variants form matched pairs that differ mainly in the training window. One pretraining seed per model.}
  \label{fig:avwmreturn}
\end{figure}
\paragraph{Compute.}
In H200 GPU-hours, Stage~2 takes about 6,400 for AVJ-WM and 225 per run for AVJ-WM-B, and a Stage~3 navigation
fine-tune about 900 for AVJ-WM and 200 for AVJ-WM-B.%

\subsection{Evaluation protocols}
\label{sec:avwmprotocols}

We read out IntPhys-X as described in \cref{sec:suppl_intphysx_protocol}. On IntPhys~v1, v1.1 and IntPhys~2, we
score each clip zero-shot by its surprise on video alone, the next-chunk video prediction error with the learned
missing-audio input in place of the audio, and choose the history length and the pooling over the clip on each
benchmark's development split (the v1.1 development pairs and the IntPhys~2 Debug split). We evaluate the released \mbox{V-JEPA}~2 and 2.1 checkpoints on IntPhys~v1 and v1.1 in the same way, using their own
predictors and $\ell_1$ training loss as the surprise measure. We select the frame rate, window, context and pooling on the
v1.1 development pairs from the grid of Garrido et al.~\cite{garrido2025intuitive}, which includes their fixed setting.
 On InfLevel, we average the
same error over windows and over contexts of 1, 3, 4, 5 and 7 chunks as a fixed rule. We evaluate AVNav-Bench as described
in \cref{sec:suppl_avnav_planner,sec:suppl_avnav_ours} and NWM's datasets as in \cref{sec:supplavwmnwmplanning}, where
ATE and RPE are in waypoint units (meters divided by the dataset's waypoint spacing). Our prediction tests (\cref{sec:supplavwmlongcontext,sec:supplavwmablations}) use one 64\,s window from each of
the 978 long HowTo100M, Ego4D and Ego-Exo4D recordings that we reserved for validation and excluded from all
world-model training.

\subsection{IntPhys v1.1: Fixing a shortcut in IntPhys~v1}
\label{sec:intphysv11}

\paragraph{Problem.}
IntPhys~v1~\cite{riochet2022intphys} renders simple scenes in matched sets of four clips. In the object-permanence
block, for example, the two implausible clips join the beginning of one plausible render to the end of the other, so
that one object turns into two, or vice versa, behind a screen. By design, this event should be the only difference between the
clips, but we found that the two renders do not share their render noise (1--2 gray levels in 255), so the noise switches at the splice point. In
occluded clips (\cref{fig:intphyssplice}b), this switch comes seconds before the screen falls and the violation becomes visible.

\paragraph{Splice-switch detector.}
This switch alone reveals the implausible clip (\cref{fig:intphyssplice}). We verified this with a detector, which finds the frame where the two clips of a pair first differ and calls the clip whose static background changes more at that frame implausible. It has no notion
of objects, motion or physics, yet it identifies the implausible clip in nearly all of the 6,480 pairs; in the remaining 12, the two clips are identical or differ only in their last four frames, which no model reads, so every model is capped at 99.8\%. This finding suggests that scores on the official release largely measure sensitivity to the splice, which would explain the near-ceiling numbers reported on it, including our own.

\paragraph{Fix.}
Our rebuild, IntPhys~v1.1, keeps the same 6,480 pairs, properties and paired scoring and changes only pixels.
Wherever a pixel of the implausible clip lies within 3 gray levels of its plausible counterpart, we copy the
counterpart's pixel over, so the render noise is continuous across the splice and only the physical event differs
(\cref{fig:intphyssplice}).

\paragraph{Results.}
We report v1 for completeness but read only v1.1 as
physics. On v1.1 (\cref{tab:avwmphysicsbreakdown}), AVJ-WM reaches the highest accuracy (\nWMvOne\%) from its video
prediction error alone, without benchmark-specific training, ahead of both \avjepa encoders and the released video
models and close to the AV-CDiT pretrained on the same data.
On IntPhys~2 (506 pairs), all these models stay inside the coin-flip band, whereas frontier multimodal models score well
above it in a harder, non-pairwise setting (GPT-6~Astra~\cite{openai2026astra} 77.4\%, Gemini-4~Argon
\cite{googledeepmind2026gemini4argon} 74.6\% on the CAIS AI Dashboard~\cite{cais2025dashboard}).

\subsection{InfLevel}
InfLevel-Lab~\cite{weihs2022inflevel} tests continuity, solidity and gravity over longer occlusions, with 1,116, 450
and 1,182 pairs (\cref{tab:avwmphysicsbreakdown}). Solidity and gravity stay inside the coin-flip band for every model.
On continuity, world models on the \avjepa-L encoder fall behind, although both encoders alone reach 78\%. AVJ-WM reaches
65.6\% against 73.2\% for AVJ-WM-B, and all three predictor sizes on the \avjepa-L encoder stay at 65.5--66.3\%
(\cref{fig:avwmladder}a), whereas WM-L and WM-H on the \avjepa-B encoder reach 73.7 and 73.8\%.%

\subsection{Visual planning}
\label{sec:supplavwmnwmplanning}
\label{sec:suppl_avwm_nwm_eval}%

To test planning on real robot video, we evaluate on NWM's four datasets with NWM's protocol
(\cref{tab:avwmplanning}).

\paragraph{Setup.}
We use NWM's test cases, four context frames, planner and candidate priors. Our models and the CDiTs on our latents score candidates by the latent distance between the predicted
final frame and the goal image, NWM's CDiTs by LPIPS on decoded frames. We fine-tune our models with Stage~3 for 10,000
updates on NWM's training datasets, SCAND~\cite{karnan2022scand}, TartanDrive~\cite{triest2022tartandrive},
RECON~\cite{shah2021recon} and HuRoN~\cite{hirose2023sacson}, with NWM's splits and our AVNav-Bench hyperparameters
(\cref{sec:suppl_avnav_ours}) without the sound terms, and fine-tune the CDiTs on our latents with their own recipe (batches of 512 samples, 10,000 updates). The videos
have no audio, so our models use the learned missing-audio input, and the pretrained AV-CDiT hears the encoding of
silence.
We retrain GNM and NoMaD with the authors' code. To reach NWM's eight waypoints, we hold the last of GNM's five.
For HuRoN, we use the public release, which lacks 23 of NWM's 100 test cases and has a lower resolution than NWM's. For the perfect cost, we rank our planner's 120 candidates by their true trajectory error and average the best five.

\paragraph{Results.}
Our world models plan better than NWM's released CDiTs on SCAND and on TartanDrive, where the CDiTs do no better
than walking straight ahead. On RECON, the released CDiTs achieve a slightly lower ATE, but AVJ-WM has the lowest relative
trajectory error (RPE). On HuRoN, we found that NWM's candidate prior is centered on standing still, so no planner that chooses among
these candidates beats walking straight ahead, not even with the perfect cost; our models still lead the released CDiTs
there. CDiTs trained on our latents trail our models on RECON and TartanDrive, and the pretrained AV-CDiT has the lowest ATE on SCAND.%
GNM and NoMaD, which predict their paths directly, do beat walking straight ahead on HuRoN, and have a far lower ATE
than every world model there and on TartanDrive but a higher one than NWM's CDiTs and ours on RECON and SCAND.

\subsection{Use of 64s context and the in-context scenarios}
\label{sec:supplavwmlongcontext}
\label{sec:suppl_avwm_context}%
AVJ-WM stays at chance on the in-context scenarios of IntPhys-X (\mainref{tab:intphysx_results}) and almost
never reaches goals in other rooms on AVNav-Bench (\cref{sec:suppl_avnav_analysis}), although both may need
information from more than a few seconds earlier. We therefore test whether it uses its 64\,s context window, on held-out clips from HowTo100M and Ego4D (\cref{sec:avwmprotocols}).

\paragraph{Returning to a scene.} In the first test, a clip shows 8\,s of a scene, then 0--48\,s of other
footage, then the scene's continuation. In a control clip, an unrelated recording replaces the first 8\,s, so the model
never sees the scene. The drop in prediction error on the continuation, relative to the control, measures how much the
model uses the scene it saw earlier (\cref{fig:avwmreturn}). For AVJ-WM, imagining the continuation 4\,s ahead is 10.6\% more accurate after a 2\,s gap and still 8.4\% after 48\,s. To isolate the role of the training window, we compare two pairs of earlier WM-B variants, which lack horizon heads, augmentation and full-resolution targets. One pair is trained with and one without the
rollout loss, and within each pair the two models differ mainly in their training window (64\,s or 10\,s). The models trained on 64\,s windows keep the gain after 48\,s (9.9\% with and 1.9\% without the rollout loss), whereas those trained on 10\,s windows lose it once the scene leaves their
10\,s of history. This suggests that the 64\,s training windows are what lets the model use older history.

\makeatletter\def\ftype@table{2}\makeatother%
\begin{table}[t]
  \color{black}
  \captionsetup{font={color=black}, labelfont={color=black}}
  \centering
  \caption{\textbf{Use of history older than 10\,s.} Older history helps mainly at returned views, and most for the longer-trained AVJ-WM. Reduction (\%) of the next-chunk video prediction error with the whole history of the 978 held-out clips instead of their last 10\,s (\cref{sec:supplavwmlongcontext}), at returned views (1.3--1.5\% of chunks) and at views seen in the last 10\,s. WM-B, WM-L and WM-H use the short schedule (\cref{tab:avwmhyper}). Three seeds for AVJ-WM-B, one for the other models.}
  \label{tab:avwmmemory}
  \footnotesize
  \setlength{\tabcolsep}{3pt}
  \begin{tabular}{@{}l l c c c@{}}%
    \toprule
    Model & Encoder & \begin{tabular}[b]{@{}c@{}}Returned\\views\end{tabular} & \begin{tabular}[b]{@{}c@{}}Seen\\recently\end{tabular} & Difference \\
    \midrule
    AVJ-WM-B & \avjepa-B & 2.3 & 0.1 & 2.2 \\
    WM-B & \avjepa-L & 2.1 & 0.1 & 2.0 \\
    WM-L & \avjepa-L & 2.3 & 0.1 & 2.1 \\
    WM-H & \avjepa-L & 2.5 & 0.3 & 2.2 \\
    \oursrow AVJ-WM & \avjepa-L & 3.5 & 0.5 & 3.0 \\
    \bottomrule
  \end{tabular}
\end{table}
\begin{table*}[t]
  \color{black}
  \captionsetup{font={color=black}, labelfont={color=black}}
  \centering
  \caption{\textbf{AVJ-WM pretraining ablations.} We observe the largest effects on end-goal success, while classic physics changes little. Each row changes one part of AVJ-WM-B's Stage~2 recipe; colored numbers give the change. IntPhys-X: means over its real and synthetic scenarios; image goal 2--6\,s: mean over the 2, 4 and 6\,s goals; \emph{Stable}: stable/run fine-tunes (\cref{sec:supplavwmablations}). IntPhys~2 and IntPhys-X stay near chance in every row, so we omit their changes. $^\ast$Also without the consistency term.}
  \label{tab:avwmablations}
  \footnotesize
  \setlength{\tabcolsep}{4pt}%
  \resizebox{\linewidth}{!}{%
  \begin{tabular}{@{}l c c c c c c c c c c@{}}%
    \toprule
    & & & & \multicolumn{2}{c}{IntPhys-X} & \multicolumn{4}{c}{AVNav-Bench} & \\
    \cmidrule(lr){5-6} \cmidrule(lr){7-10}
    & & & & & & \multicolumn{2}{c}{Image goal} & Sound goal & Closed loop & \\
    \cmidrule(lr){7-8} \cmidrule(lr){9-9} \cmidrule(lr){10-10}
    Variant & IntPhys~v1.1 & IntPhys~2 & InfLevel & Real & Synth. & 2--6\,s & End & End & 2\,s & Stable \\
    \midrule
    \oursrow AVJ-WM-B & 66.0 & 49.3 & 58.2 & 47.1 & 47.4 & 75.5 & 32.4 & 18.0 & 93.7 & 5/5 \\%
    \quad without the consistency term & 66.2~\dpos{$+$0.2} & 48.9 & 57.9~\dneg{$-$0.3} & \na & \na & 72.8~\dneg{$-$2.7} & 31.3~\dneg{$-$1.1} & 11.6~\dneg{$-$6.4} & 93.5~\dneg{$-$0.2} & 3/5 \\%
    \quad without rollouts & 65.6~\dneg{$-$0.4} & 49.3 & 58.3~\dpos{$+$0.1} & 48.2 & 47.8 & 70.9~\dneg{$-$4.6} & 30.1~\dneg{$-$2.3} & 13.2~\dneg{$-$4.8} & 93.4~\dneg{$-$0.3} & 5/5 \\%
    \quad without the horizon heads & 65.8~\dneg{$-$0.2} & 48.8 & 57.8~\dneg{$-$0.4} & 47.5 & 47.4 & 76.0~\dpos{$+$0.5} & 32.6~\dpos{$+$0.2} & 15.3~\dneg{$-$2.7} & 93.5~\dneg{$-$0.2} & 3/5 \\%
    \quad without the reconstruction term & 65.7~\dneg{$-$0.3} & 49.3 & 57.9~\dneg{$-$0.3} & 49.7 & 47.8 & 75.8~\dpos{$+$0.3} & 34.7~\dpos{$+$2.3} & 14.2~\dneg{$-$3.8} & 93.8~\dpos{$+$0.1} & 5/5 \\%
    \quad without the VICReg terms & 65.8~\dneg{$-$0.2} & 50.0 & 58.2~\dnul{$\pm$0.0} & 50.2 & 47.9 & 75.5~\dnul{$\pm$0.0} & 32.5~\dpos{$+$0.1} & 16.1~\dneg{$-$1.9} & 93.4~\dneg{$-$0.3} & 5/5 \\%
    \quad with 10\,s instead of 64\,s windows & 66.3~\dpos{$+$0.3} & 48.5 & 58.4~\dpos{$+$0.2} & 50.4 & 47.3 & 76.4~\dpos{$+$0.9} & 27.3~\dneg{$-$5.1} & 16.9~\dneg{$-$1.1} & 92.5~\dneg{$-$1.2} & 2/5 \\%
    \quad video only & 68.2~\dpos{$+$2.2} & 48.7 & 58.3~\dpos{$+$0.1} & \na & \na & 74.7~\dneg{$-$0.8} & 36.1~\dpos{$+$3.7} & \na & 92.7~\dneg{$-$1.0} & 5/5 \\%
    \quad audio only & \na & \na & \na & \na & \na & \na & \na & 4.1~\dneg{$-$13.9} & \na & 5/5 \\%
    \quad packing instead of the combiner$^\ast$ & 65.9~\dneg{$-$0.1} & 48.7 & 57.5~\dneg{$-$0.7} & \na & \na & 73.7~\dneg{$-$1.8} & 33.3~\dpos{$+$0.9} & 11.8~\dneg{$-$6.2} & 93.1~\dneg{$-$0.6} & 5/5 \\%
    \quad pooled targets (49 video, 5 audio tokens) & 66.9~\dpos{$+$0.9} & 50.1 & 57.4~\dneg{$-$0.8} & 49.7 & 48.0 & 73.0~\dneg{$-$2.5} & 30.4~\dneg{$-$2.0} & 17.9~\dneg{$-$0.1} & 91.6~\dneg{$-$2.1} & 4/5 \\%
    \quad 32 instead of 64 combiner latents & 63.6~\dneg{$-$2.4} & 48.7 & 57.5~\dneg{$-$0.7} & 48.0 & 48.3 & 71.7~\dneg{$-$3.8} & 27.4~\dneg{$-$5.0} & 11.3~\dneg{$-$6.7} & 93.4~\dneg{$-$0.3} & 5/5 \\%
    \quad 128 instead of 64 combiner latents & 67.3~\dpos{$+$1.3} & 49.3 & 58.0~\dneg{$-$0.2} & 50.3 & 47.3 & 76.7~\dpos{$+$1.2} & 34.0~\dpos{$+$1.6} & 12.8~\dneg{$-$5.2} & 94.1~\dpos{$+$0.4} & 4/5 \\%
    \bottomrule
  \end{tabular}}
  
  \vspace{-0.1cm}%
\end{table*}
\begin{table*}[t]
  \color{black}\captionsetup{font={color=black}, labelfont={color=black}}
  \centering
  \caption{\textbf{Prediction and representation without each auxiliary term.} We observe that prediction degrades with the horizon without rollouts, that the chunk states lose rank without the consistency or VICReg terms, and that they keep less of the input sound without reconstruction. Each row removes one auxiliary term from AVJ-WM-B's Stage~2 recipe, as in \cref{tab:avwmablations}. Prediction gain: over copying the last chunk; rank: effective rank of the input tokens at the end of training and of the chunk states (per latent and whole); latent content: encoder-token variance that a linear probe recovers from the chunk states. Rank and content are changes (\%)
    relative to a run of AVJ-WM-B's recipe. 978 held-out clips, three seeds. $^\dagger$In one forward
    pass. $^\ddagger$Not comparable, because the state rank changes.}
  \label{tab:avwmtermstats}
  \footnotesize
  \setlength{\tabcolsep}{3.5pt}
  \begin{tabular}{@{}l cccc cc cc cc@{}}
    \toprule
    & \multicolumn{4}{c}{Prediction gain (\%)} & \multicolumn{2}{c}{Input rank} & \multicolumn{2}{c}{State rank} & \multicolumn{2}{c}{Latent content} \\
    \cmidrule(lr){2-5} \cmidrule(lr){6-7} \cmidrule(lr){8-9} \cmidrule(l){10-11}
    Variant & 0.5\,s & 2\,s & 4\,s & 8\,s & Video & Audio & Latent & Whole & Video & Audio \\
    \midrule
    \oursrow AVJ-WM-B & 18.5 & 21.6 & 22.2 & 23.0 & \na & \na & \na & \na & \na & \na \\
    \quad without the consistency term & 19.4~\dpos{$+$0.9} & 21.8~\dpos{$+$0.2} & 22.4~\dpos{$+$0.2} & 23.1~\dpos{$+$0.1} & \textcolor{tabloss}{$-$0.3} & \textcolor{tabgain}{$+$0.2} & \textcolor{tabloss}{$-$16.3} & \textcolor{tabloss}{$-$53.0} & \na$^\ddagger$ & \na$^\ddagger$ \\
    \quad without rollouts & 18.9~\dpos{$+$0.4} & 13.8~\dneg{$-$7.8} & 7.2~\dneg{$-$15.0} & 0.7~\dneg{$-$22.3} & \textcolor{tabgain}{$+$0.4} & \textcolor{tabgain}{$+$0.2} & \textcolor{tabloss}{$-$0.4} & \textcolor{tabgain}{$+$3.0} & \textcolor{tabloss}{$-$0.4} & \textcolor{tabloss}{$-$1.4} \\
    \quad without the horizon heads & 18.9~\dpos{$+$0.4} & 21.5~\dneg{$-$0.1} & 22.2~\dnul{$\pm$0.0} & 22.9~\dneg{$-$0.1} & \textcolor{tabgain}{$+$0.3} & \textcolor{tabgain}{$+$0.5} & \textcolor{tabgain}{$+$1.8} & \textcolor{tabgain}{$+$5.4} & \textcolor{tabloss}{$-$0.9} & \textcolor{tabgain}{$+$0.1} \\
    \quad predicting with the horizon heads$^\dagger$ & \na & 21.5~\dneg{$-$0.1} & 21.9~\dneg{$-$0.3} & 22.6~\dneg{$-$0.4} & \na & \na & \na & \na & \na & \na \\
    \quad without the reconstruction term & 18.2~\dneg{$-$0.3} & 21.4~\dneg{$-$0.2} & 22.1~\dneg{$-$0.1} & 22.9~\dneg{$-$0.1} & \textcolor{tabgain}{$+$0.3} & \textcolor{tabgain}{$+$0.8} & \textcolor{tabgain}{$+$0.7} & \textcolor{tabgain}{$+$1.3} & \textcolor{tabloss}{$-$1.3} & \textcolor{tabloss}{$-$5.6} \\
    \quad without the VICReg terms & 18.8~\dpos{$+$0.3} & 21.8~\dpos{$+$0.2} & 22.4~\dpos{$+$0.2} & 23.1~\dpos{$+$0.1} & \textcolor{tabloss}{$-$44.2} & \textcolor{tabloss}{$-$51.1} & \textcolor{tabloss}{$-$5.2} & \textcolor{tabloss}{$-$18.3} & \na$^\ddagger$ & \na$^\ddagger$ \\
    \bottomrule
  \end{tabular}
\end{table*}
\makeatletter\def\ftype@table{1}\makeatother
\paragraph{Revisits in natural video.} The second test checks unedited video. We predict every chunk of the 978 held-out clips twice, from the whole history and from only the last 10\,s (\cref{tab:avwmmemory}). We mark a chunk as a
\textit{returned view} when its most similar earlier chunk, by the cosine similarity of the encoder's mean-pooled video tokens,
lies more than 10\,s back and is clearly more similar (by at least 0.05) than any chunk of the last 10\,s. Returned views are rare (1.3--1.5\% of chunks), so the whole history barely changes the average error. At returned views, however, it lowers
AVJ-WM's video prediction error by 3.5\%, about as much as the last three quarters of training lower it on these clips
(2.9\%), against 0.5\% at views also seen in the last 10\,s. Both tests show that AVJ-WM uses history well beyond the
last 10\,s, which suggests that memory alone does not explain its failures above.%

\paragraph{In-context scenarios.} We hypothesize that the in-context scenarios test matching rather than
prediction. Within one clip, the model has to bind each sound to its event (a light of the Simon device, a move on the
grid, a cup swap) and keep that binding while only the sound continues, which next-chunk prediction on natural video
rarely requires. We found that swapping an item's context for another item's changes the model's choices, which confirms it uses the context. Yet, the model expects
sounds heard earlier in the clip rather than the sound of the correct light or move, and both endings consist of such
sounds. This suggests that binding, not memory, is the obstacle.

\subsection{Pretraining ablations}
\label{sec:supplavwmablations}

We scale the world model in size and training length (\cref{fig:avwmladder}) and remove one
part of the Stage~2 recipe at a time from AVJ-WM-B (\cref{tab:avwmablations}).

\paragraph{Setup.} We train WM-B, WM-L and WM-H (\cref{tab:avwmconfigs}) with our short schedule. For training
length, we cool down the learning rate at several points of the main runs (a warmup-stable-decay schedule~\cite{hu2024minicpm,hagele2024scaling}); hollow circles in
\cref{fig:avwmladder}b mark checkpoints before a cool-down. For the navigation columns, we fine-tune every model with the same Stage~3 recipe
(\cref{sec:suppl_avnav_ours}). Unless noted otherwise, we average prediction and physics over three pretraining seeds and navigation over the
stable fine-tunes among the same five seed pairs, and use one pretraining run with three fine-tunes for AVJ-WM;
\emph{Stable} in \cref{tab:avwmablations} gives how many of the five train stably under our Stage~3 recipe. Error bars show the standard deviation over these seeds or fine-tunes. We count a fine-tune as unstable when its held-out prediction loss ends
more than 0.03 above its minimum or its median gradient norm over the last 2,000 updates exceeds 1. We found that these
two training signals identify diverging fine-tunes well, without using navigation success.%

\paragraph{Prediction.} Prediction improves with size and training. On the next 0.5\,s of the 978 held-out clips (\cref{sec:avwmprotocols}), the gain over copying the last chunk (one minus
the ratio of the two errors; comparable only within an encoder) rises from 21.0 to 22.8\% between WM-B and WM-H and
from 20.4 to 26.4\% over training.

\paragraph{Physics.} Cross-modal physics improves mostly with training. On IntPhys-X, with the readout of \mainref{tab:intphysx_results}, accuracy on its four real scenarios rises from 52\% after 4,444 updates to 69\% at the end of training, mainly on count and synchronization, but only from 53 to 56\% between WM-B and WM-H on the shorter schedule; the synthetic scenarios stay at or slightly below chance (47--49\%).  IntPhys~v1.1
improves slightly with training (67 to 71\%), and IntPhys~2 and InfLevel stay close to chance.

\paragraph{Planning.} Scaling benefits planning less clearly. While several measures stay stable across sizes and training lengths, we observe higher
success rates on image end goals with further training. On
AVNav-Bench, mean offline success on the 2, 4 and 6\,s image goals goes from 77.0 to 78.5\% between WM-B and WM-H, and over training it falls from 79.1\% to 75.8\%. One reason may be the goal cost. With exact predictions (dashed in \cref{fig:avwmladder}), our planner and goal cost reach 80.5\% on these goals with the \avjepa-L encoder and 80.8\% with \avjepa-B (\cref{tab:avnavceiling}), so the larger world models are within 2--3 points of what the cost allows. Another reason may be that every model
uses the same Stage~3 recipe, which we developed on the base size to save compute and which may therefore favor it.

On RECON, the ATE stays at 1.19--1.27 in NWM's waypoint units. We score how well each model's cost ranks the planner's candidates, from 0 (its five cheapest
are no better than five random ones) to 1 (they are the five closest to the true path). Every model scores 0.50--0.55,
and NWM's released CDiT-S 0.53, which suggests that on RECON it is also the cost rather than the world model  that limits planning.
Pretraining itself matters, however. Without it, AVJ-WM-B reaches only 27.8\% at 2\,s (\mainref{sec:avwmplanningsec}).

\paragraph{Recipe.}
Sound goals are the most sensitive to the Stage~2 recipe. Almost every change in \cref{tab:avwmablations} lowers their success, for instance, by ${\sim}$6--7 points without the consistency term or with half the combiner's latents. Classic physics, in contrast,
changes by about a point or less, except with video only and with 32 latents (about 2 points on IntPhys~v1.1).
On image goals at 2--6\,s, success drops by 4.6 points without rollouts and by 3.8 points with half the combiner's
latents. Several variants also fine-tune less reliably under our Stage~3 recipe (\emph{Stable}). Without the horizon heads,
with 10\,s windows or with 128 latents, their stable fine-tunes match AVJ-WM-B on these goals. On image end goals, 37 steps
away, 10\,s windows and half the latents each cost about 5 points, which suggests that distant goals benefit most from long
windows and the combiner's capacity. On IntPhys~2 and
IntPhys-X, all rows stay near chance after the shorter schedule (47--50\%), so these columns do not meaningfully separate the model variants.

\paragraph{Auxiliary terms.}
We also examine what the auxiliary terms change beyond the benchmarks (\cref{tab:avwmtermstats}). Without the rollout loss, the gain of the model's own rollouts over copying the last chunk falls from 22.2 to 7.2\% at 4\,s and to 0.7\% at 8\,s, which suggests that this loss limits the accumulation of errors. The horizon heads seem to offer a much
cheaper route to long-range predictions. In a single forward pass, they predict up to 8\,s ahead and match the
step-by-step rollouts, which need up to 16 passes, to within 0.4 points. Keeping them costs neither rollout nor benchmark
accuracy (\cref{tab:avwmablations,tab:avwmtermstats}). Similar to how such heads speed up
decoding in language models~\cite{gloeckle2024multitoken}, we hypothesize that our horizon heads could help plan across longer distances (in fewer steps) when conditioned on actions (\cref{sec:suppl_limitations}). Without the VICReg terms, the effective rank of the input tokens roughly halves by the end of training, and that of the whole chunk states on held-out clips drops by 18\%.  Without the consistency term,
the effective rank of the whole chunk states halves while that of the input tokens is unchanged, and sound-goal success
drops by 6.4 points. We kept the reconstruction term from an earlier recipe. We found that removing it changes neither the rank nor the prediction gain, but less of the current chunk's encoder tokens can be recovered from the chunk states (with a linear probe), especially for sound. In line with this finding, we observe that sound-goal
success drops by 3.8 points without it, while image end goals rise by 2.3 (\cref{tab:avwmablations}).%

\makeatletter\def\ftype@table{2}\makeatother%
\FloatBarrier%
}
\FloatBarrier
\vspace{-0.1cm}
\section{IntPhys-X benchmark}
\label{sec:suppl_intphysx}
\vspace{-0.1cm}

IntPhys-X is an evaluation benchmark of \emph{intuitive physics that can only be resolved through cross-modal
inference}. In every item, the evidence that determines the answer is split between sight and sound, so neither stream
alone suffices: a model must correlate, align and jointly reason over the two. Each scenario isolates one physical
regularity that links what is heard to what is seen. This appendix describes how the evaluation set was built (\cref{sec:suppl_intphysx_construction}), its
scenarios (\cref{sec:suppl_intphysx_scenarios}), %
the evaluation protocol (\cref{sec:intphysx_protocol}) and the human study (\cref{sec:protocol_subjective}).

\begin{table*}[t]
  \centering
  \providecommand{\ipxtRR}{\raggedright\arraybackslash}%
  \caption{\textbf{IntPhys-X scenarios and statistics.} Each scenario probes a
    physical regularity that has to be resolved by combining audio and video.
    Synthetic clips are rendered as minimal pairs that are identical except for
    the critical attribute; real clips are recorded. In-context scenarios are
    synthetic. Counts are evaluation examples (practice items excluded).}
  \label{tab:intphysx_tasks}
  \footnotesize
  \setlength{\tabcolsep}{4pt}
  \begin{tabular}{@{}l >{\ipxtRR}p{1.35in} >{\ipxtRR}p{1.8in} >{\ipxtRR}p{1.75in} r r@{}}
    \toprule
    Scenario & What it probes & Cross-modal cue & \leavevmode Distractor selection & \#Synthetic & \#Real \\
    \midrule
    Count           & Event numerosity
                    & Number of heard impacts must match the number of seen impacts
                    & \leavevmode Different number of impacts; the shared impacts are in sync
                    & \nIPXsynthPer & \nIPXrealPer \\
    Sync            & Temporal alignment of individual events
                    & Each impact sound must coincide with the seen contact
                    & \leavevmode Same number of impacts; a single impact out of sync
                    & \nIPXsynthPer & \nIPXrealPer \\
    Size            & Object size from the acoustic signature
                    & Larger objects sound lower (pitch and resonance)
                    & \leavevmode Impacts in sync; different object
                    & \nIPXsynthPer & \nIPXrealPer \\
    Material        & Material or instrument type
                    & Heard timbre must match the seen material
                    & \leavevmode Impacts in sync; different object
                    & \nIPXsynthPer & \nIPXrealPer \\
    Localization    & Position of the sound source
                    & Stereo interaural time and level cues must match the seen position
                    & \leavevmode Impacts in sync; the position of the source is altered
                    & \nIPXsynthPer & \na \\
    \midrule
    Simon           & In-context light--sound binding
                    & Tones must follow the light--tone mapping set up in the context
                    & \leavevmode A single different action
                    & \nIPXlongPer & \na \\
    Random Walk     & In-context binding of characters and moves to sounds
                    & Final positions must follow the sound--move mapping set up in the context
                    & \leavevmode At least one character ends at a different position
                    & \nIPXlongPer & \na \\
    3-Cups          & In-context sequential rollout
                    & The ball's final cup must follow the sound--swap mapping set up in the context
                    & \leavevmode One swap substituted and the sequence reordered
                    & \nIPXlongPer & \na \\
    \bottomrule
  \end{tabular}
  
  \vspace{-0.3cm}
\end{table*}
\begin{figure*}[tp]
  \centering
  \def\ipxEx{\FigRoot/06_intuitive_physics/intphysx_examples/panels}%
  {\small\bfseries Cross-modal matching: synthetic (left) and real (right)}\par\vspace{2pt}
  \includegraphics[width=.485\textwidth]{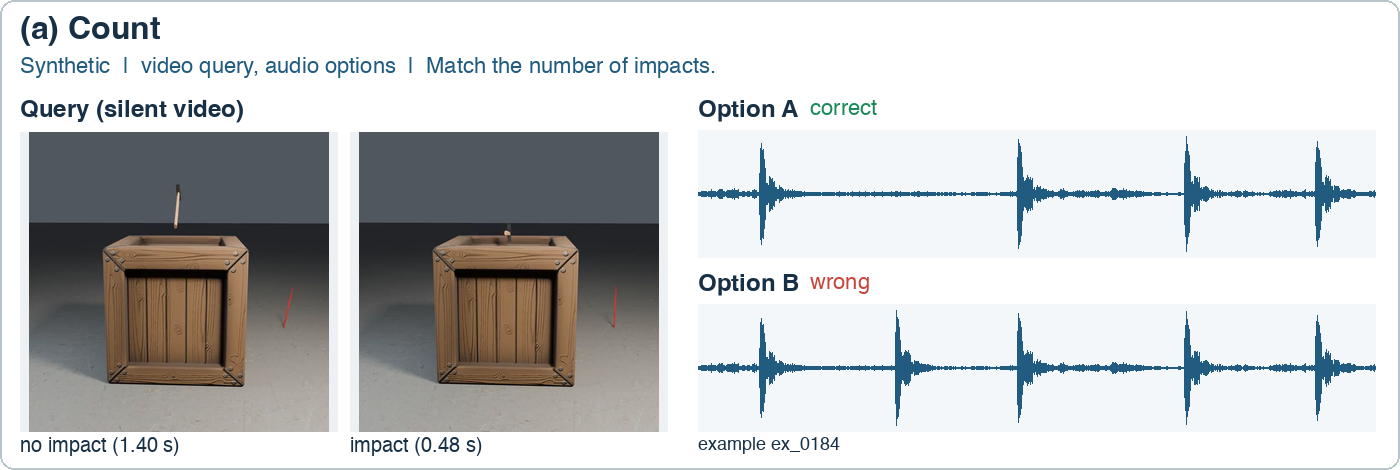}\hfill\includegraphics[width=.485\textwidth]{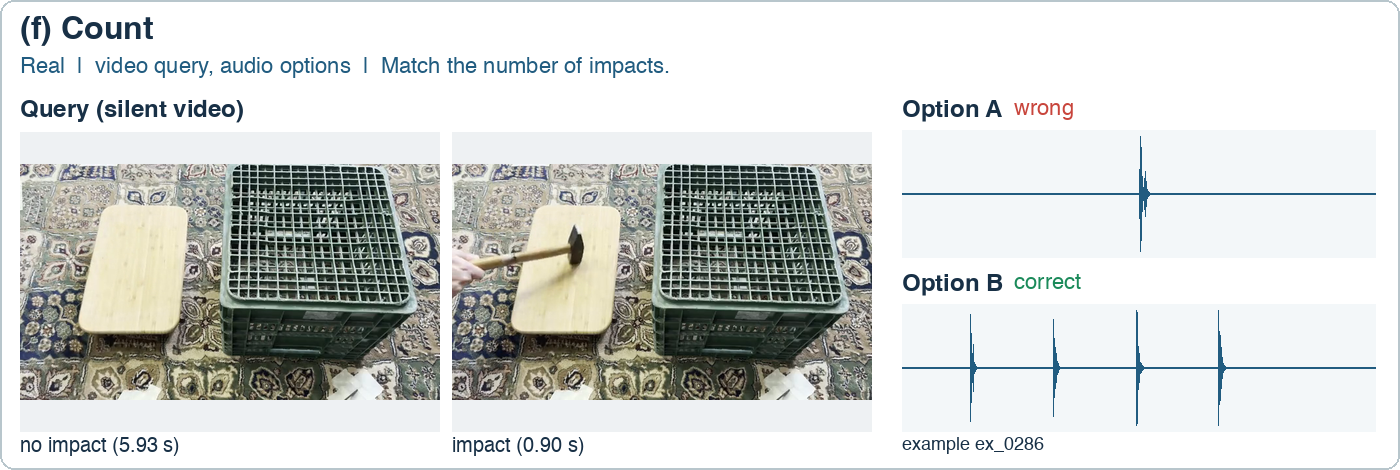}\par\vspace{2pt}
  \includegraphics[width=.485\textwidth]{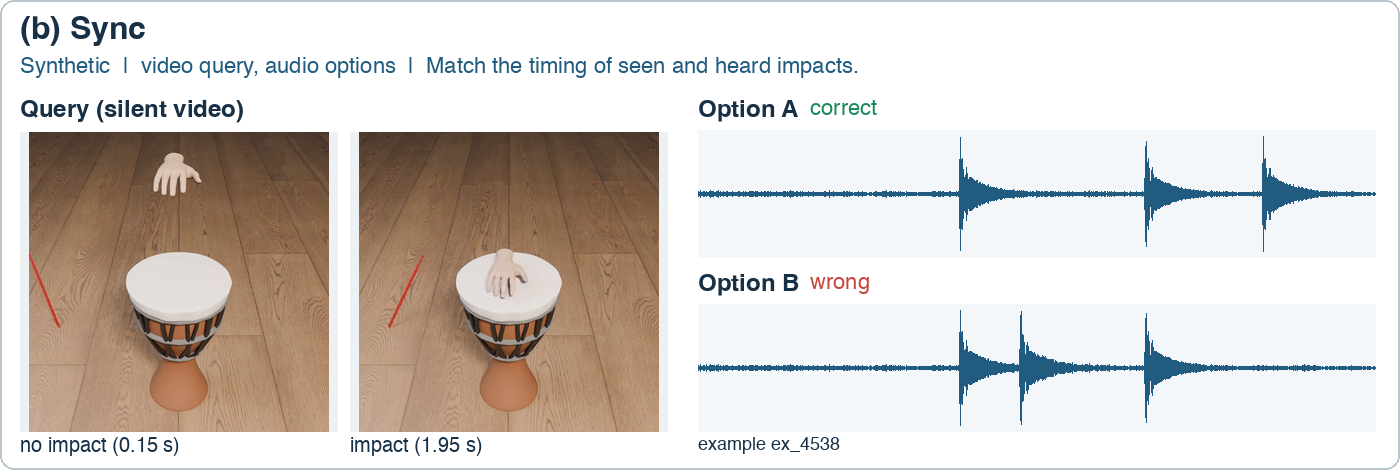}\hfill\includegraphics[width=.485\textwidth]{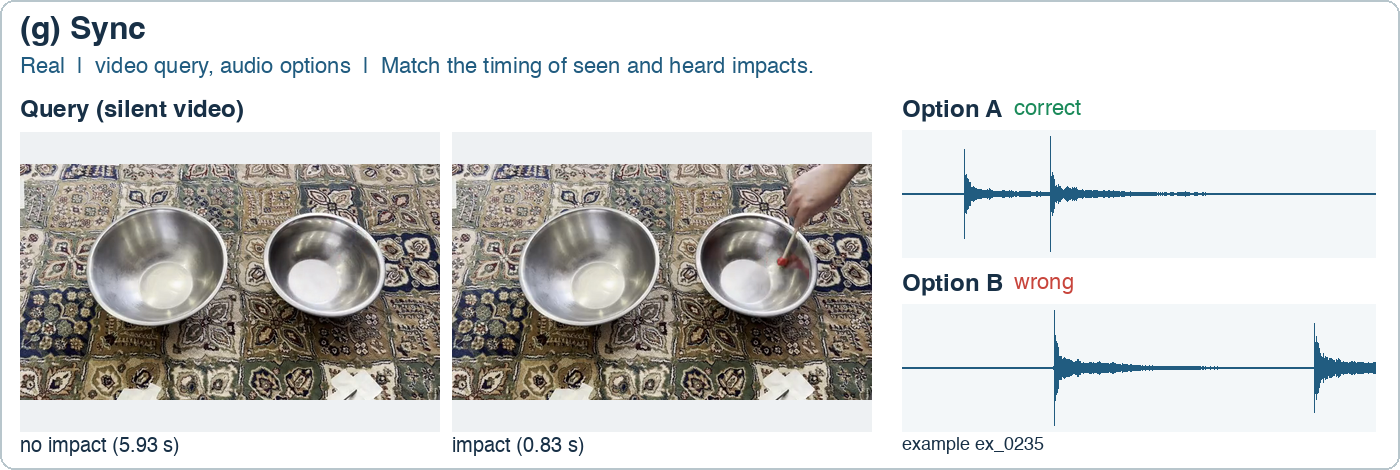}\par\vspace{2pt}
  \includegraphics[width=.485\textwidth]{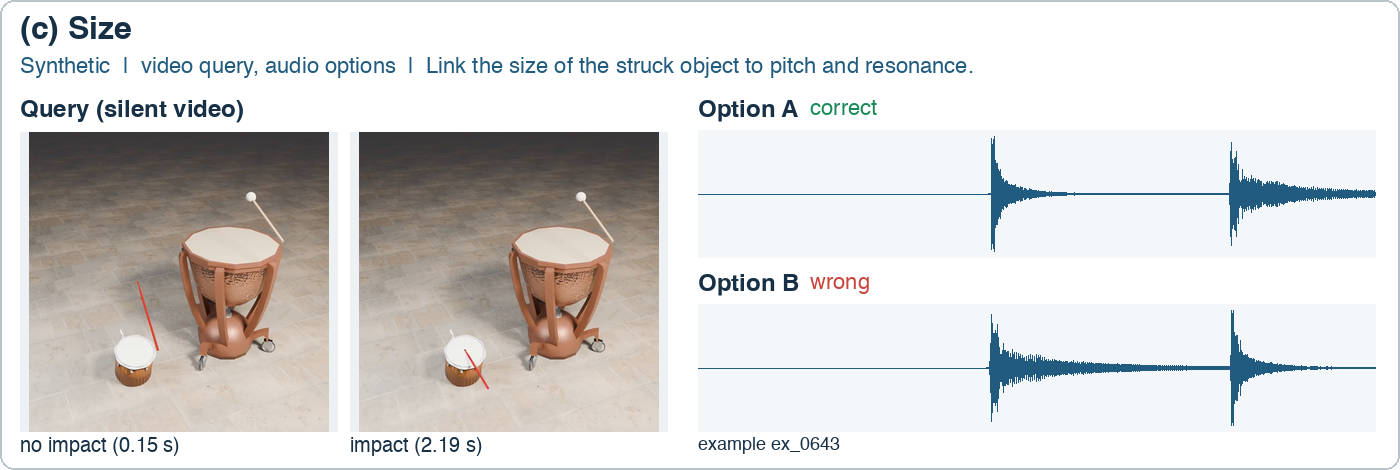}\hfill\includegraphics[width=.485\textwidth]{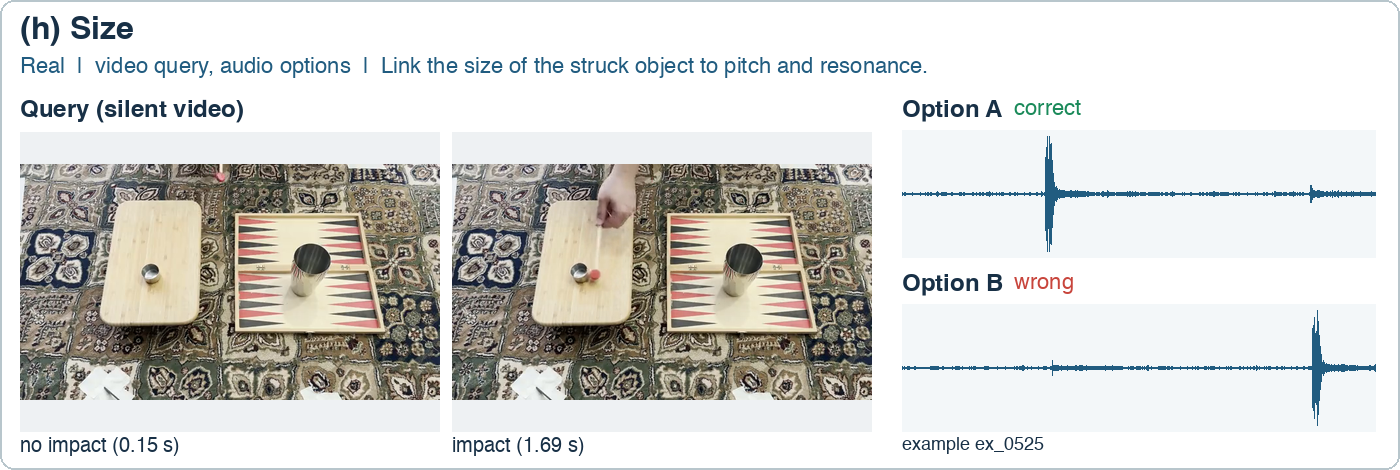}\par\vspace{2pt}
  \includegraphics[width=.485\textwidth]{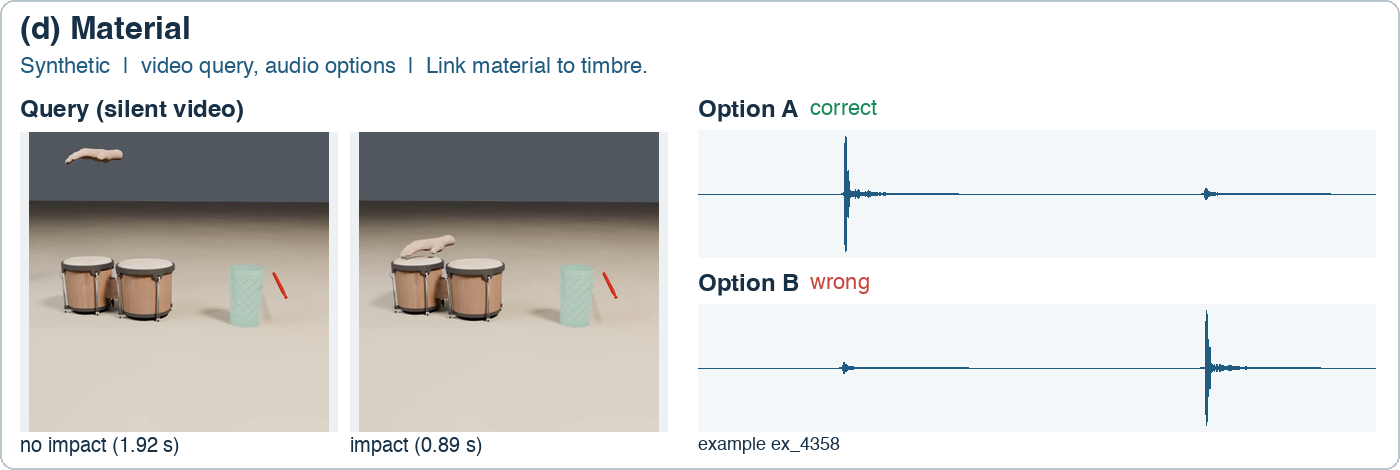}\hfill\includegraphics[width=.485\textwidth]{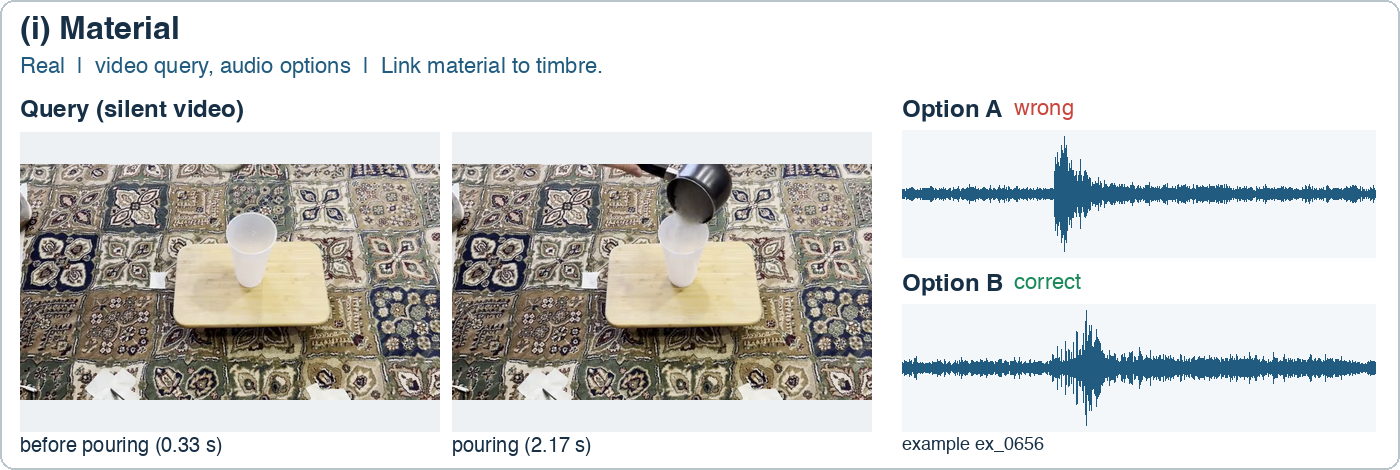}\par\vspace{2pt}
  \includegraphics[width=.485\textwidth]{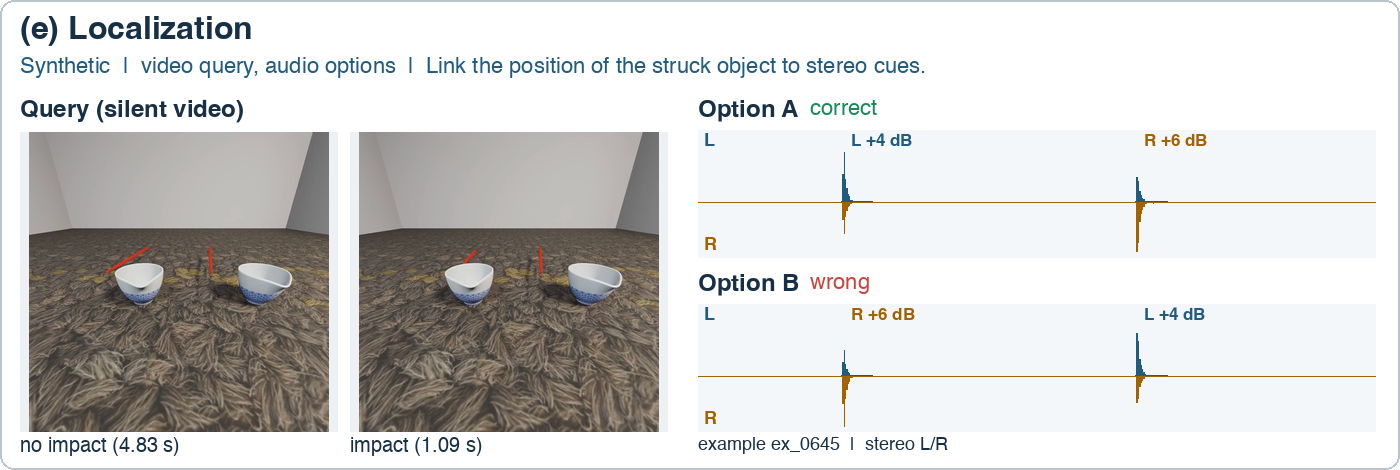}\hfill\includegraphics[width=.485\textwidth]{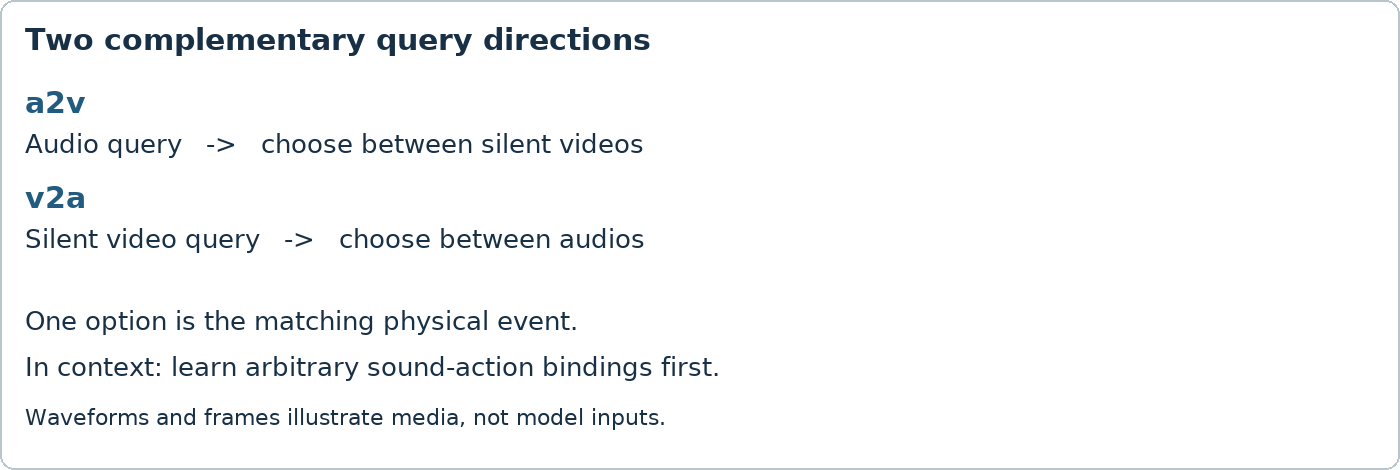}\par\vspace{4pt}
  {\small\bfseries In-context reasoning}\par\vspace{2pt}
  \includegraphics[width=.315\textwidth]{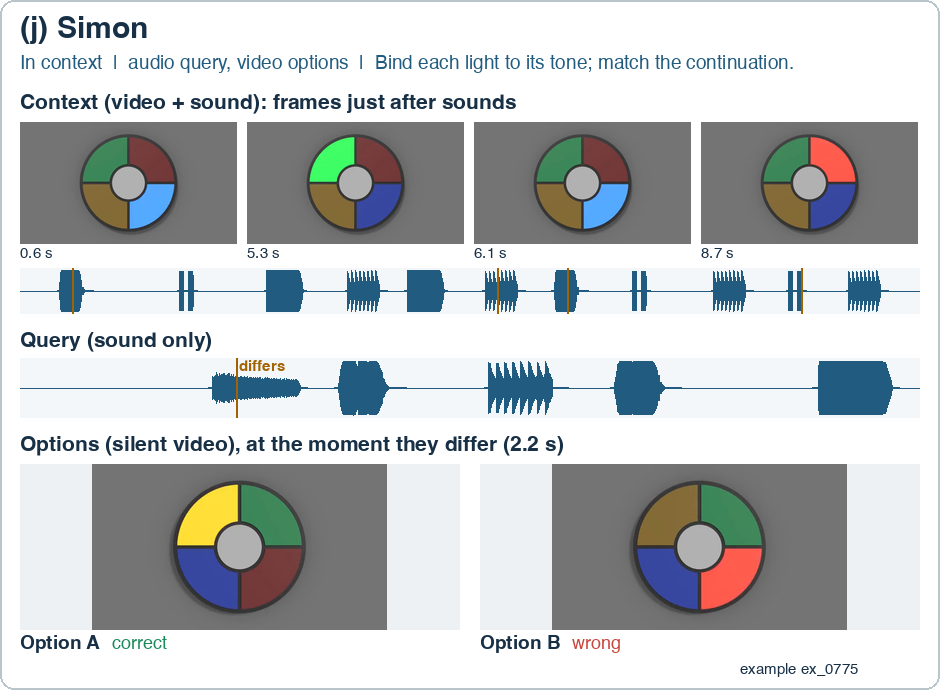}\hfill\includegraphics[width=.315\textwidth]{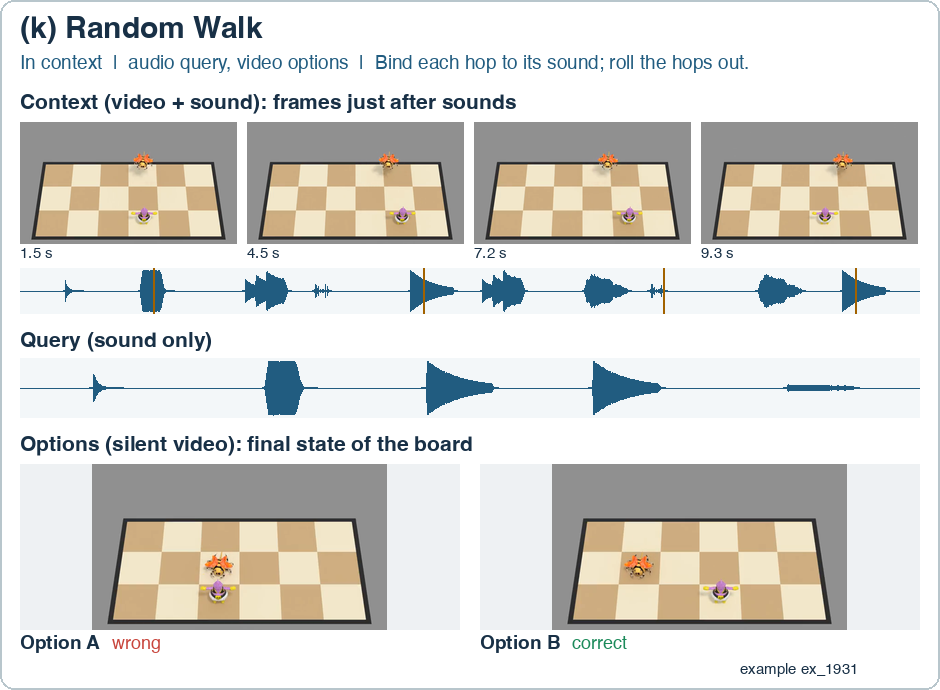}\hfill\includegraphics[width=.315\textwidth]{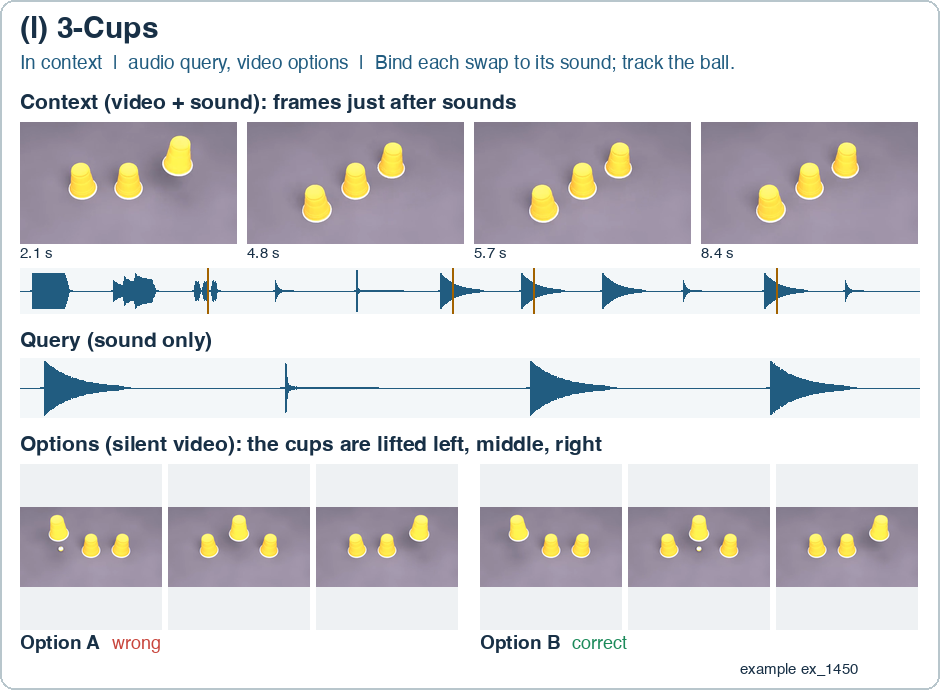}
  \caption{\textbf{\intphysx examples}, one per scenario and source. Each no-context panel shows a silent video
    query at an impact and away from impacts (for the poured real material, before and during the pour) and the
    waveforms of the two audio options; for Localization, the left (up) and right (down) channel envelopes, with the
    level difference of each impact. In-context panels show context frames just after sounds, the query sound and the two options. Examples were chosen for
    diverse objects, backgrounds and bindings, not by model scores.}%
  \label{fig:intphysx_examples}
\end{figure*}

\subsection{\texorpdfstring{Construction of the evaluation set}{Construction of the evaluation set}}
\label{sec:suppl_intphysx_benchmark}
\label{sec:suppl_intphysx_construction}

\paragraph{Minimal pairs and quadruplets.}
Every item is built from a minimal pair of clips $(A,B)$ that are identical except for a single critical attribute
(\eg, two vs.\ three drum strikes). The core constraint is salience: the contrast is made maximally perceptible
in \emph{both} modalities through the choice of assets, framing, lighting, color and audio processing. From each pair we derive a quadruplet: the congruent clips $A$ and
$B$, and the cross-muxed incongruent clips $A_vB_s$ and $B_vA_s$ (the video of one clip with the sound of the other).
Each incongruent clip therefore violates exactly the critical attribute and nothing else.

\paragraph{Task format.}
An item presents a \emph{query} $Q$ in one modality and two \emph{options} $X\in\{A,B\}$ in the other, of which exactly one stems from the same physical event.
\begin{itemize}
  \item \textsc{a2v}: an audio query with two silent video options;
  \item \textsc{v2a}: a silent video query with two audio options.
\end{itemize}
The task is a two-alternative forced choice, so chance is 50\%. Correct letters are balanced per scenario and
direction. Each no-context synthetic scenario contains \nIPXsynthPer{} evaluation examples, each in-context scenario
\nIPXlongPer{} and each real scenario \nIPXrealPer{} (\cref{tab:intphysx_tasks}), split equally between the two
directions, except that Random Walk and 3-Cups have \textsc{a2v} items only.
A small development set of 100 items (25 per direction for the no-context scenarios and for Simon)
provides label-free statistics and is the only data used to calibrate inference
(\cref{sec:intphysx_protocol}) and is released with the benchmark.
Its items are disjoint from the evaluation items, although Random Walk and 3-Cups draw on the same pool of action
sounds in both.
All sets are released at 30\,fps.

\paragraph{Rendering and recording.}
The synthetic clips are rendered with Blender/Cycles, with physically timed impact sounds and fine-grained control
over the timing of each event; each item is a 5\,s clip. The real clips were recorded manually to test whether
conclusions drawn on the synthetic set transfer to real-world data, and both sets use visual and auditory augmentations
for diversity.
\vspace{-0.1cm}
\subsection{\texorpdfstring{Scenarios}{Scenarios}}
\label{sec:suppl_intphysx_scenarios}
\vspace{-0.1cm}
Five scenarios match a query to an option without context, and three require reasoning in context
(\cref{tab:intphysx_tasks}). \Cref{fig:intphysx_examples} shows an example of each.

\paragraph{No-context scenarios.}
\Cref{tab:intphysx_tasks} lists what each scenario probes and how its distractor is built. In Count, Sync, Size and
Material, a striker (hand, drumstick or mallet) hits the objects; some real Material clips instead pour a material
onto them.
Localization is synthetic only: its stereo audio carries interaural time and level cues of the source position.

\paragraph{In-context scenarios.}
These scenarios test cross-modal in-context reasoning.
A context clip \emph{with} sound establishes an arbitrary
binding between visual entities and sounds; a query continues the episode in one modality (audio only in some), and two options in the other modality follow,
one consistent with the binding and the query. The binding appears only in the
context, so it must be learned there. Difficulty increases from memorizing the binding
(Simon), to rolling actions out from sound alone (Random Walk), to rolling out order-dependent actions
(3-Cups; \mainref{fig:intphysx_structure}).

\begin{figure*}[t]
  \centering
  \input{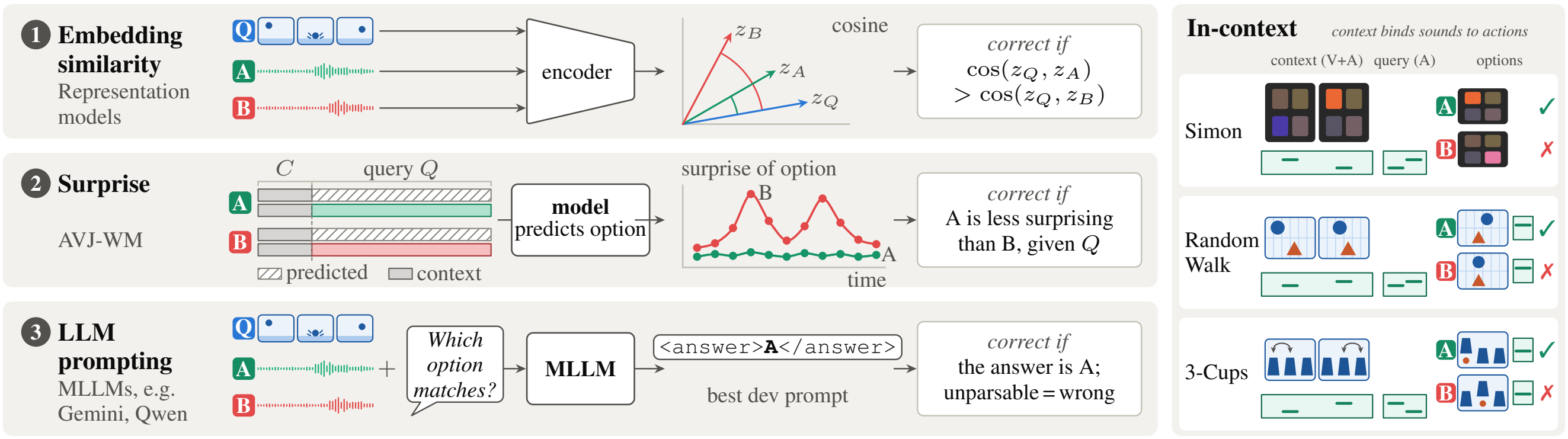}
  \caption{\textbf{IntPhys-X inference.} \emph{Left:} for a \textsc{v2a} item (video query $Q$, audio options $A$, $B$),
    (1)~representation models pick the option whose embedding is more similar to the query's, (2)~AVJ-WM the option
    whose audio is less surprising given $Q$, and (3)~MLLMs answer a prompt with the option's letter. \emph{Right:} in-context items: the
    context binds sounds to actions, an audio query continues it, and the options show the final state (Simon: the query
    in the other modality).}
  \vspace{-0.5cm}
  \label{fig:intphysx_protocols}
\end{figure*}

\paragraph{Simon.}
A front-facing Simon-like device has four colored pads and a central hub, giving five lights, each bound to a
distinct tone. In the 10\,s context, the lights flash one at a time, every 0.8--1\,s, and one light never flashes.
The continuation (query and options) opens with a 1\,s camera rotation by $\pm90^\circ$ or $180^\circ$, so the
binding must be kept with the colors of the lights rather than with their positions. The query and options then follow
the IntPhys-X format: the tones of the continuation must be matched to the lights that produce them, which only the
context reveals. The light that never flashes in the context appears at least once in the query, and the distractor
differs from the true continuation in a single action. Simon thus tests whether the model memorizes and applies an
in-context \emph{light--sound mapping}.

\paragraph{Random Walk.}
Two characters of different types stand on a $6\times3$ board and hop left or right, one at a time. Each character has
its own sound for a leftward hop and another for a rightward hop. A 12\,s context shows every (character, direction)
combination at least twice. The query is a \emph{blind} continuation: its audio only, with no video. It may also
contain, at most once, an action not seen in the context: a new sound with which one of the characters moves up or down
a row. The options are videos of the final state of the board in which each character hops in place with its sound; in
the distractor, at least one character ends at a different position, \ie, the net number of left and right hops
differs for at least one character. The model must therefore deduce the sound--action binding from the context and roll
the hops out to the characters' final positions, although counting each character's left and right hops and adding
them to its starting position also suffices.

\paragraph{3-Cups.}
Three cups hide a ball. Six actions, lifting each cup and swapping each of the three cup pairs, each have their
own sound. As in Random Walk, the query is
an audio-only continuation and the options are 2\,s animations that lift the cups one by one (left, middle,
right) and reveal where the ball ends up. Unlike
hops, swaps do not commute: the ball's final position depends on their order, so counting is
insufficient and a rollout is required.
The 10\,s context contains 8--12 actions: each of the three lifts and the three swap types (left--middle,
middle--right, left--right) at least once, followed by 2--6 further swaps. The 5\,s query contains 3--4 swaps. Each
episode binds the six actions to sounds drawn at random from 23 sound groups (93 variants), either synthesized by us
or from FSD50K~\cite{fonseca2022fsd50k}. The wrong option substitutes one swap
and reorders the sequence. The camera is fixed (70\,mm lens, 52\textdegree{} elevation).

\subsection{Evaluation protocol}
\label{sec:suppl_intphysx_protocol}
\label{sec:intphysx_protocol}

Items are two-alternative forced choices between options $X\in\{A,B\}$ for a query $Q$
(\cref{sec:suppl_intphysx_construction}, \cref{fig:intphysx_protocols}); in-context items add a context clip $C$.
No model is trained or fine-tuned on IntPhys-X. Representation models are scored by cross-modal similarity, the
world model by surprise and multimodal LLMs (MLLMs) by prompting (\cref{sec:ipx_rep,sec:ipx_wm,sec:ipx_mllm}), after
the preprocessing of \cref{sec:ipx_prep}.
Each evaluation method $m$ defines a score $s_m(Q,X)$ and predicts
\begin{equation}
  \hat X = \argmax_{X\in\{A,B\}} s_m(Q,X).
  \label{eq:intphysx_choice}
\end{equation}
Each model's free choices (a layer, a readout or a prompt) are made only on
the development set $\mathcal{D}_{\mathrm{dev}}$, which is disjoint from the evaluation set $\mathcal{D}_{\mathrm{eval}}$
and released with the benchmark:
\begin{equation}
  m^\star = \argmax_{m\in\mathcal{M}} \mathrm{Acc}\big(m;\mathcal{D}_{\mathrm{dev}}\big),
  \label{eq:intphysx_select}
\end{equation}
and we report only $\mathrm{Acc}(m^\star;\mathcal{D}_{\mathrm{eval}})$.
$\mathcal{D}_{\mathrm{dev}}$ also provides label-free statistics (\eg the dataset center of \cref{eq:ipx_tc}), but no quantity is ever estimated from the labels of $\mathcal{D}_{\mathrm{eval}}$.

\subsubsection{Preprocessing}
\label{sec:ipx_prep}
Synthetic clips are $1920\times1920$ pixels and real clips keep their recording resolution; all have 30\,fps
video and 48\,kHz stereo audio. Each clip is conformed to the input contract of the evaluated model
(\cref{tab:ipx_contracts}). Video is padded with black to the model's aspect ratio and then resized, so nothing is
cropped or distorted. It is then re-timed to the model's frame rate by the nearest source frame, padded to the model's
frame count by repeating the last frame, and normalized with the model's own statistics. Audio is resampled to the
model's rate, truncated or zero-padded to its length, and converted to its features (waveform, log-mel or filterbank);
it is not loudness-normalized.

A clip longer than the model's window $L$ is split into sliding windows with hop $L/2$, the last ending at the clip
end, and the score is averaged over time-aligned window pairs. Counting is the exception, as is \avjepa on every task:
the clip is instead time-compressed into a single window, so that the model sees every impact. Compression preserves
pitch, using a phase vocoder for the JEPA models and \texttt{atempo} for the others. For the JEPA models, clips shorter than the 4\,s window are padded,
not stretched. Query and option videos are silent, and the context soundtrack of in-context items enters as a separate
audio input. Single-channel models receive the right channel of stereo inputs (Localization), and black-box MLLMs receive the original stereo file (\cref{tab:ipx_contracts}).
For single-channel models, this discards the interaural cues on which Localization relies.

\begin{table*}[t]
  \centering
  \caption{\textbf{Per-model input contracts on IntPhys-X.} API MLLMs without native video and audio file input are
    skipped.}
  \label{tab:ipx_contracts}
  \footnotesize
  \setlength{\tabcolsep}{4pt}
  \begin{tabular}{@{}l >{\raggedright\arraybackslash}p{1.5cm} >{\raggedright\arraybackslash}p{1.3cm} >{\raggedright\arraybackslash}p{3.3cm} >{\raggedright\arraybackslash}p{4.3cm} l@{}}
    \toprule
    Model & Frame size & fps & Window $L$ (hop) & Audio & Spatial fit \\
    \midrule
    PE-AV & $336^2$ & 30 & 30\,s (15\,s) & \leavevmode48\,kHz, right channel & padded and resized \\
    \mjepa (B/L) & $224^2$, 16 frames & 4 & \leavevmode video 4\,s (2\,s), audio 10\,s per video window; count: one 4\,s window & 16\,kHz, right channel, log-mel (128 bins) & padded and resized \\
    \cavmaesync & $224^2$ & 1.6 & per frame, no windows & 16\,kHz, right channel (Localization), mono (others), fbank (128 bins, 416 frames) & padded and resized \\
    Qwen3-Omni & \leavevmode processor default & 5 ($\le$ clip fps) & whole clip & \leavevmode16\,kHz, right channel, separate audio file & processor \\
    API MLLMs (\eg Gemini) & native & native file & whole clip & \leavevmode original stereo file, separate & none \\
    \avjepa (B/L) & \leavevmode$224^2$, 16 frames & \leavevmode4 & \leavevmode one 4\,s window (time-compressed) & \leavevmode16\,kHz, right channel, waveform & padded and resized \\
    AV-Flamingo & \leavevmode$448^2$ & \leavevmode32 frames per clip & \leavevmode whole clip & \leavevmode16\,kHz, right channel, log-mel (128 bins) & \leavevmode resize \\
    \avwm & \leavevmode$224^2$ & \leavevmode4 & \leavevmode0.5\,s chunks (64\,s context) & \leavevmode16\,kHz, right channel (Localization), waveform & padded and resized \\
    \bottomrule
  \end{tabular}
\end{table*}

\subsubsection{Representation models}
\label{sec:ipx_inference}
\label{sec:ipx_rep}
A representation model encodes the query and each option in its own modality and picks the option
most similar to the query (ties go to option~A). PE-AV and \cavmaesync average the cosine over the $W$ time-aligned
windows of a clip (\cref{sec:ipx_prep}), with embedding $e(\cdot)$:
\begin{equation}
  s_{\mathrm{sim}}(Q,X)=\frac{1}{W}\sum_{w}\cos\big(e(Q_w),\,e(X_w)\big).
  \label{eq:ipx_sim}
\end{equation}
The four JEPA models (\mjepa-B/L and \avjepa-B/L) encode the query and each option with the same encoder, in two
forward passes in which the other modality is masked, with the EMA teacher weights and without a projection head or
probe.
They are not run on the in-context scenarios, as the protocol ignores the context clip.

\paragraph{Temporal centering.}
Instead of \cref{eq:ipx_sim}, the JEPA models use temporal centering. Embeddings of the same kind of clip are
dominated by static content (the scene, the object and the recording conditions), which both options of a minimal pair
share; only the dynamics carry evidence for the choice. Each 4\,s window is therefore split into $T=8$ chunks of
0.5\,s; for \mjepa, each audio patch is assigned to the video chunk it overlaps. The chunk embeddings $u_t$ are
$\ell_2$-normalized and their within-clip mean is subtracted, $\tilde u_t=u_t-\frac{1}{T}\sum_{t'}u_{t'}$, which removes
whatever is constant over the clip. A dataset center $c_m$, the mean centered vector of modality $m$ over the 300 clips of
$\mathcal{D}_{\mathrm{dev}}$ (from file paths only, recomputed for each layer), then removes the offset that would favor
options ``similar to everything'':
\begin{equation}
  \begin{aligned}
    s_{\mathrm{tc}}(Q,X)=\cos\big(&\mathrm{vec}(\tilde u^{Q}_{1:T})-c_{m_q},\\
    &\mathrm{vec}(\tilde u^{X}_{1:T})-c_{m_o}\big),
  \end{aligned}
  \label{eq:ipx_tc}
\end{equation}
where $m_q$ and $m_o$ are the modalities of the query and the options. The score thus rewards only temporal
structure that co-varies across the modalities.

\paragraph{Choices.}
The scoring method of each JEPA model (\cref{eq:ipx_sim} or \cref{eq:ipx_tc}) was chosen on
$\mathcal{D}_{\mathrm{dev}}$ and then frozen. The layer (between half depth and the last layer: raw block outputs or the normalized
last-layer output) and, where several exist, the seed are chosen on $\mathcal{D}_{\mathrm{dev}}$ by grid search.

\subsubsection{AVJ-WM}
\label{sec:ipx_wm}
\avwm is scored by violation of expectation with its action-free model: the congruent pairing of query and option
should be less surprising than the incongruent one. \avwm is trained on natural videos, so the rendered synthetic
clips are out of distribution for it by design.

\paragraph{Reference-corrected surprise.}
\avwm (\cref{sec:suppl_avwm_arch}) splits a clip into chunks of 0.5\,s. Chunk $k$ has a video stream and an
audio stream, each a set of tokens from the frozen encoder, and the model predicts the tokens of chunk $k$ from the
true chunks $1,\dots,k-1$ (teacher forcing). For an item with query $Q$ and an option $X\in\{A,B\}$, we feed the
model one stream from $Q$ and one from $X$ and assign them two roles:
the \emph{target} $Y$, whose prediction error we measure, and the \emph{condition} $S$, the other stream. No-context
items are always scored from video to audio ($Y$ audio, $S$ video):
\begin{itemize}[nosep,leftmargin=*]
  \item \textsc{v2a} (video query, audio options): $Y$ is the audio of $X$ and $S$ the video of $Q$;
  \item \textsc{a2v} (audio query, video options): $Y$ is the audio of $Q$ and $S$ the video of $X$.
\end{itemize}
In-context items first feed the context clip and then score the continuation; there $Y$ is the option stream
(the stream of $X$) and $S$ the query stream.

Let $t_{k,p}$ be token $p$ of chunk $k$ of $Y$ and $\hat t_{k,p}(S)$ its prediction from the earlier chunks of
$Y$ and $S$. The prediction error is their mean absolute difference in latent space, after each of the two is
layer-normalized (LN) separately:
\begin{equation}
\begin{aligned}
  e_{k,p}(Y\mid S)&=\delta\bigl(\hat t_{k,p}(S),\,t_{k,p}\bigr),\\
  \delta(x,y)&=\tfrac{1}{D}\textstyle\sum_{i=1}^{D}\bigl|\mathrm{LN}(x)_i-\mathrm{LN}(y)_i\bigr|,
\end{aligned}
  \label{eq:avwm_err}
\end{equation}
with $D$ the token dimension.
This raw error mostly reflects how predictable $Y$ is on its own: silence or a static scene is easy to predict
whatever $S$ is. We therefore subtract the error of the same prediction with $S$ hidden, \ie replaced at every chunk
by the learned missing-stream tokens that the model saw in training:
\begin{equation}
  m_{k,p}=e_{k,p}(Y\mid S)-e_{k,p}(Y\mid S\text{ hidden}).
  \label{eq:avwm_pmi}
\end{equation}
A negative $m_{k,p}$ means that $S$ makes token $(k,p)$ of $Y$ easier to predict, so the option with the lower $m$ fits
better. The difference $m$ is a proxy for the negative pointwise mutual information between $Y$ and
$S$. In \textsc{a2v} items, $Y$ and the reference term $e_{k,p}(Y\mid S\text{ hidden})$ are the same for both options,
so it cancels in the comparison.

\paragraph{Difference weighting.}
The two options of an item are identical almost everywhere; in Simon, for example, they differ in only one
or two of the roughly ten continuation chunks. A uniform mean of $m$ is therefore dominated by tokens on which
the options agree and which carry no evidence for the choice. As in the matched-set comparison of
IntPhys~\cite{riochet2022intphys}, we compare the options only where they differ, by weighting each token with the
distance between the teacher tokens $t^{A}_{k,p}$ and $t^{B}_{k,p}$ of the two options,
\begin{equation}
  d_{k,p}=\delta\bigl(t^{A}_{k,p},\,t^{B}_{k,p}\bigr),
  \label{eq:avwm_d}
\end{equation}
with $\delta$ as in \cref{eq:avwm_err}. The weight is the same for both options and needs
no label. The score of option $X$ is the weighted mean of $m$, negated so that the higher score wins:
\begin{equation}
  s(Q,X)=-\frac{\sum_{k,p}d_{k,p}^{\gamma}\,m_{k,p}}{\sum_{k,p}d_{k,p}^{\gamma}}.
  \label{eq:avwm_loc}
\end{equation}
The exponent $\gamma\ge0$ sets how sharply the weights focus on the most different tokens; $\gamma=0$ recovers the
uniform mean of $m$. In \textsc{a2v} items, $m$ is computed on the query audio, but the options differ in video. A
causal model first sees option-video chunk $k-1$ when it predicts audio chunk $k$, so the audio tokens of chunk $k$ are
weighted by the mean of $d$ over the video tokens of chunk $k-1$.

\paragraph{Readout selection.}
The candidate readouts (raw surprise, \cref{eq:avwm_pmi} averaged uniformly, \cref{eq:avwm_loc}, and
the predicted-to-option token distance with and without the reference term) and the selection rule
were fixed before computing any evaluation number. For the no-context scenarios, the readout was selected on the 50
no-context items of $\mathcal{D}_{\mathrm{dev}}$ with $\gamma=1$ (nothing fitted); for the in-context scenarios, on the 50 Simon
items of $\mathcal{D}_{\mathrm{dev}}$, with $\gamma=3$ chosen from a grid, the only quantity fitted to dev labels.

\begin{table*}[t]
  \centering
  \caption{\textbf{Per-direction accuracy on IntPhys-X.} \textsc{a2v}\,/\,\textsc{v2a} accuracy (\%) per scenario.
    Humans: retained ratings after quality-assurance filtering (\cref{sec:suppl_human_study}, `Quality assurance'), per frame rate; in-context
    scenarios are rated at 30\,fps only. Models: at the input frame rate of \cref{tab:ipx_contracts}, scored as in
    \mainref{tab:intphysx_results}. -: not evaluated; n/a: the direction does not exist.}%
  \label{tab:ipx_direction}
  \scriptsize
  \setlength{\tabcolsep}{2.2pt}
  \resizebox{\textwidth}{!}{%
  \begin{tabular}{@{}l cccc ccccc ccc@{}}
    \toprule
    & \multicolumn{4}{c}{Real} & \multicolumn{5}{c}{Synthetic} & \multicolumn{3}{c}{In-context} \\
    \cmidrule(lr){2-5} \cmidrule(lr){6-10} \cmidrule(l){11-13}
    & Count & Sync & Size & Mat. & Count & Sync & Size & Mat. & Loc. & Simon & R.\,Walk & 3-Cups \\
    \midrule
    Humans, 2\,fps & 91.6 / 85.9 & 81.6 / 86.0 & 87.9 / 90.0 & 92.5 / 90.8 & 88.8 / 87.4 & 68.3 / 76.1 & 86.1 / 88.3 & 95.1 / 95.7 & 93.8 / 91.6 & - & - & - \\
    Humans, 5\,fps & 98.0 / 97.4 & 91.9 / 93.5 & 87.9 / 90.7 & 92.5 / 92.3 & 98.3 / 98.8 & 93.2 / 92.7 & 86.4 / 86.4 & 96.6 / 97.6 & 93.2 / 92.7 & - & - & - \\
    Humans, 30\,fps & 99.2 / 99.4 & 94.8 / 94.1 & 90.4 / 92.2 & 93.7 / 93.4 & 98.8 / 98.7 & 95.1 / 94.8 & 89.5 / 90.4 & 96.6 / 96.6 & 93.9 / 95.2 & 90.3 & 84.7 & 84.4 \\
    \midrule
    PE-AV-L & 71.2 / 58.6 & 65.2 / 58.4 & 61.2 / 54.2 & 72.4 / 63.0 & 59.5 / 52.2 & 58.4 / 53.0 & 50.4 / 50.9 & 50.1 / 49.8 & 49.0 / 49.1 & - & - & - \\
    \cavmaesync & 37.8 / 45.8 & 48.2 / 42.4 & 47.4 / 48.6 & 48.2 / 59.4 & 48.0 / 48.4 & 42.4 / 52.4 & 50.2 / 50.5 & 48.6 / 49.2 & 50.7 / 48.7 & - & - & - \\
    \mjepa-B & 77.6 / 75.4 & 77.0 / 84.6 & 49.2 / 44.6 & 56.6 / 55.4 & 74.5 / 72.8 & 81.2 / 74.0 & 57.4 / 52.9 & 44.6 / 50.2 & 53.3 / 52.0 & - & - & - \\
    \mjepa-L & 72.8 / 61.6 & 68.8 / 69.8 & 51.6 / 51.2 & 55.0 / 49.4 & 65.5 / 62.0 & 78.7 / 74.8 & 48.3 / 52.0 & 47.5 / 48.2 & 50.5 / 52.3 & - & - & - \\
    \avjepa-B & 68.6 / 84.0 & 69.6 / 82.8 & 54.0 / 52.6 & 63.2 / 70.4 & 70.1 / 72.6 & 75.6 / 80.8 & 59.4 / 60.9 & 47.6 / 48.7 & 51.3 / 52.2 & - & - & - \\
    \avjepa-L & 72.8 / 81.0 & 72.2 / 84.6 & 60.0 / 66.8 & 69.6 / 63.6 & 71.8 / 76.5 & 81.4 / 86.0 & 51.4 / 50.4 & 43.6 / 47.4 & 49.7 / 50.4 & - & - & - \\
    \midrule
    Gemini-4 Argon & 83.0 / 73.4 & 78.2 / 74.2 & 70.4 / 68.6 & 80.8 / 77.2 & 56.4 / 55.3 & 55.2 / 55.2 & 64.1 / 67.4 & 59.4 / 57.7 & 50.0 / 49.8 & 64.7 / 67.6 & 60.9 / n/a & 50.8 / n/a \\
    Gemini 3.1 Pro & 70.8 / 73.8 & 61.4 / 69.4 & 66.2 / 62.4 & 75.8 / 81.6 & 53.7 / 51.1 & 50.6 / 52.5 & 55.3 / 57.0 & 51.6 / 51.2 & 50.3 / 49.5 & 51.3 / 53.4 & 54.0 / n/a & 50.8 / n/a \\
    Gemini 3.8 Flash & 60.8 / 58.6 & 57.4 / 59.8 & 53.8 / 51.0 & 69.4 / 66.8 & 52.9 / 50.4 & 50.6 / 50.0 & 51.3 / 51.3 & 50.0 / 51.7 & 49.9 / 49.7 & 42.7 / 46.7 & 49.5 / n/a & 49.9 / n/a \\
    Qwen3-Omni & 51.2 / 59.6 & 50.0 / 51.2 & 49.8 / 49.8 & 59.0 / 63.2 & 50.2 / 50.6 & 49.6 / 49.3 & 49.7 / 50.3 & 49.1 / 49.0 & 48.5 / 50.6 & 39.6 / 28.5 & 51.7 / n/a & 51.2 / n/a \\
    AV-Flamingo & 49.8 / 52.6 & 50.2 / 50.6 & 50.0 / 51.2 & 50.2 / 58.4 & 50.0 / 50.1 & 50.0 / 49.8 & 50.0 / 50.2 & 50.0 / 49.9 & 50.0 / 50.5 & 50.2 / 38.1 & 49.6 / n/a & 49.8 / n/a \\
    \midrule
    \avwm & 86.4 / 75.0 & 90.2 / 86.2 & 50.6 / 49.8 & 60.8 / 54.2 & 46.2 / 49.9 & 38.6 / 48.5 & 50.8 / 49.4 & 52.0 / 51.1 & 50.6 / 48.5 & 51.8 / 57.1 & 50.0 / n/a & 51.1 / n/a \\
    \bottomrule
  \end{tabular}}
\end{table*}
\subsubsection{Multimodal LLMs}
\label{sec:ipx_mllm}

We evaluate MLLMs as black boxes through text: each item becomes one conversation, answered with an option letter.

\paragraph{Input.}
The system turn states a persona and the task. The user turn presents the query, option~A and option~B; in-context
items first present the context as a silent video and its soundtrack. Audio is always a separate input.

\paragraph{Prompts.}
The in-context scenarios use one fixed, neutral prompt per scenario and direction. For the no-context scenarios, the
prompt is chosen from 25 candidates per direction. The candidates state only the generic matching task and never name the
scenario, and candidate $i$ uses the same persona and paraphrase in both directions. We rank the candidates by
accuracy on the 50 no-context items of $\mathcal{D}_{\mathrm{dev}}$ and evaluate with the best one, breaking ties by answer rate and
then by prompt index.

\paragraph{Decoding and answers.}
The open models decode greedily (temperature 0, fixed seed), and the Gemini models use the defaults of their API.
The reply must contain \texttt{<answer>A</answer>} or \texttt{<answer>B</answer>}. An unparsable or conflicting reply
gets one format retry and otherwise counts as wrong, so abstaining never pays off. When a model exposes the
log-probabilities of the answer tokens, $\sigma\big(\log p(\text{``A''})-\log p(\text{``B''})\big)$ replaces text
parsing.

\paragraph{Position bias.}
The correct letter is balanced across items (\cref{sec:suppl_intphysx_construction}), so a model that favors one
letter loses accuracy. Qwen3-Omni and AV-Flamingo answer ``A'' in 80.2\% and 81.1\% of their valid replies.
AV-Flamingo is extreme: it answers ``A'' on 99.9\% of the no-context \textsc{a2v} items but on only 6.3\%
(Random Walk) and 1.9\% (3-Cups) of the in-context items, matching its chance-level accuracy. Nearly all replies are
valid: only 22 of Qwen3-Omni's 35,000 are unparsable, and all other models' replies parse.

\paragraph{Shortcut analysis.}
To probe whether an MLLM tracks the sequence or takes a shortcut, we asked Gemini-4 Argon to explain its
answers on 400 sampled items (100 Simon items per direction and 200 Random Walk items) and hand-coded them
blind to the answer. The explanations are post hoc and need not reflect how the model reaches its answers.
On Simon, in 45\% of the items it reports checking only the step in which the two options differ, and it is then more
accurate than with full tracking (\textsc{a2v}: 78.0\% \vs 53.7\%; \textsc{v2a}: 83.7\% \vs 57.4\%). On Random Walk, it
reports tracking the full sequence in 96.5\% of the items, although counting the sounds suffices: a hint saying so raises
its accuracy from 60.9\% to 75.7\% on the same 2,000 items. It thus does not discover this shortcut on its own.

\subsubsection{Reporting}
\label{sec:suppl_intphysx_reporting}
Accuracy is reported per source, scenario and direction (and per frame rate for humans), with 95\% Wilson intervals. Near chance, their
half-width is 1.4, 3.1 and 2.2 points for a synthetic (5,000 items), real (1,000) and in-context (2,000) scenario, and
2.0, 4.4 and 3.1 points per direction (2.2 for Random Walk and 3-Cups, which have one direction).

\subsection{Human evaluation}
\label{sec:protocol_subjective}
Human accuracy serves as a reference for model performance. Audio-visual world models and MLLMs often accept video
only at low frame rates, so, to even the odds, annotators see each no-context item at 2, 5 or 30\,fps (every 15th, 6th
or every frame), with the audio unchanged. Each no-context scenario contributes 250 items per direction at each
frame rate, each rated by about five annotators, \ie, about 7,500 ratings per scenario before filtering. Each in-context
scenario contributes 250 items at 30\,fps only (Simon: 125 per direction; Random Walk and 3-Cups: \textsc{a2v} only),
with about 1,250 ratings. Before filtering, 461 distinct annotator IDs contribute (307 no-context and 154
in-context, with no overlap): 264--290 to each no-context scenario and 142--147 to each in-context one, with at most 70 and 15 ratings per ID, respectively. Unscored practice items familiarize
annotators with the task. We report the accuracy over the ratings retained after quality-assurance filtering (below),
overall and per direction (\cref{tab:ipx_direction}).

\paragraph{Quality assurance.}\label{sec:suppl_human_study}
Rejected responses (\eg, media that did not play) and blank ones are discarded: 482 of 67,422 no-context responses
and 64 in-context ones, together with one duplicate rating. In the no-context scenarios, we remove the ratings
of an annotator on a scenario if they agree with the majority of the other annotators on
fewer than 75\% of their items (with at least 10 rated items, pooled over frame rates). This removes 248
annotator--scenario pairs (7,911 ratings; 66,940 to 59,029). In the in-context scenarios, we remove the annotators
whose accuracy on a scenario is at most 50\%: 20 annotators (155 ratings) on Simon, 44 (357) on 3-Cups and 44 (339) on
Random Walk. After filtering, 303 no-context and 152 in-context annotator IDs remain, 189--284 per no-context scenario
and 96--123 per in-context one. The filtering uses only human responses and the ground truth, never model outputs.
Humans rate 250 items per direction in each no-context scenario and 250 items in each in-context scenario, each
rated by about five annotators; models are scored on the full evaluation set.

For humans, Count and Sync improve markedly from 2 to 5 and 30\,fps, whereas Size, Material and
Localization change little. Models show scenario-dependent direction gaps, not one favored query modality.

\FloatBarrier
{\color{black}
\section{AVNav-Bench}
\label{sec:suppl_avnav}

\begin{figure*}[t]
  \color{black}
  \captionsetup{font={color=black}, labelfont={color=black}}
  \centering
  \begin{minipage}[t]{0.245\textwidth}
    \centering
    \includegraphics[width=\textwidth]{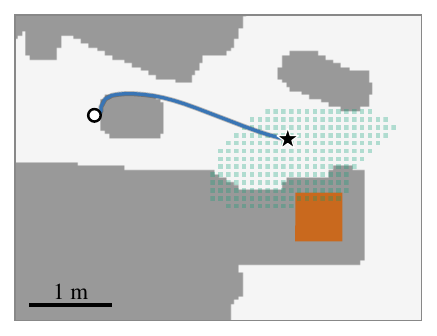}\\
    {\footnotesize (a) Walk and sound-goal region.}
  \end{minipage}
  \hfill
  \begin{minipage}[t]{0.245\textwidth}
    \centering
    \includegraphics[width=\textwidth]{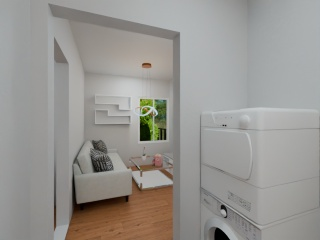}\\
    {\footnotesize (b) Walker's view.}
  \end{minipage}
  \hfill
  \begin{minipage}[t]{0.245\textwidth}
    \centering
    \includegraphics[width=\textwidth]{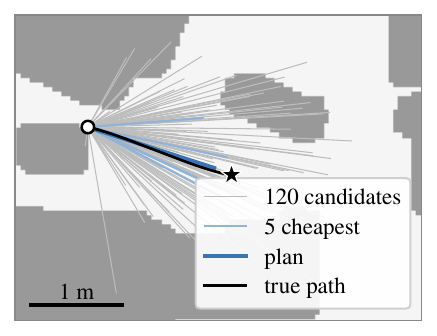}\\
    {\footnotesize (c) Offline planning (4\,s goal).}
  \end{minipage}
  \hfill
  \begin{minipage}[t]{0.245\textwidth}
    \centering
    \includegraphics[width=\textwidth]{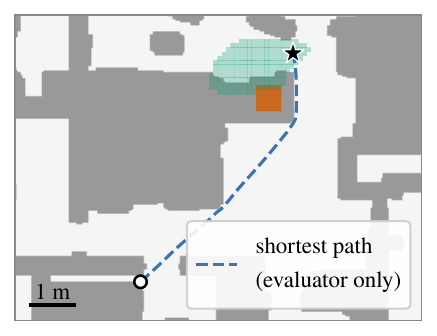}\\
    {\footnotesize (d) Closed loop (medium tier).}
  \end{minipage}
  \caption{\textbf{AVNav-Bench.} \textbf{(a)}~A walk (blue) to a sound end goal, a washer-dryer whose footprint (orange) and success region (green) define where the walker must stop. \textbf{(b)}~The view along this walk. \textbf{(c)}~Offline planning toward an image goal 4\,s ahead. AVJ-WM's cost picks the 5 cheapest (light blue) of 120 candidate paths (gray), and their mean is the plan (blue; true path in black). \textbf{(d)}~A medium closed-loop episode from another room, with the shortest path (dashed) known only to the evaluator.}
  \label{fig:avnavbench}
\end{figure*}

AVNav-Bench is a simulated indoor navigation benchmark with binaural sound (\cref{fig:avnavbench}). A walker must reach
a goal, given as an image or as the sound of a named object, either offline from a single plan or in closed loop in the
simulator. This appendix describes the benchmark, the planner and the baselines, how we fine-tune our world models, and
further results and analysis.

\subsection{Construction}
\label{sec:suppl_avnav_construction}

We generate walks through 150 HSSD houses~\cite{khanna2024hssd}, split into 92 training, 30 validation and 28 test
houses.
Each walk follows a shortest path on the house's navigation mesh~\cite{savva2019habitat}, to a pose whose view becomes
an image goal or to a sounding object. Goal walks move at 0.45--0.7\,m/s. We record the walker's pose 4 times per second, and the action of a
step is its change in pose, as in NWM~\cite{bar2025nwm}. At each step, Blender's
Cycles path tracer~\cite{blender} renders a $240 \times 320$ view from a camera 1.25\,m above the floor, and the walker
hears 16\,kHz binaural audio simulated with SoundSpaces~2.0~\cite{chen2022soundspaces2} and the SADIE~II head-related
transfer functions~\cite{armstrong2018sadie}. Sounding objects play recordings of 12 household categories, mostly from
FSD50K~\cite{fonseca2022fsd50k}, with speech from LibriTTS-R~\cite{koizumi2023librittsr} for the TVs. The splits share no
house, recording, uploader or speaker. Every design and selection choice is made on the validation houses, and the test houses serve only to report results.

\subsection{Goals and settings}
\label{sec:avnavsettings}

An \emph{image goal} is the view at a goal pose. A \emph{sound goal} is a 1--3\,s clip of a named sounding object
with its category; the goal is that object, whose footprint (the area it occupies) defines where the walker must stop.
In the \emph{offline} mode, which follows NWM~\cite{bar2025nwm}, a model receives 1\,s of context (4 frames) and
produces a single plan, which we score in free space without executing it. Its goals come from \emph{goal walks}, which
start with the goal in view, 2--4.5\,m away along a nearly straight path, and end with a final turn and a short rest at
the goal. The short goals are the views 2, 4 and 6\,s after the context. \emph{End goals} use the walk's end, where the
walker comes to rest (or the sound of the object it walked to), so the plan must also stop there; image end goals lie
about as far away as the 6\,s goals but need a smaller radius and a larger final turn.

In the \emph{closed-loop} mode, the walker executes the first second of the model's plan in the simulator, observes
and replans until it stops or its step budget runs out. Besides a 2\,s tier on the offline goals, medium and hard tiers
have goals out of sight in other rooms, 5--10\,m and 10--25\,m away along the shortest path; medium sound goals are
audible from the start. The agent knows its step budget only on the 2\,s tier, as it would reveal the shortest path elsewhere,
where running out of steps is a failure. \Cref{tab:avnavsettings} lists the context, goal,
success criterion, step budget and size of every setting.

\begin{table*}[t]
  \color{black}
  \captionsetup{font={color=black}, labelfont={color=black}}
  \centering
  \caption{\textbf{AVNav-Bench settings.} Test sizes are the episodes behind every AVNav-Bench result in this paper. Closed-loop step budgets are twice the horizon (2\,s tier) or twice the shortest path plus 24 steps (medium and hard, known only to the evaluator).}
  \label{tab:avnavsettings}
  \footnotesize
  \setlength{\tabcolsep}{5pt}
  \begin{tabular}{@{}l l l r r@{}}
    \toprule
    Setting & Context and goal & Success criterion & Val.\ size & Test size \\
    \midrule
    \multicolumn{5}{@{}l@{}}{\cellcolor{black!8}\textbf{Offline, image goals (one-shot plan, scored in free space)}} \\
    Image goal, 2\,s & 1\,s video; goal image 8 steps ahead & within 0.25\,m and $30^{\circ}$ & 1,144 windows & 1,051 \\
    Image goal, 4\,s & 1\,s video; goal image 16 steps ahead & within 0.5\,m and $30^{\circ}$ & 1,140 windows & 1,039 \\
    Image goal, 6\,s & 1\,s video; goal image 24 steps ahead & within 0.75\,m and $30^{\circ}$ & 700 windows & 628 \\
    \midrule
    \multicolumn{5}{@{}l@{}}{\cellcolor{black!8}\textbf{Offline, end goals (final resting pose, 37 steps ahead)}} \\
    Image end goal & 1\,s video; image of the final pose & within 0.5\,m and $30^{\circ}$ & 591 episodes & 532 \\
    Sound end goal & 1\,s video; 1--3\,s sound clip and category & within 1\,m, facing within $45^{\circ}$ & 553 episodes & 519 \\
    \midrule
    \multicolumn{5}{@{}l@{}}{\cellcolor{black!8}\textbf{Closed loop, in the simulator (execute, replan)}} \\
    Closed loop, 2\,s & image goal as above, executed step by step & stop within 0.25\,m and $30^{\circ}$ & 1,144 episodes & 1,051 \\
    Closed loop, medium & image or sound goal 5--10\,m away, out of sight & stop at the goal (as end goals) & 467 episodes & 443 \\
    Closed loop, hard & as medium, 10--25\,m away, at least two rooms on & stop at the goal (as end goals) & 430 episodes & 392 \\
    \bottomrule
  \end{tabular}
\end{table*}

\subsection{Scoring}
\label{sec:suppl_avnav_scoring}

Success is defined geometrically, identically for every model.
For image goals, the final pose must face the goal within $30^{\circ}$ and lie within $H/32$\,m of it for a goal $H$
steps ahead (about a quarter of the median goal distance), or 0.5\,m for end goals and on the medium and hard tiers. For sound goals, following the object-goal
criterion of \citet{batra2020objectnav}, the walker must stop within 1\,m of the object's footprint; it must also face the
object within $45^{\circ}$ and be at most 1.3\,m from it along the navigation mesh, so that standing behind a wall does not
count.
In the closed loop, the walker must also stop at the goal rather than pass through it, as is standard in embodied
navigation~\cite{anderson2018evaluation}.

\subsection{What can be evaluated}
\label{sec:suppl_avnav_models}

Because the settings are defined by goals and success criteria, different model families can be evaluated. World
models score candidate paths against the goal (\cref{sec:suppl_avnav_planner}) in both modes. Policies run in the closed
loop on the goal types they accept, and those that predict a whole path, such as GNM and NoMaD, also offline. The perfect
cost and the goal-blind walk bound every setting (\cref{sec:suppl_avnav_baselines}).

\subsection{Planner shared by the world models}
\label{sec:suppl_avnav_planner}

Every world model uses the same planner and search budget, so differences between rows reflect the models and their goal costs.

\paragraph{Offline search.} We follow NWM's planning protocol~\cite{bar2025nwm}. We draw 120 candidate end
poses from a prior fitted on training windows of the same horizon, seeding the random generator with the episode so that
every model ranks the same candidates, score each candidate with the model's cost and take the mean of the 5 cheapest
as the plan. As in NWM, we average the cost of the diffusion world models, which sample their predictions, over 3 imagined rollouts.

\paragraph{Costs.} For latent world models, such as AVJ-WM, the cost of a candidate is the distance between its
predicted final state and the goal state. For pixel and diffusion world models, we use the LPIPS
distance~\cite{zhang2018lpips} between the generated final frame and the goal frame, as NWM does. For sound goals, we
compare each ear's predicted audio at the end of a candidate with the goal clip (its closest 1\,s window), average the
two ears and add their difference, which carries the direction of the sound, with a weight set on validation.

\paragraph{Closed-loop search.} In the closed loop, the walker executes the first second of the plan, observes and
replans (model-predictive control), seeing only the views and binaural sound along its own path, never the map. Each
replanning uses the cross-entropy method~\cite{deboer2005cem,hafner2019planet}. Starting from the previous plan, we draw
32 candidates, refit their distribution to the 5 cheapest and repeat for three rounds, so 96 imagined paths per decision,
close to the 120 offline. Candidates turn, walk straight and turn again. In their prior, fitted on the training walks, the final heading of a 2\,s walk varies by only about $40^{\circ}$, so we add 7 turns in
place to the first round, to headings $45^{\circ}$ apart, letting the walker turn toward a goal beside or behind it.
Here we draw one sample per candidate from the diffusion world models, as three would triple their already long planning time.

\paragraph{Stopping.} Success requires stopping at the goal, so our planner, the \emph{stopping planner}, also
decides when to stop. In the closed loop, the walker compares at every step the predicted cost of standing still,
$C_{\text{stay}}$, with that of the best plan, $C_{\text{plan}}$, and stops if
\begin{equation}
  C_{\text{stay}} \le C_{\text{plan}} + \delta .
  \label{eq:avnavstop}
\end{equation}
We choose the margin $\delta$ per model on the 2\,s validation episodes by success, including
$\delta = -\infty$, for which the walker stops only when the 2\,s tier's step budget is used up. Offline, the planner
imagines walking the first $k = 0, 1, \dots, H$ steps of the plan and then standing still (\cref{sec:suppl_avnav_ours}), and stops after the $k$ with the cheapest outcome.

\paragraph{Medium and hard tiers.} Here the planner looks 8\,s ahead (32 steps) instead of 2\,s, and still replans
every second. As the step budget is hidden, the walker must stop by itself. On image goals, it stops once its current view, encoded like the goal image, is close enough to the goal, with a distance threshold chosen per model on medium validation episodes. On sound goals, it keeps its own stop rule (\cref{eq:avnavstop}) with $\delta = 0$. Each
training seed runs with one planner seed.%

\paragraph{Planning cost.} Our world models plan faster than every diffusion world model, and far faster than those on SD-VAE latents (\cref{tab:avnavplantime}). We time every model on the same 2\,s closed-loop validation episodes (15 steps, with a full search about every fourth step and a cheaper check whether to stop at the others) with the same search, on one H200 GPU without the simulator's time, after optimizing each implementation as well as we could (for example with CUDA graphs) without changing its predictions. AVJ-WM-B plans a 2\,s episode in 6.7\,s and
AVJ-WM in 14.2\,s, against 23--36\,s for the CDiTs on frozen features and 4--37\,min for those on SD-VAE latents. AVJ-WM-B
is thus 3.4 times faster than the fastest CDiT of the same size, and 85 and 329 times faster than NWM's CDiT-B and CDiT-XL,
mostly because it needs one network pass per imagined step instead of 8 or 250.

\begin{table}[t]
  \color{black}
  \captionsetup{font={color=black}, labelfont={color=black}}
  \centering
  \caption{\textbf{Planning cost in the closed loop.} Our world models plan faster than the CDiTs on frozen features and far faster than those on SD-VAE latents. We report median seconds per search and \emph{Time}, the seconds to plan a 2\,s episode, on one H200 GPU with the same search for every model (\cref{sec:suppl_avnav_planner}). Models as in \mainref{tab:avnavbench}; --: no pretraining; \emph{Sampling}: network passes per imagined step.}
  \label{tab:avnavplantime}
  \footnotesize
  \setlength{\tabcolsep}{3pt}
  \begin{tabular}{@{}l l l c r r@{}}%
    \toprule
    Model & Latents & Data & Sampling & Search (s) & Time (s) \\
    \midrule
    CDiT-XL & SD-VAE & NWM\textsuperscript{+} & 250 & 463 & 2,203 \\
    CDiT-B & SD-VAE & NWM & 250 & 129 & 570 \\
    AV-CDiT & SD-VAE, DAC & NWM & 250 & 66.0 & 325 \\
    CDiT & SD-VAE & \na & 250 & 50.1 & 261 \\
    AV-CDiT & \avjepa-B & Ours & 8 & 8.01 & 35.7 \\
    AV-CDiT & \avjepa-B & \na & 8 & 5.31 & 26.7 \\
    CDiT & \avjepa-B & \na & 8 & 4.60 & 22.6 \\
    CDiT & \vjepatwoone-B & \na & 8 & 4.61 & 22.7 \\
    \oursrow AVJ-WM & \avjepa-L & Ours & 1 & \second{1.38} & \second{14.2} \\%
    \oursrow AVJ-WM-B & \avjepa-B & Ours & 1 & \best{0.60} & \best{6.7} \\
    \bottomrule
  \end{tabular}
\end{table}

\subsection{Baselines}
\label{sec:suppl_avnav_baselines}

We train all learned baselines in the training houses, on their 9,702 walks or, for reinforcement learning, in the simulator, and average three training seeds, except for CDiT-XL and the policies trained by reinforcement learning, which are single runs, as is CDiT-B's
closed loop. CDiT-XL has 1.0B parameters and CDiT-B 195M, against
249M for AVJ-WM-B including its frozen encoder. \Mainref{tab:avnavbench} groups them by model
family, and \cref{tab:avnavvariants} adds training variants of the diffusion baselines.

\paragraph{Floors without a learned model.}
The goal-blind walk continues the motion of the context, and random actions replay a random action segment of the
training walks; in the closed loop, both stop after a fixed time per tier, chosen on validation houses. The random cost
gives random costs to the candidates and their stopping points within our planner, the chance level of every world
model.

\paragraph{Visual navigation policies.}
GNM~\cite{shah2023gnm} and NoMaD~\cite{sridhar2024nomad} predict the next waypoints towards an image goal from the
last few frames. We retrain both with the authors' code and recipes on NWM's four training datasets and fine-tune them on
the AVNav-Bench training walks, with later frames of a walk as goals. Offline, their predicted path is the plan, extended
by repeating its last step in cases where it is shorter than the goal's horizon (GNM predicts 1.25\,s). In the closed loop, they
take their first predicted waypoint as the next move and stop when their predicted distance to the goal falls below a
threshold chosen on validation. As they take no audio, they run only on image goals, and in the closed loop only on the
2\,s tier. NoMaD predicts no final heading, which caps its success on the 6\,s and end goals.

\paragraph{Audio-visual navigation policies.}
AV-Nav~\cite{chen2020soundspaces} and SAVi~\cite{chen2021savi}, the standard audio-visual navigation policies, choose
the next action with a recurrent network from the current view and binaural audio. We give both the goal as an extra
input (the goal image, or the goal sound with its category) and train each in two ways. With reinforcement learning in the
simulator, we follow the published recipes (80M steps for AV-Nav, 300M in two stages for SAVi). To match GNM and NoMaD, we
also train them by imitation on the training walks, with three ideas from these policies. We stack the goal image with
the current view, predict the remaining distance to the goal, and take 10\% of the training goals from other houses with
``do not stop'' as target, which raises imitation success on the 2\,s tier from 11.3 to 35.4\% for AV-Nav and from 21.0
to 34.7\% for SAVi. Both act one step at a time, without an offline plan.

\paragraph{World models on SD-VAE latents.}
CDiT, NWM's conditional diffusion transformer~\cite{bar2025nwm}, predicts the SD-VAE latents of future frames. We
fine-tune NWM's released CDiT-B on the training walks with batches of 16 contexts and 4 goals each for 120k updates,
which beats NWM's own recipe (batches of 1,024 contexts, goals up to 16\,s away) on AVNav-Bench.
AV-CDiT~\cite{wang2025avwm} adds binaural audio to CDiT. We reimplement its three stages, starting its video-only first
stage from NWM's weights, with a 16\,kHz DAC codec~\cite{kumar2023dac} trained from scratch on the training audio and applied to each
ear, which largely preserves each ear's level and the level difference between the ears (\cref{tab:avnavdac,fig:avnavdac}).
\Cref{tab:avnavvariants} also reports both CDiT-B recipes, the first stage alone from NWM's or random weights, and NWM's
two released CDiT-XL models, without and with Ego4D.
\makeatletter\def\ftype@table{1}\makeatother
\begin{table}[t]
  \color{black}
  \captionsetup{font={color=black}, labelfont={color=black}}
  \centering
  \caption{\textbf{Training variants of the diffusion baselines.} Our small-batch fine-tuning beats NWM's recipe at 2 and 4\,s, and neither NWM's weights nor three stages help the AV-CDiT at 2--6\,s. \emph{Data}: pretraining on NWM's datasets (NWM; NWM\textsuperscript{+} adds Ego4D) or none (--); shaded: the variants in \mainref{tab:avnavbench}. $^\ast$NWM's fine-tuning recipe. We average three training seeds for the offline rows, except the CDiT-XL rows and CDiT-B with NWM's recipe (one seed).}
  \label{tab:avnavvariants}
  \scriptsize
  \setlength{\tabcolsep}{2.2pt}
  \begin{tabular}{@{}l l l ccc cc c@{}}
    \toprule
    & & & \multicolumn{3}{c}{Image goal} & \multicolumn{2}{c}{End goal} & CL \\
    \cmidrule(lr){4-6} \cmidrule(lr){7-8} \cmidrule(l){9-9}
    Model & Latents & Data & 2\,s & 4\,s & 6\,s & Image & Sound & 2\,s \\
    \midrule
    \rowcolor{black!6}CDiT-B~\cite{bar2025nwm} & SD-VAE & NWM & 70.9 & 65.0 & 54.2 & 24.6 & \na & 89.7 \\
    CDiT-B$^\ast$ & SD-VAE & NWM & 67.7 & 60.6 & 49.0 & 21.1 & \na & 84.2 \\
    CDiT-XL & SD-VAE & NWM & 66.1 & 62.5 & 50.2 & 22.4 & \na & 87.1 \\
    CDiT-XL$^\ast$ & SD-VAE & NWM & 57.8 & 50.9 & 41.6 & 17.3 & \na & \na \\
    \rowcolor{black!6}CDiT-XL & SD-VAE & NWM\textsuperscript{+} & 78.5 & 71.4 & 61.1 & 26.9 & \na & 93.1 \\
    CDiT-XL$^\ast$ & SD-VAE & NWM\textsuperscript{+} & 77.2 & 70.0 & 57.0 & 27.4 & \na & \na \\
    \midrule
    AV-CDiT stage~1 & SD-VAE & NWM & 69.0 & 61.1 & 48.4 & 20.6 & \na & 90.1 \\
    AV-CDiT stage~1 & SD-VAE & \na & 71.9 & 64.5 & 50.2 & 16.9 & \na & 92.2 \\
    \rowcolor{black!6}AV-CDiT, 3 stages & SD-VAE, DAC & NWM & 66.4 & 60.7 & 46.1 & 17.6 & 2.4 & 90.0 \\
    \midrule
    \rowcolor{black!6}AV-CDiT, 1 stage & \avjepa-B & \na & 81.2 & 71.9 & 55.0 & 12.7 & 4.5 & 95.3 \\
    AV-CDiT, 3 stages & \avjepa-B & \na & 80.7 & 69.6 & 50.4 & 13.7 & 5.1 & 95.6 \\
    \bottomrule
  \end{tabular}
\end{table}

\begin{table}[t]
  \color{black}
  \captionsetup{font={color=black}, labelfont={color=black}}
  \centering
  \caption{\textbf{Audio codec of our AV-CDiT.} The codec largely preserves the binaural cues that sound goals rely on, and its level errors stay below what listeners notice (about 1\,dB~\cite{jesteadt1977intensity}; 0.6\,dB for the level difference between the ears~\cite{klockgether2016jnd}). Windows of
    0.26\,s along the test paths with a sound end goal.}
  \label{tab:avnavdac}
  \footnotesize
  \setlength{\tabcolsep}{4pt}
  \begin{tabular}{@{}l c@{}}
    \toprule
    Louder ear unchanged (\% of windows) & 99.8 \\
    Level rises towards the source (\% of walks) & \\
    \quad before / after the codec & 84.8 / 84.6 \\
    \midrule
    Mean level change (dB) & $+$0.03 \\
    Level error per window (dB) & 0.35 \\
    Level-difference error per window (dB) & 0.33 \\
    \quad for scale, spread of the level difference (dB) & 1.89 \\
    Time difference within one sample (\% of windows) & 90.4 \\
    \bottomrule
  \end{tabular}
\end{table}
\begin{figure}[t]
  \color{black}
  \captionsetup{font={color=black}, labelfont={color=black}}
  \centering
  \includegraphics[width=\linewidth]{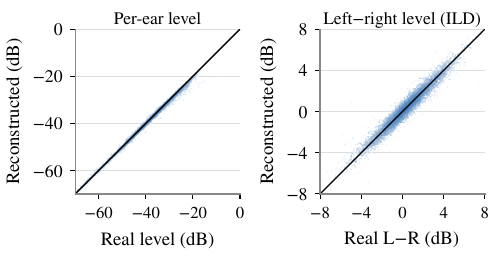}
  \caption{\textbf{Audio codec of our AV-CDiT.} The codec largely preserves each ear's level (left) and the level difference between the ears (right), on 0.26\,s windows along the test paths with a sound end goal. The line marks equal values.}
  \label{fig:avnavdac}
\end{figure}

\paragraph{World models on JEPA latents.}
We also train the CDiT and AV-CDiT from random initialization on frozen JEPA features, following the planning
recipe of \vjepatwoone~\cite{murlabadia2026vjepa21}: clean-feature prediction, 8 DDIM steps~\cite{song2021ddim} and a
latent goal cost. The CDiT uses the video features of the Base \avjepa encoder or of the released \vjepatwoone ViT-B. For
the AV-CDiT, we encode each frame with the frozen \avjepa encoder as a static two-frame clip, and the 0.5\,s of sound
before it separately for each ear, and the model denoises the next frame's video and audio tokens together. It keeps
AV-CDiT's audio-visual blocks and conditions on the four previous frames, the action and the time offset of the target. We train it in one stage or in AV-CDiT's
three.

\paragraph{Pretraining the AV-CDiT.}
To test whether pretraining also helps the CDiT, we pretrain the one-stage AV-CDiT without actions on AVJ-WM's
Stage~2 data (\mainref{sec:avwm}) for 708k updates of 1,024 samples, matching AVJ-WM's data exposure, with
QK-norm~\cite{henry2020qknorm} for stability. Without actions, it is conditioned on the context frames and the time
offset of the target, as NWM does for Ego4D, and learns action inputs in fine-tuning.

\paragraph{Toplines with privileged information.}
The perfect cost replaces a model's cost by the true distance to the goal pose within the same planner, a perfect
world model for these candidates. It stays below 100\% because the planner averages the five cheapest candidates, whose
mean can miss the goal, although at least one of the 120 candidates reaches it in 98--99\% of the image-goal episodes.
Perfect pose moves straight to the goal pose offline and follows the shortest path in the closed loop.

\begin{table}[t]
  \color{black}
  \captionsetup{font={color=black}, labelfont={color=black}}
  \centering
  \caption{\textbf{AVJ-WM Stage~3 hyperparameters.} Navigation fine-tuning on the 9,702 AVNav-Bench training
    walks.}
  \vspace{-0.2cm}
  \label{tab:avwmnavhyper}
  \fontsize{7.5}{9}\selectfont
  \setlength{\tabcolsep}{2pt}
  \begin{tabular}{
      @{}p{0.28\columnwidth}
      @{\hspace{6pt}}
      >{\centering\arraybackslash}p{\dimexpr0.72\columnwidth-6pt\relax}
      @{}
    }
    \toprule
    Configuration & AVJ-WM and AVJ-WM-B \\
    \midrule
    \multicolumn{2}{@{}l@{}}{\cellcolor{black!8}\textbf{Model}} \\
    Trained & predictor, readout, action adapters, ear embeddings \\
    Frozen & encoder, AV-Combiner \\
    Action adapters & residual MLP of width 128 after each block \\
    Adapter output & zero-initialized \\
    Binaural audio & ears encoded separately; video and both ears predicted \\
    \midrule
    \multicolumn{2}{@{}l@{}}{\cellcolor{black!8}\textbf{Training windows}} \\
    Window & 20\,s (40 chunks), half with up to 64\,s of history \\
    Context & 2--35 chunks, probability $\propto 1/\text{length}$ \\
    Rollout & 4--27 chunks, probability $\propto \text{length}^{-1/2}$ \\
    Standing still & after a drawn number of chunks (possibly 0) \\
    Still until & 4\,s after the window's end \\
    Augmentation & left--right mirroring (ears swapped), probability 0.5 \\
    \midrule
    \multicolumn{2}{@{}l@{}}{\cellcolor{black!8}\textbf{Objective (weights)}} \\
    Prediction & 1, teacher-forced and rolled out \\
    Audio & 0.25, none when standing still \\
    Two-ear term & 0.25, $\ell$ on the left-minus-right ear tokens \\
    Action-contrastive & 0.05, temperature 0.02 \\
    Altered actions & speed, turns, mirroring or timing \\
    Planner-shaped path & turn, straight walk, turn; 1 and 0.05 \\
    \midrule
    \multicolumn{2}{@{}l@{}}{\cellcolor{black!8}\textbf{Optimization}} \\
    Optimizer & AdamW, weight decay 0.01 \\
    Learning rate & $1.41 \times 10^{-4}$ \\
    Gradient clipping & norm 1 \\
    Updates, batch & 26,713, 32 windows \\
    Cool-down & last 2,463 updates \\
    Evaluated weights & EMA, decay 0.9998 \\
    \bottomrule
  \end{tabular}
  \vspace{-0.2cm}
\end{table}
\begin{table*}[t]
  \color{black}
  \captionsetup{font={color=black}, labelfont={color=black}}
  \centering
  \caption{\textbf{Stage~3 ablations on AVNav-Bench.} Rollout training, long rollouts and stand-still training are essential, and sound goals need both binaural changes. Each row changes one part of AVJ-WM-B's Stage~3 recipe; colored numbers give the change. \emph{Stable}: stable/run fine-tunes (\cref{sec:supplavwmablations}). $^\dagger$No stable fine-tune, so we report the mean of all five.}%
  \label{tab:avwmnavrecipes}
  \footnotesize
  \setlength{\tabcolsep}{4pt}
  \begin{tabular}{@{}l ccc cc c c@{}}
    \toprule
    & \multicolumn{3}{c}{Image goal} & \multicolumn{2}{c}{End goal} & Closed loop & \\
    \cmidrule(lr){2-4} \cmidrule(lr){5-6} \cmidrule(lr){7-7}
    Variant & 2\,s & 4\,s & 6\,s & Image & Sound & 2\,s & Stable \\
    \midrule
    \oursrow AVJ-WM-B & 82.7 & 77.6 & 66.2 & 32.4 & 18.0 & 93.7 & 5/5 \\%
    \quad without the action-contrastive term & 76.0~\dneg{$-$6.7} & 68.7~\dneg{$-$8.9} & 58.8~\dneg{$-$7.4} & 31.4~\dneg{$-$1.0} & 11.3~\dneg{$-$6.7} & 93.8~\dpos{$+$0.1} & 5/5 \\%
    \quad without the straight-turn rollout loss & 82.0~\dneg{$-$0.7} & 76.3~\dneg{$-$1.3} & 62.7~\dneg{$-$3.5} & 34.4~\dpos{$+$2.0} & 16.4~\dneg{$-$1.6} & 93.2~\dneg{$-$0.5} & 4/5 \\%
    \quad without both of these terms & 78.9~\dneg{$-$3.8} & 72.3~\dneg{$-$5.3} & 57.6~\dneg{$-$8.6} & 20.1~\dneg{$-$12.3} & 8.9~\dneg{$-$9.1} & 93.5~\dneg{$-$0.2} & 5/5 \\%
    \quad without the rollout loss on recorded actions & 82.5~\dneg{$-$0.2} & 78.0~\dpos{$+$0.4} & 65.9~\dneg{$-$0.3} & 31.1~\dneg{$-$1.3} & 15.9~\dneg{$-$2.1} & 92.6~\dneg{$-$1.1} & 5/5 \\%
    \quad without rollout training (teacher-forced only) & 45.1~\dneg{$-$37.6} & 27.6~\dneg{$-$50.0} & 17.8~\dneg{$-$48.4} & 4.4~\dneg{$-$28.0} & 3.2~\dneg{$-$14.8} & 41.7~\dneg{$-$52.0} & 5/5 \\%
    \quad with short (2\,s) rollouts & 82.9~\dpos{$+$0.2} & 51.2~\dneg{$-$26.4} & 31.2~\dneg{$-$35.0} & 6.4~\dneg{$-$26.0} & 8.4~\dneg{$-$9.6} & 94.9~\dpos{$+$1.2} & 5/5 \\%
    \quad with short (10\,s) training windows & 83.3~\dpos{$+$0.6} & 78.3~\dpos{$+$0.7} & 67.2~\dpos{$+$1.0} & 32.4~\dnul{$\pm$0.0} & 18.5~\dpos{$+$0.5} & 94.7~\dpos{$+$1.0} & 5/5 \\%
    \quad without the two-ear term & 83.8~\dpos{$+$1.1} & 78.7~\dpos{$+$1.1} & 67.8~\dpos{$+$1.6} & 34.1~\dpos{$+$1.7} & 12.7~\dneg{$-$5.3} & 94.5~\dpos{$+$0.8} & 5/5 \\%
    \quad without both sound changes (\cref{sec:suppl_avnav_ours}) & 82.8~\dpos{$+$0.1} & 78.3~\dpos{$+$0.7} & 68.7~\dpos{$+$2.5} & 36.8~\dpos{$+$4.4} & 6.2~\dneg{$-$11.8} & 94.5~\dpos{$+$0.8} & 5/5 \\%
    \quad without walks before standing still$^\dagger$ & 35.2~\dneg{$-$47.5} & 42.4~\dneg{$-$35.2} & 32.2~\dneg{$-$34.0} & 13.2~\dneg{$-$19.2} & 11.6~\dneg{$-$6.4} & 33.2~\dneg{$-$60.5} & 0/5 \\%
    \quad without stand-still training & 69.7~\dneg{$-$13.0} & 75.1~\dneg{$-$2.5} & 62.2~\dneg{$-$4.0} & 6.4~\dneg{$-$26.0} & 13.2~\dneg{$-$4.8} & 16.6~\dneg{$-$77.1} & 2/5 \\%
    \quad with the world model frozen (no two-ear term) & 74.2~\dneg{$-$8.5} & 67.3~\dneg{$-$10.3} & 53.4~\dneg{$-$12.8} & 25.1~\dneg{$-$7.3} & 4.2~\dneg{$-$13.8} & 90.6~\dneg{$-$3.1} & 5/5 \\%
    \quad with the predictor blocks frozen & 76.4~\dneg{$-$6.3} & 68.1~\dneg{$-$9.5} & 54.0~\dneg{$-$12.2} & 23.7~\dneg{$-$8.7} & 10.4~\dneg{$-$7.6} & 91.3~\dneg{$-$2.4} & 4/5 \\%
    \bottomrule
  \end{tabular}
\end{table*}

\subsection{Our world models}
\label{sec:suppl_avnav_ours}
\begin{table}[t]
  \color{black}
  \captionsetup{font={color=black}, labelfont={color=black}}
  \centering
  \caption{\textbf{Training data and pretraining.} Pretraining matters more for AVJ-WM-B than for the one-stage AV-CDiT on \avjepa-B latents. Both are trained on all or 10\% of the training walks, with pretraining on our data (Ours) or without (--).}
  \label{tab:avnavdata}
  \scriptsize
  \setlength{\tabcolsep}{2.0pt}
  \begin{tabular}{@{}l l l ccc cc ccc@{}}
    \toprule
    & & & \multicolumn{3}{c}{Image goal} & \multicolumn{2}{c}{End goal} & \multicolumn{3}{c@{}}{Closed loop} \\
    \cmidrule(lr){4-6} \cmidrule(lr){7-8} \cmidrule(l){9-11}
    Model & Pretr. & Walks & 2\,s & 4\,s & 6\,s & Image & Sound & 2\,s & Med. & Hard \\
    \midrule
    AV-CDiT, 1 stage & \na & all & 81.2 & 71.9 & 55.0 & 12.7 & 4.5 & 95.3 & 1.2 & 0.2 \\
    AV-CDiT, 1 stage & \na & 10\,\% & 76.6 & 63.2 & 43.7 & 9.1 & 5.1 & 94.5 & 1.4 & 0.1 \\
    AV-CDiT, 1 stage & Ours & all & 81.5 & 70.9 & 53.2 & 13.2 & 5.0 & 94.9 & 0.8 & 0.0 \\
    \midrule
    \oursrow AVJ-WM-B & Ours & all & 82.7 & 77.6 & 66.2 & 32.4 & 18.0 & 93.7 & 1.4 & 0.1 \\
    AVJ-WM-B & Ours & 10\,\% & 81.4 & 73.7 & 58.4 & 26.3 & 13.3 & 92.5 & 0.4 & 0.1 \\%
    AVJ-WM-B & \na & all & 27.8 & 21.9 & 9.6 & 1.1 & 5.6 & 25.2 & 0.0 & 0.0 \\
    \bottomrule
  \end{tabular}
\end{table}
\paragraph{Models.}
AVJ-WM-B and AVJ-WM are adapted to actions with Stage~3 (\cref{sec:suppl_avwm_arch}) and the recipe of
\cref{tab:avwmnavhyper}. AVJ-WM-B matches the CDiTs in encoder and size and is pretrained with the shorter schedule of
our ablations (\cref{sec:avwmrecipe}); AVJ-WM is our full-scale model, WM-H on the \avjepa-L encoder. Without world-model pretraining, we train the predictor on the AVNav-Bench walks only. To test data efficiency, we also fine-tune AVJ-WM-B and the one-stage AV-CDiT on a random 10\% of the training walks
(\cref{tab:avnavdata}). As in \cref{sec:supplavwmablations}, we average the stable fine-tunes among the same five seed pairs.

\paragraph{Stage~3 objective.} We fine-tune on 20\,s windows of the training walks, half of them with up to
64\,s of earlier history and half mirrored left to right (\cref{tab:avwmnavhyper}). Besides the prediction loss
$\Loss_{\mathrm{pred}}$ on teacher-forced chunks and the rollout loss $\Loss_{\mathrm{roll}}$ of Stage~2 under the recorded
actions, now over 4--27 chunks, two terms teach the model to tell actions apart. A straight-turn rollout loss rewrites a
rollout's actions as a turn, a straight walk and a turn to the same end pose (i.e., the form of our planner's candidates), and
compares the predicted end with the target. An action-contrastive term furthermore requires the prediction under the true actions to
lie closer to the target than predictions under altered actions (other speeds or turns, mirrored or shifted in time), for
both forms of the actions. Audio terms are weighted by 0.25.

\paragraph{Learning to stop.}
A goal image is encoded as a still clip, the image held over the whole 2\,s window, whereas the chunks of a walk
carry motion, so a predicted chunk can match the goal only once the walker has stood still there for 2\,s. World models on single frames, such as the CDiTs, do not have this mismatch. For our world models only, the planner therefore scores each candidate after 2\,s of standing still at its end (\cref{sec:suppl_avnav_planner}), and every training rollout walks a drawn number of chunks, possibly none, and then
stands still, with the last frame held, until 4\,s after its end. As the training walks hold their last audio while real sound keeps changing,
standing-still chunks get no audio loss (\cref{tab:avwmnavhyper},
\cref{tab:avwmnavrecipes}).

\paragraph{Binaural adaptation.}
AVNav-Bench is binaural, and we adapt to it through a binaural port of the same model rather than through the
adapters alone. The two streams of Stage~2, video and mono audio, become three, $m \in \{v, a\mathrm{L}, a\mathrm{R}\}$. We
encode each ear's waveform with the model's own frozen audio path and add a learned ear embedding to its tokens,
\[
  \tilde{e}^{a,r}_t = e^{a,r}_t + u_r, \qquad r \in \{\mathrm{L}, \mathrm{R}\},
\]
where $e^{a,r}_t$ are the audio tokens of ear $r$ and $u_r$ is broadcast over them. The combiner reads the video tokens
with both ears' tokens, and the readout reads out the second ear with its own queries, initialized from the first ear's,
so the model predicts both ears and the direction between them. We add to the Stage~3 objective, now over the three streams, a two-ear term, $\ell$ on the difference between the ears' tokens, because this difference, which carries the sound's direction, is small and would otherwise contribute little to the loss. Goals enter only through the planning cost. The ear
embeddings and second-ear queries train with the adapters and the predictor.

\subsection{Additional results}
\label{sec:suppl_avnav_full}

\Cref{tab:avnavvariants} reports the training variants of the diffusion baselines (\cref{sec:suppl_avnav_baselines}),
and \cref{tab:avnavdata} AVJ-WM-B and the AV-CDiT trained on 10\% of the walks and with or without pretraining;
\cref{sec:suppl_avnav_analysis} discusses both.

\subsection{Analysis of the navigation results}
\label{sec:suppl_avnav_analysis}

\begin{table}[!htb]%
  \color{black}
  \captionsetup{font={color=black}, labelfont={color=black}}
  \centering
  \caption{\textbf{Offline planning with exact predictions.} Even exact predictions leave the planner far below the perfect cost. Success (\%) on image goals with our planner and goal cost and the same candidates in every row. \emph{Exact predictions}: each candidate's predicted end state is replaced by the encoding of the view rendered at its end pose.}
  \label{tab:avnavceiling}
  \footnotesize
  \setlength{\tabcolsep}{4pt}
  \begin{tabular}{@{}l l c c c c@{}}
    \toprule
    & Encoder & 2\,s & 4\,s & 6\,s & End \\
    \midrule
    Random cost & \na & 10.7 & 5.7 & 3.5 & 2.2 \\
    \midrule
    AVJ-WM-B & \avjepa-B & 82.7 & 77.6 & 66.2 & 32.4 \\
    WM-B & \avjepa-L & 83.4 & 77.9 & 69.7 & 38.5 \\%
    WM-L & \avjepa-L & 83.7 & 78.3 & 70.2 & 37.0 \\%
    WM-H & \avjepa-L & 84.2 & 79.8 & 71.5 & 36.9 \\%
    AVJ-WM & \avjepa-L & 82.4 & 76.6 & 68.3 & 39.6 \\
    \midrule
    Exact predictions & \avjepa-B & 85.4 & 83.6 & 73.4 & 39.7 \\
    Exact predictions & \avjepa-L & 85.3 & 82.4 & 73.7 & 37.2 \\
    Exact predictions & \vjepatwoone-B & 84.3 & 82.1 & 71.3 & 37.0 \\%
    Exact predictions & \vjepatwoone-L & 84.3 & 82.8 & 71.7 & 36.1 \\
    Exact predictions & \vjepatwo-L & 61.5 & 52.9 & 40.8 & 17.7 \\%
    \midrule
    Perfect cost & \na & 98.0 & 96.2 & 85.8 & 84.8 \\
    \bottomrule
  \end{tabular}
\end{table}
\begin{figure*}[t]
  \color{black}
  \captionsetup{font={color=black}, labelfont={color=black}}
  \centering
  \begin{tabular}{@{}c@{\hspace{0.3125in}}c@{}}
    \includegraphics[width=3.28in]{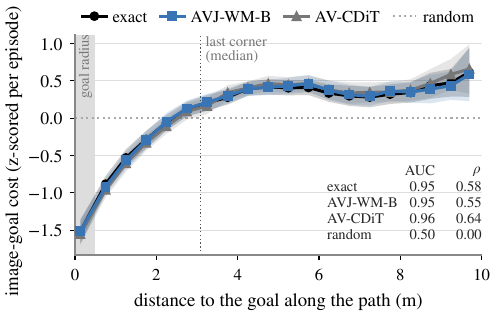} &
    \includegraphics[width=3.28in]{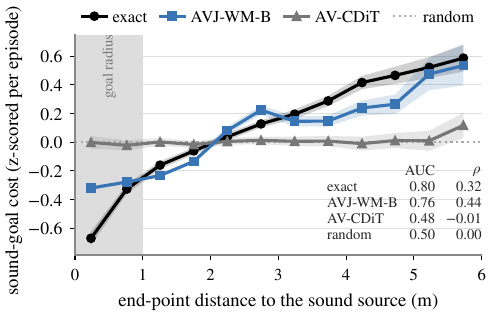} \\
    \small (a) Image goals, medium tier & \small (b) Sound end goals \\[1.5mm]
    \multicolumn{2}{@{}c@{}}{\includegraphics[width=\textwidth,height=1.15in,keepaspectratio]{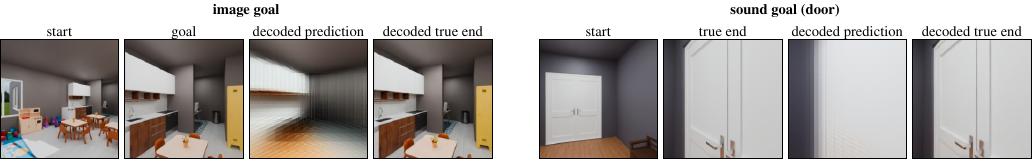}} \\
    \multicolumn{2}{@{}c@{}}{\small (c) Predicted end states, decoded}
  \end{tabular}
  \caption{\textbf{Goal costs and predicted end states on AVNav-Bench.} \textbf{(a)}~On the medium tier, the image-goal cost
    falls only within about 4\,m of the goal, for the exact current view and both world models alike. \textbf{(b)}~AVJ-WM-B's
    predicted sound ranks the candidate plans almost as well as the exact sound, whereas the AV-CDiT's cost hardly depends on
    where a plan ends. \textbf{(c)}~Decoded for visualization, AVJ-WM's predicted end states keep the coarse layout of the goal view,
    but not its details. Costs are standardized within each episode (means with 95\% intervals). In the small tables, AUC is the probability that a pose (a) or plan (b) at the goal costs less than one elsewhere, and $\rho$ the median rank correlation of cost with distance.}
  \label{fig:avwmnavpred}
\end{figure*}
\begin{figure*}[t]
  \color{black}
  \captionsetup{font={color=black}, labelfont={color=black}}
  \centering
  \includegraphics[width=\textwidth]{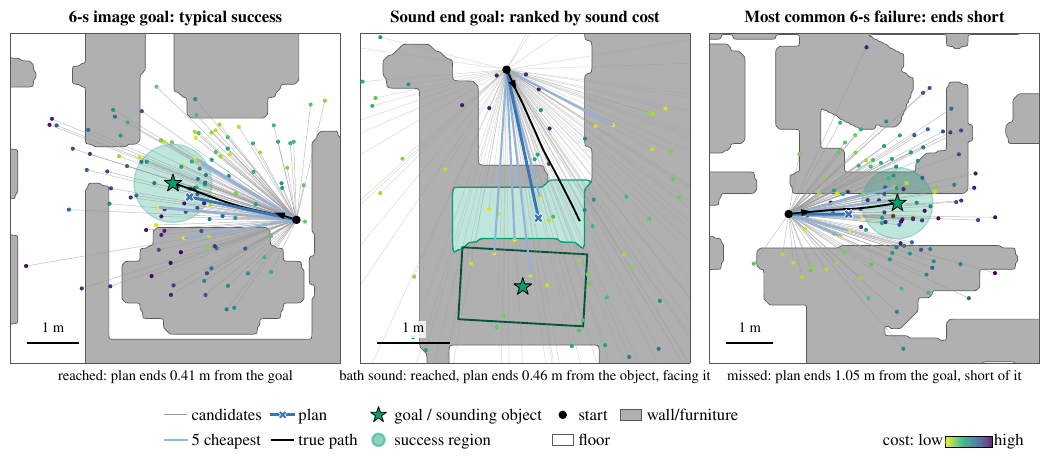}\\[-1mm]
  {\small (a)~Offline planning}\\[2mm]
  \includegraphics[width=\textwidth]{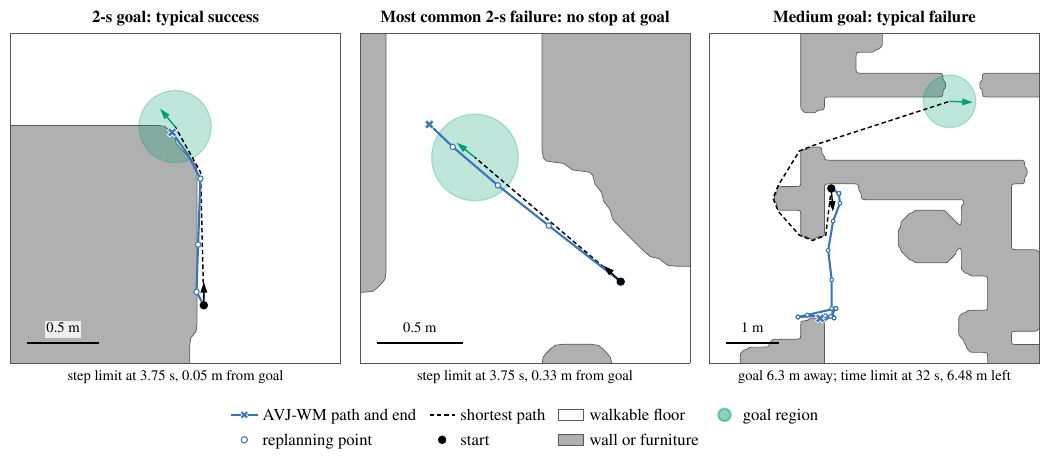}\\[-1mm]
  {\small (b)~Closed loop}
  \caption{\textbf{Planning with AVJ-WM on AVNav-Bench} (validation houses). \textbf{(a)}~Candidate plans colored by goal cost in offline planning, for a 6\,s success, a sound end goal and the most common 6\,s failure, a plan that ends short of the goal. \textbf{(b)}~Closed-loop paths for a typical 2\,s success, the most common 2\,s failure, which passes through the goal without stopping, and a medium-tier failure.}
  \label{fig:avwmnavqual}
\end{figure*}
\paragraph{Offline planning.} The further the goal, the larger our world models' lead
(\mainref{tab:avnavbench}). Close to the best diffusion world model at 2\,s, AVJ-WM-B and AVJ-WM lead the diffusion world
models by 5.1 and 7.2 points at 6\,s, and on end goals, 37 steps away, AVJ-WM reaches 39.6\% against at most 26.9\%. GNM is
strong at 2\,s but, like NoMaD, falls behind at longer horizons, partly because it predicts only 1.25\,s of path and NoMaD
no final heading. \Cref{fig:avwmnavqual}a shows a typical 6\,s success and the most common failure, a plan that ends short of the goal because the cheapest candidates themselves stop early. Offline, we score
only where each plan ends, in free space; the closed loop executes every move in the simulated house, walls included
(\cref{fig:avwmnavqual}b).
\paragraph{What limits offline planning.} The goal cost limits offline planning more than the world model
does (\cref{tab:avnavceiling}). To separate the two, we replace each candidate's predicted end state with the encoding of
the view rendered at its end pose. Even with these exact predictions, the planner stays well below a perfect cost (85
against 98\% at 2\,s), with either of our encoders and with the released \vjepatwoone encoders alike, whereas the patch
features of \vjepatwo-L plan far worse. Our world models come within a few points of exact predictions at 2--6\,s, so a
better world model alone can improve offline planning only within this margin. As most remaining failures end close to
the goal, we hypothesize that the cost ranks nearby views too coarsely.

\paragraph{Pretraining.} Pretraining (Stage~2 for AVJ-WM-B, action-free for the AV-CDiT) matters more for
our world model than for the CDiT (\cref{tab:avnavdata}). Without world-model pretraining, AVJ-WM-B reaches only 27.8\% at
2\,s, and with 10\% of the training walks it loses 7.8 points at 6\,s, against 11.3 for the AV-CDiT. Pretraining the
AV-CDiT without actions on our Stage~2 data, in turn, does not improve it.

\paragraph{Diffusion baselines.} Our small-batch fine-tuning beats NWM's own recipe on its released CDiTs at 2
and 4\,s, by up to 12 points, except for CDiT-XL pretrained with Ego4D, where the two are about level
(\cref{tab:avnavvariants}). Neither NWM's weights nor three training stages raise the AV-CDiT's success at 2--6\,s.

\paragraph{Fine-tuning recipe.} We remove one part of our Stage~3 recipe at a time from AVJ-WM-B
(\cref{tab:avwmnavrecipes}). Rollout training, long training rollouts and stand-still training are essential: without any
one of them, success falls sharply at 2\,s, at 6\,s or in the closed loop. Sound goals need both binaural changes of
\cref{sec:suppl_avnav_ours} (18.0 against 6.2\% without them). Short training windows match our recipe on these tasks, but we found that they reach medium goals even less often.

\paragraph{Closed loop.} On the 2\,s tier, our world models are close to the best baselines
(\mainref{tab:avnavbench}) and plan faster than every diffusion world model (\cref{tab:avnavplantime}). Most of AVJ-WM's
successes end when the step limit stops the walker inside the goal region, and its most common failure is to pass
through the goal region without stopping (\cref{fig:avwmnavqual}b).

\paragraph{Medium and hard tiers.} No world model solves the medium and hard tiers, whose goals lie in other
rooms. All reach at most 1.7\% on medium and 0.3\% on hard (we do not run the costly SD-VAE world models there). All their successes are image goals; no world model reaches a single medium or hard sound goal. With the
true distance as its cost and plans of 8\,s, the same planner reaches 46.5 and 11.0\%, which suggests that the goal
costs, more than the planner, limit planning across rooms. To see why, we walk along the shortest path to 112 medium test goals and compute the goal cost at every
step (\cref{fig:avwmnavpred}a). We compute it from the current view itself (exact) and from AVJ-WM-B's and the AV-CDiT's
predictions of standing still at that point. All three costs fall only within about 4\,m of the goal, roughly
once the walker has turned the last corner, and stay flat farther away. Policies trained by reinforcement learning do
better, almost only on sound goals, which they hear at every step. We examine how far AVJ-WM uses its 64\,s context
window in \cref{sec:supplavwmlongcontext}.

\paragraph{Sound goals.} Among the world models, only ours plan towards sound goals well above chance (18.0
and 19.1\% on sound end goals, against at most 5.0\% for the diffusion world models and 3.9\% for random costs). The
AV-CDiT's audio codec largely preserves the cues these goals rely on (\cref{tab:avnavdac}), which suggests that the codec
is not the main cause of this gap. Instead, the sound AVJ-WM-B predicts ranks the 120 candidate plans almost as well as
the exact sound, whereas the AV-CDiT's cost hardly depends on where a plan ends (\cref{fig:avwmnavpred}b). Along the
validation walks, too, the cost of AVJ-WM's predictions rises with the distance to the source in 75\% of the walks,
against 65\% for the AV-CDiT's predictions. When we decode them to pixels with a separately trained decoder, for visualization only, we observe that AVJ-WM's
predicted end states keep the coarse layout of the goal view but lose its details (\cref{fig:avwmnavpred}c; an image and a sound goal,
each the median of 20 random validation episodes). Still, in 90\% of the image-goal episodes on the validation houses,
the predicted end state is closer to the goal view than to any other view of the same house.

\begin{figure*}[t]
  \color{black}
  \captionsetup{font={color=black}, labelfont={color=black}}
  \centering
  \includegraphics[width=\textwidth]{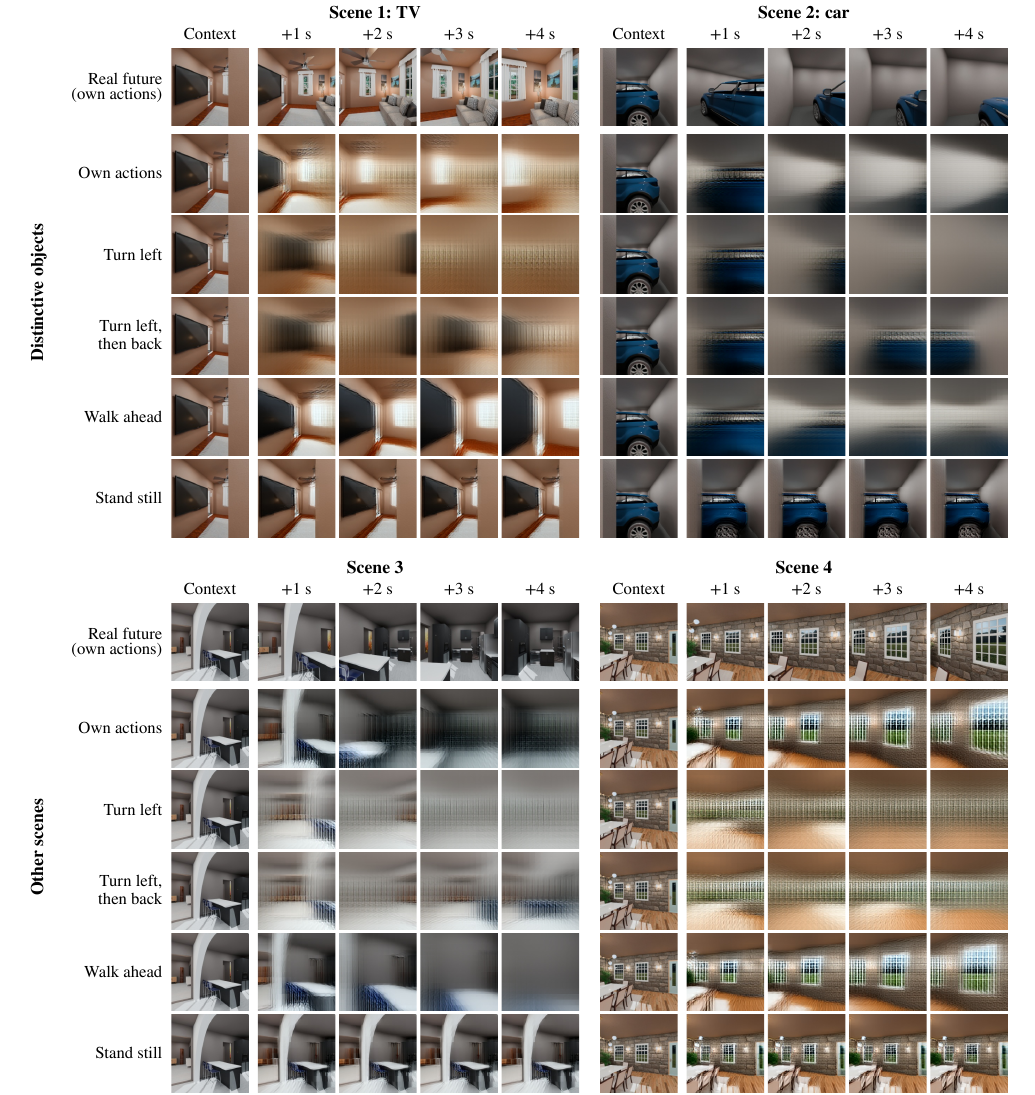}
  \caption{\textbf{Counterfactual rollouts on AVNav-Bench} (validation houses). From 1\,s of a walk, AVJ-WM imagines the next 4\,s under five action plans, shown below the real frames and decoded to pixels for visualization only. We show four of 30 scenes drawn at random, one per validation house, and picked the top two for a distinctive object (a TV, a car).}
  \label{fig:avwmrollouts}
\end{figure*}
\paragraph{Counterfactual rollouts.} We roll the same second of context out under
different action plans (\cref{fig:avwmrollouts}). AVJ-WM's predictions follow the walk's own actions more closely than a copy of the context (cosine similarity of the
tokens 0.80 against 0.52 at 2\,s, over 30 scenes),
standing still keeps the view, and objects that the walker turns away from and back to reappear roughly where they were.

\subsection{Relation to prior benchmarks}
\label{sec:suppl_avnav_prior}

\begin{table*}[t]
  \color{black}
  \captionsetup{font={color=black}, labelfont={color=black}}
  \centering
  \caption{\textbf{AVNav-Bench among navigation benchmarks with sound or world models.}
    \emph{Fixed walks}: a fixed set of recorded walks with actions to learn from (otherwise agents learn by interacting with the simulator, or need no training); \emph{Offline}: plans are scored without executing them; \emph{Closed loop}: the agent acts in the simulator, or on a robot, until it stops; \emph{WM plans}: a world model chooses the actions; MP3D: Matterport3D. \textsuperscript{a}Only its language parts are pretrained on fixed
    instruction data. \textsuperscript{b}A world model looks ahead for a pretrained policy. \textsuperscript{c}Audio
    only.}
  \label{tab:avnavprior}
  \footnotesize
  \setlength{\tabcolsep}{4pt}
  \providecommand{\stackh}[2]{\begin{tabular}[b]{@{}c@{}}#1\\#2\end{tabular}}%
  \begin{tabular}{@{}l l >{\raggedright\arraybackslash}p{212.2pt} c c c c@{}}%
    \toprule
    Benchmark & Scenes & Goal & \stackh{Fixed}{walks} & \stackh{Off-}{line} & \stackh{Closed}{loop} & \stackh{WM}{plans} \\
    \midrule
    \multicolumn{7}{@{}l@{}}{\cellcolor{black!8}\textbf{Audio-visual navigation}} \\
    Look, Listen, and Act~\cite{gan2020look} & 7 AI2-THOR apartments & a sound source & \xmark & \xmark & \cmark & \xmark \\
    SoundSpaces~\cite{chen2020soundspaces} & Replica, MP3D & a sound source, optionally with its position & \xmark & \xmark & \cmark & \xmark \\
    SAVi~\cite{chen2021savi} & MP3D & a sounding object; its sound stops & \xmark & \xmark & \cmark & \xmark \\
    SAAVN~\cite{yu2022saavn} & Replica, MP3D & a sound source, with an adversarial distractor & \xmark & \xmark & \cmark & \xmark \\
    Moving sources~\cite{younes2023catch} & Replica, MP3D & a static or moving sound source, with distractors & \xmark & \xmark & \cmark & \xmark \\
    AVLEN~\cite{paul2022avlen} & MP3D & a sounding object, with language help on request & \xmark & \xmark & \cmark & \xmark \\
    CAVEN~\cite{liu2024caven} & MP3D & a sounding object, with dialogue & \xmark\textsuperscript{a} & \xmark & \cmark & \xmark \\
    SoundSpaces~2.0~\cite{chen2022soundspaces2} & SoundSpaces episodes & a sound source & \xmark & \xmark & \cmark & \xmark \\
    Sonicverse~\cite{gao2023sonicverse} & iGibson, MP3D; a robot & a sounding object; its sound stops & \xmark & \xmark & \cmark & \xmark \\
    BeDAViN~\cite{shi2025enmus} & MP3D & an object of a given category, alone or with a distractor and noise & \xmark & \xmark & \cmark & \xmark \\
    SAVN-CE~\cite{zeng2026savnce} & MP3D & a sounding object; its sound starts late and stops & \xmark & \xmark & \cmark & \xmark \\
    LH-AVLN~\cite{chen2026lhavln} & MP3D & 2--4 goals, by category, text or image, with sounds & \xmark & \xmark & \cmark & \xmark \\
    \midrule
    \multicolumn{7}{@{}l@{}}{\cellcolor{black!8}\textbf{World models for navigation}} \\
    NWM~\cite{bar2025nwm} & robot logs; no audio & an image of the goal & \cmark & \cmark & \xmark & \cmark \\
    AV-CDiT (AVW-4k)~\cite{wang2025avwm} & MP3D & one fixed sound source & \cmark & \xmark & \cmark & \cmark\textsuperscript{b} \\
    AudioWorldSim~\cite{zerkowski2026audioworldsim} & Replica, MP3D & none (a dataset) & \cmark\textsuperscript{c} & \xmark & \xmark & \xmark \\
    \midrule
    \oursrow AVNav-Bench (ours) & HSSD~\cite{khanna2024hssd} & an image at the goal, or the sound of a named object & \cmark & \cmark & \cmark & \cmark \\
    \bottomrule
  \end{tabular}
\end{table*}
\Cref{tab:avnavprior} compares AVNav-Bench with existing benchmarks. AV navigation benchmarks, from SoundSpaces~\cite{chen2020soundspaces,chen2022soundspaces2} onwards, evaluate agents trained mainly by reinforcement learning
in the simulator, or that need no training, and vary the goal, from a sound source to a sounding object whose sound
stops~\cite{chen2021savi,zeng2026savnce}, moving~\cite{younes2023catch} or distracting sounds~\cite{yu2022saavn,shi2025enmus},
language help~\cite{paul2022avlen,liu2024caven} and several goals~\cite{chen2026lhavln}. NWM~\cite{bar2025nwm} plans offline
with a video-only world model on real robot logs, and AV-CDiT~\cite{wang2025avwm} is evaluated on audio-visual prediction
and as a lookahead for a pretrained policy navigating to one fixed sound source.
AudioWorldSim~\cite{zerkowski2026audioworldsim} generates binaural audio with actions for training audio world models but
defines no task.

AVNav-Bench instead specifies the goal itself, as an image taken at the goal or the sound of a named object, and
tests a world model as planner; to our knowledge, its closed loop is the first audio-visual navigation setting in which the
world model itself decides, with no policy in the loop. A fixed set of 9,702 training walks with actions (72\,h), as in
NWM and AVW-4k, lets latent and pixel world models and policies learn from the same data. All world models rank the same
candidates in each episode, so they share the chance level and the ceiling, and the score mainly reflects how well a model judges plans. The same model is tested offline and in the closed loop, on image and sound goals up to several rooms away, with
path-traced frames and 48\,kHz binaural and first-order ambisonic audio. AVNav-Bench does not test distractor, moving or competing sources, sounds that stop, or language help, though these would be valuable future additions.

} %

\FloatBarrier%
\makeatletter\def\ftype@table{2}\makeatother%

\FloatBarrier

\end{document}